\documentclass[twoside,11pt]{article}

\usepackage{jmlr2e_arxiv}

\usepackage[breaklinks=true]{hyperref}
\usepackage{breakurl}

\usepackage{amsmath}
\usepackage{amssymb}
\usepackage{dsfont}
\usepackage[latin1]{inputenc}
\usepackage{textcomp}
\usepackage{url}
\usepackage{framed}
\usepackage{graphics,overpic,color}
\usepackage{overpic}
\usepackage{color}
\usepackage[font=small]{caption}
\usepackage{wrapfig}
\usepackage[hang,footnotesize]{subfigure}
\usepackage[arrow, matrix, curve]{xy}
\usepackage{array} 
\usepackage{longtable}
\usepackage{algorithm}
\usepackage{algorithmic}
\usepackage{capt-of}
\usepackage{multirow}
\usepackage{rotating}
\usepackage{rotfloat}
\usepackage{tikz}
\usetikzlibrary{matrix}
\usepackage{paralist}
\usepackage[emulate=units]{siunitx} 

\newcommand{\RM}[1]{\MakeUppercase{\romannumeral #1}}

\definecolor{rot}{RGB}{255,0,0}
\definecolor{gruen}{RGB}{0,255,0}
\definecolor{blau}{RGB}{0,0,255}
\definecolor{violetred}{RGB}{208,32,144}

\DeclareSymbolFont{wideparensymbol}{OMX}{yhex}{m}{n}
\DeclareMathAccent{\wideparen}{\mathord}{wideparensymbol}{"F3}

\newlength{\fixboxwidth}
\newcommand{\N}{\mathbb{N}}

\newcommand{\R}{\mathbb{R}}
\newcommand{\Rd}{\mathbb{R}^d}

\newcommand{\E}{\mathbb{E}}

\newcommand{\RR}{\mathcal{R}}

\newcommand{\D}{\mathrm{D}}

\newcommand{\fcl}{\wideparen{f}}
\newcommand{\fB}{f^*_{L,\P}}
\newcommand{\Lb}{\Lambda}

\newcommand{\ind}{\mathds{1}}
\newcommand{\cl}[1]{\wideparen{#1}}

\newcommand{\RLP}{\mathcal{R}_{L,\P}}

\newcommand{\RLPj}{\mathcal{R}_{L_{j},\P}}
\newcommand{\RLPJ}{\mathcal{R}_{L_J,\P}}
\newcommand{\RLPJT}{\mathcal{R}_{L_{J_T},\P}}
\newcommand{\RLPT}{\mathcal{R}_{L_{T},\P}}

\newcommand{\RLD}{\mathcal{R}_{L,\D}}
\newcommand{\RLDj}{\mathcal{R}_{L_{j},\D}}

\newcommand{\RB}{\mathcal{R}_{L,\P}^*}
\newcommand{\RBj}{\mathcal{R}_{L_{j},\P}^*}
\newcommand{\RBJT}{\mathcal{R}_{L_{J_T},\P}^*}

\newcommand{\RBJ}{\mathcal{R}_{L_J,\P}^*}
\newcommand{\RBT}{\mathcal{R}_{L_T,\P}^*}

\newcommand{\NN}{\mathcal{N}}
\newcommand{\I}{\mathcal{I}}

\renewcommand{\L}{\mathcal{L}}

\newcommand{\bs}{\boldsymbol}
\newcommand{\TrP}[1]{{\P_X}_{|#1}}
\newcommand{\Cpp}{C\nolinebreak\hspace{-.05em}\raisebox{.4ex}{\tiny\bf +}\nolinebreak\hspace{-.10em}\raisebox{.4ex}{\tiny\bf +}}

\newcommand{\argmin}[1]{\underset{#1}{\operatorname{arg\,min}}}

\renewcommand{\a}{\alpha}
\renewcommand{\b}{\beta}
\newcommand{\g}{\gamma}
\newcommand{\G}{\Gamma}
\renewcommand{\d}{\delta}
\newcommand{\e}{\varepsilon}

\newcommand{\vt}{\vartheta}

\newcommand{\lb}{\lambda}
\newcommand{\x}{\xi}

\renewcommand{\t}{\tau}

\renewcommand{\P}{\mathrm{P}}

\DeclareMathOperator{\supp}{supp}

\jmlrheading{?}{????}{?-??}{?/??}{??/??}{Mona Eberts and Ingo Steinwart}

\ShortHeadings{Optimal Learning Rates for Localized SVMs}{Eberts and Steinwart}
\firstpageno{1}

\begin{document}

\title{Optimal Learning Rates for Localized SVMs}

\author{\name Mona Eberts \email mona.eberts@mathematik.uni-stuttgart.de \\
	\name Ingo Steinwart \email ingo.steinwart@mathematik.uni-stuttgart.de \\
	\addr Institute for Stochastics and Applications\\ 
	University of Stuttgart\\
	70569 Stuttgart, Germany}

\editor{}

\maketitle

\begin{abstract}
 One of the limiting factors of using support vector machines (SVMs) in large scale applications
	are their super-linear computational requirements in terms of the number of training 
	samples. To address this issue,  several approaches that train SVMs on many small chunks of large data sets separately
	have been proposed in the literature. So far, however, almost all these approaches have only been 
	empirically investigated. In addition,  
	their motivation was always based on computational requirements.
 In this work, we consider  a localized SVM approach based upon a 
 partition of the input space. For this  local SVM, we derive a general oracle inequality. 
 Then we apply this oracle inequality to 
 least squares regression using Gaussian kernels and deduce local learning rates that are 
 essentially minimax optimal under some standard  smoothness assumptions on 
 the regression function. 
	This gives the first 
	motivation for using local SVMs that is not based on computational requirements but on theoretical 
	predictions on the generalization performance.
	We further introduce a data-dependent parameter selection 
 method for our local SVM approach and show that this method achieves the same learning rates as before. 
 Finally, we present some larger scale experiments for our localized SVM showing that it achieves
	essentially the same test performance as a global SVM for a fraction of the computational requirements.
	In addition, it turns out that the computational requirements for the local SVMs are similar to those of 
	a vanilla random chunk approach, while the achieved test errors are significantly better.
%
%
\end{abstract}


\allowdisplaybreaks

\section{Introduction}

Based on a training set $D:=((x_1,y_1),\ldots,(x_n,y_n))$ of i.i.d.~input/output 
observations drawn from an unknown distribution $\P$ on $X\times Y$, where $X\subset\Rd$ 
and $Y\subset\R$, the goal of non-parametric regression is to find a function 
$f_D : X\to\R$ such that important characteristics of the conditional distribution 
$\P(Y|x)$, $x\in X$, can be recovered. 
For instance, an $f_D$ approximating the conditional mean $\E(Y|x)$, $x\in X$, 
is sought in the non-parametric least squares regression.
This classical non-parametric regression problem has been extensively studied in the 
literature, where a general reference is the book \citep{GyKoKrWa02}, presenting 
plenty of results concerning the non-parametric least squares regression.

In the literature, 
there are many learning methods that solve the non-parametric regression problems, some 
of them are e.g.~described in \citep{GyKoKrWa02,Koenker05,Sim96}. 
In this paper, we utilize some kernel-based regularized empirical risk minimizers, 
also known as support vector machines (SVMs), which 
solve the regularized problem
\begin{align}\label{standard SVM}
 f_{\D,\lb} 
\in \arg\min_{f\in H} \lb\left\|f\right\|_{H}^2+\RLD\left(f\right) \,.
\end{align}
Here, $\lb>0$ is a fixed real number and $H$ is a reproducing kernel Hilbert space (RKHS) 
over $X$ with reproducing kernel $k:X\times X \to\R$, see e.g.~\citep{Aronszajn50a,BeTA04, StCh08}. 
Besides, $\RLD\left(f\right)$ denotes 
the empirical risk of a function $f:X\to\R$, that is
\begin{align*}
 \RLD\left(f\right)
 =\frac{1}{n}\sum_{i=1}^{n}L\left(x_i,y_i,f\left(x_i\right)\right) \,,
\end{align*}
where $\D$ is the empirical measure associated to the data $D$ defined by 
$\D:=\frac{1}{n}\sum_{i=1}^n\d_{(x_i,y_i)}$ with Dirac measure $\d_{(x_i,y_i)}$ at 
$(x_i,y_i)$. 
Note that the empirical SVM solution $ f_{\D,\lambda} $ exists and is unique 
\citep[cf.][Theorem 5.5]{StCh08} whenever the loss $L$ is convex in its last argument. 
Moreover, an SVM is $L$-risk consistent under a few assumptions on the RKHS $H$ and the 
regularization parameter $\lb$, see \citep[Section 6.4]{StCh08} for more details. 
Besides, it is worth mentioning that the ability to choose the RKHS $H$ as well as 
the loss function $L$ in \eqref{standard SVM} provides the possibility to flexibly 
apply SVMs to various learning problems. Namely,  the learning target is 
modeled by the loss function, e.g.~the least squares loss is used to estimate the 
conditional mean. Moreover, since RKHSs 
are defined on arbitrary $X$,  data types that are not $\Rd$-valued can be handled, too.
Furthermore, SVMs are enjoying great popularity, since they can be implemented and 
applied in a relatively simple way and only have a few free parameters that can usually be
determined by cross validation.

\begin{wrapfigure}{r}{0.45\textwidth}
\begin{scriptsize}
    \begin{center}
      \begin{overpic}[scale=0.48]{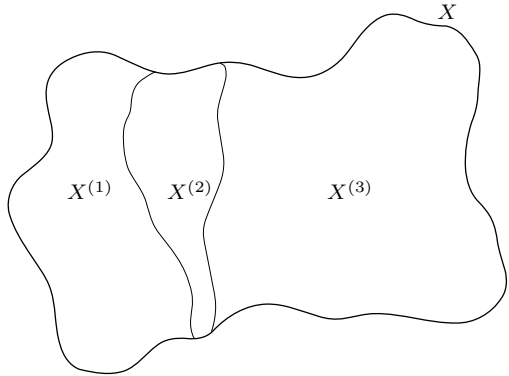}
	\put(86,71){$X$}
	\put(12,35){$X^{(1)}$}
	\put(32,35){$X^{(2)}$}
	\put(64,35){$X^{(3)}$}
      \end{overpic}
      \caption{The input space $X$ is partitioned by $X^{(1)}$, $X^{(2)}$, 
      and $X^{(3)}$ such that the regression function $f^*$ is less smooth on $X^{(2)}$ 
      compared to $X^{(1)}$ and $X^{(3)}$. However, it is desirable to achieve locally optimal 
      learning rates.}
      \label{Bild_Gebiet}
    \end{center}
\end{scriptsize}\vspace{-0.5cm}
\end{wrapfigure}

An essential theoretical task, which has attracted many considerations, is the investigation of 
learning rates for SVMs. For example, such rates for SVMs using the least squares loss
and generic kernels can be found in
 \citep{CuSm02a,DeCaRo05a,SmZh07a,CaDe07a,MeNe08a, StHuSc09b} and the references therein,
while similar rates for 
SVMs using the pinball loss can be found in \citep{StCh08c,StCh11a}. 
At this point, we 
do not want to take a closer look at these results, instead we relegate 
to \citep{ EbSt13}, where a detailed discussion can be found. 
More important for our purposes is the fact that 
\citet{EbSt11,EbSt13} establish (essentially) asymptotically optimal
learning rates for least squares SVMs (LS-SVMs) using Gaussian RBF kernels.
More precisely, for a domain $X\subset B_{\ell_2^d}$, $Y:=[-M,M]$ with $M>0$, 
a distribution $\P$ on $X\times Y$ such that $\P_X$ has a bounded Lebesgue density on 
$X$, and for $f^*$ contained in the Sobolev space $W_2^{\a}(\P_X)$, $\a\in\N$, or in 
the Besov space $B_{2,\infty}^\a(\P_X)$, $\a\geq 1$, respectively, the LS-SVM using 
Gaussian kernels learns for all $\xi>0$ with rate $n^{-\frac{2\a}{2\a+d}+\xi}$ with a 
high probability. 
Although these rates are essentially asymptotically optimal, 
they depend on the  order of smoothness of the regression function on the 
\emph{entire} input space $X$.
 That is, if the regression function $f^*$ is on 
some area of $X$ smoother 
than on another area, the learning rate is determined by the part of $X$, where the 
regression function $f^*$ is least smooth (cf.~Figure \ref{Bild_Gebiet}).
 In contrast 
to this, it would be desirable to achieve  a learning rate on every region of $X$ that 
corresponds with the order of smoothness of $f^*$ on this region. 
Therefore, one of our goals of this paper is to modify the standard SVM approach such that we 
achieve local learning rates that are asymptotically optimal. 
Our technique to achieve such local learning rates is a special 
 local SVM approach. Local SVMs
have been extensively investigated in the literature to speed-up the training time, see 
for instance, the early works 
\citep{BoVa92,VaBo93}. The basic idea of many local approaches  is to \emph{a)} split the 
training data and just consider a few examples near a testing sample, \emph{b)} train on this 
small subset  of the training data, and \emph{c)} use the solution for a prediction w.r.t.~the test 
sample. Here, many up-to-date investigations use SVMs to train on the local data set 
but, yet there are different ways to split the whole training data set into smaller, 
local sets. For example, \citet{ChGuLiLu10, WuBeCrSh99, BeBl98} use decision
trees while in \citep{Ha13, SeBl10, SeBl08, BlMe08, BlBr07a, BlBr07b, ZhBeMaMa06} 
local subsets are built considering $k$ nearest neighbors. The latter approaches further vary, for example, 
\citet{ZhBeMaMa06, BlBr07a, Ha13} consider different metrics w.r.t.~the input space 
whereas  \citet{SeBl08, BlMe08, BlBr07b} consider metrics w.r.t.~the feature space. 
Nonetheless, the basic idea of all these articles is that an SVM problem based on $k$ training samples is 
solved for \emph{each} test sample. Another approach using $k$ nearest neighbors is 
investigated in \citep{SeBl10}. Here, $k$-neighborhoods consisting of training samples 
and collectively covering the training data set are constructed and an SVM is calculated 
on each neighborhood. The prediction for a test sample is then made according to the 
nearest training sample that is a center of a $k$-neighborhood. 
As for the other nearest neighbor approaches, however, the results 
are mainly experimental. An exception to this rule is \citep{Ha13}, where universal 
consistency for   localized versions of SVMs, or more precisely,  a large class of regularized 
kernel methods, is proven. 
Another article presenting theoretical results for localized versions of learning methods 
is \citep{ZaRi09}. Here, the authors show that a consistent learning method behaves 
locally, i.e.~the prediction is essentially influenced by  close by samples. However, this 
result is based on a localization technique considering only training samples contained 
in a neighborhood with a fixed radius and center $x$ when an estimate in $x$ is sought. 
Probably closest to our approach is the one examined in \citep{ChTaJi10} and 
\citep{ChTaJi07}, where the training data is splitted into clusters and then an SVM is
trained on each cluster. However, the presented results are only of experimental 
character. 

In this article, we partition the input space $X$ according to a cover of $X$ with 
radius $r_n$ and build an SVM model for each partition cell.
The following section is dedicated to the detailed description of this method. Section 
\ref{general results} then presents   some theoretical results that enable the 
analysis of this new method. For example, we examine extensions and direct sums of 
RKHSs. At the end of Section \ref{general results}, we finally present a first oracle 
inequality for the localized  SVM.  
In Section \ref{Sec entropy estim}, we focus 
on RKHSs using Gaussian RBF kernels and, in conjunction with that, we study some entropy 
estimates. After that, Section \ref{Sec LS loss} 
concentrates on the least squares loss and introduces  
an oracle inequality and learning rates for our localized SVM method using 
Gaussian kernels. Moreover, a data-dependent parameter selection method is 
studied that induces the same rates. Section \ref{Sec Experiments} then presents 
some experimental results w.r.t.~the localized SVM technique. All proofs can be found in Section \ref{sec proofs},
and the appendix contains various tables displaying detailed results of our experiments.

\section{Description of the Localized SVM Approach}\label{VP-SVM illustration}

In this section, we introduce some general notations and assumptions. 
Based on the latter we modify the standard SVM approach.
Let us start with the probability measure $\P$ on $X\times Y$, where $X\subset\Rd$ is 
non-empty and $Y:=[-M,M]$ for some $M>0$. 
Depending on the learning target one chooses a loss 
function $L$, i.e.~a function $L : X\times Y\times\R\to[0,\infty)$ that is measurable. 
Then, for a measurable function $f:X\to\R$, the $L$-risk is defined by
\begin{align*}
 \RR_{L,\P}(f)=\int_{X\times Y} L(x,y,f(x)) \,d\P(x,y) 
\end{align*}
and the optimal $L$-risk, called the Bayes risk with respect to $\P$ and $L$, is given by
\begin{align*}
 \RR^*_{L,\P}:=\inf\left\{\RR_{L,\P}\left(f\right) \ |
 \ f : X\to \R \text{ measurable}\right\} \,.
\end{align*}
A measurable function $f^*_{L,\P} : X\to\R$ with $\RR_{L,\P}(f^*_{L,\P})=\RR^*_{L,\P}$ 
is called a Bayes decision function. 
For the commonly used losses such as the least squares loss 
treated in Section \ref{Sec LS loss} 
the Bayes decision function $\fB$ is $\P_X$-almost surely 
$[-M,M]$-valued, since $Y=[-M,M]$. In this case, it seems 
obvious to consider estimators with values in $[-M,M]$ on $X$. To this end, we now 
introduce the concept of clipping the decision function. Let $\cl{t}$ be the clipped 
value of some $t\in \R$ at $\pm M$ defined by
\begin{align*}
 \cl{t}:=\begin{cases}
	-M  & \text{if }t<-M\\
	t  & \text{if }t\in [-M,M]\\
	M & \text{if }t>M \,.
\end{cases}
\end{align*}
Then a loss is called clippable at $M>0$ if, for all $(x,y,t)\in X\times Y\times\R$, 
we have
\begin{align*}
 L(x,y,\cl{t}\,)\leq L(x,y,t) \,.
\end{align*}
Obviously, the latter implies
\begin{align*}
 \RLP(\fcl\, ) \leq \RLP(f) 
\end{align*}
for all $f : X \to \mathbb{R}$. In other words, restricting the decision function to 
the interval $[-M,M]$ containing our labels cannot worsen the risk, in fact, clipping 
this function typically reduces the risk. Hence, we consider the clipped version $\fcl_D$ 
of the decision function as well as the risk $\RLP(\fcl_D)$ instead 
of the risk $\RLP(f_D)$ of the unclipped decision function. Note, this clipping idea 
does \emph{not} change the learning method since it is performed \emph{after} the 
training phase.

To modify the standard SVM approach \eqref{standard SVM}, we assume that 
$(A_{j})_{j=1,\ldots,m}$ is a partition of $X$ such that $\mathring{A}_j\neq\emptyset$ 
for every $j\in\{1,\ldots,m\}$. Obviously, this implies $A_{j_1}\cap A_{j_2}=\emptyset$ for all 
$j_1,j_2\in\{1,\ldots,m\}$ with $j_1\neq j_2$ and
\begin{align*}
 X = \bigcup_{j=1}^{m} A_{j} \,.
\end{align*}
Now, the basic idea of the approach developed in this paper is to consider for each set 
of the partition $(A_{j})_{j=1,\ldots,m}$ an individual SVM. 
To describe this approach in a mathematically rigorous way, we have to introduce some more definitions and notations. 
Let us begin with the index set 
\begin{align*}
 I_{j}:=\{i\in\{1,\ldots,n\}: x_i\in A_{j}\} \,,\qquad\qquad j=1,\ldots,m\,,
\end{align*}
indicating the samples of $D$ contained in $A_{j}$, as well as the corresponding data 
set
\begin{align*}
 D_{j}:=\{(x_i,y_i)\in D : i\in I_{j}\} \,,\qquad\qquad j=1,\ldots,m\,.
\end{align*}
Moreover, for every $j\in\{1,\ldots,m\}$, we define a (local) loss function 
$L_j:X\times Y\times\R\to[0,\infty)$ by
\begin{align}
 L_j(x,y,t):=\ind_{A_{j}}(x)L(x,y,t) \,, \label{L_j}
\end{align}
where $L : X\times Y\times\R\to[0,\infty)$ is the loss that corresponds to our learning 
problem at hand. We further assume that 
$H_j$ is an RKHS over $A_{j}$ with kernel $k_j : A_{j}\times A_{j}\to\R$. 
Note that every function $f\in H_j$ is only defined on $A_j$ even though a function 
$f_D : X\to\R$ is finally sought.
To this end, for $f\in H_j$, we define a function $\hat{f} : X\to\R$ by
\begin{align*}
 \hat{f}(x) := \begin{cases}
                f(x)\,, & x\in A_j\,, \\
                0\,, & x\notin A_j\,.
               \end{cases}
\end{align*}
Then the space $\hat{H}_j:=\{\hat{f} : f\in H_j\}$ equipped with the norm 
\begin{align*}
 \|\hat{f}\|_{\hat{H}_j}:=\|f\|_{H_j}\,,\qquad\qquad \hat{f}\in\hat{H}_j\,,
\end{align*}
is an RKHS on $X$ (cf.~Lemma \ref{Lemma extended RKHS}). That is, $\hat{H}_j$ is an 
isometrically isomorphic extension of the RKHS $H_j$ on $A_j$ to an RKHS on $X$.
After all, we are now able to formulate a modified SVM approach. 
To this end, for every $j\in\{1,\ldots,m\}$, consider the local SVM optimization 
problem
\begin{align}\label{SVM_j}
 f_{\D_{j},\lb_j}=\arg\min_{\hat{f}\in \hat{H}_j}\lb_j\|\hat{f}\|_{\hat{H}_j}^2 
 + \frac{1}{n}\sum_{i=1}^nL_j(x_i,y_i,\hat{f}(x_i)) \,,
\end{align}
where $\lb_j>0$ for every $j\in\{1,\ldots,m\}$.
Based on these empirical SVM solutions, 
we then define the decision function $f_{\D,\bs\lb} : X \to \R$ by
\begin{align} \label{VP-SVM}
 f_{\D,\bs\lb}(x) := \sum_{j=1}^{m} f_{\D_{j},\lb_j}(x)
               = \sum_{j=1}^{m} \ind_{A_{j}}(x)f_{\D_{j},\lb_j}(x) \,,
\end{align}
where $\bs\lb:=(\lb_1,\ldots,\lb_m)$. Here, clipping $f_{\D,\bs\lb}$ at $M$ yields 
\begin{align*}
 \fcl_{\D,\bs\lb}(x) = \sum_{j=1}^{m} \ind_{A_{j}}(x)\fcl_{\D_{j},\lb_j}(x)
\end{align*}
for every $x\in X$.
Note that the empirical SVM solutions $f_{\D_{j},\lb_j}$ in \eqref{SVM_j} exist and are 
unique by \citep[Theorem 5.5]{ StCh08} and that, for arbitrary $j\in\{1,\ldots,m\}$, 
$f_{\D_j,\lb_j}=0$ if $x_i \notin A_j$ for all $i\in\{1,\ldots,n\}$. 
In addition, the SVM optimization problem \eqref{SVM_j} equals the SVM optimization 
problem \eqref{standard SVM} using $H_j$, $D_j$, and the regularization parameter 
$\tilde{\lb}_j:=\frac{n}{|I_j|}\lb_j$, since, for $\hat{f}\in \hat{H}_j$ and 
$f:=\hat{f}_{|A_j}$, we have
\begin{align*}
   \lb_j\|\hat{f}\|_{\hat{H}_j}^2 + \frac{1}{n}\sum_{i=1}^nL_j(x_i,y_i,\hat{f}(x_i)) 
 &= \lb_j\|f\|_{H_j}^2 + \frac{1}{n}\sum_{i\in I_j}^nL(x_i,y_i,f(x_i)) \\
 &= \frac{|I_j|}{n} \left(\tilde{\lb}_j\|f\|_{H_j}^2 + \RR_{L,\D_j}(f)\right)\,.
\end{align*}
That is, $f_{\D_{j},\lb_j}$ as in \eqref{SVM_j} and $h_{\D_{j},\tilde{\lb}_j}:=
\arg\min_{f\in H_j}\tilde{\lb}_j\|f\|_{H_j}^2 + \RR_{L,\D_j}(f)$ satisfy
$$ h_{\D_{j},\tilde{\lb}_j} = {f_{\D_{j},\lb_j}}_{|A_j}\,. $$

For the sake of completeness, we briefly examine the Bayes risks w.r.t.~$\P$ and $L_j$. 
To this end, let $X\subset\Rd$, $Y\subset\R$, 
$L: X\times Y\times\R\to [0,\infty)$ be a loss function and $\P$ be a distribution on 
$X\times Y$ such that a Bayes decision function $\fB:X\to\R$ exists. Then, for all 
$j\in\{1,\ldots,m\}$ and losses $L_j$ defined by \eqref{L_j}, it is easy to show
\begin{align*}
 \RLPj(\fB)=\RBj \,,
\end{align*}
whenever $\fB$ exists. 
In other words, a Bayes decision function $\fB$ w.r.t.~$\P$ and $L$ additionally is a 
Bayes decision function w.r.t.~$\P$ and $L_j$.
Moreover, for function spaces $\mathcal{F}_1,\ldots,\mathcal{F}_m$ over $X$, we have
\begin{align}\label{swapping sum and min}
 \sum_{j=1}^m \min_{f_j\in\mathcal{F}_j} \RR_{L_j,\D}(f_j)
 = \min_{f_1\in\mathcal{F}_1,\ldots,f_m\in\mathcal{F}_m} \sum_{j=1}^m \RR_{L_j,\D}(f_j)
\end{align}
by the construction of the loss $L_j$.

Let us now present an advantageous characteristic of our modified SVM, 
namely the required computing time. Solving an usual SVM problem has a computational 
cost of $\mathcal{O}(n^q)$ where $q\in[2,3]$ and $n$ is the sample size. 
For the new approach we consider $m$ working sets of size $n_1,\ldots,n_m$ where 
$n_i\approx n_j$ for all $i,j\in\{1,\ldots,m\}$, i.e. $n_i\approx \tfrac{n}{m}$. Then 
for each working set an usual SVM problem has to be solved such that, altogether, the
modified SVM induces a computational cost of 
$\mathcal{O}\left(m\left(\frac{n}{m}\right)^q\right)$. 
That is, for some $\b>0$ and $m\approx n^\b$ our approach is computationally 
cheaper than a traditional SVM.
Note that our strategy using a partition of the input space is a typical way to 
speed-up algorithms and handle large data sets. Other techniques that possess similar 
properties are e.g.~applied in the articles cited in the introduction. Besides, we 
refer to \citep{TsKoKw07} and \citep{TsKwCh05} using enclosing ball problems to solve an 
SVM, to \citep{GrCoBoDuVa05} presenting an model of multiple filtering SVMs and 
to \citep{CoBeBe01} investigating a mixture of SVMs based on several subsets of the 
training set.

\begin{figure}[tbp]
  \begin{scriptsize}  
  \hfill\begin{minipage}[t]{0.45\textwidth}
	\begin{center}
         \begin{overpic}[scale=0.48]{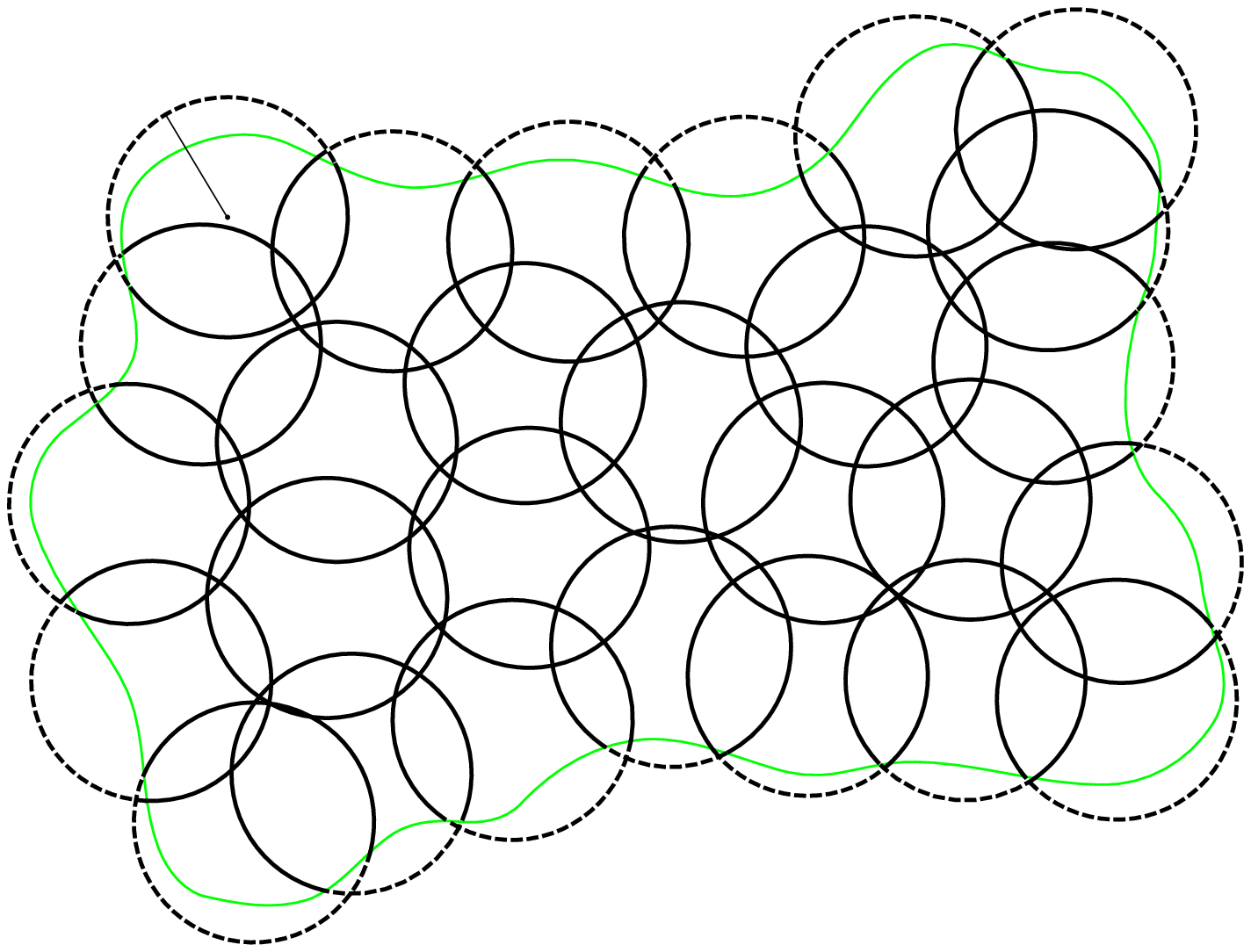}
	  \put(12,61){$r$}
	  \put(18.5,59){$z_j$}
	  \put(14,70){$B_{j}$}
	  \put(86,71){{\color{gruen}$X$}}
	 \end{overpic}
	\caption{Cover $(B_{j})_{j=1,\ldots,m}$ of $X$, where 
                 $B_{1},\ldots,B_{m}$ are balls with radius $r$ and centers 
                 $z_j \ (j=1,\ldots,m)$.}
        \label{Bild_cover}
	\end{center}
  \end{minipage}\hfill
  \begin{minipage}[t]{0.45\textwidth}
	\begin{center}
         \begin{overpic}[scale=0.48]{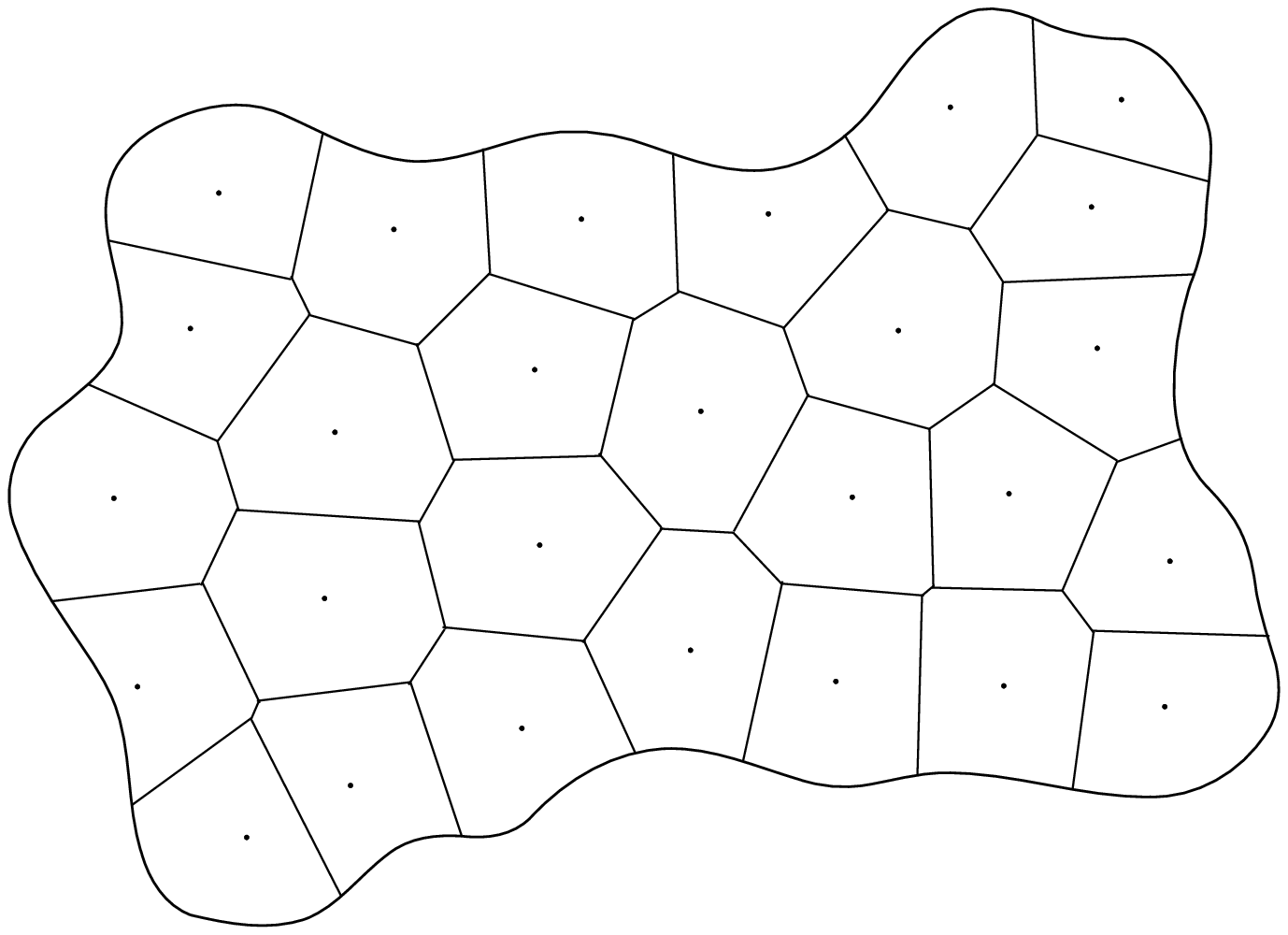}
	  \put(86,71){$X$}
          \put(18,57){$z_j$}
          \put(16,61){$A_{j}$}
	 \end{overpic}
	\caption{Voronoi partition $(A_{j})_{j=1,\ldots,m}$ of $X$ defined by 
                 \eqref{Def. VP A_{j}}, where $A_{j}\subset B_{j}$ for every 
                 $j\in\{1,\ldots,m\}$.}
	\label{Bild_Partition}
	\end{center}
  \end{minipage}\hfill
  \end{scriptsize}
\end{figure}

To describe the above SVM approach $(A_{j})_{j=1,\ldots,m}$ only has to 
be some partition of $X$. However, for the theoretical investigations concerning 
learning rates of our new approach, we have to further specify the partition. To this end, 
we denote by $B_{\ell_2^d}$ the closed unit ball in the 
$d$-dimensional Euclidean space $\ell_2^d$ and we define
balls $B_{1},\ldots,B_{m}$ with radius $r>0$ and mutually distinct centers 
$z_1,\ldots,z_m \in X$ by
\begin{align}\label{Def. B_j}
 B_j := B_{r}(z_j) 
     := \{x\in X : \|x-z_j\|_2\leq r\} \,,  \qquad\qquad j\in\{1,\ldots,m\}\,,
\end{align}
where $\|\,\cdot\,\|_2$ is the Euclidean norm in $\Rd$. 
Moreover, choose $r$ and $z_1,\ldots,z_m$ such that 
\begin{align*}
 \bigcup_{j=1}^m B_j = X\,,
\end{align*}
i.e.~such that the balls $B_{1},\ldots,B_{m}$ cover $X$ (cf.~Figure \ref{Bild_cover}). 
The following well-known lemma relates the radius of such a cover with the 
number of centers.
\begin{lemma}\label{Lem relation r and m}
Let $X\subset \Rd$ be a bounded subset, i.e.~$X\subset c B_{\ell_2^d}$ for some 
constant $c>0$.
Then there exist balls $(B_{j})_{j=1,\ldots,m}$ with radius $r>0$ covering $X$ such that
\begin{align*}
 r \leq 8 c m^{-\frac{1}{d}} \,.
\end{align*}
\end{lemma}
For simplicity of notation, we assume in the following that $X\subset B_{\ell_2^d}$, 
i.e.~according to Lemma \ref{Lem relation r and m} there exists a cover 
$(B_{j})_{j=1,\ldots,m}$ with
\begin{align}\label{ex. Ueberdeckung}
 r \leq 8 m^{-\frac{1}{d}} \,. 
\end{align}
Finally, we can specify the partition $(A_{j})_{j=1,\ldots,m}$ of $X$ by the following 
assumption.
\begin{itemize}
 \item[\textbf{(A)}] Let $(A_{j})_{j=1,\ldots,m}$ be a partition of 
            $X\subset B_{\ell_2^d}$ such that $\mathring{A}_j\neq\emptyset$ for every 
            $j\in\{1,\ldots,m\}$ and such that there exist mutually distinct 
            $z_1,\ldots,z_m\in X$ with $A_j \subset B_r(z_j) =: B_j$, where 
            $(B_j)_{j=1,\ldots,m}$ is a cover of $X$ satisfying \eqref{ex. Ueberdeckung}.
\end{itemize}
In the remaining sections we will frequently refer to Assumption (A). 
However, the results hold as well if we merely assume $z_1,\ldots,z_m\in B_{\ell_2^d}$ 
instead of $z_1,\ldots,z_m\in X\subset B_{\ell_2^d}$ in (A).
The following example illustrates that (A) is indeed a natural assumption.

\begin{example}\label{ex::voronoi_partition}
For some $r>0$, let us consider an $r$-net $z_1,\ldots,z_m$ of $X$, where 
$z_1,\ldots,z_m$ are mutually distinct. 
Based on these $z_1,\ldots,z_m$, a Voronoi partition $(A_{j})_{j=1,\ldots,m}$ of $X$ is 
defined by
\begin{align}\label{Def. VP A_{j}}
 A_{j}:=\left\{x\in X : j=\min\{\argmin{k\in\{1,\ldots,m\}}\|x-z_k\|_2\} \right\}\,,
\end{align}
cf.~Figure \ref{Bild_Partition}.
That is, $A_{j}$ contains all $x\in X$ such that the center $z_j$ is the nearest 
center to $x$, and if there exist $j_1$ and $j_2$ with $j_1<j_2$ and
\begin{align*}
 \|x-z_{j_1}\|_2=\|x-z_{j_2}\|_2<\|x-z_k\|_2 
\end{align*}
for all $k\in\{1,\ldots,m\}\backslash\{j_1,j_2\}$, then $x\in A_{j_1}$ since 
$j_1<j_2$. 
In other words, they are resolved in favor of the smallest index of the involved 
centers.
Moreover, it is obvious that $\mathring{A}_j\neq\emptyset$, 
$A_{j}\subset B_r(z_j)$ for all $j\in\{1,\ldots,m\}$, $A_{j_1}\cap A_{j_2}=\emptyset$ for 
all $j_1,j_2\in\{1,\ldots,m\}$ with $j_1\neq j_2$, and $X = \bigcup_{j=1}^{m} A_{j}$.
In other words, a Voronoi partition based on an $r$-net $z_1,\ldots,z_m$ of $X$ satisfies 
condition (A), if $r$ and $m$ fullfil \eqref{ex. Ueberdeckung}.
\end{example}

Following Example \ref{ex::voronoi_partition}, we call the learning method producing 
$f_{\D,\bs\lb}$ given by \eqref{VP-SVM} a \textit{Voronoi partition support vector 
machine}, in short VP-SVM. Nevertheless, we just take a partition 
$(A_{j})_{j=1,\ldots,m}$ satisfying (A) as basis here instead of requesting 
$(A_{j})_{j=1,\ldots,m}$ to be a Voronoi partition. 

Recall that our goal is to derive not only global but also local learning rates for 
this VP-SVM approach. To this end, we additionally consider an arbitrary measurable  
set $T \subset X$ such that $\P_X(T)>0$. Then we examine the 
learning rate of the VP-SVM on this subset $T$ of $X$. To formalize this, it is 
necessary to introduce some basic notations related to $T$. Let us define the 
index set $J_{T}$ by
\begin{align}\label{Def. J_T}
 J_{T}:=\{j\in\{1,\ldots,m\}: A_{j} \cap  T\neq\emptyset\}
\end{align}
specifying every set $A_{j}$ that has at least one common point with $T$. 
Note that, for every non-empty set $T\subset X$, the index set $J_T$ is non-empty, too, 
i.e.~$|J_{T}|\geq1$.
Besides, deriving local rates on $T$ requires us to investigate the excess risk of the 
VP-SVM with respect to the distribution $\P$ and the loss 
$L_{T} : X\times Y\times\R\to[0,\infty)$ defined by
\begin{align}\label{Def. L_T allg}
 L_{T}(x,y,t):=\ind_{T}(x)L(x,y,t)\,.
\end{align}
\begin{wrapfigure}{r}{0.45\textwidth}\vspace{-0.5cm}
\begin{scriptsize}
\begin{center}
	\begin{overpic}[scale=0.48]{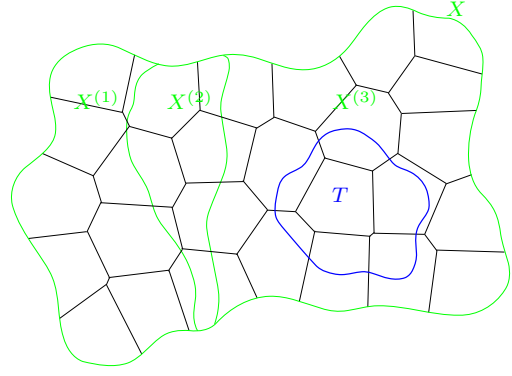}
          \put(64,35){{\color{blau}$T$}}
	  \put(86,71){{\color{gruen}$X$}}
	  \put(14,53){{\color{gruen}$X^{(1)}$}}
	  \put(32,53){{\color{gruen}$X^{(2)}$}}
	  \put(64,53){{\color{gruen}$X^{(3)}$}}
	\end{overpic}
\end{center}
\vspace{-0.5cm}
\caption{The input space $X$ with the corresponding partition 
$(A_{j})_{j=1,\ldots,m}$ and the subset $T$, where the local learning rate should 
be examined.}
\label{Bild_Partition_T}
\end{scriptsize}\vspace{0.5cm}
\end{wrapfigure}
However, to manage the analysis we additionally need the loss
$L_{J_T} : X\times Y\times\R\to[0,\infty)$ given by
\begin{align}\label{Def. L_{J_T} allg}
 L_{J_T}(x,y,t):=\ind_{\bigcup_{j\in J_T}A_{j}}(x)L(x,y,t)
\end{align}
which may only be nonzero, if $x$ is contained in some set $A_j$ 
with $j\in J_T$. 
Note that the risiks $\RR_{L_T,\P}(f)$ and
$\RR_{L_{J_T},\P}(f)$ quantify the quality of some 
function $f$ just on $T$ and 
\begin{align*}
 A_T:=\bigcup_{j\in J_T}A_j\supset T\,,
\end{align*}
respectively. Hence, examining the excess risks
\begin{align*}
 \RR_{L_T,\P}(\fcl_{\D,\bs\lb})-\RR_{L_T,\P}^* \leq 
 \RR_{L_{J_T},\P}(\fcl_{\D,\bs\lb})-\RR_{L_{J_T},\P}^*
\end{align*} 
leads to learning rates on $A_T$ and implicitly on $T$. 
Recapitulatory, let us declare a second set of assumptions.
\begin{itemize}
 \item[\textbf{(T)}] For $T\subset X$, we define an index set $J_T$ by \eqref{Def. J_T}, 
                     loss functions $L_T,L_{J_T}:X\times Y\times \R \to[0,\infty)$ by 
                     \eqref{Def. L_T allg} and \eqref{Def. L_{J_T} allg}, and the set $A_T:=\bigcup_{j\in J_T}A_j$.
\end{itemize}

\section{An Oracle Inequality for VP-SVMs}\label{general results} 

In this section, we first focus on RKHSs and direct sums of RKHSs. 
Then we present a lemma that relates the 
risk of a function w.r.t.~the general loss $L$ to the risks w.r.t.~the losses $L_j$. 
Finally, we establish a first oracle inequality for VP-SVMs. 

Let us begin with some basic notations. 
For $q\in[1,\infty]$ and a measure $\nu$, we denote by $L_q(\nu)$ the 
Lebesgue spaces of order $q$ w.r.t.~$\nu$ and for the Lebesgue measure $\mu$ on 
$X\subset\Rd$ we write $L_q(X):=L_q(\mu)$. In addition, for a measurable space $X$, the 
set of all real-valued measurable functions on $X$ is given by 
$\L_0(X) := \{ f:X\to\R \, | \, f \text{ measurable} \}$.
Moreover, for a measure $\nu$ on $X$ and measurable $\widetilde{X}\subset X$, we define 
the trace measure $\nu_{|\widetilde{X}}$ of $\nu$ in $\widetilde{X}$ by 
$\nu_{|\widetilde{X}}(A)=\nu(A\cap\widetilde{X})$ for every $A\subset X$. 

Our first goal is to show that $f_{\D,\bs\lb}$ in \eqref{VP-SVM} is actually an ordinary 
SVM solution. To this end, we
consider an RKHS on some $A\subsetneq X$ and extend it to an RKHS on $X$ by the following 
lemma, where we omit the obvious proof.

\begin{lemma}\label{Lemma extended RKHS}
Let $A\subset X$ and $H_A$ be an RKHS on $A$ with corresponding kernel $k_A$.
Denote by $\hat{f}$ the extension of $f\in H_A$ to $X$ defined by
\begin{align*}
 \hat{f}(x):=\begin{cases}
              f(x) \,, & \text{for } x\in A \,, \\
	      0 \,,    & \text{for } x\in X\backslash A \,.
             \end{cases}
\end{align*}
Then the space $\hat{H}_A := \{\hat{f} : f \in H_A\}$
equipped with the norm
\begin{align*}
 \|\hat{f}\|_{\hat{H}_A}:=\|f\|_{H_A} 
\end{align*}
is an RKHS on $X$ and its reproducing kernel is given by
\begin{align}\label{extended kernel}
 \hat{k}_A(x,x'):=\begin{cases}
                   k_A(x,x')\,, & \text{if }x,x'\in A\,, \\
		   0\,,		& \text{else.}
                  \end{cases}
\end{align}
\end{lemma}

Based on this lemma, we are now able to construct an 
RKHS by a direct sum of 
RKHSs $\hat{H}_A$ and $\hat{H}_B$ with $A,B\subset X$ and $A\cap B=\emptyset$. Here, we 
skip the proof once more, since the assertion follows immediately using, 
for example, orthonormal bases of $\hat{H}_A$ and $\hat{H}_B$.

\begin{lemma}\label{Lemma RKHS as direct sum}
For $A,B\subset X$ such that $A\cap B=\emptyset$ and $A\cup B\subset X$, let $H_A$ 
and $H_B$ be RKHSs of $k_A$ and $k_B$ over $A$ and $B$, respectively.
Furthermore, let $\hat{H}_A$ and $\hat{H}_B$ be the RKHSs of all functions of $H_A$ and 
$H_B$ extended to $X$ in the sense of Lemma \ref{Lemma extended RKHS} and let $\hat{k}_A$ 
and $\hat{k}_B$ given by \eqref{extended kernel} be the associated reproducing kernels.
Then $\hat{H}_A\cap\hat{H}_B=\{0\}$ and hence the direct sum
\begin{align}\label{RKHS as direct sum}
 H := \hat{H}_A \oplus \hat{H}_B
\end{align}
exists.
For $\lb_A, \lb_B>0$ and $f\in H$, let $\hat{f}_A\in \hat{H}_A$ and 
$\hat{f}_B\in \hat{H}_B$ be the unique functions such that $f = \hat{f}_A + \hat{f}_B$. 
Then we define the norm $\|\cdot\|_H$ by
\begin{align}\label{norm RKHS as direct sum}
 \|f\|_H^2 := \lb_A\|\hat{f}_A\|_{\hat{H}_A}^2+\lb_B\|\hat{f}_B\|_{\hat{H}_B}^2
\end{align}
and $H$ equipped with the norm $\|\cdot\|_H$ is again an RKHS for which 
\begin{align*}
k(x,x') := \lb_A^{-1} \hat{k}_A(x,x') + \lb_B^{-1} \hat{k}_B(x,x')\,, \qquad x,x'\in X\,, 
\end{align*}
is the reproducing kernel.
\end{lemma}

To relate Lemmas \ref{Lemma extended RKHS} and \ref{Lemma RKHS as direct sum} with 
\eqref{VP-SVM}, we have to introduce some more notations. 
For pairwise disjoint sets $A_1,\ldots,A_m\subset X$, let $H_j$ be an RKHS on $A_j$ 
for every $j\in\{1,\ldots,m\}$. 
Then, based on RKHSs $\hat{H}_1,\ldots,\hat{H}_m$ on $X$ defined by Lemma 
\ref{Lemma extended RKHS}, the joined RKHSs can be designed analogously to Lemma 
\ref{Lemma RKHS as direct sum}. That is, for an arbitrary index set 
$J\subset\{1,\ldots,m\}$ and a vector $\bs\lb=(\lb_j)_{j\in J}\in(0,\infty)^{|J|}$, 
the direct sum 
\begin{align*}
 H_J := \bigoplus_{j\in J} \hat{H}_j 
      = \left\{ f=\sum_{j\in J} f_j : f_j\in\hat{H}_j \text{ for all } j\in J \right\} 
\end{align*}
is again an RKHS equipped with the norm
\begin{align}\label{norm H_J}
 \|f\|_{H_J}^2 = \sum_{j\in J} \lb_j\|f_j\|_{\hat{H}_j}^2\,.
\end{align}
If $J=\{1,\ldots,m\}$ we simply write 
\begin{align}\label{Def. H}
  H &:= H_J 
\end{align}
Note that $H$ contains 
inter alia $f_{\D,\bs\lb}$ given by \eqref{VP-SVM}. 
Summarizing, we can define another assumption set.
\begin{itemize}
 \item[\textbf{(H)}] For every $j\in\{1,\ldots,m\}$, let $H_j$ be a separable RKHS of 
                     measurable kernels $k_j$ over $A_j$, where 
                     $A_1,\ldots,A_m\subset X$ are pairwise disjoint and
                     $$\|k_j\|_{L_2(\TrP{A_j})}^2 := \int_X k_j(x,x) d\TrP{A_j}(x) < \infty\,.$$
                     Then we define 
                     RKHSs $\hat{H}_1,\ldots,\hat{H}_m$ by Lemma 
                     \ref{Lemma extended RKHS} and the joined RKHS $H$ by 
                     \eqref{Def. H} equipped with the norm \eqref{norm H_J} 
                     for fixed $\lb_1,\ldots,\lb_m>0$.
\end{itemize}

Having designed a joined RKHS as above, a crucial property of its function's risks is 
expressed by the following lemma.

\begin{lemma}\label{Lem RLP splitting}
Let $\P$ be a distribution on $X\times Y$ and $L:X\times Y\times\R\to[0,\infty)$ be a 
loss function. 
For $A,B\subset X$ such that $A\cup B =X$ and $A\cap B=\emptyset$, define loss functions 
$L_A,L_B:X\times Y\times\R\to[0,\infty)$ by 
$L_A(x,y,t)=\ind_A(x)L(x,y,t)$ and $L_B(x,y,t)=\ind_B(x)L(x,y,t)$, respectively.
Furthermore, let $f_A:X\to\R$ as well as $f_B:X\to\R$ be measurable functions and 
$f:X\to\R$ be defined by $f(x)=\ind_A(x)f_A(x)+\ind_B(x)f_B(x)$ for all $x\in X$.
Then we have
\begin{align*}
 \RLP(f) = \RR_{L_A,\P}(f_A) + \RR_{L_B,\P}(f_B)\,.
\end{align*}
as well as
\begin{align*}
 \RLP(f) - \RB  
 = \left(\RR_{L_A,\P}(f_A)-\RR_{L_A,\P}^*\right) 
 + \left( \RR_{L_B,\P}(f_B)-\RR_{L_B,\P}^*\right)\,.
\end{align*}
\end{lemma}
Note that Lemma \ref{Lem RLP splitting} can be transferred to finite, pairwise 
disjoint unions. To be more precise, let us consider an arbitrary index set 
$J\subset\{1,\ldots,m\}$ and define the corresponding loss function 
$L_J : X\times Y\times\R\to[0,\infty)$ by
\begin{align*}
 L_J(x,y,t):=\ind_{\bigcup_{j\in J}A_{j}}(x)L(x,y,t) \,. 
\end{align*}
Now, it is straightforward to show 
\begin{align*}
\RLPJ(f) = \sum_{j\in J} \RLPj(f) 
\end{align*}
for every function $f:X\to\R$.
Based on this generalization and the whole index set $J=\{1,\ldots,m\}$, let 
us briefly consider Lemma \ref{Lem RLP splitting} for the empirical measure $\D$ 
and for $f_{\D,\bs\lb}=\sum_{j=1}^n \ind_{A_{j}}f_{\D_{j},\lb_j}$, where 
$f_{\D_{j},\lb_j}$, $j=1,\ldots,m$, are defined by \eqref{SVM_j}. Then, for an arbitrary 
$f \in H$, it immediately follows
\begin{align}
 \left\|f_{\D,\bs\lb}\right\|^2_H + \RLD(\fcl_{\D,\bs\lb}) 
 & = \sum_{j=1}^{m}\left( \lb_j\left\|f_{\D_j,\lb_j}\right\|_{\hat{H}_j}^2 
     + \RLDj(\fcl_{\D,\bs\lb}) \right) \nonumber\\
 & \leq \sum_{j=1}^{m} \left( \lb_j \left\|\ind_{A_j}f\right\|_{\hat{H}_j}^2 
        + \RLDj(f) \right) \nonumber\\
 & = \left\|f\right\|^2_H + \RLD(f)\,. \label{equivalence VP-SVM - SVM using H}
\end{align}
That is, $f_{\D,\bs\lb}$ is the decision function of an SVM using $H$ and $L$  as 
well as the regularization parameter $\tilde{\lb}=1$. In other words, the latter SVM 
equals the VP-SVM given by \eqref{VP-SVM}. This will be a key insight used in our 
analysis.

To derive an oracle inequality, i.e.~an appropriate upper bound for the excess risk 
$\RLPJ(\fcl_{\D,\bs\lb})-\RBJ$ for some index set $J\subset\{1,\ldots,m\}$, we 
have to introduce a few more notations. 
Let $\P$ be a distribution on $X\times Y$ such that a Bayes decision function 
$\fB : X\to[-M,M]$ exists, for some constant $M>0$ at which $L$ can be clipped. 
Moreover, we denote by $L\circ f$ the function $(x,y)\mapsto L(x,y,f(x))$. 
If there exist constants $B>0$, $\vt\in[0,1]$, and $V\geq B^{2-\vt}$ 
such that we have
\begin{align}
  L(x,y,t)&\leq B \,, \label{supbound} \\
 \E_{\P}\left(L\circ f-L\circ f^*_{L,\P}\right)^2 
 &\leq V \cdot \left(\E_{\P}\left(L\circ f-L\circ f^*_{L,\P}\right)\right)^\vt \,,
 \label{varbound}
\end{align}
for all $(x,y)\in X\times Y$, $t\in[-M,M]$, and $f:X\to[-M,M]$, we say that the supremum 
bound \eqref{supbound} and the variance bound \eqref{varbound}, respectively, is 
fulfilled. Actually, \eqref{supbound} immediately yields
\begin{align*}
 L_J(x,y,t) = \ind_{\bigcup_{j\in J}A_{j}}(x)L(x,y,t) \leq L(x,y,t) \leq B 
\end{align*}
for all $(x,y)\in X\times Y$ and $t\in[-M,M]$, i.e.~the supremum bound is also satisfied 
for $L_J$. 
Moreover, if \eqref{varbound} holds for all $f:X\to[-M,M]$, the variance bound using the
loss $L_J$ is satisfied, too. Indeed, by the use of 
$\widetilde{f}(x) := \ind_{\bigcup_{j\in J} A_j}(x)f(x) 
+ \ind_{X\backslash\left(\bigcup_{j\in J} A_j\right)}(x)\fB(x)$ for all $x\in X$, we have
\begin{align*}
 \E_{\P}\left(L_J\circ f-L_J\circ f^*_{L,\P}\right)^2 
 & = \E_{\P}\left(L_J\circ \widetilde{f}-L_J\circ f^*_{L,\P}\right)^2 \\
 & = \E_{\P}\left(L\circ \widetilde{f}-L\circ f^*_{L,\P}\right)^2 \\
 & \leq V \cdot \left(\E_{\P}\left(L\circ \widetilde{f}-L\circ f^*_{L,\P}\right)\right)^\vt \\
 & \leq V \cdot \left(\E_{\P}\left(L_J\circ f-L_J\circ f^*_{L,\P}\right)\right)^\vt 
\end{align*}
for all $f:X\to[-M,M]$.
Let us quickly define a third assumption set.
\begin{itemize}
 \item[\textbf{(P)}] 
                     Let $\P$ be a distribution on $X\times Y$ such that the variance 
                     bound \eqref{varbound} is satisfied for 
                     constants $\vt\in[0,1]$, $V\geq B^{2-\vt}$, and all 
                     functions $f:X\to[-M,M]$.
\end{itemize}

Up to now, there is still missing a classical tool that is used to derive learning 
rates, namely entropy numbers, see \citep{ CaSt90} or 
\citep[Definition A.5.26]{ StCh08}. 
Recall that, for normed spaces $(E,\|\,\cdot\,\|_E)$ and $(F,\|\,\cdot\,\|_F)$ as well as 
an integer $i\geq 1$, the $i$-th (dyadic) entropy number of a bounded, linear operator 
$S : E\to F$ is defined by
\begin{align*}
 e_i(S : E\to F) & := e_i(SB_E,\|\,\cdot\,\|_F) \\
 & := \inf\Biggl\{\e>0:\exists s_1,\ldots,s_{2^{i-1}}\in SB_E 
\text{ such that } SB_E \subset \bigcup_{j=1}^{2^{i-1}}(s_j + \e B_F)\Biggr\}\,,
\end{align*}
where we use the convention $\inf\emptyset:=\infty$, and $B_E$ as well as $B_F$ denote 
the closed unit balls in $E$ and $F$, respectively.
Finally, we present a first oracle inequality involving an upper bound 
for the excess risk $\RLPJ(\fcl_{\D,\bs\lb})-\RBJ$, where $J\subset\{1,\ldots,m\}$ is 
an arbitrary index set.

\begin{theorem}\label{main thm.}
Let $L: X\times Y\times\R\to [0,\infty)$ be a locally Lipschitz continuous loss that 
can be clipped at $M>0$ and that satisfies the supremum bound \eqref{supbound} for some 
$B>0$. 
Based on a partition $(A_{j})_{j=1,\ldots,m}$ of $X$, where $\mathring{A}_j\neq\emptyset$ 
for every $j\in\{1,\ldots,m\}$, we assume (H). Furthermore, for an arbitrary index set 
$J\subset\{1,\ldots,m\}$, we suppose (P). 
Assume that, for fixed $n\geq 1$, there exist constants $p\in(0,1)$ and $a_1,\ldots,a_m > 0$ 
such that for all $j\in \{1,\ldots,m\}$
\begin{align} \label{entropy assumption}
 e_i(\mathrm{id} : H_j \to L_2(\TrP{A_j})) \leq a_j\, i^{-\frac{1}{2p}} 
 \,,\qquad\qquad i\geq1\,. 
\end{align}
Finally, fix an $f_0\in H$ and a constant $B_0\geq B$ such that 
$\|L_J\circ f_0\|_\infty\leq B_0$. Then, for all fixed $\t>0$, 
$\bs\lb=(\lb_1,\ldots,\lb_m)>0$, and 
\begin{align*}
 a:= \max\left\{c_p \sqrt{m} \left(\sum_{j=1}^m\lb_{j}^{-p} a_j^{2p}\right)^{\frac{1}{2p}}, B\right\}\,,
\end{align*}
the VP-SVM given by \eqref{VP-SVM} using $\hat{H}_1,\ldots,\hat{H}_m$ and $L_J$ satisfies
\begin{align*}
  &\sum_{j=1}^m\lb_j\|f_{\D_{j},\lb_j}\|^2_{\hat{H}_j}+\RLPJ(\fcl_{\D,\bs\lb})-\RBJ \nonumber\\ 
  &\leq  9 \!\left(\sum_{j=1}^m\!\lb_j\|\ind_{A_j}f_0\|^2_{\hat{H}_j}\!
  +\!\RLPJ(f_0) -\!\RBJ\!\right) \!
  +\! C\! \left(a^{2p}n^{-1}\right)^{\frac{1}{2-p-\vt+\vt p}} \!
  +\! 3 \left(\frac{72 V\t}{n}\right)^{\frac{1}{2-\vt}} \! + \!\frac{15B_0\t}{n} 
\end{align*}
with probability $\P^n$ not less than $1-3e^{-\t}$, where 
$C>0$ is a constant only depending on $p$, $M$, $V$, $\vt$, and $B$.
\end{theorem}

The above theorem deals with the case of a partition with quite a few sets 
$A_j$, $j\in\{1,\ldots,m\}$. However, if we consider a partition consisting of just one 
set $A_1$, i.e.~$A_1=X$, Theorem \ref{main thm.} is supposed to provide an oracle 
inequality that is comparable to the already known ones. To make that sure, let us 
briefly consider the case $m=1$ and hence $A_1:=X$, $\lb_1:=\bs\lb$ as well as RKHSs 
$H_1=\hat{H}_1=H$ over X with $\|\,\cdot\,\|_H^2=\bs\lb \|\,\cdot\,\|_{H_1}^2$. Note that 
in this case we have $f_{\D,\bs\lb}=f_{\D_{1},\lb_1}$. If \eqref{entropy assumption} holds 
for $H_1$, Theorem \ref{main thm.} yields that an SVM using $H$ and $L_J=L$ satisfies 
\begin{align*}
  &\bs\lb\|f_{\D,\bs\lb}\|_{H_1}^2+\RLP(\fcl_{\D,\bs\lb})-\RB \\ 
  &\leq 9\left(\bs\lb\|f_0\|^2_{H_1}+\RLP(f_0)-\RB\right) 
  + C \left(\tilde{c}_p\frac{a_1^{2p}}{\bs\lb^{p}n}\right)^{\frac{1}{2-p-\vt+\vt p}} 
  + 3\left(\frac{72 V\t}{n}\right)^{\frac{1}{2-\vt}} + \frac{15B_0\t}{n} 
\end{align*}
with probability $\P^n$ not less than $1-3e^{-\t}$ for fixed $\t>0$. Note that this 
oracle inequality indeed matches with the one stated in \citep[Theorem 7.23]{ StCh08} 
apart from the constant $\tilde{c}_p$, which is, however, only depending on $p$, $\vt$, 
and $B$.

In the folllowing section, we focus on RKHSs using Gaussian RBF kernels and examine
the associated entropy numbers to specify \eqref{entropy assumption}. 
Subsequently in Section \ref{Sec LS loss}, we additionally consider 
the least squares loss and adapt the oracle inequality of Theorem \ref{main thm.}.

\section{Entropy Estimates for Local Gaussian RKHSs}\label{Sec entropy estim}

In this section, we refine assumption \eqref{entropy assumption}. More precisely, 
in the subsequent theorem we determine an upper bound for the entropy numbers of the 
operator $\mathrm{id} : H_\g(A) \to L_2(\TrP{A})$, where $H_\g(A)$ is the RKHS over 
$A$ of the Gaussian RBF kernel $k_\g$ on $A\subset\Rd$ defined by
\begin{align*}
 k_\g(x,x'):=\exp\left(-\g^{-2}\|x-x'\|_2^2\right)\,, \qquad\qquad x,x'\in A\,,
\end{align*}
for some width $\g>0$.

\begin{theorem}\label{entropy number H_g}
Let $X\subset\Rd$, $\P_X$ be a distribution on $X$ and $A\subset X$ be such that
$\mathring{A}\neq\emptyset$ and such that there exists an Euclidean ball $B\subset\Rd$ 
with radius $r>0$ containing $A$, i.e.~$A\subset B$.
Moreover, for $0<\g\leq r$, let $H_\g(A)$ be the RKHS of the Gaussian RBF kernel $k_\g$ 
over $A$. 
Then, for all $p\in(0,1)$, there exists a constant $c_p>0$ such that
 \begin{align*}
   e_i(\mathrm{id} : H_\g(A) \to L_2(\TrP{A})) 
   \leq  c_p \sqrt{\P_X(A)}\, r^\frac{d+2p}{2p}\g^{-\frac{d+2p}{2p}} i^{-\frac{1}{2p}}
   \,, \qquad  \qquad i\geq 1\,.
 \end{align*}
\end{theorem}

Obviously, this theorem specifies assumption \eqref{entropy assumption}. 
Now, for the Gaussian case we elaborate assumption (H) and introduce the following 
additional set of assumptions.
\begin{itemize}
 \item[\textbf{(G)}]
Let $A_1,\ldots,A_m$ be pairwise disjoint subsets of $X$ with non-empty interior 
such that, for some fixed $r>0$ and every $j\in\{1,\ldots,m\}$, 
$\sup_{x,x'\in A_j} \|x-x'\|_2\leq 2r$ is satisfied. Furthermore, for every 
$j\in\{1,\ldots,m\}$, let $H_j:=H_{\g_j}(A_j)$ be the RKHS of the Gaussian kernel 
$k_{\g_j}$ with width $\g_j\in(0,r]$ over $A_j$. Consequently, for 
$\bs\lb:=(\lb_1,\ldots,\lb_m)\in(0,\infty)^m$, we define the 
joined RKHS $H:=\bigoplus_{j=1}^m \hat{H}_{\g_j}(A_j)$ by \eqref{Def. H} equipped with  
the norm \eqref{norm H_J}.
\end{itemize}

Since we do not consider SVMs with a fixed kernel, 
we use a more detailed notation than \eqref{SVM_j} and \eqref{VP-SVM} in the 
following specifying the kernel width $\g_j$ of the RKHS $H_{\g_j}(A_j)$ at hand. 
For all $j\in\{1,\ldots,m\}$ and $\bs\g:=(\g_1,\ldots,\g_m)$, we thus write
\begin{align*}
 f_{\D_{j},\lb_j,\g_j} 
 = \arg\min_{f\in \hat{H}_{\g_j}(A_j)}\lb_j\|f\|_{\hat{H}_{\g_j}(A_j)}^2 
 + \frac{1}{n}\sum_{i=1}^nL_j(x_i,y_i,f(x_i)) \,,
\end{align*}
and
\begin{align*}
 f_{\D,\bs\lb,\bs\g} := \sum_{j=1}^{m} f_{\D_{j},\lb_j,\g_j}
\end{align*}
instead of $f_{\D_{j},\lb_j}$ and $f_{\D,\bs\lb}$ in the remainder of this work.

In the subsequent section, we consider the least squares loss which, together with 
Assumption (G) and Theorem \ref{entropy number H_g}, allows us to elaborate the
oracle inequality stated in Theorem \ref{main thm.} so that we finally obtain 
learning rates.

\section{Learning Rates for Least Squares VP-SVMs}\label{Sec LS loss}

In this section, the non-parametric least squares regression problem is considered 
using the least squares loss $L:Y\times\R\to[0,\infty)$ defined by 
$L(y,t):=(y-t)^2$. It is well known that, in this case, the Bayes decision function 
$\fB:\Rd\to\R$ is given by $\fB(x)=\E_\P(Y|x)$ for $\P_X$-almost all $x\in\Rd$. Moreover, 
this function is unique up to zero-sets. Besides, for the least squares loss the equality 
\begin{align*}
 \RLP(f)-\RB = \left\| f - \fB \right\|_{L_2(\P_X)}^2
\end{align*}
can be shown by some simple, well-known transformations. 
Recall that $T$ is a non-empty subset of $X$, where the index set $J_T$ defined by 
\eqref{Def. J_T} indicates every set $A_{j}$ of the partition $(A_{j})_{j=1,\ldots,m}$ of 
$X$ that shares at least one point with $T$. The associated loss function 
$L_{J_T} : X\times Y\times\R\to[0,\infty)$ is defined by \eqref{Def. L_{J_T} allg}.

\subsection{Basic Oracle Inequalities for LS-VP-SVMs}

To formulate oracle inequalities and derive rates for VP-SVMs using the least 
squares loss,
the target function $\fB$ is assumed to satisfy certain smoothness 
conditions. To this end, we initially recall the modulus of smoothness, a device to 
measure the smoothness of functions (see e.g.~\citep[p.~44]{DeVore}, 
\citep[p.~398]{DeVorePopov}, and \citep[p.~360]{DeVoreArticle}). Denote by 
$\left\|\, \cdot \,\right\|_2$ the Euclidean norm and let $X\subset\mathbb{R}^d$ be a 
subset with non-empty interior, $\nu$ be an arbitrary measure on $X$, $p\in(0,\infty]$, 
and $f:X\to\R$ be contained in $L_p\left(\nu\right)$. Then, for $s\in\N$, 
the $s$-th modulus of smoothness of $f$ is defined by
\begin{align*}
 \omega_{s,{L_p\left(\nu\right)}}\left(f,t\right) = 
 \sup_{\left\|h\right\|_2\leq t}\left\|\triangle_h^s\left(f,\,\cdot\,\right)
 \right\|_{L_p\left(\nu\right)} \ , \qquad\qquad t\geq 0 \ ,
\end{align*}
where $\triangle_h^s\left(f,\, \cdot\,\right)$ denotes the $s$-th difference of $f$ 
given by
\begin{align*}
 \triangle_h^s\left(f,x\right)
 = \begin{cases}
    \sum_{j=0}^{s}\binom{s}{j}\left(-1\right)^{s-j} f\left(x+jh\right) 
    & \text{ if } x \in X_{s,h} \\
    0 & \text{ if } x \notin X_{s,h}
   \end{cases}
\end{align*}
for $h=(h_1,\ldots,h_d)\in[0,\infty)^d$ and 
$X_{s,h}:=\left\{x\in X : x+th\in X \mbox{ f.a. }t\in[0,s]\right\}$.
Based on the modulus of smoothness, we introduce Besov spaces, i.e.~function spaces 
that provide a finer scale of smoothness than the commonly used Sobolev spaces and that 
will thus be assumed to contain the target function later on. To this end, let 
$1\leq p,q\leq\infty$, $\a>0$, $s:=\left\lfloor \a \right\rfloor +1$, and $\nu$ be an 
arbitrary measure. Then the Besov space $B_{p,q}^\alpha\left(\nu\right)$ is defined by 
\begin{align*}
 B_{p,q}^\alpha\left(\nu\right):=\left\{f\in L_p\left(\nu\right) : 
 \left|f\right|_{B_{p,q}^\alpha\left(\nu\right)} <\infty\right\} \,,
\end{align*}
where the seminorm $\left|\,\cdot\,\right|_{B_{p,q}^\alpha\left(\nu\right)}$ is given by
\begin{align*}
 \left|f\right|_{B_{p,q}^\alpha\left(\nu\right)}:=
 \left(\int_0^\infty\left(t^{-\alpha}\omega_{s,L_p(\nu)}\left(f,t\right)\right)^q 
 \frac{dt}{t}\right)^\frac{1}{q} \,,\qquad\qquad 1\leq q<\infty\,,
\end{align*}
or
\begin{align*}
 \left|f\right|_{B_{p,\infty}^\alpha\left(\nu\right)}:=
 \sup_{t>0}\left(t^{-\alpha}\omega_{s,L_p(\nu)}\left(f,t\right)\right) \,,
\end{align*}
see e.g.~\citep[Section 7]{AdFo03} and \citep[Sections 2 and 3]{Triebel83}.
Note that $\left\|f\right\|_{B_{p,q}^\alpha\left(\nu\right)}:=
\left\|f\right\|_{L_p\left(\nu\right)}+\left|f\right|_{B_{p,q}^\alpha\left(\nu\right)}$
actually describes a norm of $B_{p,q}^\alpha\left(\nu\right)$ for all $q\in[1,\infty]$, 
see e.g.~\citep[pp.~54/55]{DeVore} and \citep[p.~398]{DeVorePopov}. 
Again, if $\nu$ is the Lebesgue measure on $X$, we write 
$B_{p,q}^\alpha\left(X\right):=B_{p,q}^\alpha\left(\nu\right)$. 
For the sake of completeness, recall from e.g.~\citep[Section 3]{AdFo03} and 
\citep[Sections 2 and 3]{Triebel83} the scale of Sobolev spaces $W_p^\a(\nu)$
defined by
\begin{align*}
 W_p^\a\left(\nu\right):=\left\{ f\in L_p\left(\nu\right) : \partial^{\left(\b\right)}
f \in L_p\left(\nu\right) \text{ exists for all } \b\in \mathbb{N}_0^d \text{ with } 
\left|\b\right|\leq \a\right\} \,,
\end{align*}
where $\a\in\N_0$, $1\leq p\leq\infty$, $\nu$ is an arbitrary measure, and 
$\partial^{\left(\b\right)}$ is the $\b$-th weak derivative for a multi-index 
$\b=\left(\b_1,\ldots,\b_d\right)\in\mathbb{N}_0^d$ with 
$\left|\b\right|=\sum_{i=1}^d\b_i$. 
That is, $W_p^\a(\nu)$ is the space of all functions in $L_p(\nu)$, whose weak 
derivatives up to order $\a$ exist and are contained in $L_p(\nu)$. 
Moreover, the Sobolev space is equipped with the Sobolev norm
\begin{align*}
 \left\|f\right\|^p_{W_p^\a\left(\nu\right)}:=\sum_{\left|\b\right|\leq \a}
\left\|\partial^{\left(\b\right)}f\right\|_{L_p\left(\nu\right)}^p \, ,
\end{align*}
\citep[cf.][page 60]{AdFo03}. We write $W_p^0(\nu)=L_p(\nu)$ and, for the 
Lebesgue measure $\mu$ on $X\subset\mathbb{R}^d$, we define 
$W_p^\a(X):=W_p^\a\left(\mu\right)$.
It is well-known, see e.g.~\citep[p.~25 and p.~44]{EdTr96}, that 
the  Sobolev spaces $W_p^\alpha(\Rd)$ fall into the scale of Besov spaces, namely
\begin{align*}
 W_p^\alpha(\Rd) \subset B_{p,q}^\alpha(\Rd) 
\end{align*}
for $\alpha\in\N$, $p\in(1,\infty)$, and $\max\{p,2\}\leq q\leq \infty$. 
Moreover, for $p=q=2$ we actually have equality, that is 
$W_2^\alpha(\Rd) = B_{2,2}^\alpha(\Rd)$ with equivalent norms.

Based on the least squares loss and RKHSs using Gaussian kernels over 
the partition sets $A_{j}$, the subsequent theorem refines the oracle inequality 
stated in Theorem \ref{main thm.}.

\begin{theorem}\label{LS oracle ineq}
Let $Y:=\left[-M,M\right]$ for $M>0$, $L: Y\times\R\to [0,\infty)$ be the least squares 
loss  and $\P$ be a distribution on $\Rd \times Y$. We write $X:=\supp{\P_X}$.
Furthermore, let (A) and (G) be satisfied. In addition, for an arbitrary subset 
$T\subset X$, we assume (T). 
Moreover, let $f^*_{L,\P}:\Rd\to\R$ be a Bayes decision function such that 
$f^*_{L,\P}\in L_2(\Rd) \cap L_\infty(\Rd)$ as well as 
$f_{L,\P}^*\in B_{2,\infty}^{\a}(\TrP{A_T})$ for some $\a \geq 1$. 
Then, for all $p \in \left(0,1\right)$, 
$n\geq 1$, $\t\geq1$, $\bs\g=(\g_1,\ldots,\g_m)\in(0,r]^m$, and 
$\bs\lb=(\lb_1,\ldots,\lb_m)>0$, the VP-SVM given by 
\eqref{VP-SVM} using $\hat{H}_{\g_1}(A_{1}),\ldots,\hat{H}_{\g_m}(A_{m})$, and the 
loss $L_{J_T}$ satisfies 
\begin{align*}
 & \sum_{j=1}^m \lb_j\|f_{\D_{j},\lb_j,\g_j}\|^2_{\hat{H}_{\g_j}(A_j)}
 +\RLPJT(\fcl_{\D,\bs\lb,\bs\g}) - \RBJT \nonumber\\ 
 & \leq C_{M,\a,p}\!\left( \sum_{j \in J_T}\! \lb_j \g_j^{-d} \! +\! \!
        \left(\frac{\max_{j\in J_T}\g_{j}}{\min_{j\in J_T}\g_{j}}\right)^d \!
        \max_{j\in J_T}\g_{j}^{2\a}
        \! +\! r^{2p}\! \left( \sum_{j=1}^m \!
        \lb_{j}^{-1} \g_j^{-\frac{d+2p}{p}} \P_X(A_{j})\! \right)^p\! n^{-1} 
        \!\! +\! \t n^{-1}\! \right) 
\end{align*}
with probability $\mathrm{P}^n$ not less than $1-e^{-\t}$, where $C_{M,\a,p}>0$ is a 
constant only depending on $M$, $\a$, $p$, $d$, $\|\fB\|_{L_2(\R^d)}$, 
$\|\fB\|_{L_\infty(\R^d)}$, and $\|\fB\|_{B_{2,\infty}^{\a}(\TrP{A_T})}$.
\end{theorem}

Using this oracle inequality, we derive learning rates w.r.t.~the loss $L_{J_T}$ 
for the learning method described by \eqref{SVM_j} and \eqref{VP-SVM} in the 
following theorem.

\begin{theorem}\label{Thm LS rate}
Let $\t\geq 1$ be fixed and $\b\geq \frac{2\a}{d}+1$. 
Under the assumptions of Theorem \ref{LS oracle ineq} and with
\begin{align}
 r_n & = c_1 n^{-\frac{1}{\b d}} \,, \label{Def. r} \\
 \lb_{n,j} & = c_2 r^d n^{-1}\,, \label{Def. lb_n} \\
 \g_{n,j} & = c_3 n^{-\frac{1}{2\a+d}}\,, \label{Def. g_n}
\end{align}
for every $j\in \{1,\ldots,m_n\}$, we have, for all $n\geq 1$ and $\x>0$,
\begin{align*}
 \RLPJT(\fcl_{\D,\bs\lb_n,\bs\g_n})-\RBJT \leq C\t n^{-\frac{2\a}{2\a+d}+\x} 
\end{align*}
with probability $\P^n$ not less than $1-e^{-\t}$, where 
$\bs\lb_n:=(\lb_{n,1},\ldots,\lb_{n,m_n})$ as well as 
$\bs\g_n:=(\g_{n,1},\ldots,\g_{n,m_n})$ and $C,c_1,c_2,c_3$ are positive constants 
with $ c_3\leq c_1$.
\end{theorem}

In the latter theorem the condition  $\b\geq \frac{2\a}{d}+1$ is required to ensure 
$\g_{n,j}\leq r_n$, $j=1,\ldots,m_n$, which in turn is a prerequisite arising from 
Theorem \ref{entropy number H_g} and the used entropy estimate. 
Let us briefly examine the extreme case $\b = \frac{2\a}{d}+1$. Using 
$r_n \approx n^{-\frac{1}{\b d}}$ and \eqref{ex. Ueberdeckung} leads to covering 
numbers of the form $m_n \approx n^{\frac{d}{2\a+d}}$ and computational costs of 
$\mathcal{O}\bigl(m_n\bigl(\frac{n}{m_n}\bigr)^q\bigr)
= \mathcal{O}\bigl(n^{\frac{2\a q+d}{2\a+d}}\bigr)$ which is actually less than the 
computational cost of order $n^q$, $q\in[2,3]$, of an usual SVM. Note that for 
increasing $\b$ the computational cost of an VP-SVM is increasing as well. However, 
for $\b > \frac{2\a}{d}+1$, $r_n \approx n^{-\frac{1}{\b d}}$, and 
$m_n \approx n^{\frac{1}{\b}}$, a VP-SVM has costs of 
$\mathcal{O}\bigl(n^{\frac{1+(\b-1)q}{\b}}\bigr)$ which still is less that 
$\mathcal{O}\left(n^q\right)$.

Let us finally take a closer look at the VP-SVM given by \eqref{VP-SVM} and the 
considerations related to \eqref{equivalence VP-SVM - SVM using H}, where 
$f_{\D,\bs\lb}\in H = \bigoplus_{j=1}^m \hat{H}_j$ solves the minimization problem
\begin{align*}
  f_{\D,\bs\lb} 
 = \underset{f_1\in \hat{H}_1,\ldots,f_m\in \hat{H}_m}{\operatorname{arg\,min}} 
 \sum_{j=1}^m \lb_j\left\|f_j\right\|_{\hat{H}_j}^2+\RLD\bigl(\sum_{j=1}^m f_j\bigr) \,.
\end{align*}
Choosing $\lb_1=\ldots=\lb_m$, the VP-SVM problem can be understood as 
$\ell_2$-multiple kernel learning (MKL) problem using the RKHSs 
$\hat{H}_1,\ldots,\hat{H}_m$. Learning rates for MKL have been treated, for example, 
in \citep{Su11} and \citep{KlBl12}. 
Assuming $\fB\in H$, the learning rate achieved in \citep{Su11} is $mn^{-\frac{1}{1+s}}$ 
for dense settings, where $s$ is the so-called spectral decay coefficient. 
In addition, \citet{KlBl12} obtain 
 essentially the same rates under these assumptions.
Let us therefore briefly investigate the above 
rate of \citep{Su11}. For RKHSs that are continuously embedded in a Sobolev space 
$W_2^\a(X)$, we have $s=\frac{d}{2\a}$ such that the 
learning rate reduces to $mn^{-\frac{2\a}{2\a+d}}$. Note that this learning rate is $m$ 
times the optimal learning rate $n^{-\frac{2\a}{2\a+d}}$, where the number $m=m_n$ of kernels 
may increase with the sample size $n$. 
In particular, if $m_n\to \infty$ polynomially, then the rates obtained in \citep{Su11}
become substantially worse than the optimal rate.
%
%
%
In contrast, due to the special choice of the RKHSs, this is not the case for our VP-SVM problem, provided that $m_n$ does not grow faster than
$n^{1/\b}$.

\begin{wrapfigure}{R}{0.45\textwidth}\vspace{-0.25cm}
\begin{scriptsize}
	\begin{center}
	 \begin{overpic}[scale=0.4]{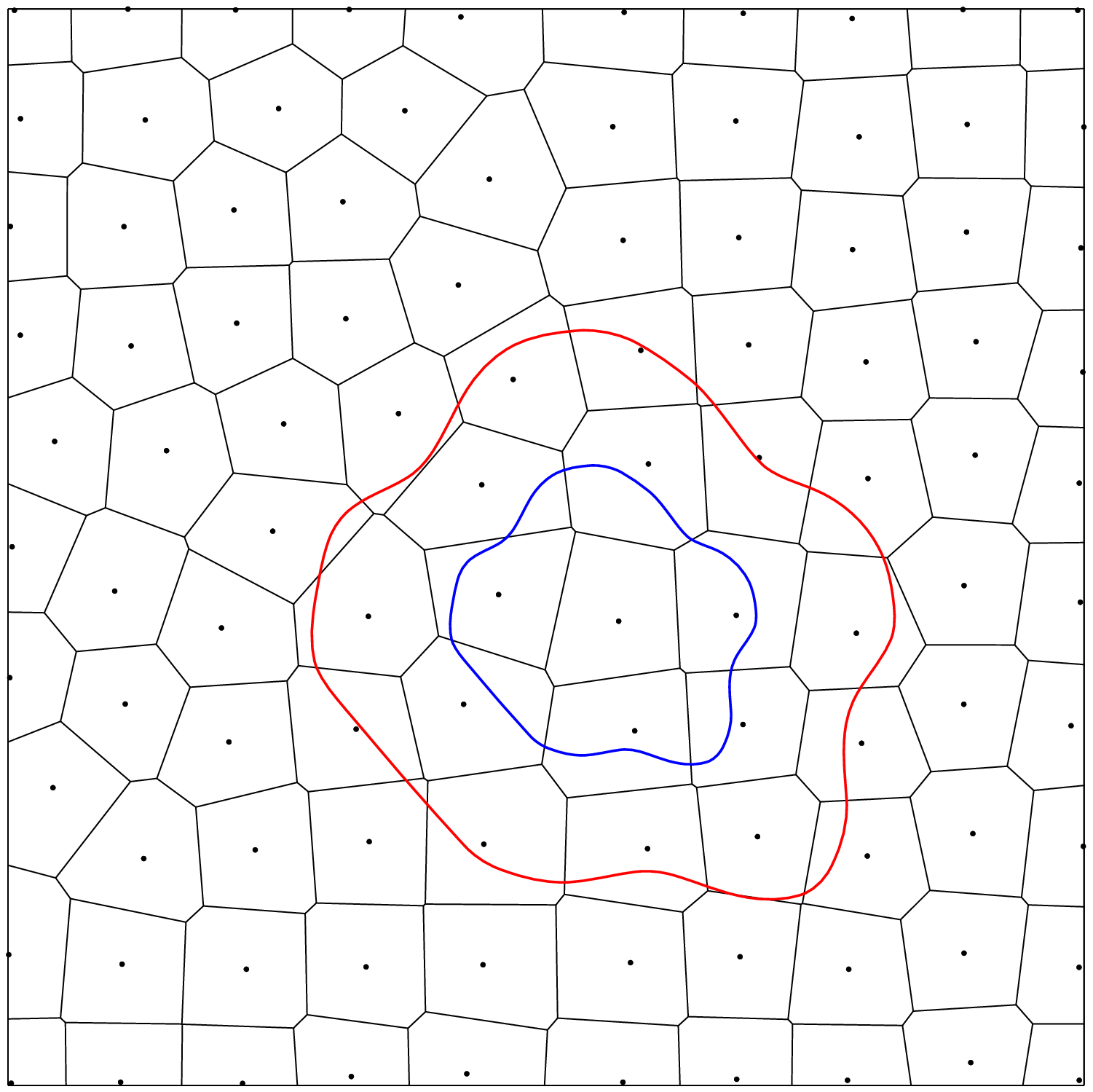}
           \put(60,54){{\color{blau}$T$}}
           \put(64,62){{\color{rot}$T^{+\d}$}}
	   \put(91,95){$X$}
	 \end{overpic}
	 \caption{An input space $X$ with the corresponding Voronoi partition as well as a subset 
$T\subset X$ enlarged by an $\d$-tube to $T^{+\d}$.}
	 \label{Bild_Partition_T^{+d}}
	\end{center}
\end{scriptsize}\vspace{-1cm}
\end{wrapfigure}
Note that the oracle inequalities and learning rates achieved in Theorems 
\ref{LS oracle ineq} and \ref{Thm LS rate} require 
$\fB\in B_{2,\infty}^{\a}(\TrP{\bigcup_{j\in J_T}A_{j}})$. However, for an 
increasing sample size $n$, the sets $A_j$ shrink and the index set $J_T$, indicating 
every set $A_j$ such that $A_j\cap T\neq\emptyset$ and $T\subset \bigcup_{j\in J_T}A_j$, 
increases. In particular, this also involves that the set $\bigcup_{j\in J_T}A_j$ 
covering $T$ changes in tandem with $n$. Since this is very inconvenient and since it 
would be desirable to assume a certain level of smoothness of the target function on a 
fixed region for all $n\in\N$, we consider the set $T$ enlarged by an $\d$-tube.
To this end, for $\d>0$, we define $T^{+\d}$ by
\begin{align}\label{T enlarged by tube}
 T^{+\d} \!:=\! \left\{ x\!\in\! X : \exists t\!\in\! T \text{ such that } \|x-t\|_2\leq\d\right\}\,\!,
\end{align}
which implies $T\subset T^{+\d}\subset X$, cf.~Figure \ref{Bild_Partition_T^{+d}}.
Note that, for every $\d>0$, there exists an 
$n_\d\in\N$ such that for every $n\geq n_\d$ the union of all partition sets $A_j$, 
having at least one common point with $T$, is contained in $T^{+\d}$, i.e.
\begin{align}\label{ex. n_d}
 \forall \d>0 \quad \exists n_\d\in\N \quad \forall n\geq n_\d \quad : \quad
 \bigcup_{j\in J_T}A_j\subset T^{+\d}\,,
\end{align}
where $J_T:=\{j\in\{1,\ldots,m_n\} : A_j\cap T\neq\emptyset\}$. Collectively, this implies 
\begin{align*}
 T \subset \bigcup_{j\in J_T}A_j\subset T^{+\d} 
\end{align*}
for all $n\geq n_\d$. 
Furthermore, since every set $A_j$ is contained in a ball with radius 
$r_n=cn^{-\frac{1}{\b d}}$, the lowest sample size $n_\d$ in \eqref{ex. n_d} can be 
determined by choosing the smallest $n_\d\in\N$ such that
$\d \geq 2 r_{n_\d}$ with $r_{n_\d}$ as in \eqref{ex. Ueberdeckung}, that is 
\begin{align*}
 n_\d = \left\lceil \left(\frac{2c}{\d}\right)^{\b d} \right\rceil \,.
\end{align*}
This leads to the following corollary where we present an oracle inequality and 
learning rates assuming the smoothness level $\a$ of the target function on a 
fixed region.

\begin{corollary}\label{rate for smoothness von T^{+d}}

Let $Y:=\left[-M,M\right]$ for $M>0$, $L: Y\times\R\to [0,\infty)$ be the least squares 
loss,  and $\P$ be a distribution on $\Rd \times Y$. We write $X:=\supp{\P_X}$. 
Furthermore, let (A) and (G) be satisfied. In addition, for an arbitrary subset 
$T\subset X$, we assume (T). 
Moreover, let $f^*_{L,\P}:\Rd\to\R$ be a Bayes decision function such that 
$f^*_{L,\P}\in L_2(\Rd) \cap L_\infty(\Rd)$ as well as 
\begin{align*}
 \fB\in B_{2,\infty}^{\a}(\TrP{T^{+\d}})
\end{align*}
for $\a \geq 1$ and some $\d>0$.
Then, for all $p \in \left(0,1\right)$, $n\geq n_\d$, $\t\geq1$, 
$\bs\g=(\g_1,\ldots,\g_m)\in(0,r]^m$, and $\bs\lb=(\lb_1,\ldots,\lb_m)>0$, 
the VP-SVM given by \eqref{VP-SVM} using 
$\hat{H}_{\g_1}(A_{1}),\ldots,\hat{H}_{\g_m}(A_{m})$, and the loss $L_T$ satisfies 
\begin{align*}
 & \sum_{j=1}^m \lb_j\|f_{\D_{j},\lb_j,\g_j}\|^2_{\hat{H}_{\g_j}(A_j)}
 +\RLPT(\fcl_{\D,\bs\lb,\bs\g}) - \RBT \nonumber\\ 
 & \leq C_{M,\a,p}\!\left( \sum_{j \in J_T} \!\lb_j \g_j^{-d} \!+\! 
        \left(\frac{\max_{j\in J_T}\g_{j}}{\min_{j\in J_T}\g_{j}}\right)^d \!
        \max_{j\in J_T}\g_{j}^{2\a}
        \! +\! r^{2p}\! \left( \sum_{j=1}^m \!
        \lb_{j}^{-1} \g_j^{-\frac{d+2p}{p}} \P_X(A_{j}) \!\right)^p \!\!n^{-1} \!\!
        +\! \t n^{-1} \!\right) 
\end{align*}
with probability $\P^n$ not less than $1-e^{-\t}$, where $C_{M,\a,p}>0$ is the same 
constant as in Theorem \ref{LS oracle ineq}.

Additionally, let $\b\geq \frac{2\a}{d}+1$ as well as, for every 
$j\in \{1,\ldots,m_n\}$, $r_n$, $\lb_{n,j}$, and $\g_{n,j}$ be as in \eqref{Def. r}, 
\eqref{Def. lb_n}, and \eqref{Def. g_n}, respectively, 
where $c_1,c_2,c_3$ are user-specified positive 
constants with $ c_3\leq c_1$. Then, for all 
$n\geq n_\d= \left\lceil \left(\frac{2c_1}{\d}\right)^{\b d} \right\rceil$ and $\x>0$,
we have
\begin{align*}
 \RLPT(\fcl_{\D,\bs\lb_n,\bs\g_n})-\RBT \leq C\t n^{-\frac{2\a}{2\a+d}+\x} 
\end{align*}
with probability $\P^n$ not less than $1-e^{-\t}$, where 
$\bs\lb_n:=(\lb_{n,1},\ldots,\lb_{n,m_n})$, $\bs\g_n:=(\g_{n,1},\ldots,\linebreak\g_{n,m_n})$, 
and $C$ is a positive constant.

\end{corollary}


Note that the assumption $f^*_{L,\P}\in B_{2,\infty}^{\a}(\TrP{T^{+\d}})$ made in Corollary \ref{rate for smoothness von T^{+d}}
is satisfied if, for example, $f^*_{L,\P}\in B_{2,\infty}^{\a}(T^{+2\d})$ and $\P_X$ has a bounded 
Lebesgue density on $T^{+\d}$. Moreover, if this density is even bounded away from 0, 
it is well-known that the 
minmax rate  is $n^{-\frac{2\a}{2\a+d}}$ for $\a>d/2$ and target functions 
$\fB\in W_2^\alpha\left(T\right)$. 
Modulo $\x$, our rate is therefore asymptotically optimal in a minmax sense on $T$.
In addition, for $\a>d$, the learning rates obtained for 
$\fB\in B_{2,\infty}^\alpha(T)$ are again asymptotically optimal modulo $\x$ on $T$.

\subsection{Data-Dependent Parameter Selection for VP-SVMs}\label{TV-approach LS}

Note that in the previous theorems the choice of the regularization parameters 
$\lb_{n,1},\ldots,\linebreak\lb_{n,m_n}$ and the kernel widths $\g_{n,1},\ldots,\g_{n,m_n}$ 
requires us to know the smoothness parameter $\a$. 
Unfortunately, in practice, we usually do know neither 
this value nor its existence. In this subsection, we thus show that a 
training/validation approach similar to the one examined in 
\citep[Chapters 6.5, 7.4, 8.2]{ StCh08} and \citep{ EbSt13} achieves the same rates 
adaptively, i.e.~without knowing $\a$.
For this purpose, let $\Lb:=(\Lb_n)$ and $\G:=(\G_n)$ be sequences of finite subsets 
$\Lb_n\subset(0,r_n^d]$ and $\G_n\subset(0,r_n]$. For a data set 
$D:=((x_1,y_1),\ldots,(x_n,y_n))$, we define
\begin{align*}
 D_1 & :=((x_1,y_1),\ldots,(x_l,y_l))\,,\\
 D_2 & :=((x_{l+1},y_{l+1}),\ldots,(x_n,y_n))\,,
\end{align*}
where $l:=\lfloor\frac{n}{2}\rfloor+1$ and $n\geq4$. We further split these sets in data 
sets 
\begin{align*}
 D_{j}^{(1)} & :=\left\{(x_i,y_i)\in D_1 : x_i\in A_{j}\right\}\,,
 \qquad\qquad j\in\{1,\ldots,m_n\}\,, \\
 D_{j}^{(2)} & :=\left\{(x_i,y_i)\in D_2 : x_i\in A_{j}\right\}\,,
 \qquad\qquad j\in\{1,\ldots,m_n\} \,,
\end{align*}
and define $l_j:=|D_{j}^{(1)}|$ for all $j\in\{1,\ldots,m_n\}$ such 
that $\sum_{j=1}^{m_n}l_j=l$. For every $j\in\{1,\ldots,m_n\}$, we basically use 
$D_{j}^{(1)}$ as a training set, i.e.~based on $D_1$ in combination with the loss 
function $L_j:=\ind_{A_{j}}L$ we compute SVM decision functions
\begin{align*}
 f_{\D_j^{(1)},\lb_j,\g_j} := \argmin{f\in \hat{H}_{\g_j}(A_{j})}
 \lb_j \|f\|_{\hat{H}_{\g_j}(A_{j})}^2 +\RR_{L_j,\D_1}(f) \,, 
 \qquad \qquad (\lb_j,\g_j)\in\Lb_n\times\G_n\,.
\end{align*}
Again, note that $f_{\D_j^{(1)},\lb_j,\g_j}=0$ if $D_j^{(1)}=\emptyset$. 
Next, for each $j$, we use $D_2$ in tandem with $L_j$ (or essentially 
$D_{j}^{(2)}$) to determine a pair $(\lb_{\D_2,j},\g_{\D_2,j})\in\Lb_n\times\G_n$ 
such that 
\begin{align*}
 \RR_{L_j,\D_2}\left(\fcl_{\D_j^{(1)},\lb_{\D_2,j},\g_{\D_2,j}}\right) = 
 \min_{(\lb_j,\g_j)\in\Lb_n\times\G_n} 
 \RR_{L_j,\D_2}\left(\fcl_{\D_j^{(1)},\lb_j,\g_j}\right)\,.
\end{align*}
Finally, combining the decision functions $f_{\D_j^{(1)},\lb_{\D_2,j},\g_{\D_2,j}}$ for 
all $j\in\{1,\ldots,m_n\}$, and defining 
$\bs\lb_{\D_2}:=(\lb_{\D_2,1},\ldots,\lb_{\D_2,m_n})$ and 
$\bs\g_{\D_2}:=(\g_{\D_2,1},\ldots,\g_{\D_2,m_n})$, we obtain a function 
\begin{align*}
 f_{\D_1,\bs\lb_{\D_2},\bs\g_{\D_2}} 
 := \sum_{j=1}^{m_n} f_{\D_j^{(1)},\lb_{\D_2,j},\g_{\D_2,j}}
 = \sum_{j=1}^{m_n}\ind_{A_{j}} f_{\D_j^{(1)},\lb_{\D_2,j},\g_{\D_2,j}}\,,
\end{align*}
and we call every learning method that produces these resulting decision functions 
\linebreak$f_{\D_1,\bs\lb_{\D_2},\bs\g_{\D_2}}$ a \textit{training validation 
Voronoi partition support vector machine} (TV-VP-SVM)  w.r.t.~$\Lb\times\G$.
Moreover, using \eqref{swapping sum and min} we have, for 
$\bs\lb:=(\lb_1,\ldots,\lb_{m_n})$ and $\bs\g:=(\g_1,\ldots,\g_{m_n})$,
\begin{align*}
 \RR_{L,\D_2}\left(\fcl_{\D_1,\bs\lb_{\D_2},\bs\g_{\D_2}}\right) 
 & = \sum_{j=1}^{m_n}\RR_{L_j,\D_2}\left(\fcl_{\D_j^{(1)},\lb_{\D_2,j},\g_{\D_2,j}}\right)\\
 & = \sum_{j=1}^{m_n}\min_{(\lb_j,\g_j)\in\Lb_n\times\G_n} 
     \RR_{L_j,\D_2}\left(\fcl_{\D_j^{(1)},\lb_j,\g_j}\right) \\
 & = \min_{(\bs\lb,\bs\g)\in(\Lb_n\times\G_n)^{m_n}} \sum_{j=1}^{m_n}
     \RR_{L_j,\D_2}\left(\fcl_{\D_j^{(1)},\lb_j,\g_j}\right) \\
 & = \min_{(\bs\lb,\bs\g)\in(\Lb_n\times\G_n)^{m_n}}\RR_{L,\D_2}\left(\fcl_{\D_1,\bs\lb,\bs\g}\right)\,,
\end{align*}
where $f_{\D_1,\bs\lb,\bs\g} := \sum_{j=1}^{m_n} f_{\D_j^{(1)},\lb_j,\g_j}$ with
$(\lb_j,\g_j)\in\Lb_n\times\G_n$ for all $j\in\{1,\ldots,m_n\}$. In other words, the 
function $\fcl_{\D_1,\bs\lb_{\D_2},\bs\g_{\D_2}}$ really minimizes the empirical risk 
$\RR_{L,\D_2}$ w.r.t.~the validation data set $D_2$ and the loss $L$, where the minimum 
is taken over all functions $\fcl_{\D_1,\bs\lb,\bs\g}$ with 
$(\bs\lb,\bs\g)\in(\Lb_n\times\G_n)^{m_n}$.

The following theorem presents learning rates for the above described TV-VP-SVM.

\begin{theorem}\label{LS oracle ineq data-dep param}
Let $r_n:=cn^{-\frac{1}{\b d}}$ with constants $c>0$ and $\b>1$.
Under the assumptions of Theorem \ref{LS oracle ineq} we fix sequences $\Lb:=(\Lb_n)$ 
and $\G:=(\G_n)$ of finite subsets $\Lb_n\subset(0,r_n^d]$ and $\G_n\subset(0,r_n]$ 
such that $\Lb_n$ is an $(r_n^d\e_n)$-net of $(0,r_n^d]$ and $\G_n$ is a $\d_n$-net of 
$(0,r_n]$ with $\e_n\leq n^{-1}$ and $\d_n\leq n^{-\frac{1}{2+d}}$. Furthermore, assume 
that the cardinalities $|\Lb_n|$ and $|\G_n|$ grow polynomially in $n$. Then, for all 
$\x>0$, $\t\geq 1$, and $\a < \frac{\b-1}{2}d$, the TV-VP-SVM producing the decision 
functions $f_{\D_1,\bs\lb_{\D_2},\bs\g_{\D_2}}$ satisfies
\begin{align*}
 \P^n\left( \RLPJT(\fcl_{\D_1,\bs\lb_{\D_2},\bs\g_{\D_2}}) - \RBJT  
 \leq c \t n^{-\frac{2\a}{2\a+d}+\x}\right) \geq 1-e^{-\t}  \,,
\end{align*}
where $c>0$ is a constant independent of $n$ and $\t$.
\end{theorem}

Once more, we can replace the assumption 
$\fB\in B_{2,\infty}^{\a}(\TrP{A_T})$ by 
$\fB\in B_{2,\infty}^{\a}(\TrP{T^{+\d}})$ for some $\d>0$ and obtain the same 
learning rate as in Theorem \ref{LS oracle ineq data-dep param} for all $n\geq n_\d$ 
although $T^{+\d}$ is fixed for all $n\in\N$.
Note that, if $\P_X$ has a Lebesgue density that is bounded away from $0$ and $\infty$
and either $\fB\in W_2^\alpha\left(T\right)$ for $\a>d/2$ or
$\fB\in B_{2,\infty}^\alpha(T)$ for $\a>d$, 
these learning rates are again asymptotically optimal modulo $\x$ on 
$T$ in a minmax sense.
However, the condition $\a<\frac{\b-1}{2}d$ restricts the set of $\a$-values where we
obtain learning rates adaptively. To be more precise, there is a trade-off between 
$\a$ and $\b$. On the one hand, for small values of $\b$ only a small number of 
possible values for $\a$ is covered. On the other hand, for larger values of $\b$ 
the set of $\a$-values where we achieve rates adaptively is increasing 
but the savings in terms of computing time is decreasing.

\section{Experimental Results}\label{Sec Experiments}

In the previous sections we defined VP-SVMs and derived 
local learning rates that are essentially optimal.
So far, it is, however, not clear if the theoretical results 
suggesting a generalization performance not worse than that of  global SVMs can be empirically confirmed
and if the predicted advantages of VP-SVMs in terms of computational 
costs are preserved in practice. Note that the latter is not as obvious as it may seem to be, since 
VP-SVMs create an overhead when generating the working sets, and the working sets themselves do not need to be as
balanced as we assumed in our na\"ive analysis.
In this section, we thus investigate the performance of 
VP-SVMs empirically. Namely, we carry out some experiments using the least squares loss 
with the objective to answer the subsequent questions:
\begin{enumerate}[\bf(1)]
 \item How do different radii affect the performance of VP-SVMs? In particular, what is the impact 
 on the training time and the VP-SVM's test error?
 \item How do the VP-LS-SVMs perform compared to the usual LS-SVMs 
 in terms of the test error? What is the speed-up?
%
 \item How does the performance of  VP-SVMs compare to vanilla data splitting approaches such as 
 random chunking \emph{(RC-SVM)}, in which the data set is devided into a random partition 
 with equally sized subsets, and the final decision function is the average of the 
 SVMs computed on each subset?
 \item How does the VP-LS-SVM behave compared to the global 
 LS-SVM, if the regression function has interruptions of its smoothness on zero sets?
\end{enumerate}

\begin{figure}[htbp]
\begin{scriptsize}
\begin{minipage}[t]{0.49\textwidth}
\subfigure[Artificial data Type \RM{1}: step function\label{fig::experiments::basic_fct_1}]{\includegraphics[angle=270, scale=0.28]{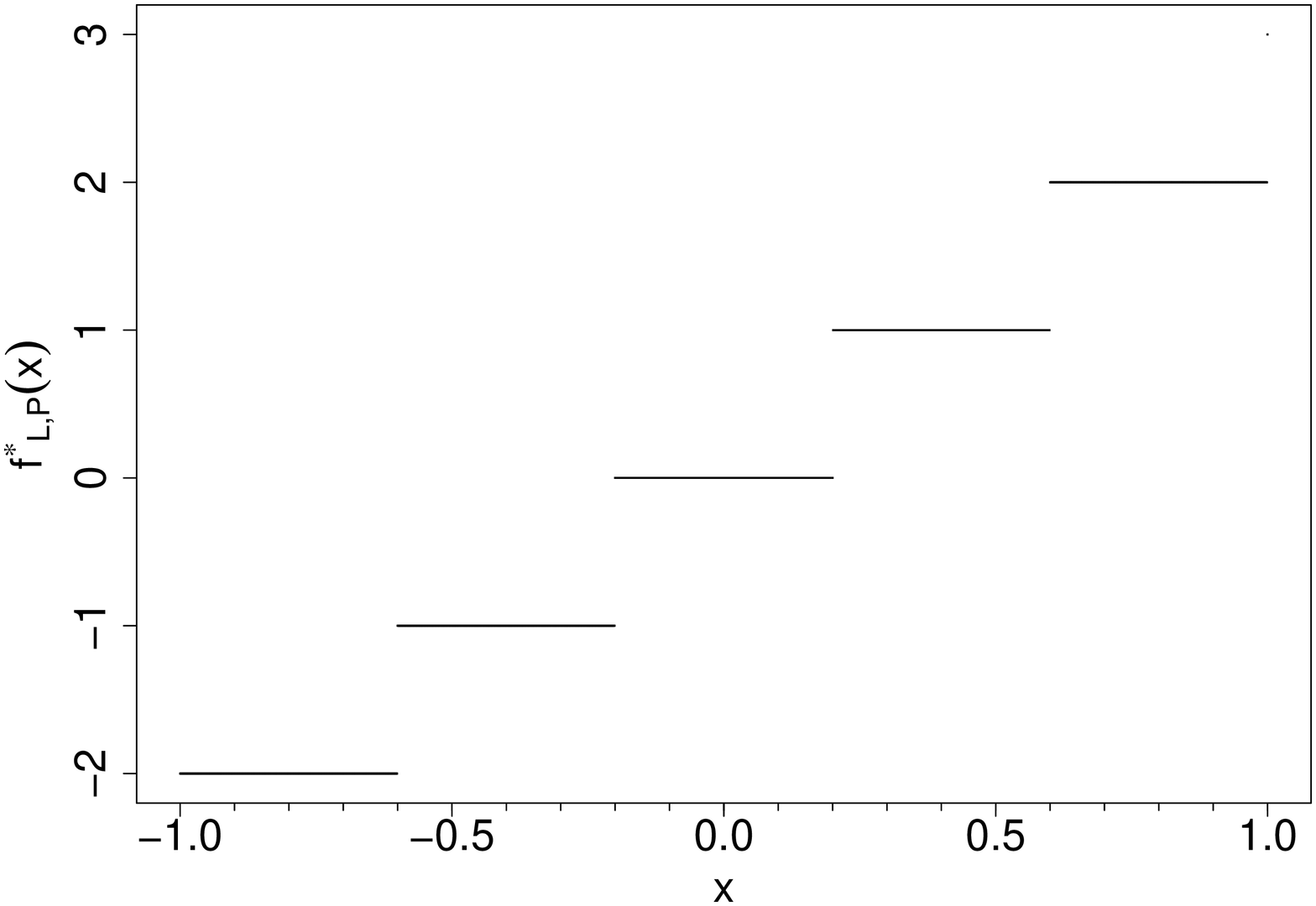}}\\
\subfigure[Artificial data Type \RM{2}: cracked function\label{fig::experiments::basic_fct_2}]{\includegraphics[angle=270, scale=0.28]{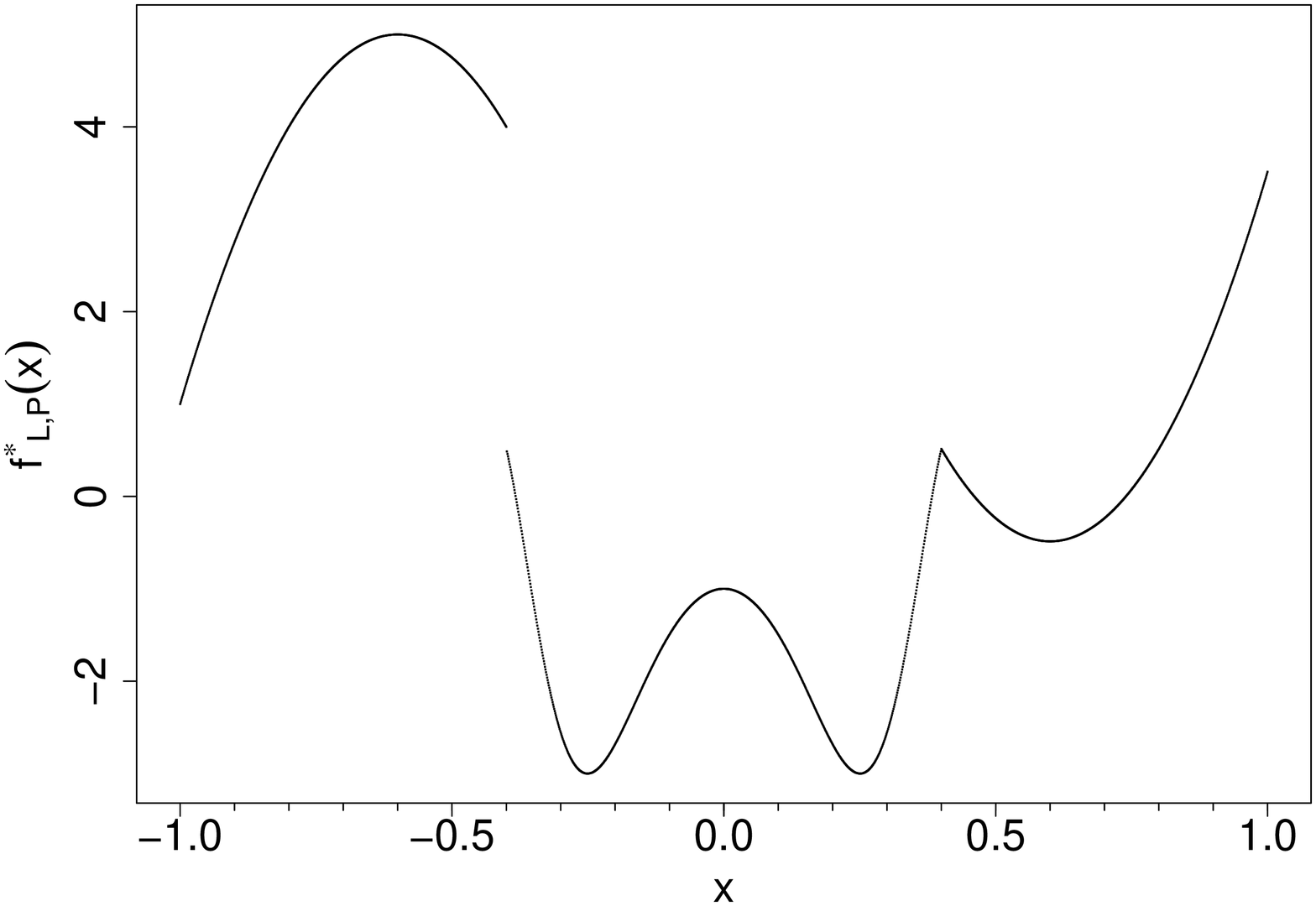}}\\
\subfigure[Artificial data Type \RM{3}: jagged function\label{fig::experiments::basic_fct_3}]{\includegraphics[angle=270, scale=0.28]{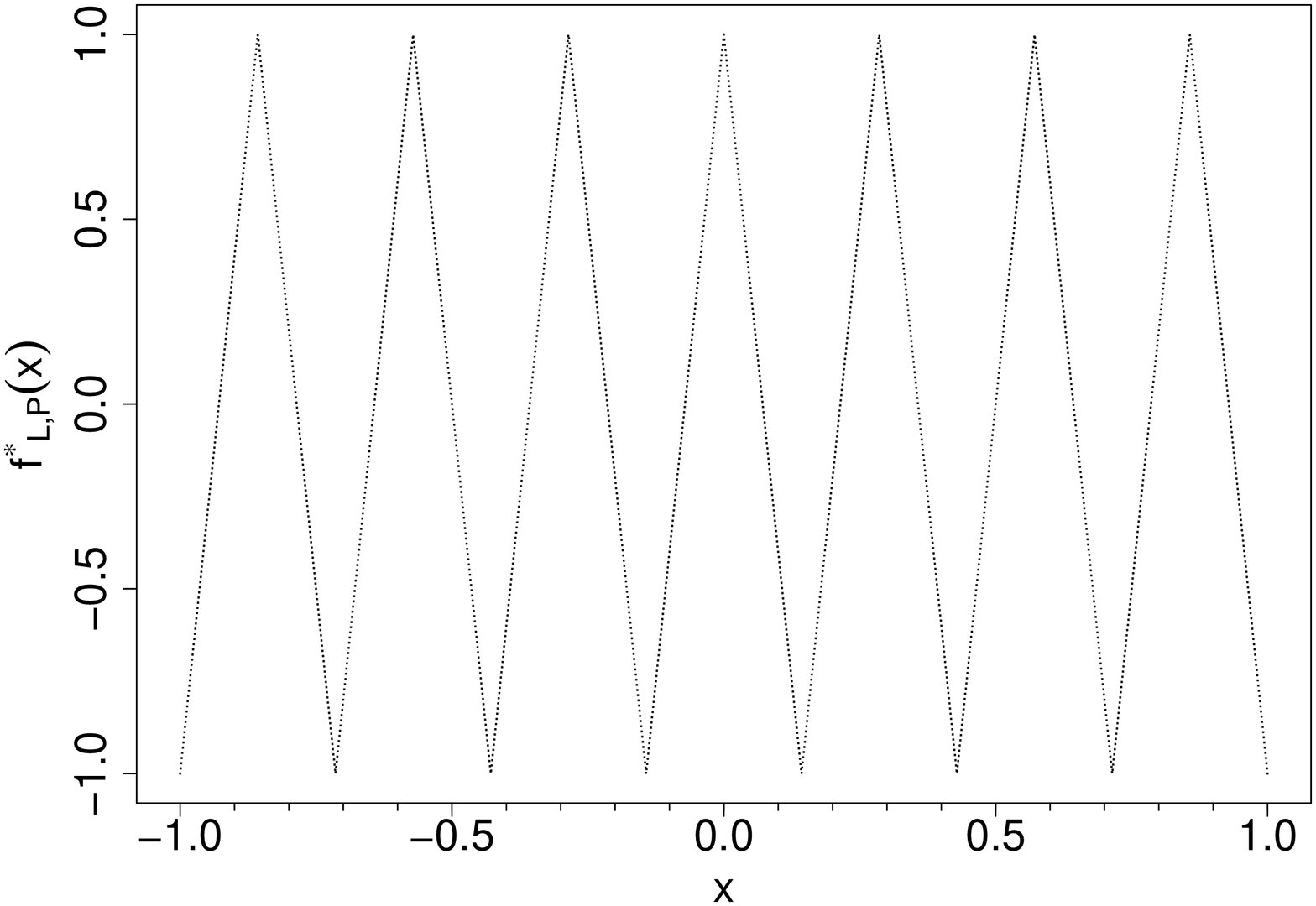}}
\end{minipage}\hfill
\begin{minipage}[t]{0.49\textwidth}
\subfigure[Artificial data Type \RM{4}: circular step function\label{fig::experiments::basic_fct_4}]{\includegraphics[angle=270, scale=0.28]{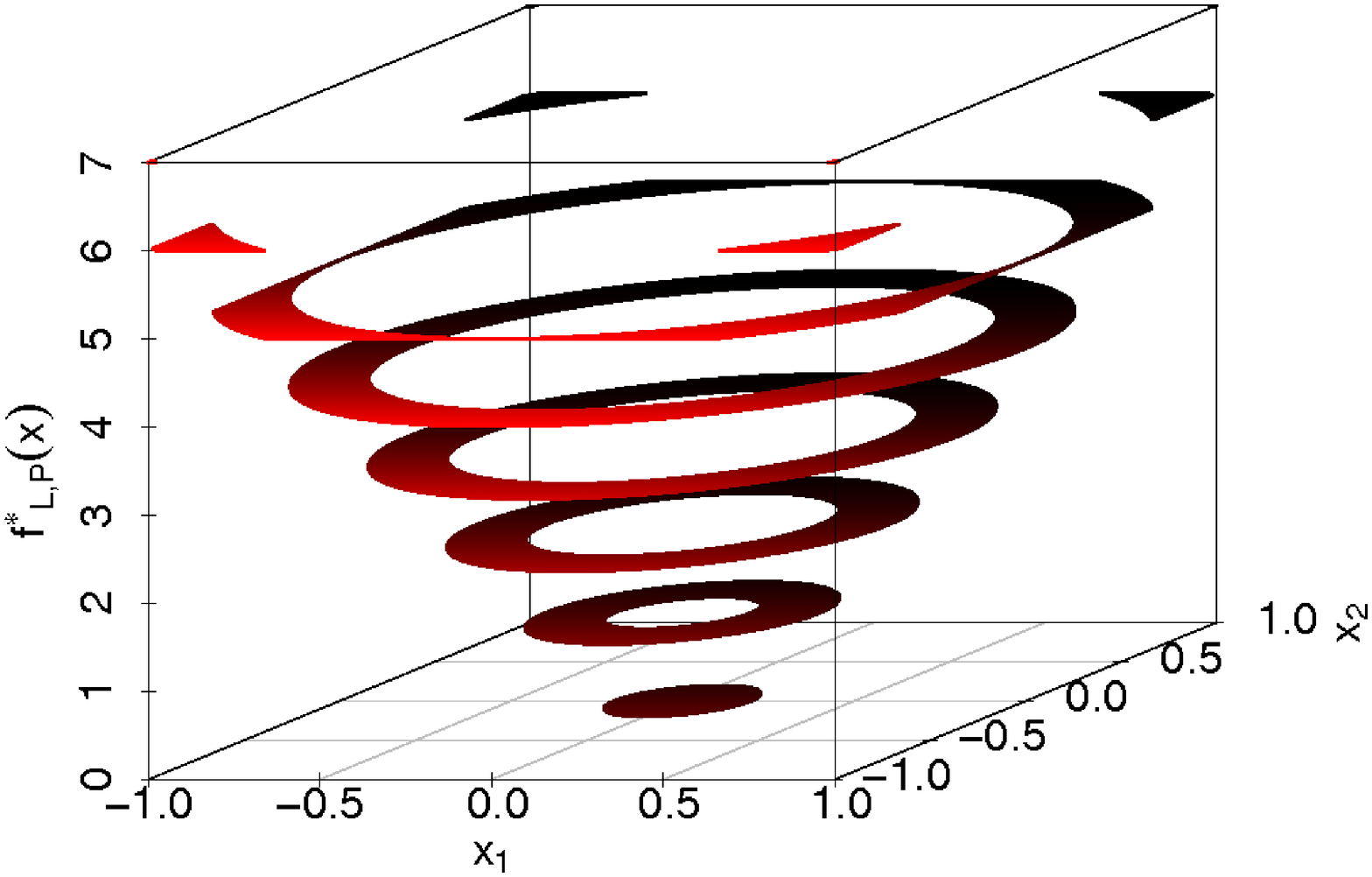}}\\
\subfigure[Artificial data Type \RM{5}: 2-dimensional Euclidean norm\label{fig::experiments::basic_fct_5}]{\includegraphics[angle=270, scale=0.28]{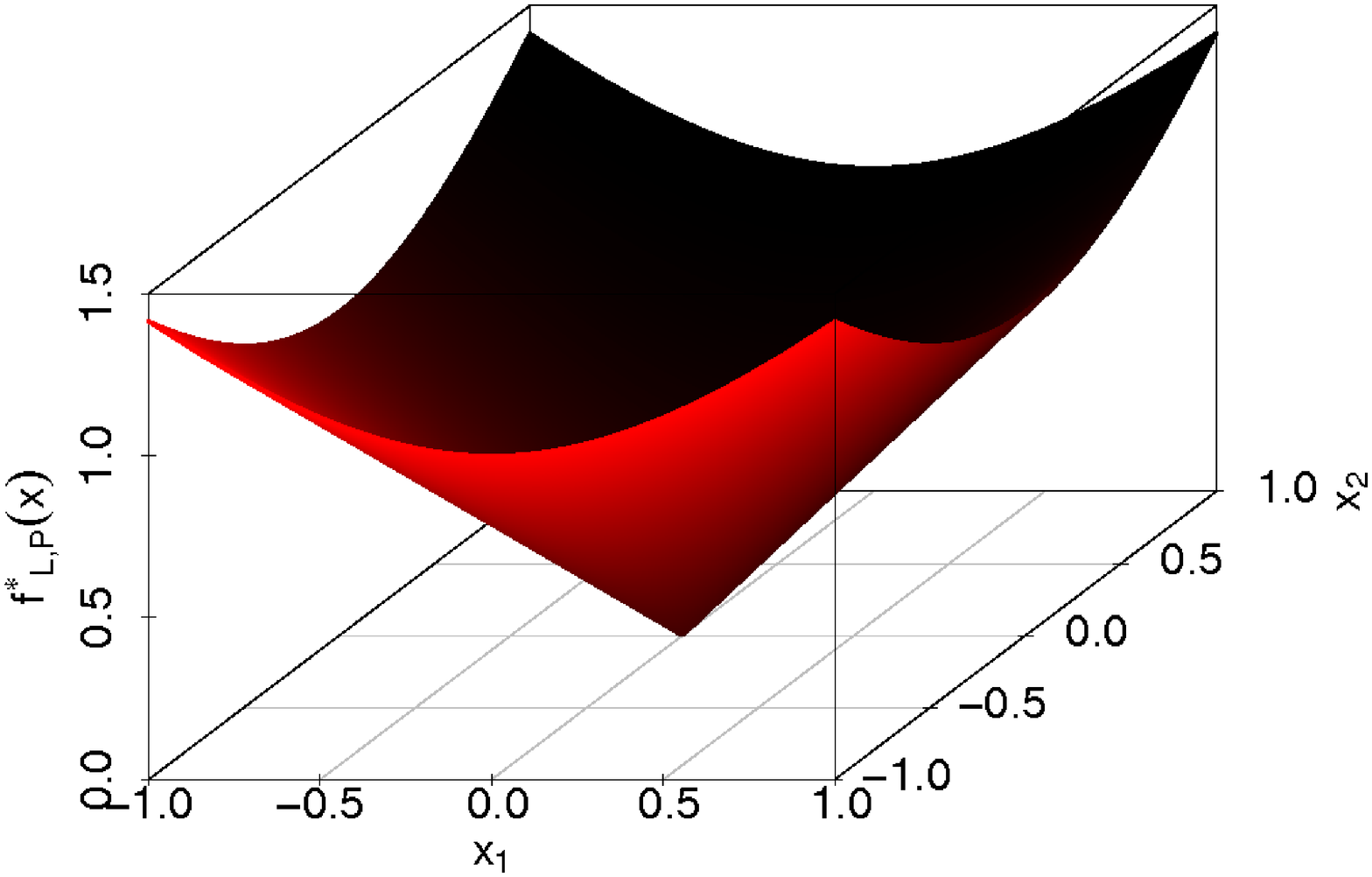}}   
\end{minipage}
\caption{Unscaled basic functions used to generate the artificial data sets.}
\label{fig::experiments::basic_fcts}
\end{scriptsize}
\end{figure}

To address these questions we utilize two kinds of data sets. On the one hand, to answer 
questions (1), (2), and (3), we  
examine the three real data sets \textsc{covtype}, \textsc{ijcnn1}, and \textsc{cod-rna}, 
which we obtained from LIBSVM's homepage, see \citep{LIBSVM}. 
Table \ref{tab::experiments::dataset_characteristics} summarizes some characteristics 
of these data sets. 
\begin{table}[bht]
  \centering
      \begin{tabular}{l|ccc}      
	data set type & full data set size & dimension & number of labels \\ \hline\hline
	\textsc{covtype} & $581\,012$ & $54$ & $2$ \\
	\textsc{cod-rna} & $488\,565$ &  $8$ & $2$ \\
	\textsc{ijcnn1}  & $141\,691$ & $22$ & $2$ 
      \end{tabular}
    \caption{Characteristics of the considered LIBSVM data sets.}
    \label{tab::experiments::dataset_characteristics}
\end{table}
On the other hand, we generated several artificial data sets to address the last question. 
In order to prepare the data sets for the experiments, we edited the data sets 
from LIBSVM in the following manner. If for a real-world data set type the raw data set was 
already split, we first merged these sets so that we obtained one data set for each data 
set type. 
In a next step, we scaled the data componentwise such that all samples including 
labels lie in $[-1,1]^{d+1}$, where $d$ is the dimension of the input data. Finally, for 
each data set type, we generated random subsets that were afterwards randomly splitted into 
a training and a test data set. In this manner, we obtained, for each of the three 
LIBSVM data set types, training sets consisting of $n = 1\,000,\ 2\,500,\ 5\,000,\ 10\,000,\ 
25\,000,\ 50\,000,\ 100\,000$ samples.
Additionally, for the data sets \textsc{covtype} and \textsc{cod-rna}, we created training sets 
of sizes $250\,000$ and $500\,000$, and of sizes $250\,000$ and $400\,000$, 
respectively. The test data sets associated to the various training sets consist of 
$n_\text{test}=50\,000$ random samples, apart from the training sets with  
$n_\text{train}\leq5\,000$, for which we took $n_\text{test}=10\,000$ test samples.

%

For the artificial data, we proceeded in a slightly different way. To generate the 
data sets we took as fundament the five regression functions pictured in Figure 
\ref{fig::experiments::basic_fcts}
and as noise, the sum of two uniform distributions on $[-c(x),c(x)]$, where 
$c(x)=\tfrac{1}{4}\left(3\sin\left(\tfrac{\pi}{2}|x|\right)+1\right)$ 
for the one-dimensional data sets and 
$c(x)=\tfrac{1}{4}\left(\sin\left(\tfrac{\pi}{4}\left(|x_1|+|x_2|\right)\right)+1\right)$ 
for the two-dimensional data sets. Thus, we produced five different types of artificial 
data sets, where the various data set types 
are named according to their type numbers as in Figure \ref{fig::experiments::basic_fcts}. 
Initially, we created two sets, namely one training and one test data set, 
each consisting of $10\,000$ random input samples contained in $[-1,1]$ and $[-1,1]^2$, 
respectively. Then, for each artificial data set type, we determined the labels belonging 
to the input data as sum of the corresponding functional value and the noise
and, finally, scaled all $20\,000$ labels to $[-1,1]$. In a last step, we randomly 
built subsets of the training sets of size $n=1\,000,\ 2\,500,\ 5\,000$. In this way, we 
altogether obtained, for each type of artificial data, four 
training data sets of size $n=1\,000,\ 2\,500,\ 5\,000,\ 10\,000$ and a corresponding 
test data set of size $n_\text{test}=10\,000$. Based on the test data sets the Bayes risks 
can be determined, see Table \ref{tab::experiments::bayes_risks} where the Bayes risks are 
summarized for the various artificial data set types.
\begin{table}[thb]
  \centering
      \begin{tabular}{lccccc}      
	&Type \RM{1} & Type \RM{2} & Type \RM{3} & Type \RM{4} & Type \RM{5} \\ \hline
	Bayes risk & 0.0254 & 0.0137 & 0.0529 & 0.0083 & 0.0634\\ 

      \end{tabular}
    \caption{Bayes risks w.r.t.~test data sets for the various artificial data set types.}
    \label{tab::experiments::bayes_risks}
\end{table}

To minimize random effects, 
we repeated the 
experiment for each setting several times. 
Since experiments using large data sets entail long run times, we reran every 
experiment using a training set of size $n \geq 50\,000$ only three times while for 
training sets of size $n=10\,000,\ 25\,000$ we performed ten repetitions and for smaller 
training sets, namely of size $n=1\,000,\ 2\,500,\ 5\,000$, even $100$ runs. 
An exception 
are the experiments using artificial training sets of size $n=10\,000$, where we realized 
$100$ repetitions for the sake of uniformity.

To approach the above problems we used the least squares loss and Gaussian kernels for all 
experiments. We implemented an LS-SVM-solver in \Cpp~similar to the one in 
\citep{ StHuSc11a}. Around this solver, we then built the routines for the VP-SVM and the RC-SVM.
The compilation of the three programmes was executed by LINUX's gcc. To produce 
comparable results in terms of run time, all real-world data experiments were realized 
by the same professional compute server\footnote{On this occasion, we would 
like to thank the Institute for Applied Analysis and Numerical Simulation of the University of 
Stuttgart, who placed the above mentioned compute server at our disposal and, 
thus, enabled us to realize our experiments on large real-world data sets. In consequence, the overall
time available for our experiments was limited.} 
equipped with four INTEL XEON E7-4830 (2.13 GHz) 8-core processor, 
256 GB RAM, and a 64 bit version of Debian GNU/Linux 6.0.7. In order that we can indeed 
compare their run time, we used eight cores to pre-compute the kernel matrix 
and to evaluate the final decision functions on the test set,
and one core 
for the subsequent solver for every real data experiment. Since the artificial 
data sets consist of at most $10\,000$ samples we performed the according experiments by 
a computer equipped with one INTEL CORE i7-3770K (3.50 GHz) quad core processor, 
16 GB RAM, and a 64 bit version of Debian GNU/Linux 6.0.7. For all artificial data 
experiments we used four cores to pre-compute the kernel matrix and
to evaluate the final decision functions on the test set,
and again one core for the 
solver. 
Even with pre-computed kernel matrices, our experiments on the real-world data altogether 
required almost 810 hours (approximately 34 days) 
for training and additionally almost 4 days for testing. Moreover, the experiments on the 
artificial data took nearly 43 hours for training and 168 minutes for testing. 
Without pre-computing the kernel matrices, e.g.~by applying a standard caching approach, 
preliminary experiments suggested a multiplcation of the training time, which would have rendered the 
experiments infeasible.
%
Besides, our experiments will show that the available amount of RAM does not 
restrict the size of the training sets used by an VP-SVMs as severely as the ones used by 
LS-SVMs.

\begin{algorithm}[t!]
\caption{Determine a Voronoi partition of the input data}
\label{alg::experiments::vp_creation}
\begin{algorithmic}[1]
 \REQUIRE{Input data set $D_X=\{x_1,\ldots, x_n\}$ with sample size $n\in\N$ and some radius $r>0$.}
 \ENSURE{Working sets indicating a Voronoi partition of $D_X$.}
 \STATE Pick an arbitrary $z\in D_X$
 \STATE $\text{\textsl{Cover}}_1\leftarrow z$
 \STATE $m \leftarrow 1$
 \WHILE{$\max_{x\in D_X}\|x-\text{\textsl{Cover}}\|_2>r$}
 \STATE $z \leftarrow \arg\max_{x\in D_X}\|x-\text{\textsl{Cover}}\|_2$
 \STATE $m \leftarrow m + 1$
 \STATE $\text{\textsl{Cover}}_m \leftarrow z$
 \STATE{$\text{\textsl{WorkingSet}}_m \leftarrow \emptyset$}
 \ENDWHILE
 \FOR{$i=1$ \TO $n$}
 \STATE{$k \leftarrow \arg\min_{j\in\{1,\ldots,m\}} \|x_i -  \text{\textsl{Cover}}_j\|_2$}
 \STATE{$\text{\textsl{WorkingSet}}_k \leftarrow \text{\textsl{WorkingSet}}_k  \cup \{x_i\}$}
 \ENDFOR
 \RETURN $\text{\textsl{WorkingSet}}_1,\ldots,\text{\textsl{WorkingSet}}_m$
\end{algorithmic}
\end{algorithm}
Let us quickly illustrate the routines of the VP- and the RC-SVM implemented around the 
LS-solver. For the VP-SVM, we first 
split the training set by Algorithm \ref{alg::experiments::vp_creation} 
in several working sets representing a Voronoi partition w.r.t.~the user-specified 
radius. For this purpose, Algorithm \ref{alg::experiments::vp_creation} initially determines a 
cover of the input data applying the farthest first traversal algorithm, see 
\citep{Da08} and \citep{Go85} for more details. Note that this procedure induces working sets 
whose sizes may be considerably varying.
In the case of an 
RC-SVM the working sets are created randomly, where their sizes are basically equal 
and the number of working sets is predefined by the user. Then, for the 
VP-SVM- as well as for the RC-SVM-algorithm the implemented LS-solver is applied on 
every working set.
For each working set, we randomly split the respective training data set of size 
$n_\text{train}$ in five folds to apply 5-fold cross-validation in order to deal with 
the hyper-parameters $\lb$ and $\g$ taken from an $10$ by $10$ grid geometrically 
generated in 
$[0.001\cdot n_\text{train}^{-1},0.1]\times[0.5\cdot n_\text{train}^{-1/d},10]$.
Finally, we 
obtain one decision function for each working set. 
To further process these decision functions the VP-SVM-algorithms picks exactly one 
decision function depending on the working set affiliation of the input value. On the 
contrary, the RC-SVM-algorithm simply takes the average of all the decision functions. 
Moreover, since we scaled the labels of all data sets to $[-1,1]$, the computed decision 
functions are clipped at $\pm1$.
Altogether, note that the usual LS-SVM-algorithm can be interpreted as special case of 
both the VP-SVM- and the RC-SVM-algorithm using one working set.

The experimental results for the three real data sets are summarized in Tables 
\ref{tab::experiments::covtype_ls_vp} to \ref{tab::experiments::ijcnn1}. These tables 
as well as Tables \ref{tab::experiments::art_data_1_2} to 
\ref{tab::experiments::art_data_5_2}, containing the results for the experiments on the
artificial data sets, can be found in the Appendix.
In addition to the average run times of the training and test phases, these tables reflect 
inter alia the average test errors of the empirical SVM solutions. Additionally, 
the $L_2$-errors of the empirical SVM solutions, i.e.~the value of
\begin{align*}
 \sqrt{\frac{1}{n_\text{test}} \sum_{i=1}^{n_\text{test}}
 \left(\fcl_{\D,\lb,\g}({x_\text{test}}_i)-\fB({x_\text{test}}_i)\right)^2}\,,
\end{align*}
is determined for the artificial data sets.
Moreover, note that some of the result tables are incomplete for 
very large real-world training data sets. In these cases, the kernel 
matrix, whose size depends on the training set size, did not fit into the 
RAM of the used computer and, thus, these experiments were left out. 

\subsection{Experiments on Real-World Data}

In this subsection, we adress questions (1), (2), and (3) by
examining the results for the real-world data sets \textsc{covtype}, \textsc{cod-rna}, and 
\textsc{ijcnn1}, which are composed in Figures 
\ref{fig::experiments::covtype}--\ref{fig::experiments::ijcnn1} and Tables 
\ref{tab::experiments::covtype_ls_vp}--\ref{tab::experiments::ijcnn1}.

\subsubsection{Comparison of VP-SVMs Using Different Radii}\label{subsubsec::exp::VP-SVM-results}

In the following, we focus on the VP-SVMs using four different radii for the 
various real-world data sets, where 
the experimental results are summarized in Tables 
\ref{tab::experiments::covtype_ls_vp}--\ref{tab::experiments::ijcnn1} 
as well as in Subfigures 
\subref{fig::experiments::covtype_vp_time}--\subref{fig::experiments::covtype_vp_error}
of Figures \ref{fig::experiments::covtype}--\ref{fig::experiments::ijcnn1}.
Examining the achieved training times for each data set type, we observe that, for increasing 
training set sizes, the radius that leads to the shortest training time
typically decreases. More precisely, for the real data sets with sample size 
$n_\text{train}>10\,000$, the VP-SVMs using the smallest radius 
always train fastest, while for the data sets with $n_\text{train}\leq 10\,000$, we can not 
make a uniform statement.
Clearly, this finding is not surprising, since an SVM for a small data set trains considerably faster 
than an SVM for a large data set, such that splitting the large data set and running an 
SVM for each of the small data sets may altogether still be faster. Recall additionally the 
considerations in terms of the computational cost made in Section \ref{VP-SVM illustration}.

Let us now consider the VP-SVM results in terms of the realized test errors. As expected, 
for the real-world data sets, the test errors achieved by VP-SVMs with fixed radii 
decrease with increasing sample size of the used data sets except twice. In addition, for
the real data sets \textsc{covtype} and \textsc{cod-rna}, the test errors decrease for 
increasing radii, cf.~Figures 
\ref{fig::experiments::covtype_vp_error} and \ref{fig::experiments::cod-rna_vp_error}.
Here, however, the test errors achieved for the various radii get close to each other with 
increasing training sample size. The same behavior of the test errors appears for the 
\textsc{ijcnn1}
data set, though, for $n_\text{train}\geq5\,000$, both intermediate radii yield even smaller 
empirical risks than the largest radius, see Figure \ref{fig::experiments::ijcnn1_vp_error}. 
In consequence, it is not straightforward to draw any conclusion on the relation between 
radius and test error. Nonetheless, we can say that VP-SVMs using small radii
enjoy test errors that are never significantly larger and somethimes even smaller than those of 
VP-SVMs using the largest of the applied radii.

Besides, Tables \ref{tab::experiments::covtype_ls_vp} and 
\ref{tab::experiments::cod-rna} or Figures 
\ref{fig::experiments::covtype_vp_time_all},\subref{fig::experiments::covtype_vp_error} and 
\ref{fig::experiments::cod-rna_vp_time_all},\subref{fig::experiments::cod-rna_vp_error}
contain an additional finding. For large data sets, namely for the \textsc{covtype} data set 
of size $500\,000$ or for the \textsc{cod-rna} data sets of size $250\,000$ and $400\,000$, 
the VP-SVMs with large radii did not yield any solution, since they failed due to the 
technical requirements caused by the used computer. More precisely, in these cases, there was 
at least one working set such that its kernel matrix did not fit into the RAM any more. 
Fortunately, the working sets of VP-SVMs using smaller radii were small enough such that we 
still received an outcome. What is more, these VP-SVMs yielded a 
better empirical risk in partially less training time compared to VP-SVMs with large radii and 
training sample sizes that still allowed a successful performance.  
That is, using a small radius for the VP-SVM and a training set that is oversized for VP-SVMs 
with a larger radius reduces the test error.
More precisely, a large training set is crucial for a small empirical risk, where 
the possibly arising computational restrictions can be eluded by a VP-SVM with an 
appropriate radius.

All in all, localized SVMs using some small radius lead in substantially less training time 
to either negligble worse or even better test errors than VP-SVMs with large radii, 
if the training sample size is adequate, i.e.~$n_\text{train}\geq5\,000$.
In addition, the real data sets with a large sample size demonstrate that VP-SVMs with small 
radii are able to conquer the technical restrictions caused by the used computer
and thus yield a better empirical risk than VP-SVMs with bigger radii can attain at all.

\subsubsection{Comparing VP-SVMs with Global LS-SVMs}\label{subsubsec::exp::VP-LS-SVMcomparison}

In the following, we compare the results of the VP-SVM using different radii to 
the standard LS-SVM. 
For the real-world data sets \textsc{cod-rna} and \textsc{ijcnn1}, the VP-SVMs, based on 
the largest of the applied radii, use only one
working set. Thus, they coincide with the standard LS-SVM modulo different values generated by the random number generator.
To verify this fact, we compare 
for the real data sets \textsc{cod-rna} and \textsc{ijcnn1} the results of the 
VP-SVMs using one working set to the results of the standard LS-SVMs, see
Tables \ref{tab::experiments::cod-rna} and \ref{tab::experiments::ijcnn1}. Here, 
we note that the LS-SVM test errors typically decrease with increasing training sample size.
The same holds for the VP-SVMs using one working set. 
Moreover, the latter VP-SVM and the LS-SVMs perform equally well in terms of training time and 
empirical risk, however, for $n_\text{train}\geq25\,000$ the VP-SVMs train slower.


In practice, a crucial problem is caused by the run time required by an algorithm. Hence,
for each data set type, we compare hereafter the LS-SVM to the VP-SVM that trains fastest 
for the largest training data set. 
The required average training times and the average test errors of these SVMs are 
illustrated in Subfigures 
\subref{fig::experiments::covtype_time}--\subref{fig::experiments::covtype_error}
of Figures \ref{fig::experiments::covtype}--\ref{fig::experiments::ijcnn1}.
First, we notice that the selected VP-SVM uses the smallest of 
the applied radii for each data set type. Besides,
the LS-SVM's test errors are lower than those of the VP-SVMs. However, with increasing 
training set size the VP-SVM's test errors get close to the ones of the LS-SVM. Moreover, for 
the \textsc{ijcnn1} data set of size $100\,000$, their empirical risks even coincide, 
cf.~Figure \ref{fig::experiments::ijcnn1_error}. Besides, the VP-SVMs train considerably faster 
than the LS-SVMs. In particular, for $n_\text{train}= 100\,000$ the VP-SVMs require at most 
$8.5\%$ of the LS-SVM's training times, see Figures \ref{fig::experiments::covtype_time_all}, 
\ref{fig::experiments::cod-rna_time_all}, and \ref{fig::experiments::ijcnn1_time_all}. 
Finally, recall that, for data sets of size 
$n_\text{train}\geq250\,000$, the LS-SVM problem is infeasible with our computer, just like
the VP-SVMs using the largest of the applied radii.
In contrast, for $n_\text{train}\geq250\,000$, VP-SVMs using small radii usually train 
considerably faster and achieve lower test errors than the LS-SVMs for 
$n_\text{train}= 100\,000$, cf.~Figures 
\ref{fig::experiments::covtype_time_all}--\subref{fig::experiments::covtype_error} and 
\ref{fig::experiments::cod-rna_time_all}--\subref{fig::experiments::cod-rna_error}.

Concluding, we have seen that the application of a VP-SVM using a 
small radius instead of the standard LS-SVM reduces the run time considerably entailing
at most a negligible worsening or even an improvement of the test errors. 
Moreover, applying VP-SVMs with sufficiently small radii enables us to use large data sets 
and, thus, to elude the computational restrictions to sufficiently small data sets.
As a result, handling really large data sets with the help of suitable VP-SVMs can lead to significantly 
improved test errors compared to an  LS-SVM setting with memory constraints.

\subsubsection{Comparison of VP-SVMs with RC-SVMs}\label{subsubsec::exp::RC-SVM-results}

\begin{figure}[p]
\begin{scriptsize}
\vspace{-1cm}
\subfigure[Average training time of the various RC-SVMs for $n_\text{train}\leq10\,000$\label{fig::experiments::covtype_rc_time}]{\includegraphics[angle=270, scale=0.26]{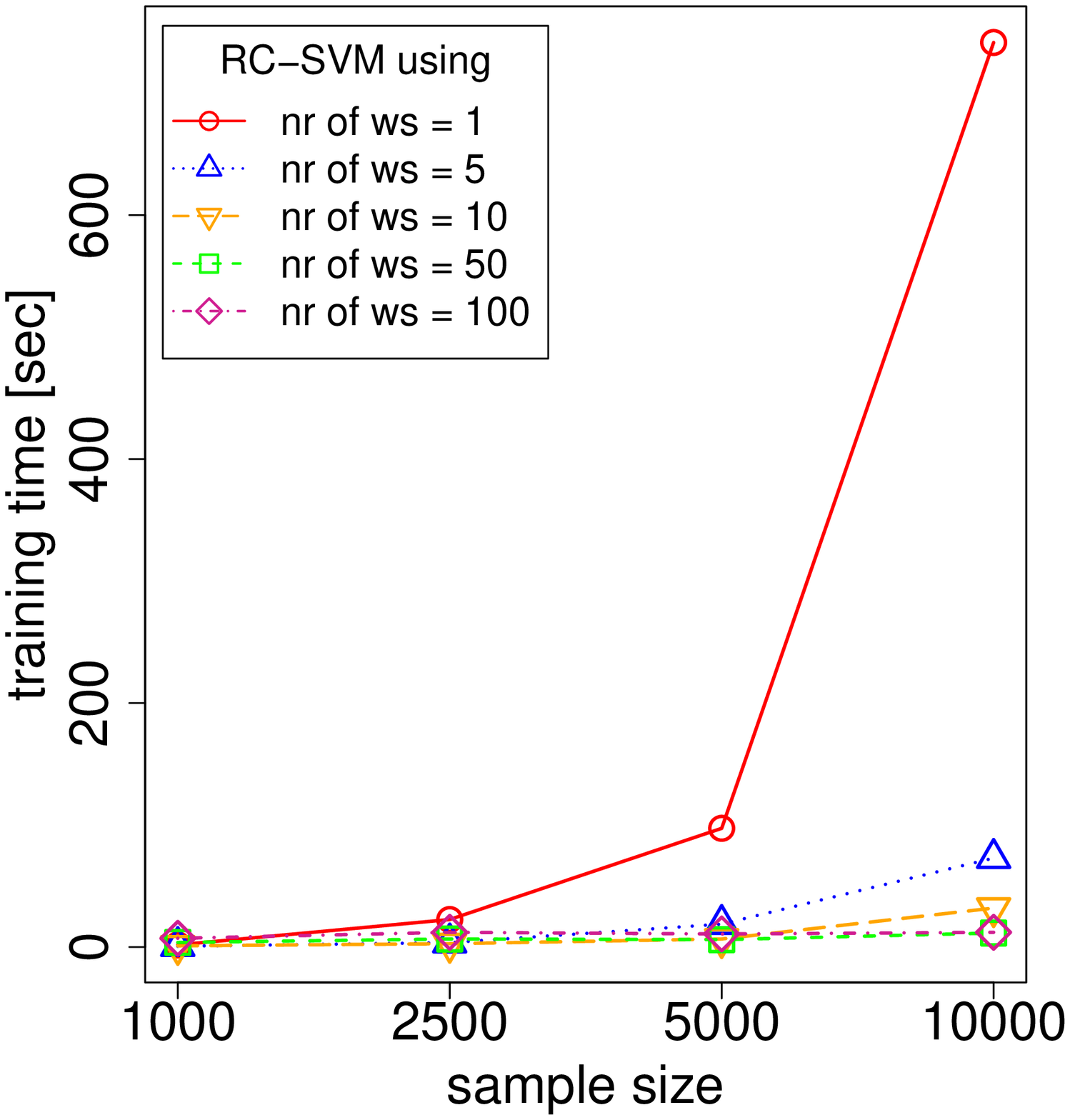}}
\hfill\subfigure[Average training time of the various RC-SVMs for $n_\text{train}\geq5\,000$\label{fig::experiments::covtype_rc_time_all}]{\includegraphics[angle=270, scale=0.26]{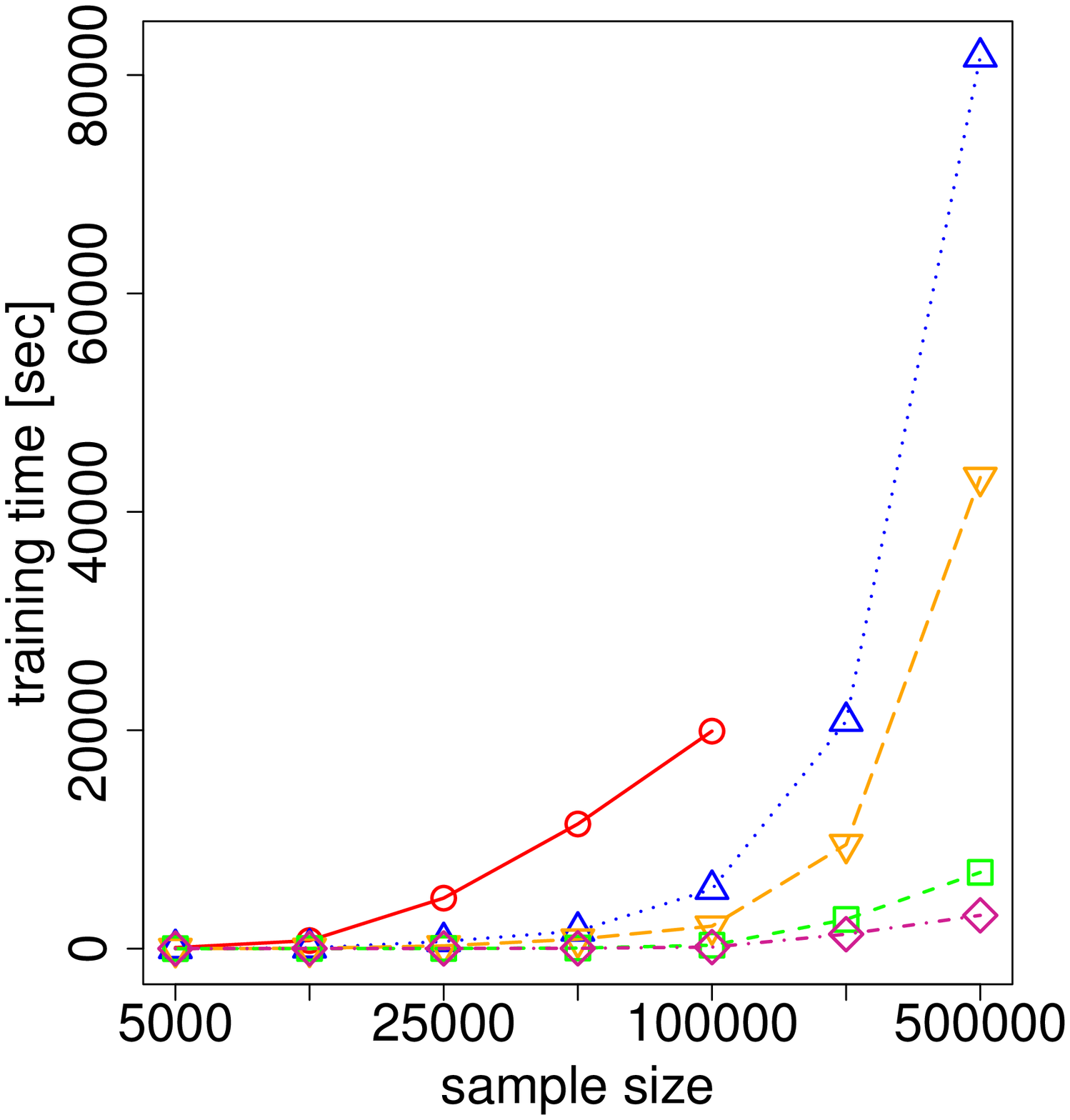}}
\hfill\subfigure[Average empirical risk of the various RC-SVMs\label{fig::experiments::covtype_rc_error}]{\includegraphics[angle=270, scale=0.26]{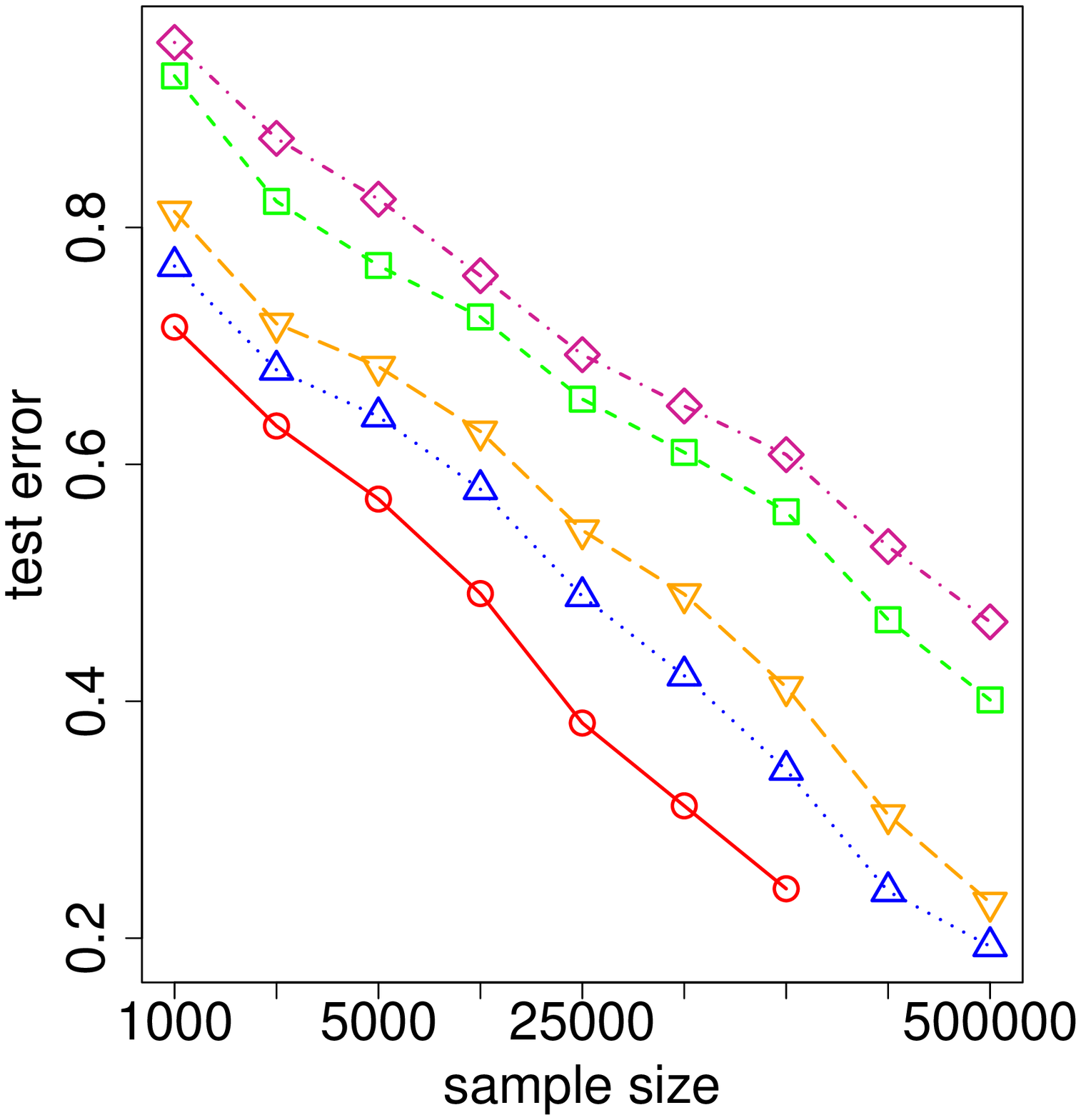}}\vspace{-0.7cm}\\
\subfigure[Average training time of the various VP-SVMs for $n_\text{train}\leq10\,000$\label{fig::experiments::covtype_vp_time}]{\includegraphics[angle=270, scale=0.26]{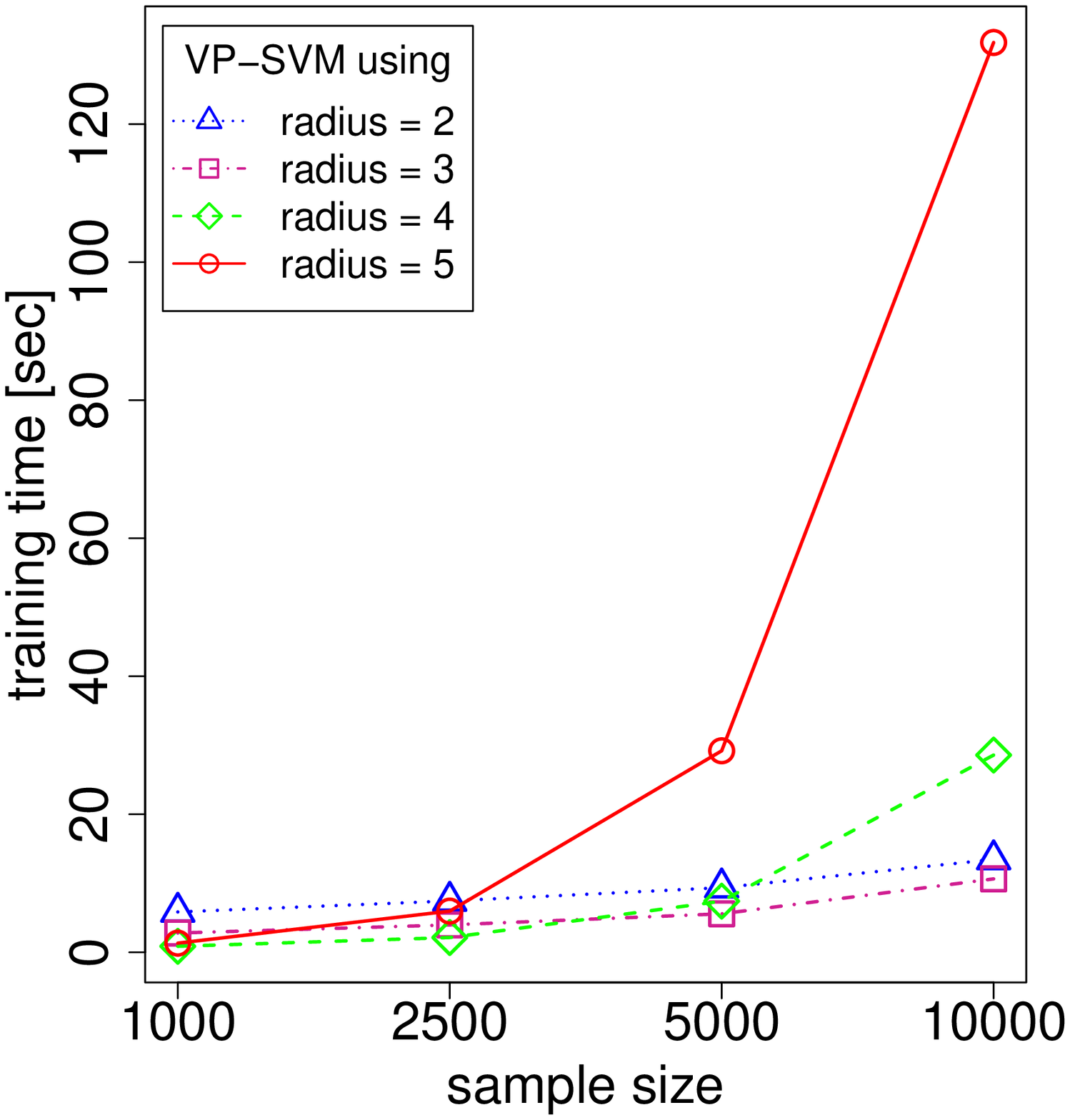}}
\hfill\subfigure[Average training time of the various VP-SVMs for $n_\text{train}\geq5\,000$\label{fig::experiments::covtype_vp_time_all}]{\includegraphics[angle=270, scale=0.26]{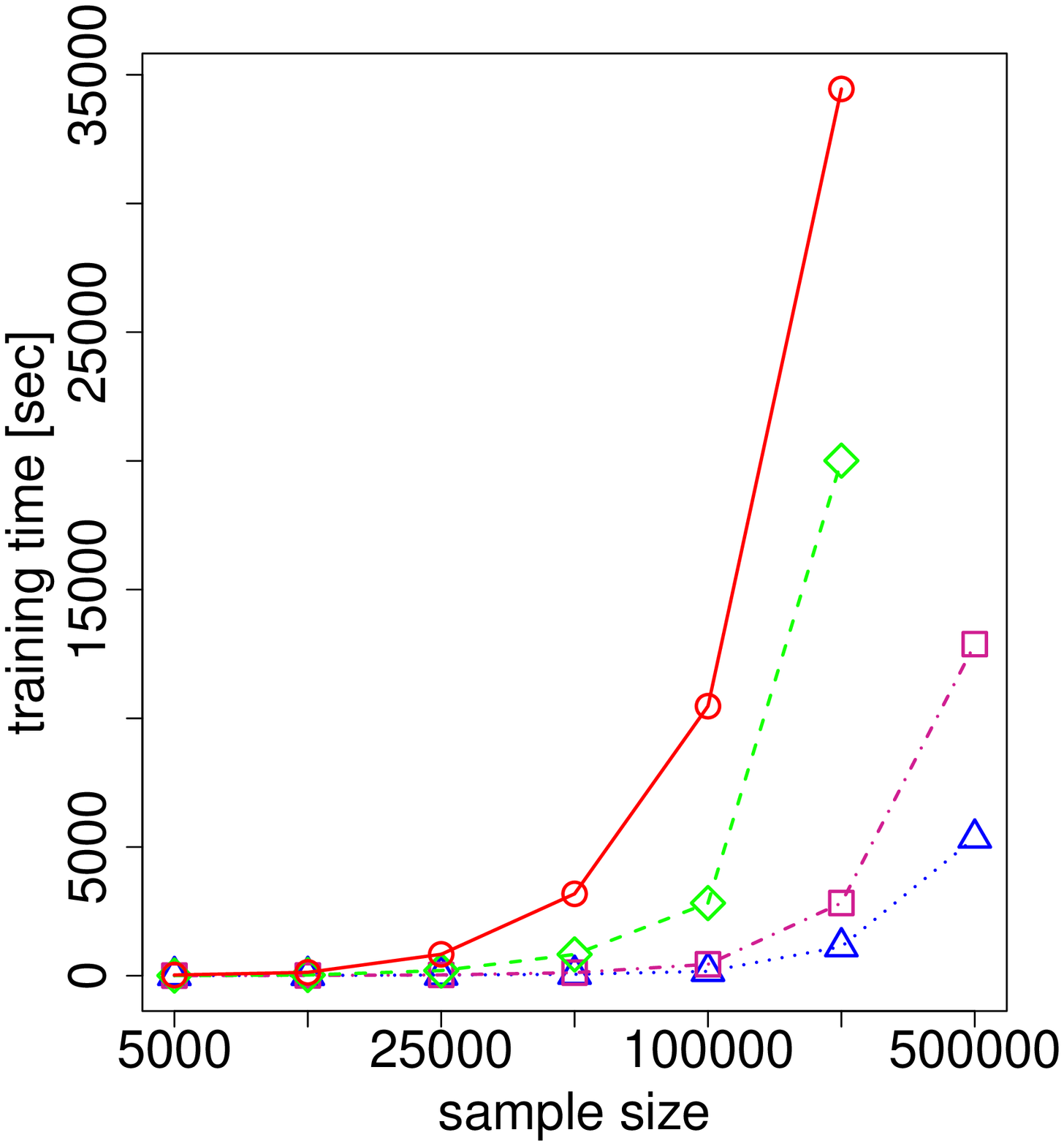}}
\hfill\subfigure[Average empirical risk of the various VP-SVMs\label{fig::experiments::covtype_vp_error}]{\includegraphics[angle=270, scale=0.26]{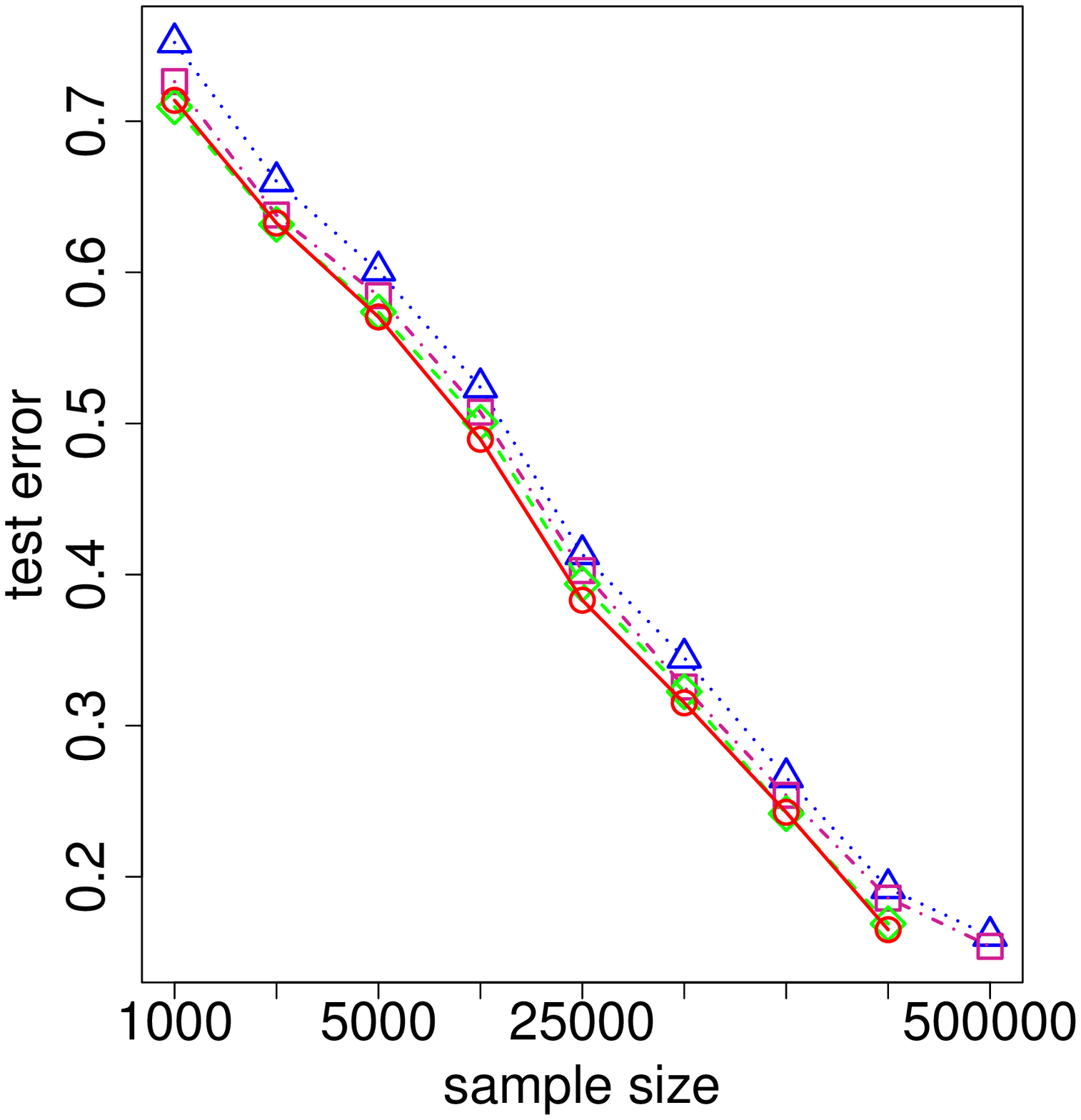}}\vspace{-0.7cm}\\
\subfigure[Average training time of \mbox{LS-,} VP-, and RC-SVMs for $n_\text{train}\leq10\,000$\label{fig::experiments::covtype_time}]{\includegraphics[angle=270, scale=0.26]{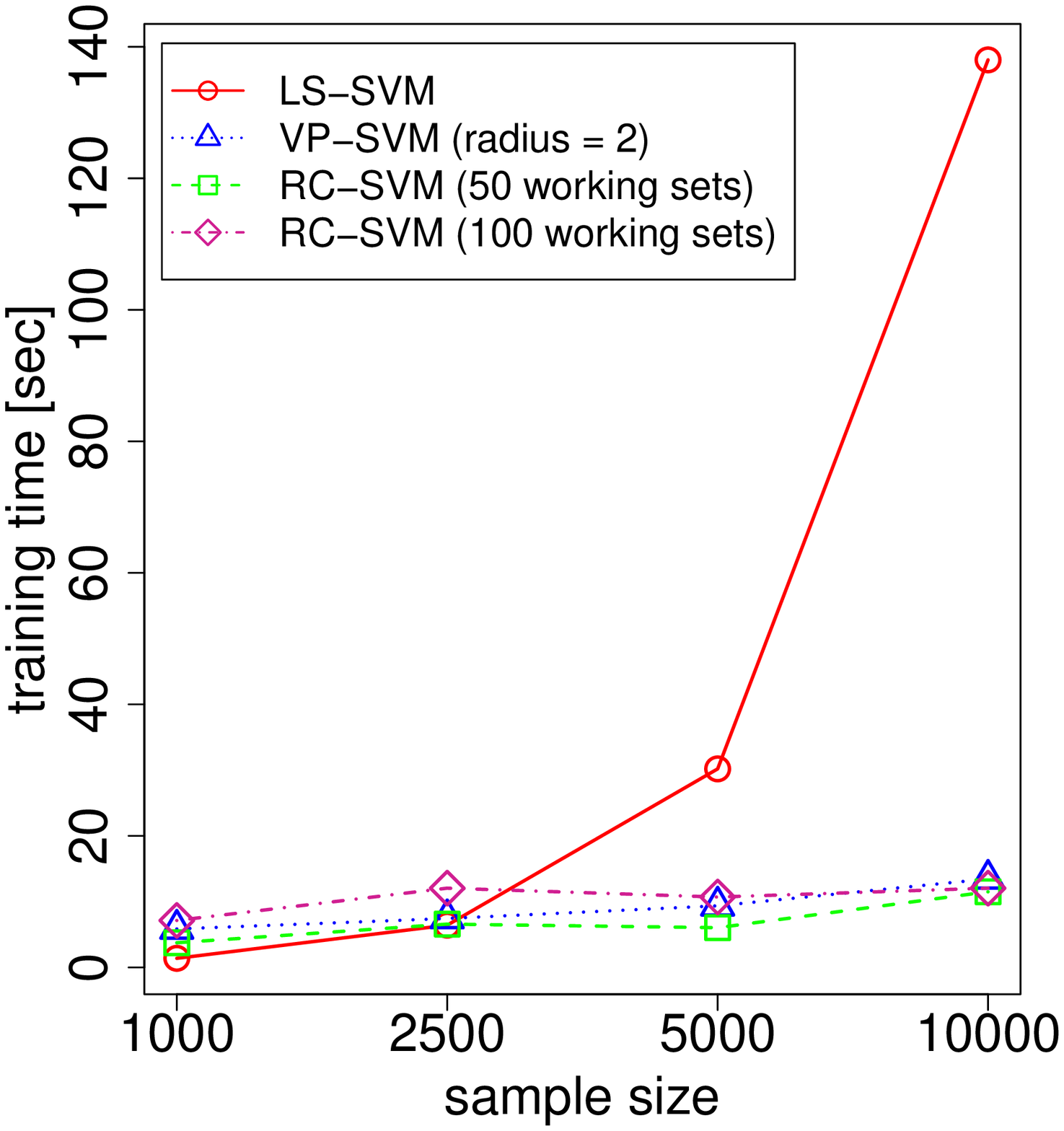}}
\hfill\subfigure[Average training time of \mbox{LS-,} VP-, and RC-SVMs for $n_\text{train}\geq5\,000$\label{fig::experiments::covtype_time_all}]{\includegraphics[angle=270, scale=0.26]{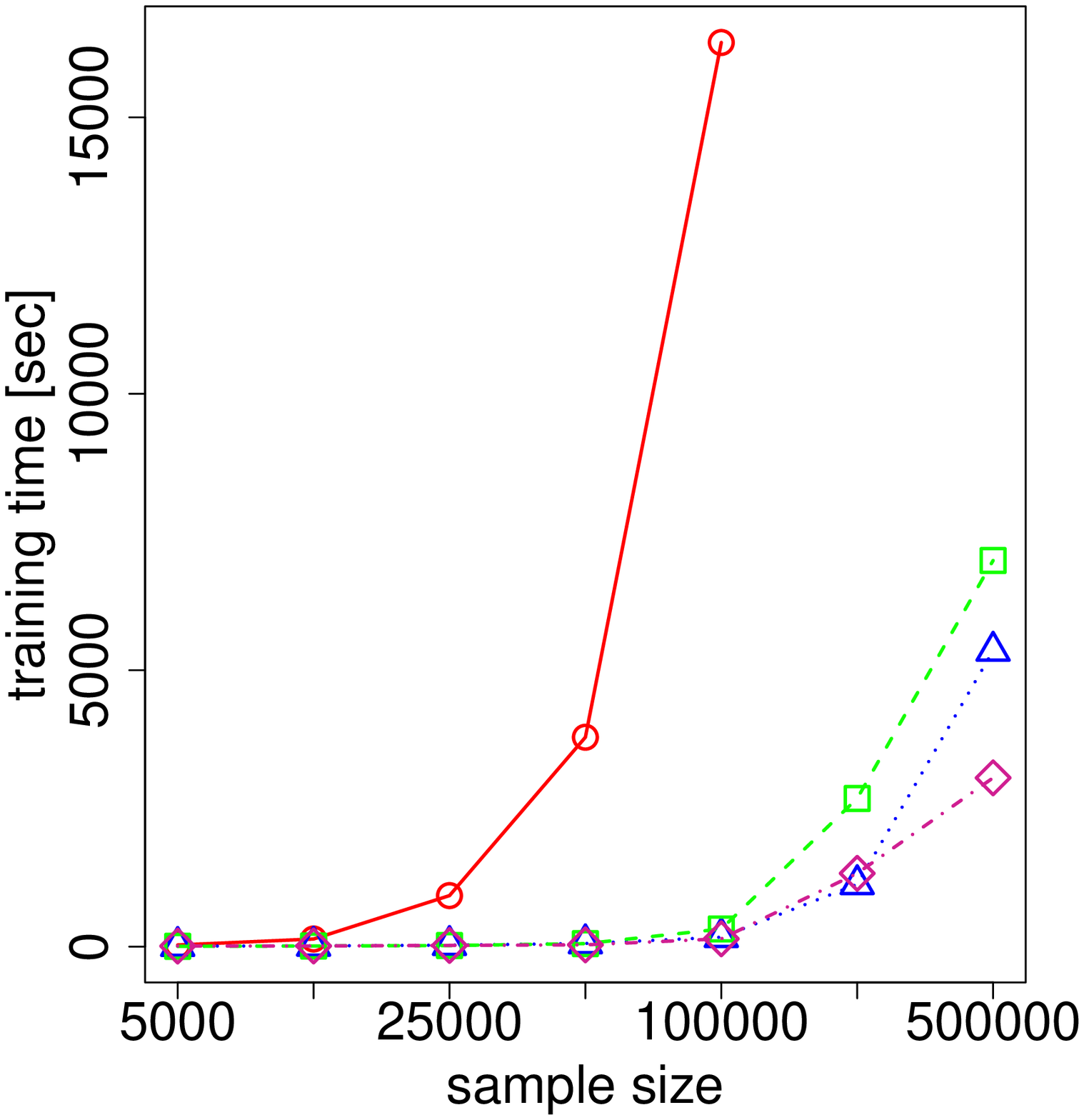}}
\hfill\subfigure[Average empirical risk of \mbox{LS-,} VP-, and RC-SVMs\label{fig::experiments::covtype_error}]{\includegraphics[angle=270, scale=0.26]{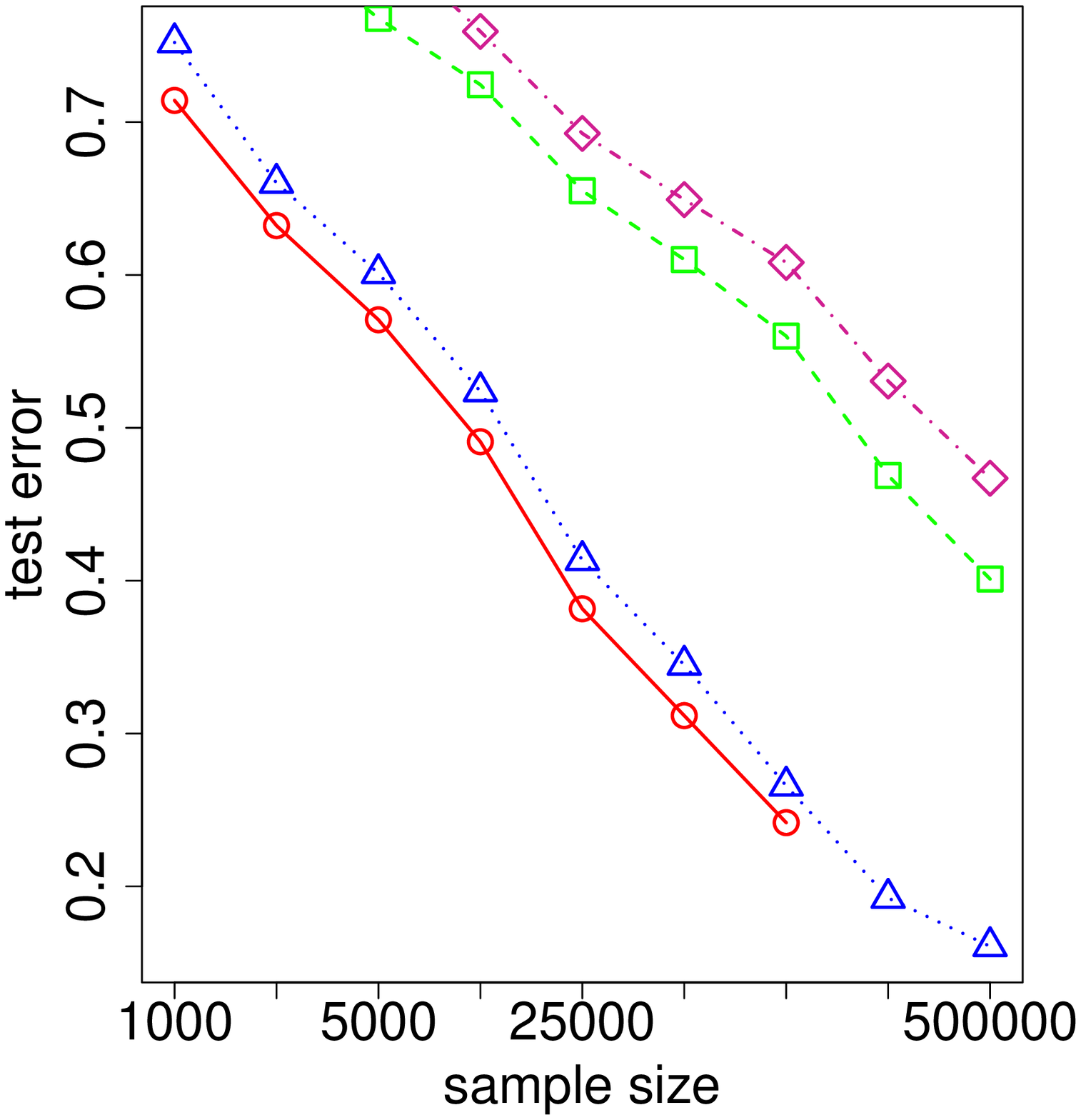}}\vspace{-0.3cm}
\caption{Average training time and test error of LS-, VP-, and RC-SVMs for the 
real-world data \textsc{covtype} depending on the 
training set size $n_\text{train}=1\,000,\ldots,500\,000$. Subfigures 
\subref{fig::experiments::covtype_rc_time}--\subref{fig::experiments::covtype_rc_error} 
show the results for RC-SVMs using different numbers of working sets and Subfigures 
\subref{fig::experiments::covtype_vp_time}--\subref{fig::experiments::covtype_vp_error} 
illustrate the results for VP-SVMs using various radii. At the bottom, Subfigures 
\subref{fig::experiments::covtype_time}--\subref{fig::experiments::covtype_error} contain the 
average training times and the average test errors of the LS-SVM, one VP-SVM and 
two RC-SVMs. Here, the VP-SVM is the one which trains fastest for $n_\text{train}=500\,000$
and the two RC-SVMs are those which achieve for $n_\text{train}=500\,000$ roughly the same 
training time as the chosen VP-SVM. Here, note that, for $n_\text{train}=10\,000$, the RC-SVM 
using one working set trains substantially slower than the LS-SVM, even though this RC-SVM 
is basically an LS-SVM. As a reason for this phenomenon, we conjecture that the used compute 
server was busy because of other influences.}
\label{fig::experiments::covtype}
\end{scriptsize}
\end{figure}

\begin{figure}[p]
\begin{scriptsize}\vspace{-0.6cm}
\subfigure[Average training time of the various RC-SVMs for $n_\text{train}\leq10\,000$\label{fig::experiments::cod-rna_rc_time}]{\includegraphics[angle=270,  scale=0.26]{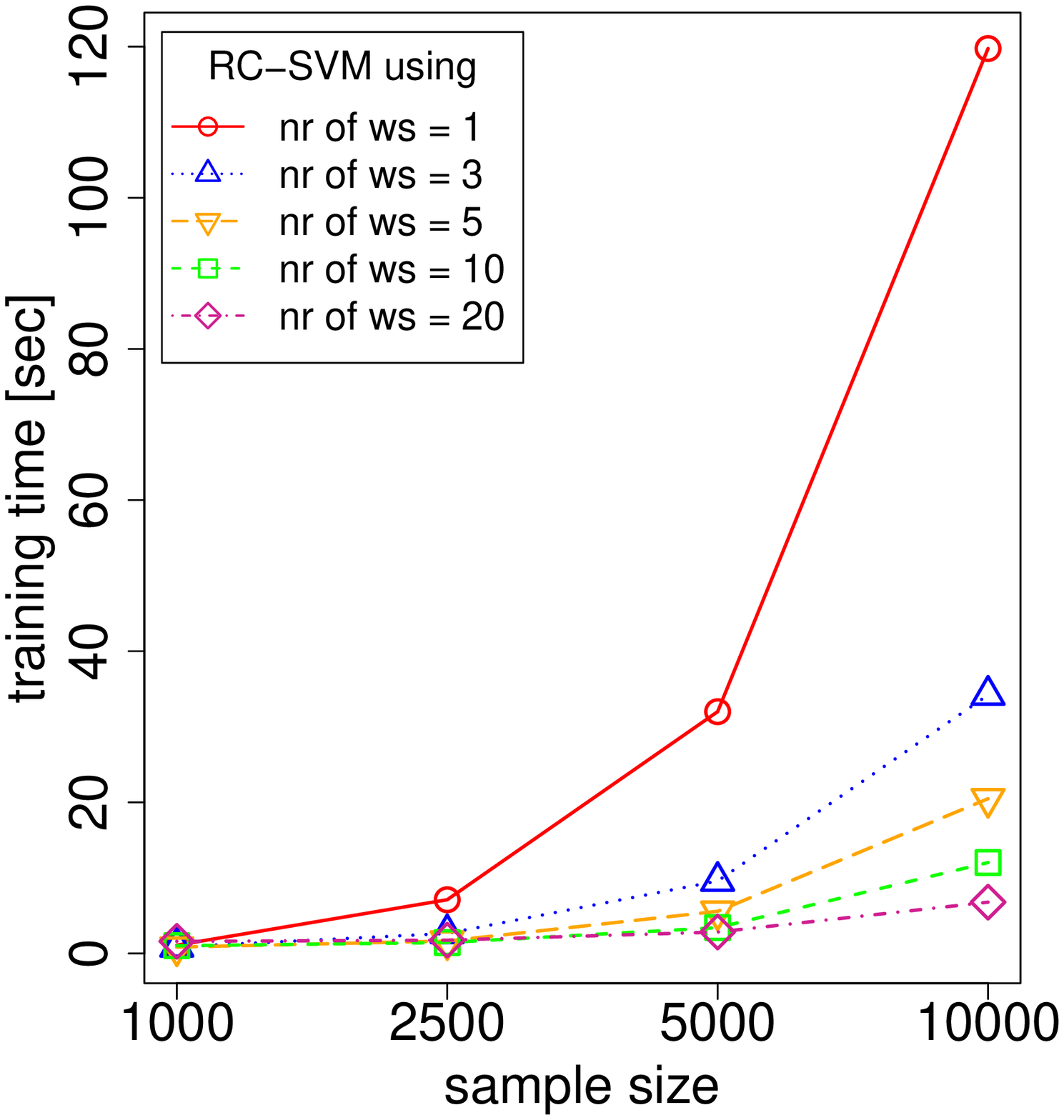}}
\hfill\subfigure[Average training time of the various RC-SVMs for $n_\text{train}\geq5\,000$\label{fig::experiments::cod-rna_rc_time_all}]{\includegraphics[angle=270,  scale=0.26]{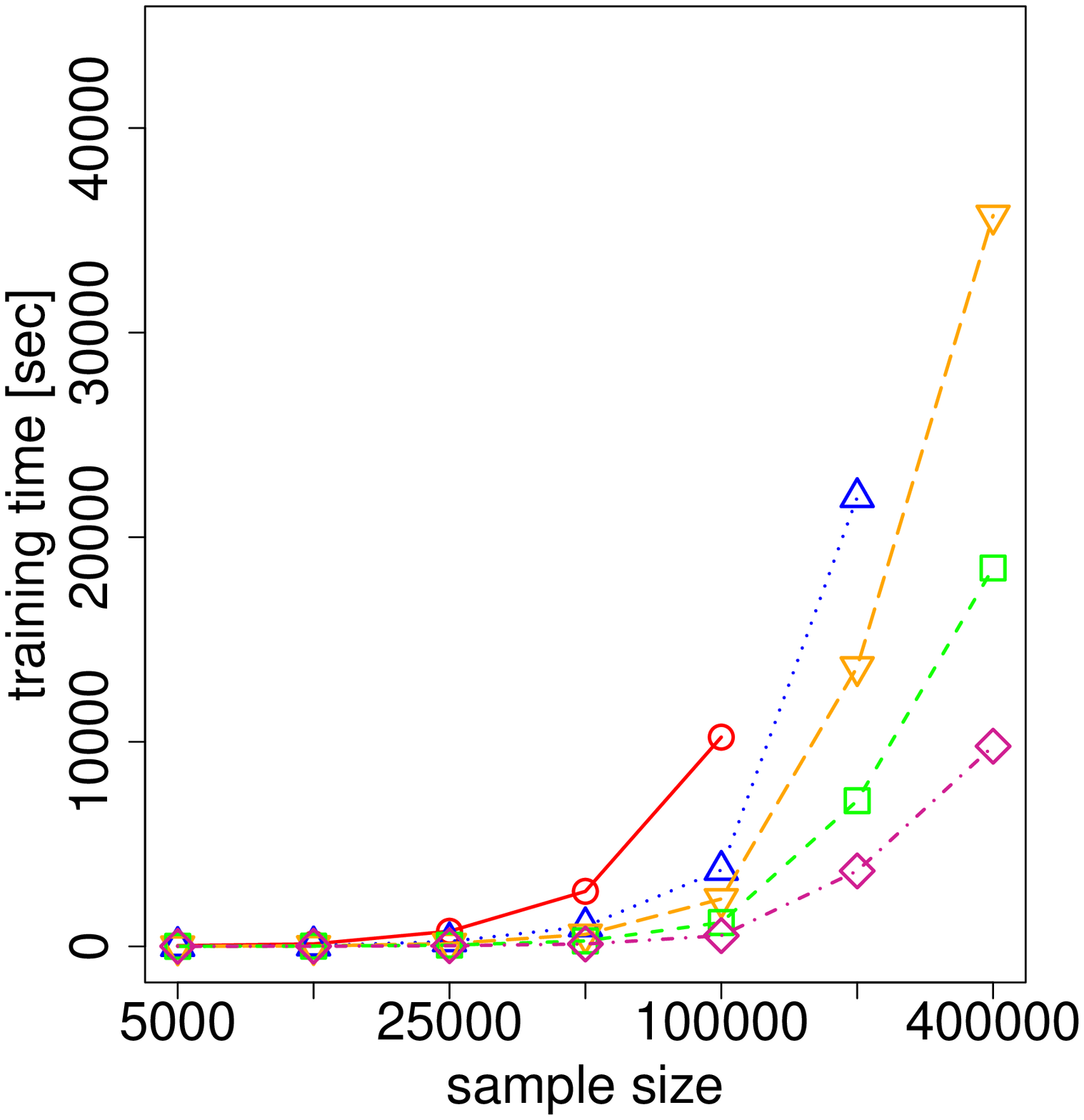}}
\hfill\subfigure[Average empirical risk of the various RC-SVMs\label{fig::experiments::cod-rna_rc_error}]{\includegraphics[angle=270,  scale=0.26]{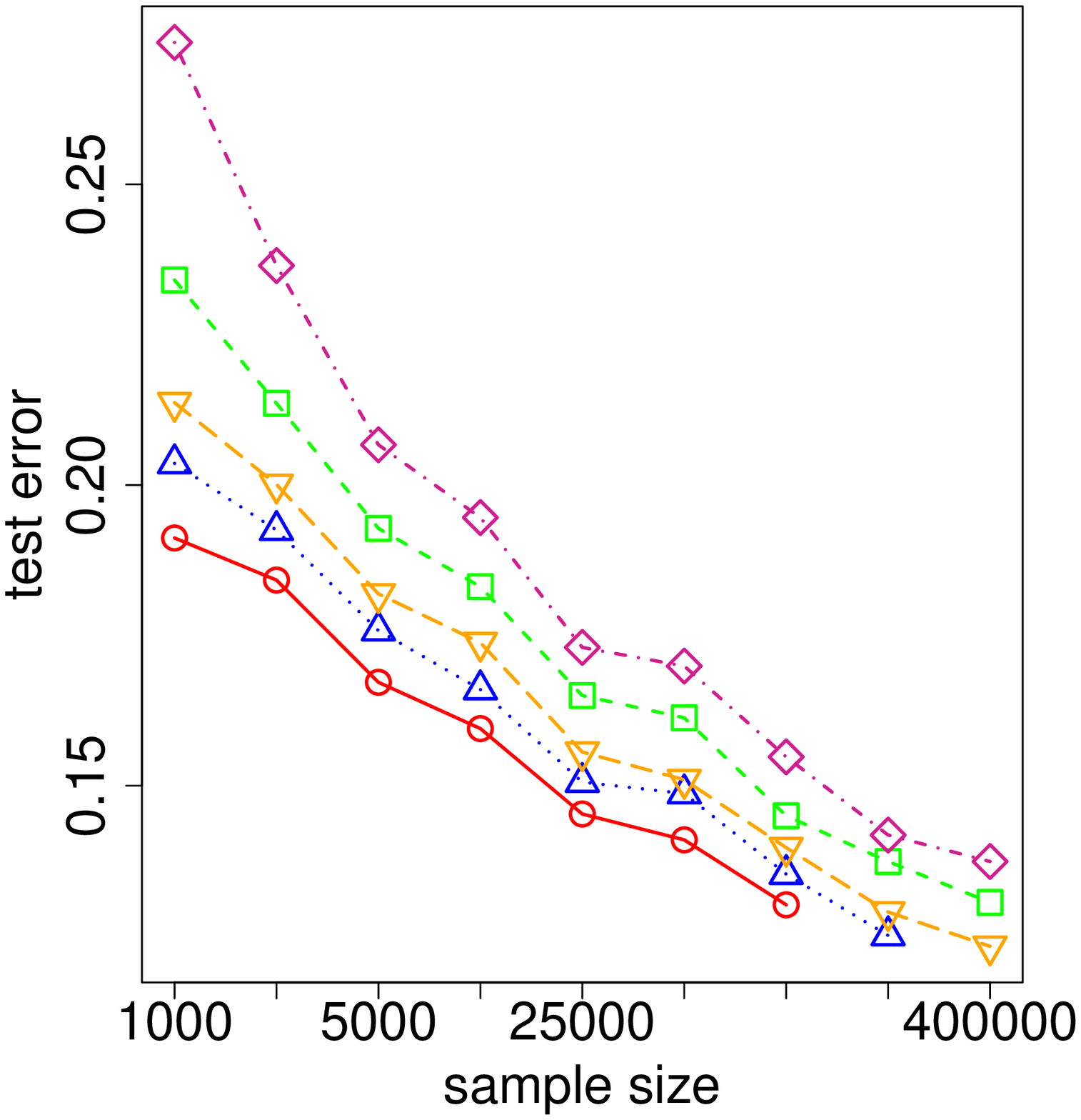}}\vspace{-0.6cm}\\
\subfigure[Average training time of the various VP-SVMs for $n_\text{train}\leq10\,000$\label{fig::experiments::cod-rna_vp_time}]{\includegraphics[angle=270,  scale=0.26]{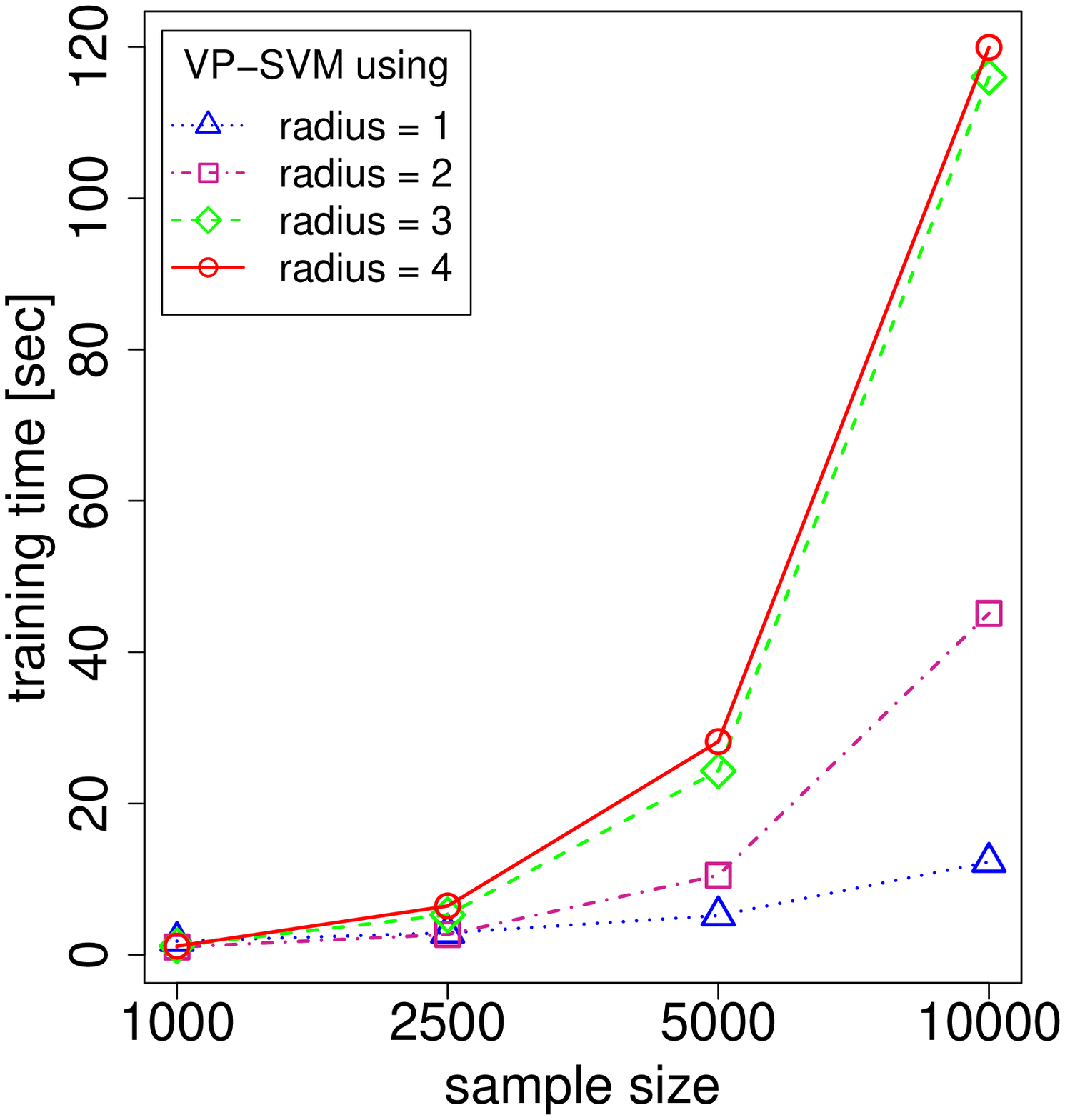}}
\hfill\subfigure[Average training time of the various VP-SVMs for $n_\text{train}\geq5\,000$\label{fig::experiments::cod-rna_vp_time_all}]{\includegraphics[angle=270,  scale=0.26]{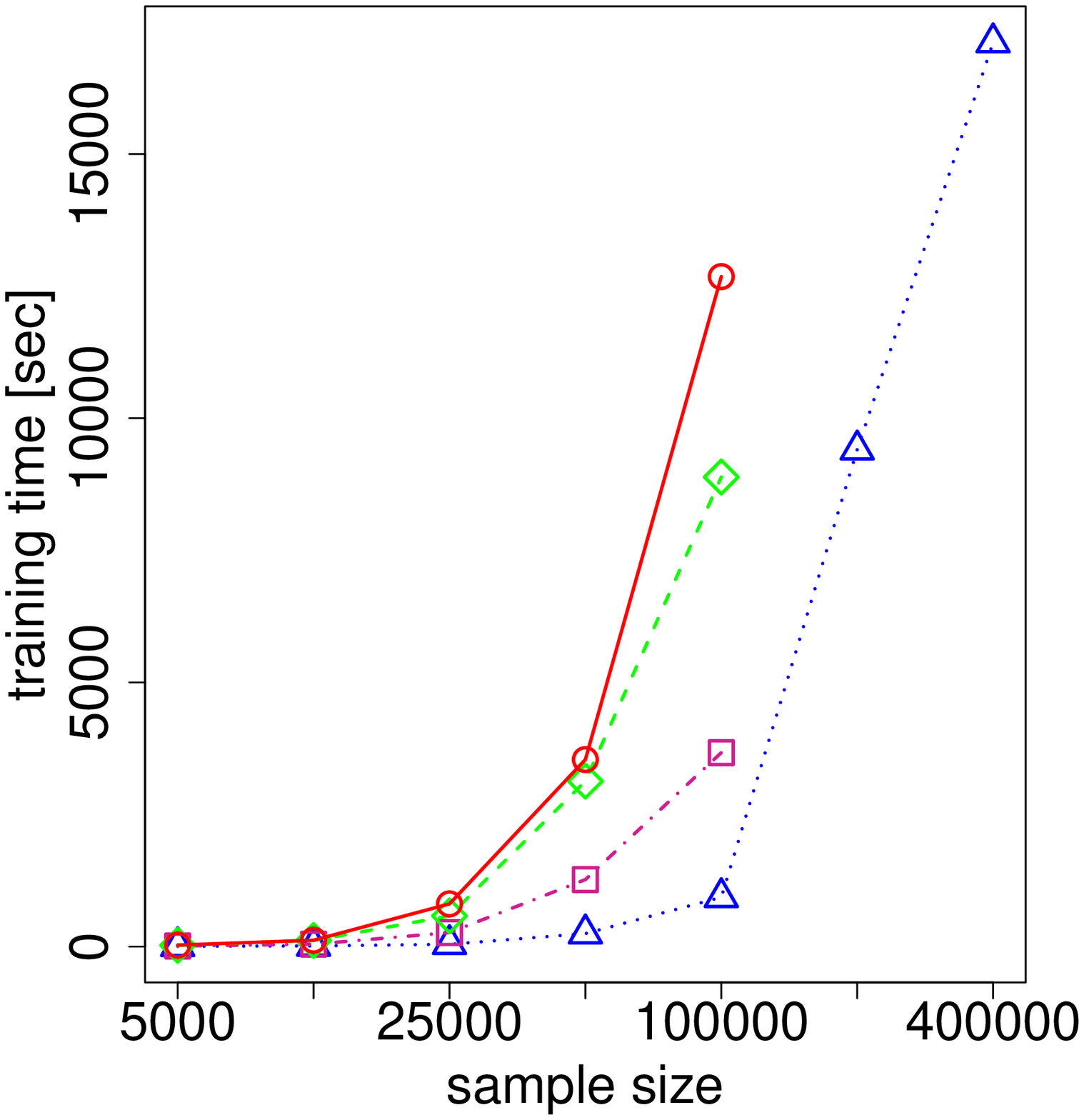}}
\hfill\subfigure[Average empirical risk of the various VP-SVMs\label{fig::experiments::cod-rna_vp_error}]{\includegraphics[angle=270,  scale=0.26]{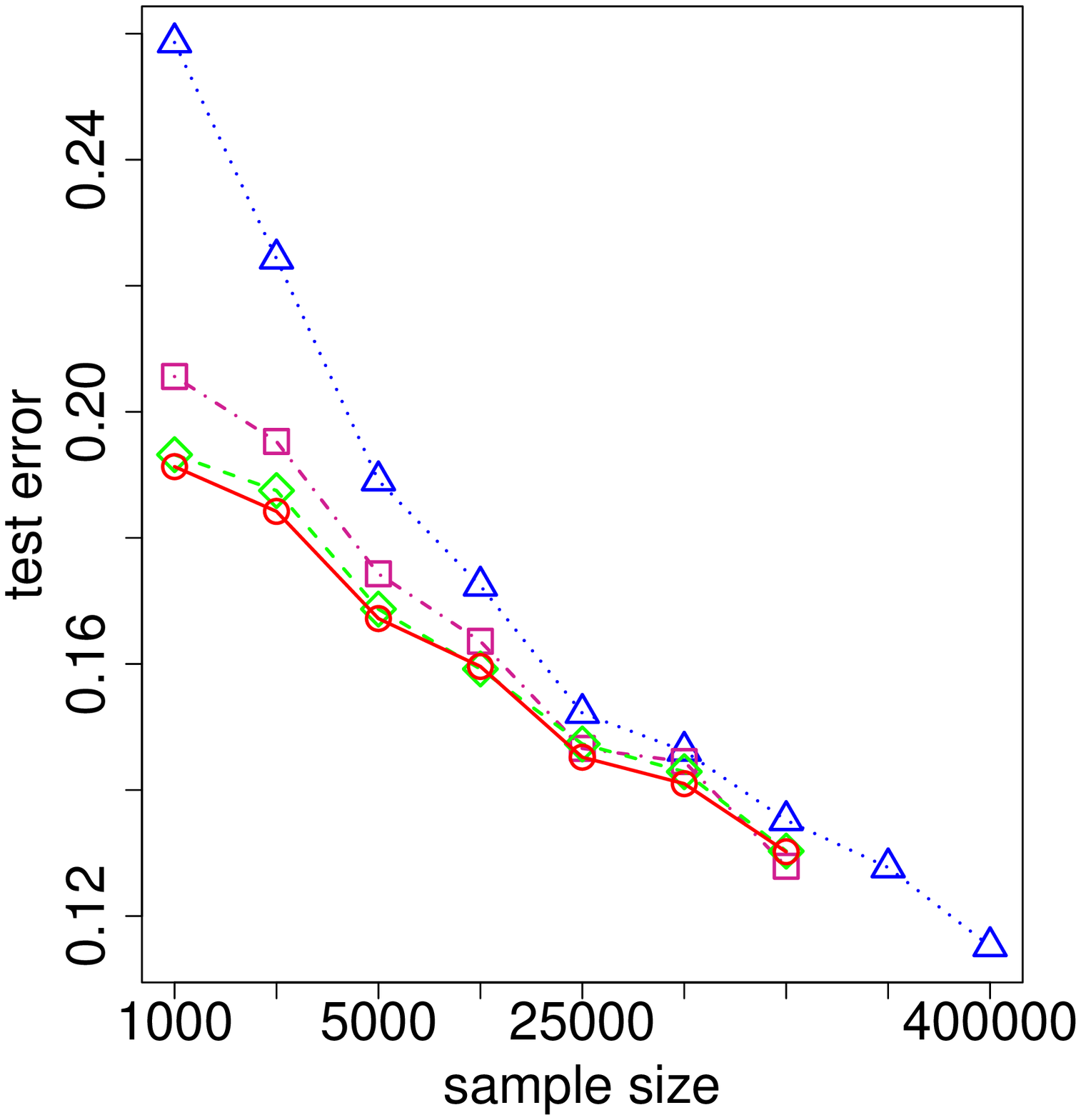}}\vspace{-0.6cm}\\
\subfigure[Average training time of \mbox{LS-,} VP-, and RC-SVMs for $n_\text{train}\leq10\,000$\label{fig::experiments::cod-rna_time}]{\includegraphics[angle=270,  scale=0.26]{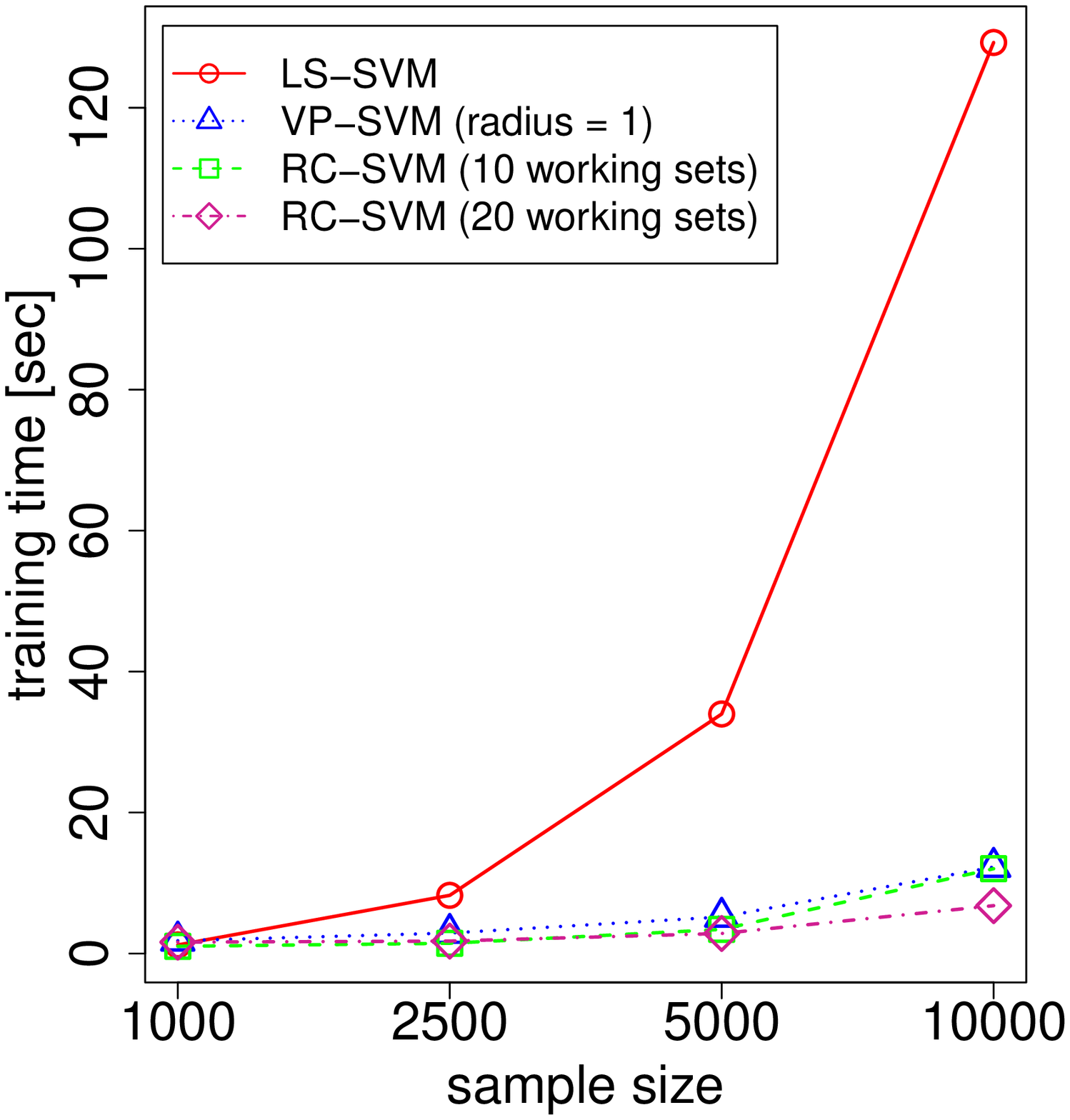}}
\hfill\subfigure[Average training time of \mbox{LS-,} VP-, and RC-SVMs for $n_\text{train}\geq5\,000$\label{fig::experiments::cod-rna_time_all}]{\includegraphics[angle=270,  scale=0.26]{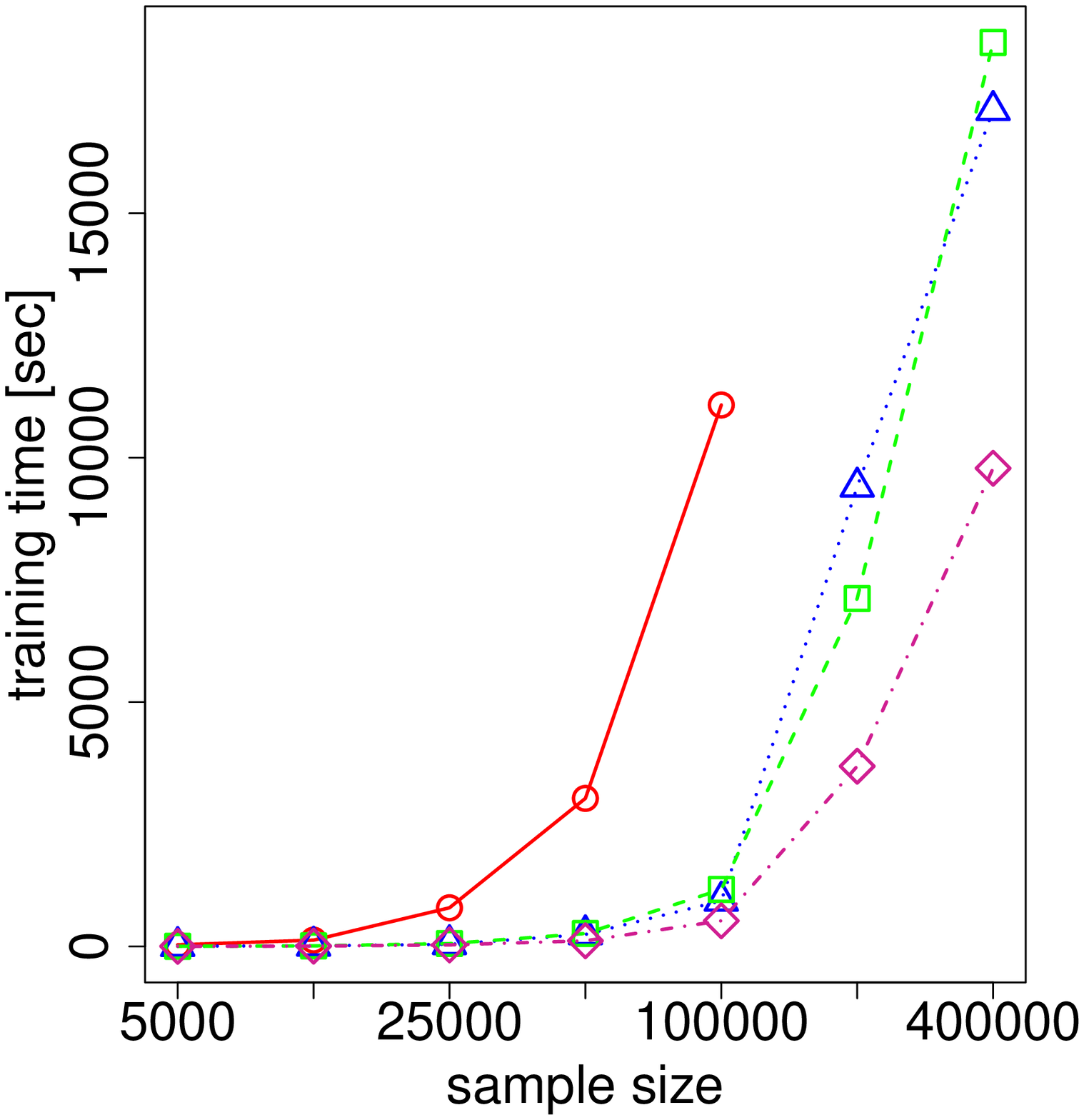}}
\hfill\subfigure[Average empirical risk of \mbox{LS-,} VP-, and RC-SVMs\label{fig::experiments::cod-rna_error}]{\includegraphics[angle=270,  scale=0.26]{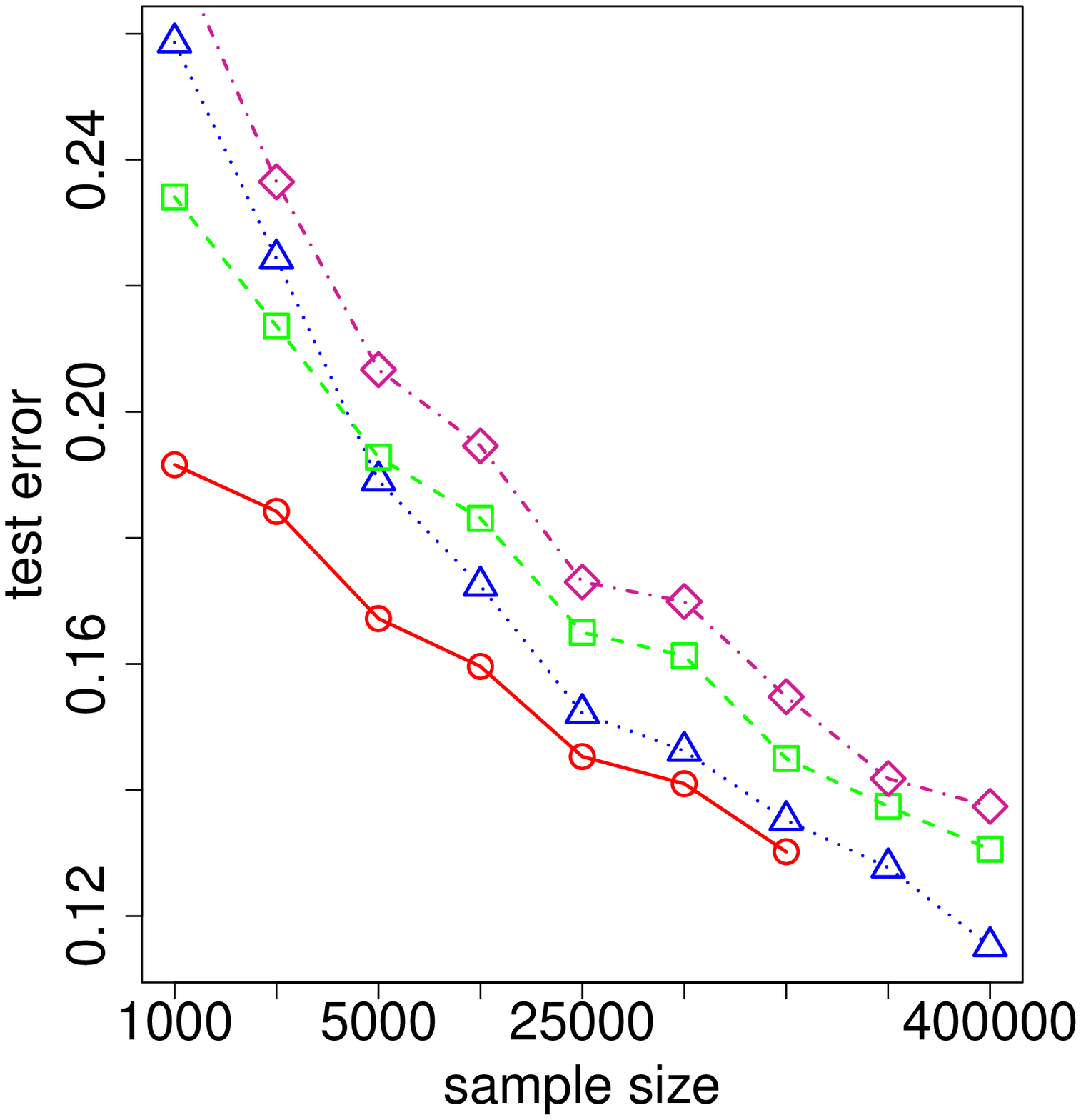}}\vspace{-0.2cm}
\caption{Average training time and test error of LS-, VP-, and RC-SVMs for the 
real-world data \textsc{cod-rna} depending on the 
training set size $n_\text{train}=1\,000,\ldots,400\,000$. Subfigures 
\subref{fig::experiments::cod-rna_rc_time}--\subref{fig::experiments::cod-rna_rc_error} 
show the results for RC-SVMs using different numbers of working sets and Subfigures 
\subref{fig::experiments::cod-rna_vp_time}--\subref{fig::experiments::cod-rna_vp_error} 
illustrate the results for VP-SVMs using various radii. At the bottom, Subfigures 
\subref{fig::experiments::cod-rna_time}--\subref{fig::experiments::cod-rna_error} contain the 
average training times and the average test errors of the LS-SVM, one VP-SVM and 
two RC-SVMs. Here, the VP-SVM is the one which trains fastest for $n_\text{train}=400\,000$
and the two RC-SVMs are those which achieve for $n_\text{train}=400\,000$ roughly the same 
training time as the chosen VP-SVM.}
\label{fig::experiments::cod-rna}
\end{scriptsize}
\end{figure}

\begin{figure}[p]
\begin{scriptsize}\vspace{-0.6cm}
\subfigure[Average training time of the various RC-SVMs for $n_\text{train}\leq10\,000$\label{fig::experiments::ijcnn1_rc_time}]{\includegraphics[angle=270,  scale=0.26]{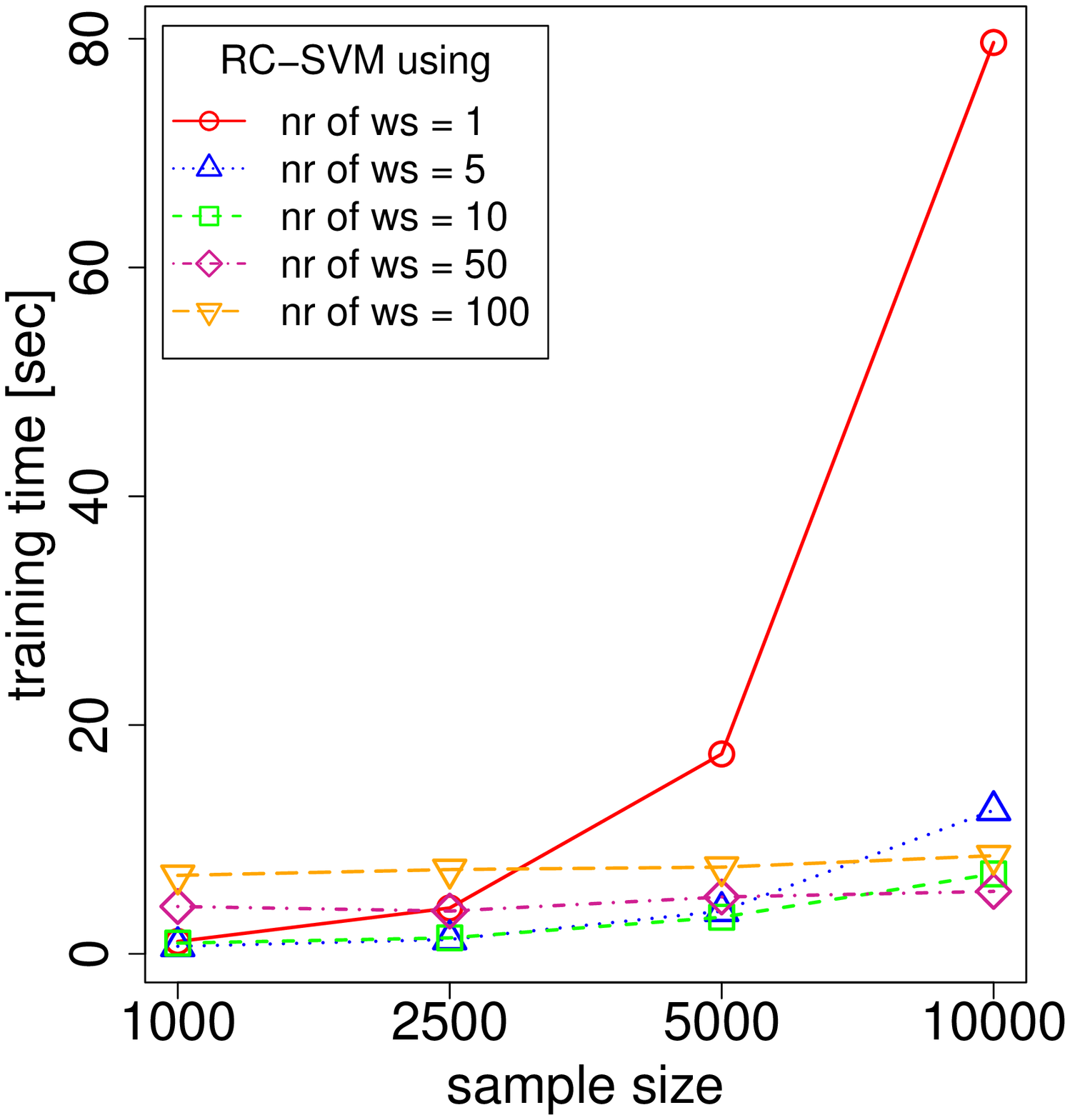}}
\hfill\subfigure[Average training time of the various RC-SVMs for $n_\text{train}\geq5\,000$\label{fig::experiments::ijcnn1_rc_time_all}]{\includegraphics[angle=270,  scale=0.26]{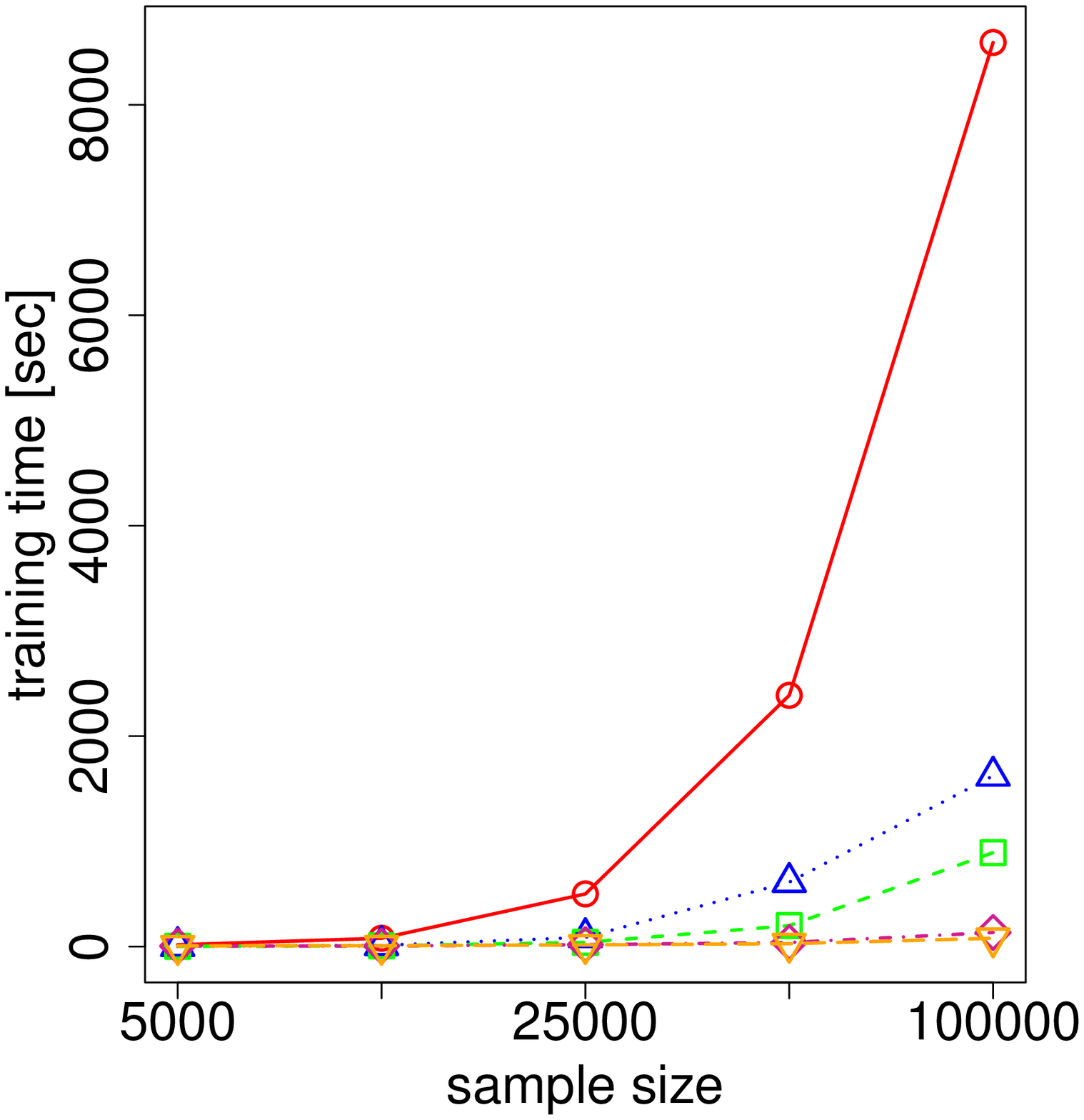}}
\hfill\subfigure[Average empirical risk of the various RC-SVMs\label{fig::experiments::ijcnn1_rc_error}]{\includegraphics[angle=270,  scale=0.26]{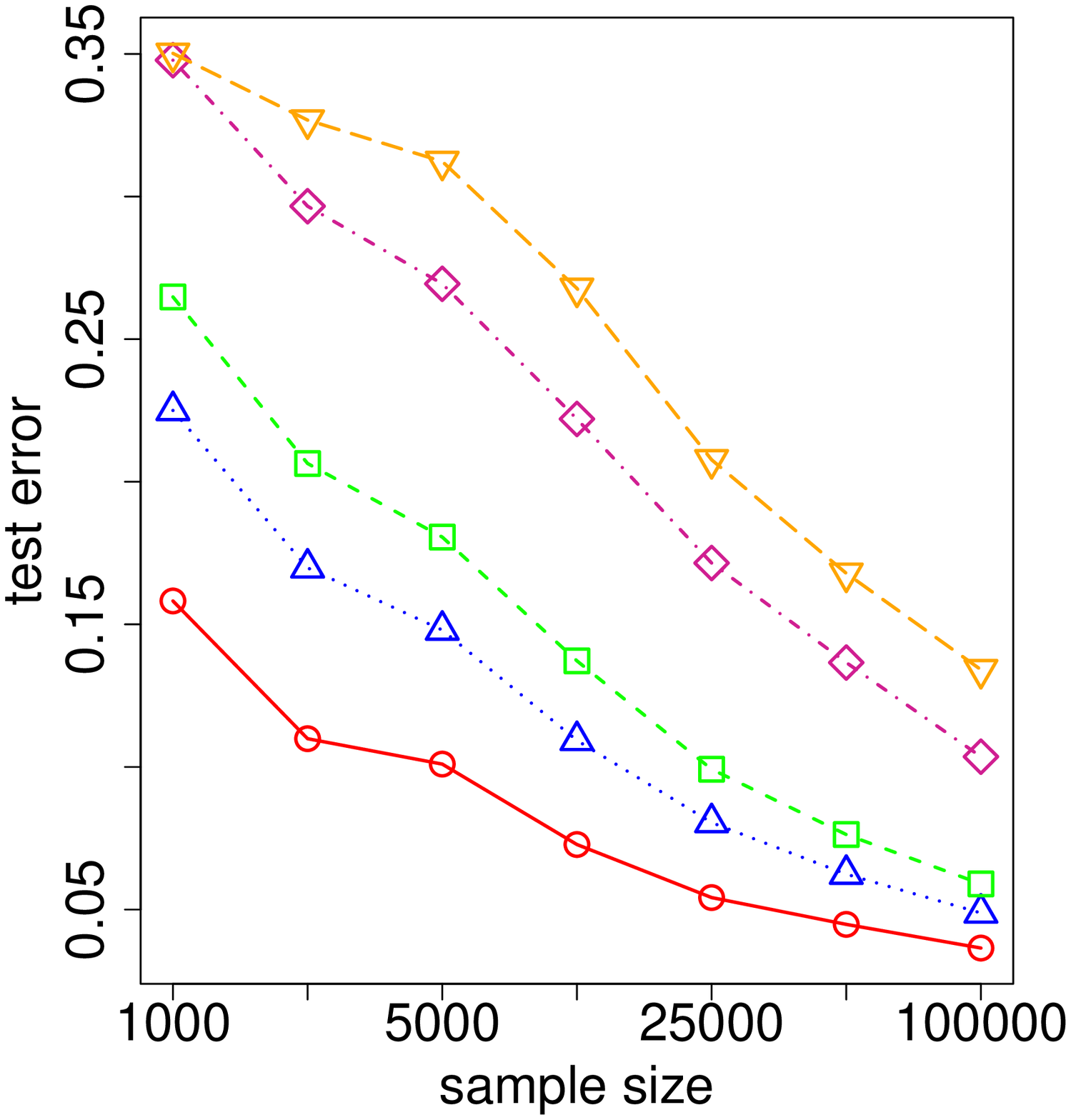}}\vspace{-0.6cm}\\
\subfigure[Average training time of the various VP-SVMs for $n_\text{train}\leq10\,000$\label{fig::experiments::ijcnn1_vp_time}]{\includegraphics[angle=270,  scale=0.26]{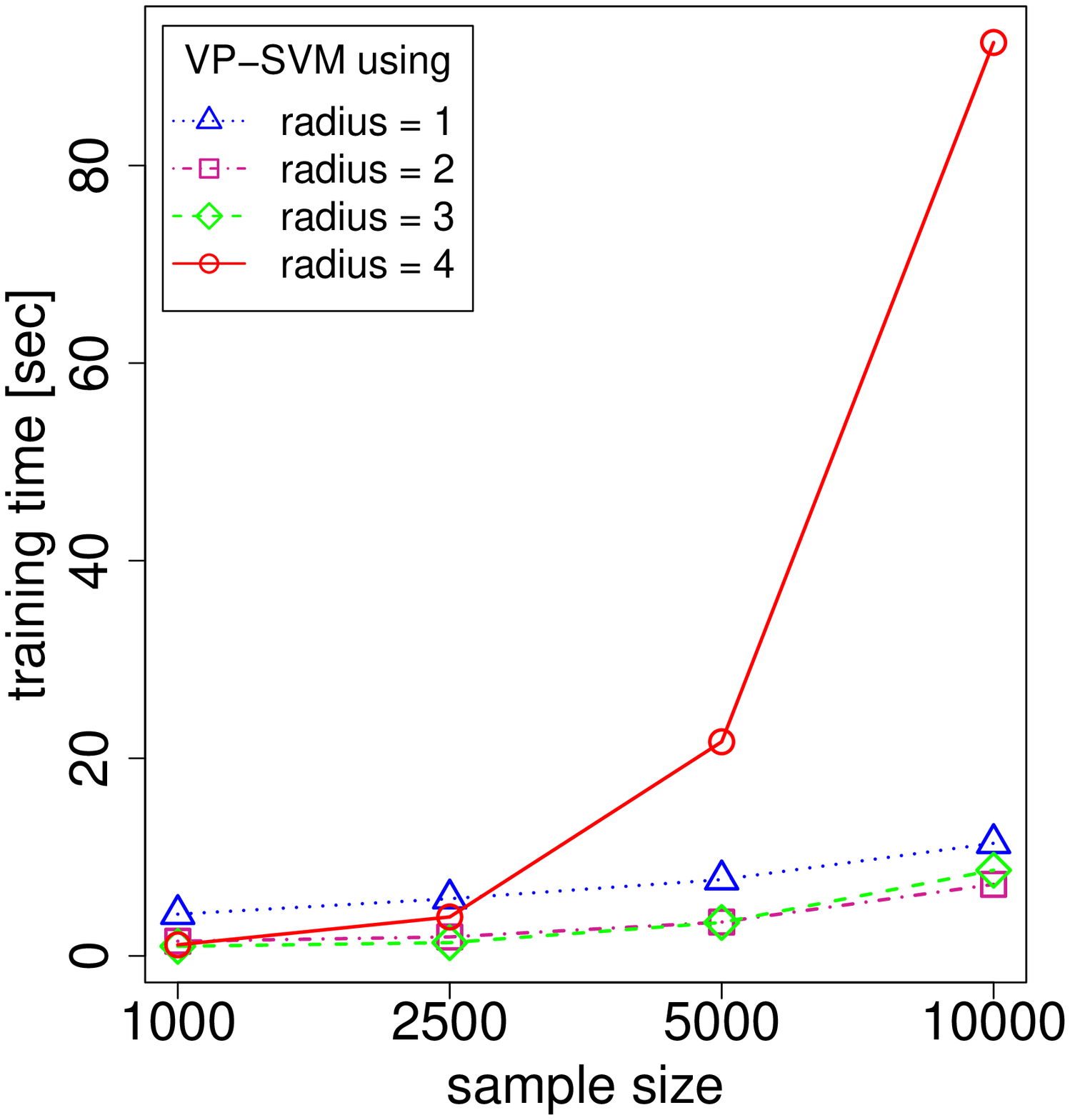}}
\hfill\subfigure[Average training time of the various VP-SVMs for $n_\text{train}\geq5\,000$\label{fig::experiments::ijcnn1_vp_time_all}]{\includegraphics[angle=270,  scale=0.26]{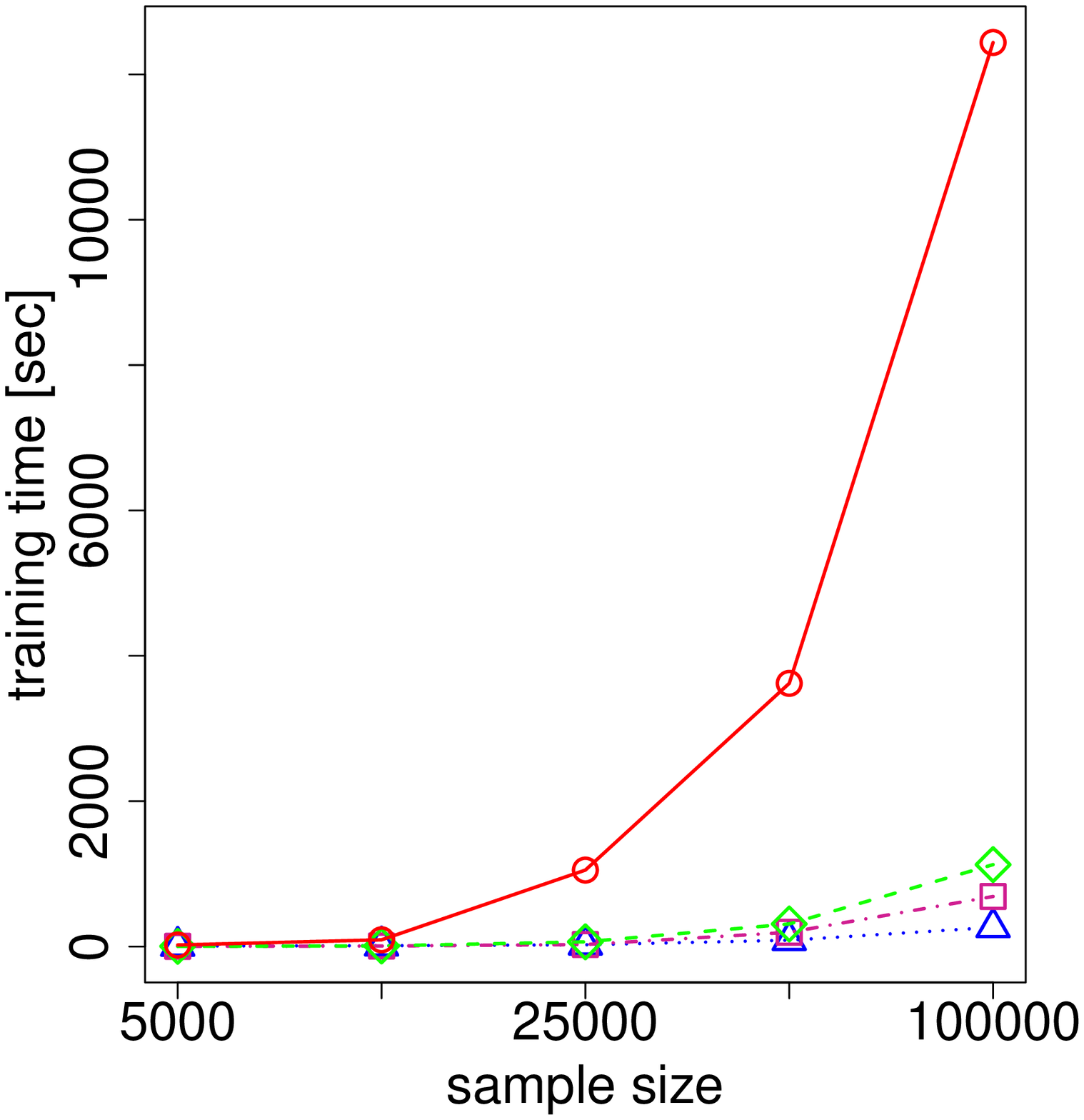}}
\hfill\subfigure[Average empirical risk of the various VP-SVMs\label{fig::experiments::ijcnn1_vp_error}]{\includegraphics[angle=270,  scale=0.26]{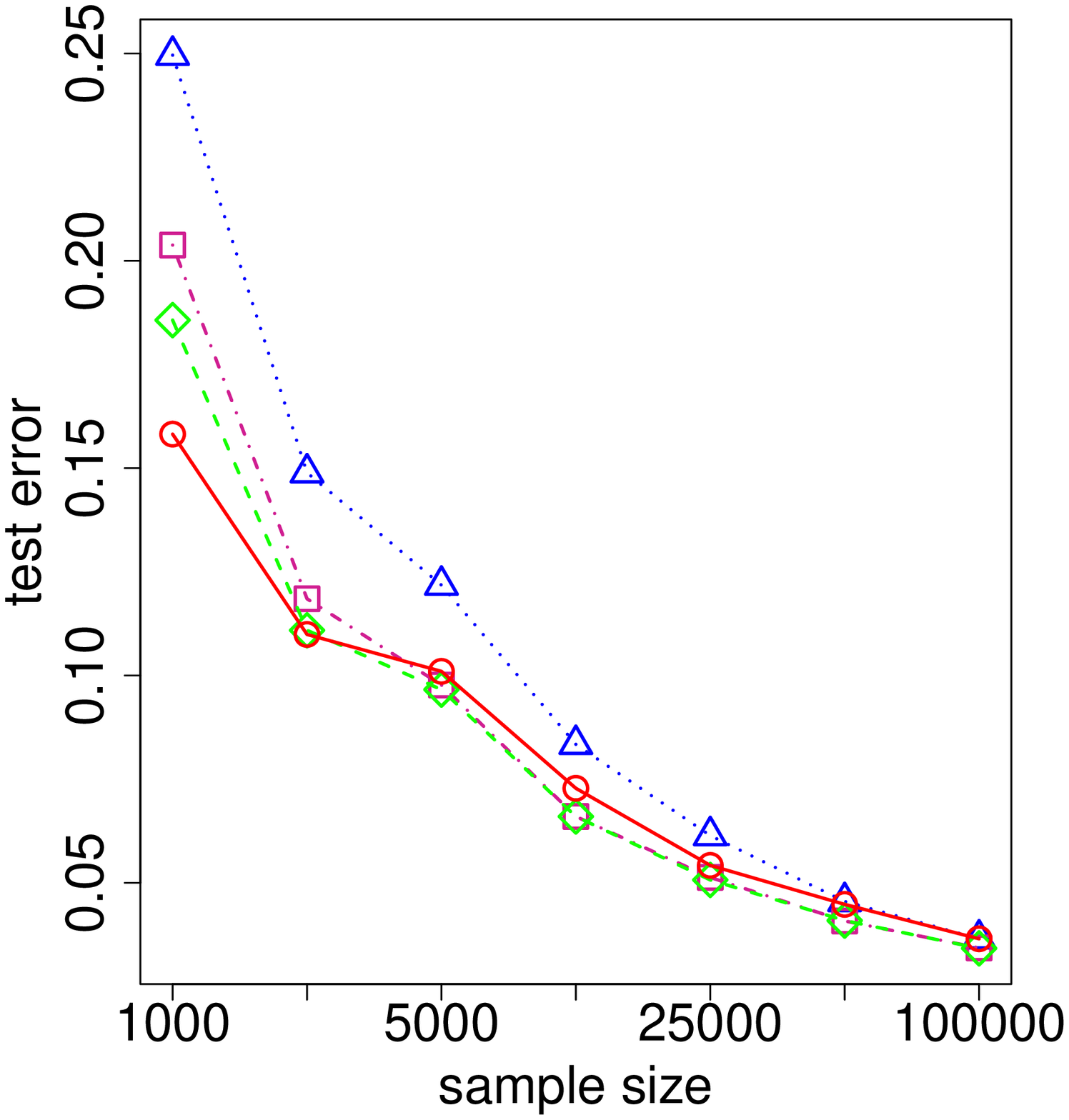}}\vspace{-0.6cm}\\
\subfigure[Average training time of \mbox{LS-,} VP-, and RC-SVMs for $n_\text{train}\leq10\,000$\label{fig::experiments::ijcnn1_time}]{\includegraphics[angle=270,  scale=0.26]{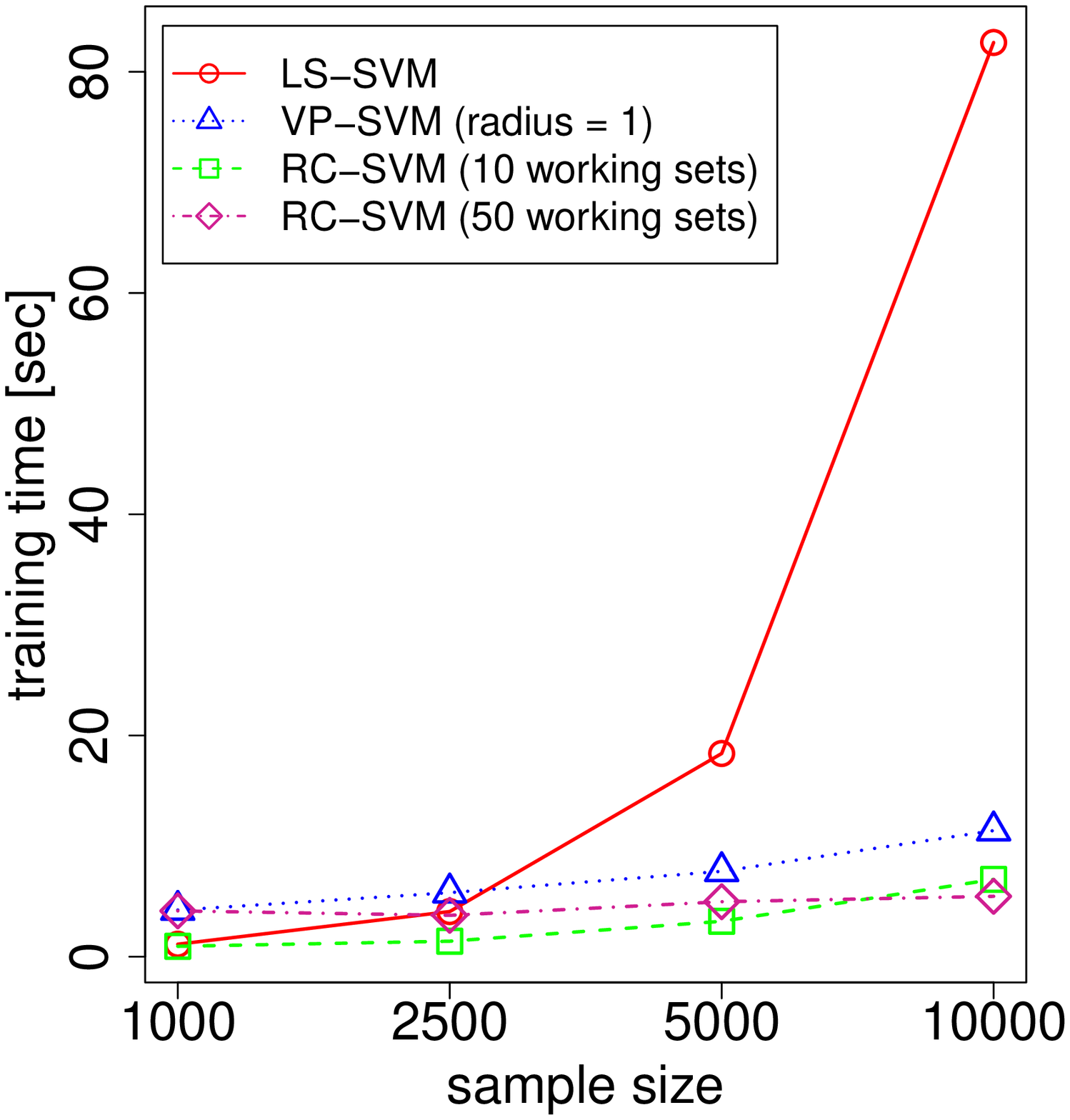}}
\hfill\subfigure[Average training time of \mbox{LS-,} VP-, and RC-SVMs for $n_\text{train}\geq5\,000$\label{fig::experiments::ijcnn1_time_all}]{\includegraphics[angle=270,  scale=0.26]{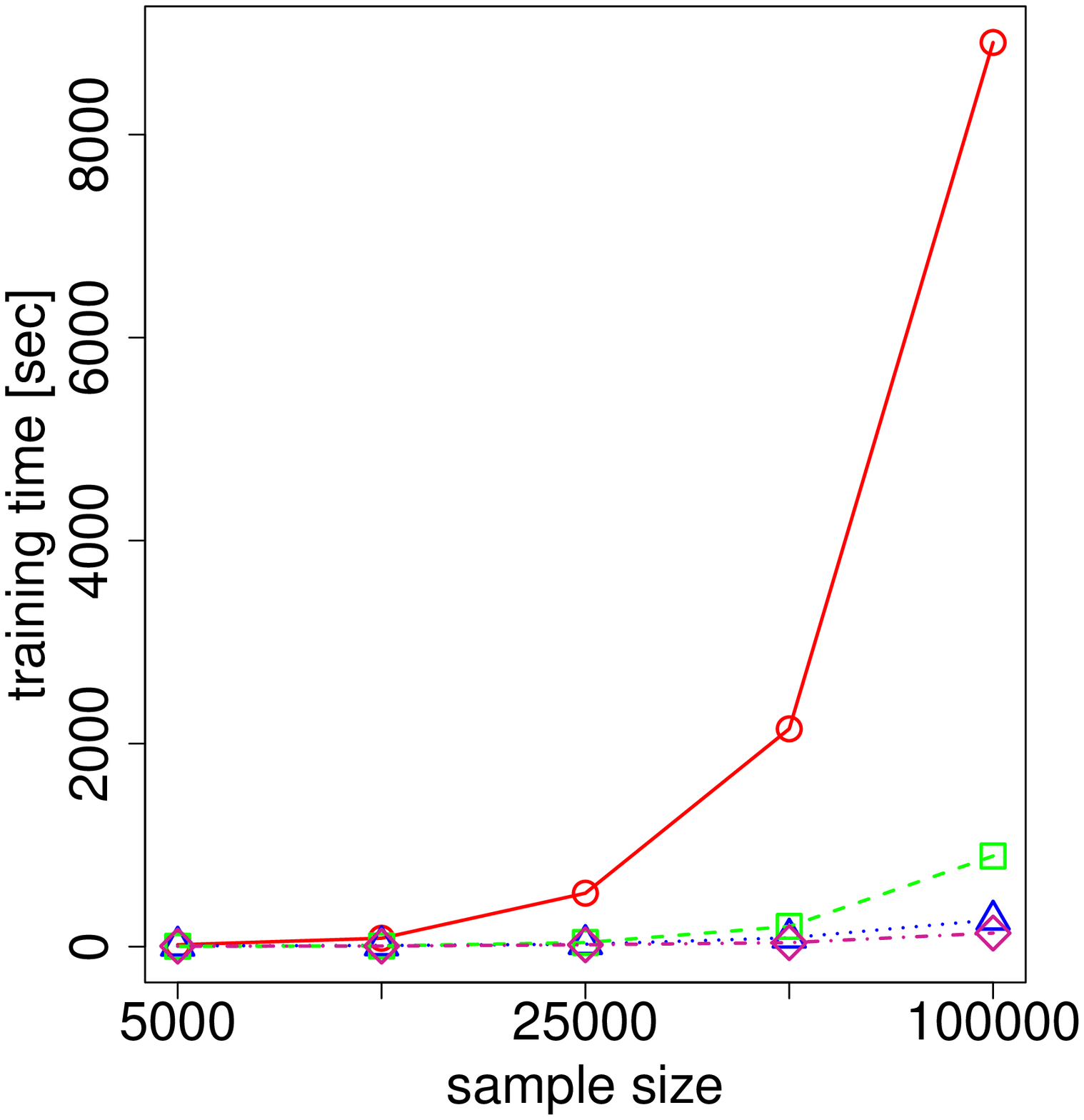}}
\hfill\subfigure[Average empirical risk of \mbox{LS-,} VP-, and RC-SVMs\label{fig::experiments::ijcnn1_error}]{\includegraphics[angle=270,  scale=0.26]{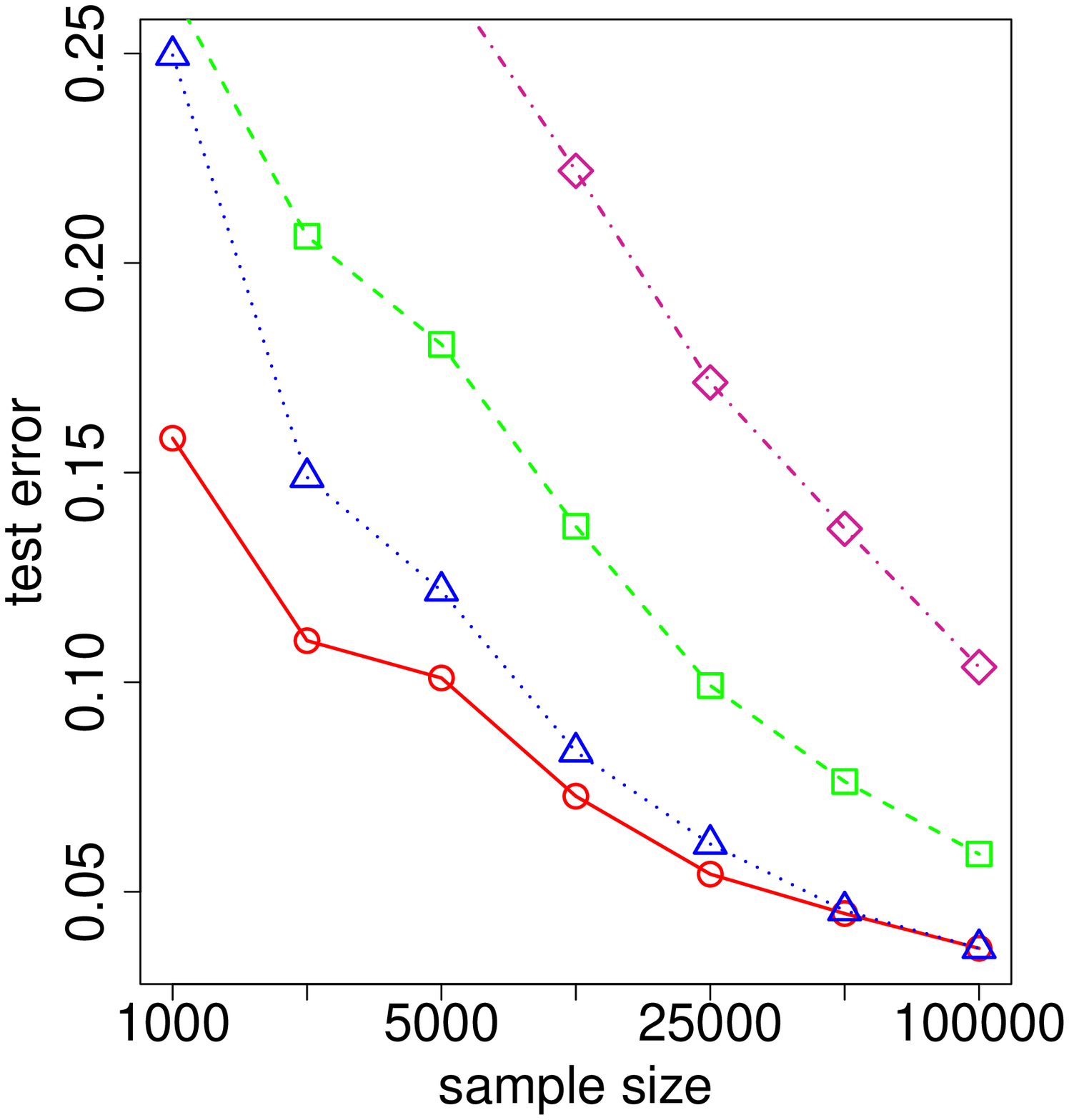}}\vspace{-0.2cm}
\caption{Average training time and test error of LS-, VP-, and RC-SVMs for the 
real-world data \textsc{ijcnn1} depending on the 
training set size $n_\text{train}=1\,000,\ldots,100\,000$. Subfigures 
\subref{fig::experiments::ijcnn1_rc_time}--\subref{fig::experiments::ijcnn1_rc_error} 
show the results for RC-SVMs using different numbers of working sets and Subfigures 
\subref{fig::experiments::ijcnn1_vp_time}--\subref{fig::experiments::ijcnn1_vp_error} 
illustrate the results for VP-SVMs using various radii. At the bottom, Subfigures 
\subref{fig::experiments::ijcnn1_time}--\subref{fig::experiments::ijcnn1_error} contain the 
average training times and the average test errors of the LS-SVM, one VP-SVM and 
two RC-SVMs. Here, the VP-SVM is the one which trains fastest for $n_\text{train}=100\,000$
and the two RC-SVMs are those which achieve for $n_\text{train}=100\,000$ roughly the same 
training time as the chosen VP-SVM.}
\label{fig::experiments::ijcnn1}
\end{scriptsize}
\end{figure}

First of all, let us investigate the RC-SVM results that are composed in 
Tables \ref{tab::experiments::covtype_rc}--\ref{tab::experiments::ijcnn1} 
as well as in Subfigures 
\subref{fig::experiments::covtype_rc_time}--\subref{fig::experiments::covtype_rc_error}
of Figures \ref{fig::experiments::covtype}--\ref{fig::experiments::ijcnn1}. For the real data sets 
\textsc{covtype} we considered ten, for the data sets \textsc{cod-rna} nine, and for the 
data sets \textsc{ijcnn1} eight different numbers of working sets. In each case, we started 
with an RC-SVM using one working set, i.e.~with an RC-SVM that corresponds to the global LS-SVM
modulo different values generated by the random generator,
%
cf.~Tables \ref{tab::experiments::cod-rna} and \ref{tab::experiments::ijcnn1}.
Comparing for every data set the RC-SVMs 
using various numbers of working sets, we observe that the number of working sets, 
minimizing the RC-SVMs training time, increases in tandem with the sample size.
Moreover, the RC-SVM using one working set never trains fastest compared to the other 
RC-SVMs using more than one working set. 
Furthermore, the average test errors for the applied RC-SVMs usually 
decrease for a decreasing number of working sets and, hence, are minimized by the smallest 
possible number of working sets.
Of course, all these findings are not surprising, since RC-SVMs are typically used to reduce 
the training time.

Let us now compare the results of VP- and RC-SVMs using roughly the same 
number of working sets, cf.~Tables 
\ref{tab::experiments::covtype_ls_vp}--\ref{tab::experiments::ijcnn1}. 
Initially note that, even though we consider VP- and RC-SVM based on the same 
number of working sets, the RC-SVM working sets are about the same size whereas the VP-SVM 
working sets may have different sizes with a large range. That is, the VP-SVMs often 
deal with a few substantially larger working sets than the RC-SVMs. Consequently, the 
RC-SVMs often perform faster than the VP-SVMs, which require up to five times the RC-SVM's 
training time for $n_\text{train}=100\,000$. Contrarily, the average empirical risks achieved 
by the VP-SVMs are substantially lower than those of the RC-SVMs. 
Besides, in a few cases the VP-SVMs possess at least one working set which is oversized for 
the computer's RAM, so that these VP-SVM problems are infeasible, whereas the comparable 
RC-SVMs avoid this conflict. Here, consider e.g.~the 
RC-SVM using seven working sets and 
the VP-SVM with radius $r=4$ for the \textsc{covtype} data set of size $500\,000$.

In Section \ref{subsubsec::exp::VP-LS-SVMcomparison}, we compared for each data set type the 
LS-SVM with the VP-SVM that trains fastest for the largest training data set. 
Here, we additionally compare this VP-SVM to the RC-SVMs.
To be able to draw a fair 
comparison in terms of the achieved test errors, we choose those RC-SVMs that train roughly 
as fast as the VP-SVM for the largest training set, i.e.~the slowest RC-SVM training faster 
and the fastest RC-SVM training slower than the above VP-SVM. Subfigures 
\subref{fig::experiments::covtype_time}--\subref{fig::experiments::covtype_error}
of Figures \ref{fig::experiments::covtype}--\ref{fig::experiments::ijcnn1} illustrate
the average training times and the average test errors of these RC-SVMs, the above VP-SVM, and 
the LS-SVM. Considering the RC-SVMs, the faster of the two requires for 
$n_\text{train}= 100\,000$ between $51\%$ and $83\%$ of the VP-SVMs training time and trains 
at most seven minutes faster than the VP-SVM. However, at least for 
$n_\text{train}\geq 5\,000$, both considered RC-SVMs induce substantially 
higher test errors than the VP- and LS-SVM.
Finally, note that VP-SVMs for $n_\text{train}\geq250\,000$ considerably outperform LS-SVMs 
for $n_\text{train}= 100\,000$, while RC-SVMs for $n_\text{train}\geq250\,000$ lead to even 
worse test errors than the considered LS-SVMs.

Summarizing, we record that RC-SVMs using as few as possible working sets achieve the smallest 
RC-SVM test errors, however, those using more working sets perform faster. Furthermore, 
compared to VP-SVMs using roughly the same number of working sets as the RC-SVMs, the latter 
ones may learn faster though not as good as the VP-SVMs.
Moreover, considering RC-SVMs that require roughly the same training time as the fastest 
VP-SVM, we saw that the RC-SVMs lead to much higher empirical risks.
That is, if the required training time is a hard constraint, then the VP-SVM that satisfies 
this constraint achieves a better test error than a RC-SVM that also trains fast enough.

\subsection{Experiments on Artificial Data}
\label{subsec::exp::smoothness_interruptions}

\begin{figure}[p]
\begin{scriptsize}\vspace{-0.6cm}
\subfigure[Average training time of the various RC-SVMs\label{fig::experiments::art_data_1_2_rc_time}]{\includegraphics[angle=270, scale=0.26]{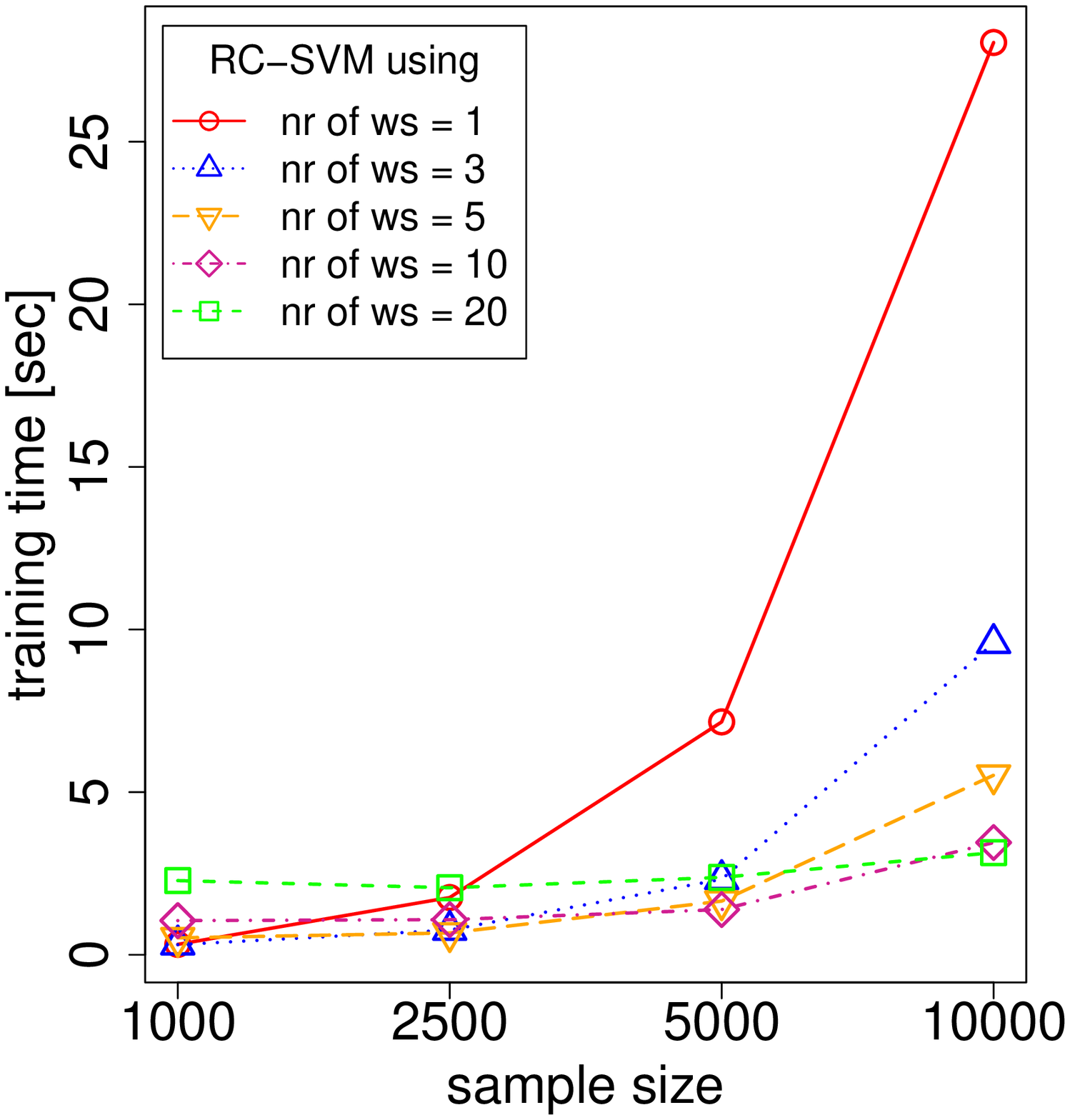}}
\hfill\subfigure[Average empirical risk of the various RC-SVMs\label{fig::experiments::art_data_1_2_rc_error}]{\includegraphics[angle=270, scale=0.26]{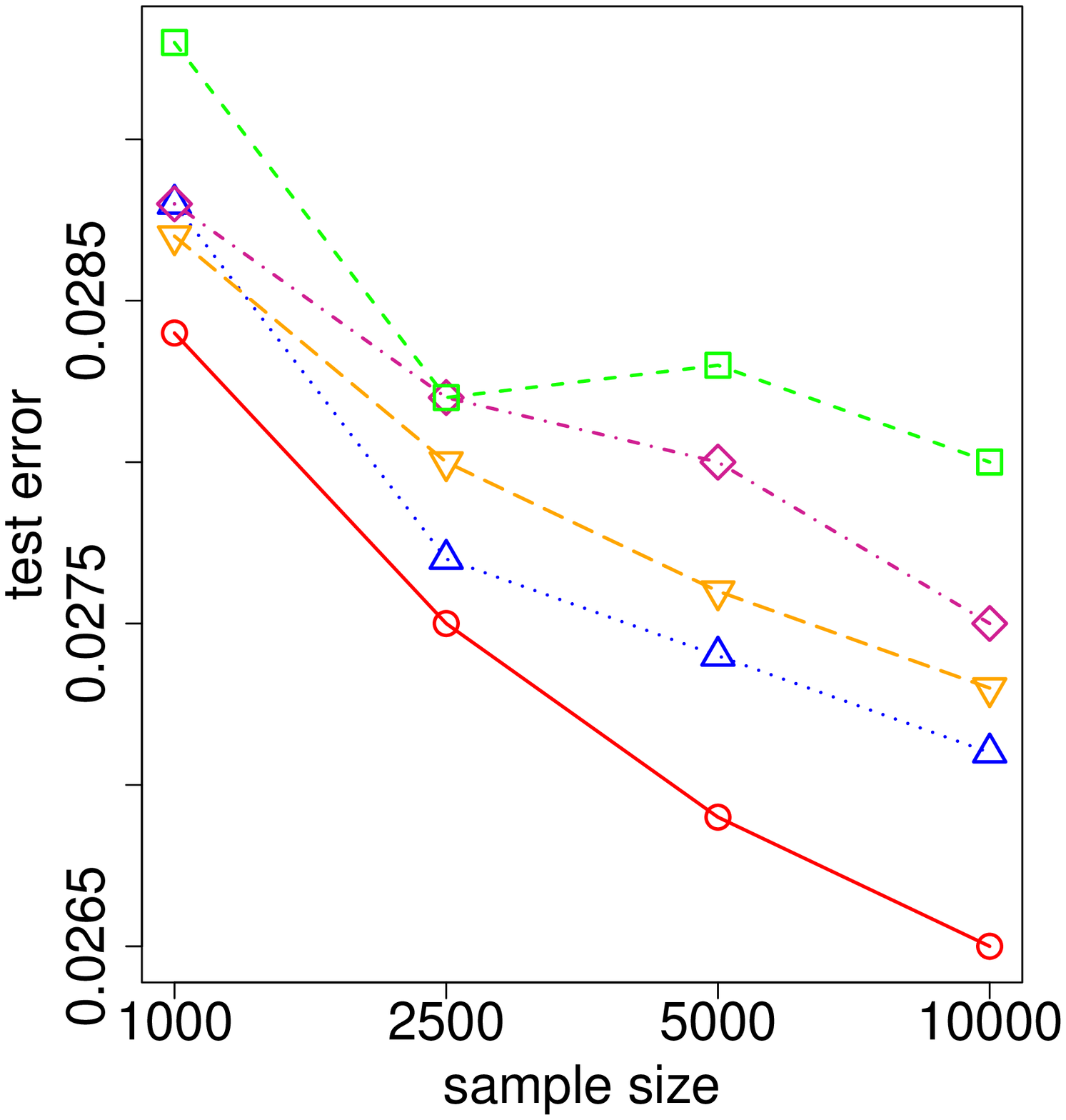}}
\hfill\subfigure[Average empirical $L_2$-error of the various RC-SVMs\label{fig::experiments::art_data_1_2_rc_L2error}]{\includegraphics[angle=270, scale=0.26]{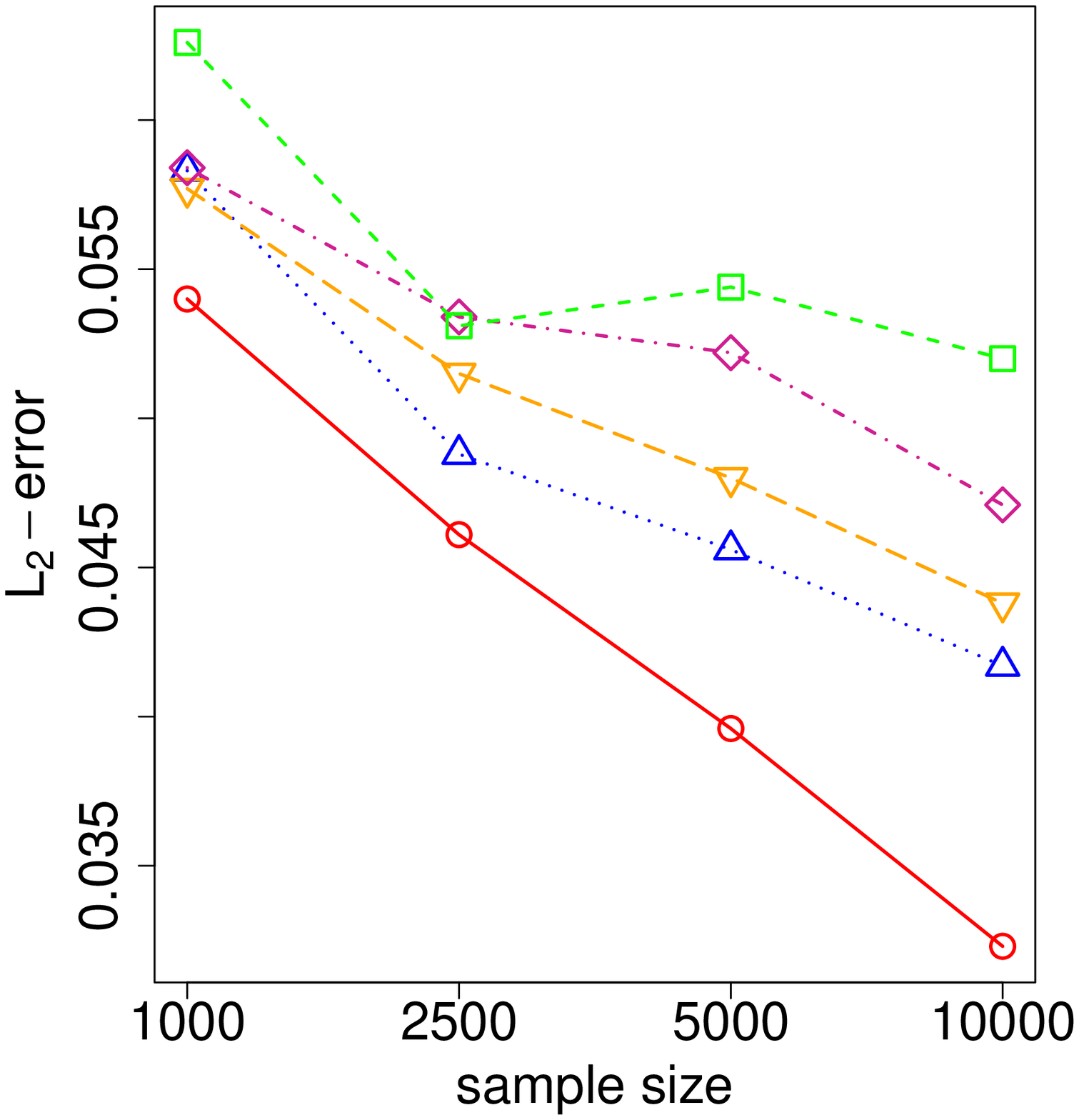}}\vspace{-0.5cm}\\
\subfigure[Average training time of the various VP-SVMs\label{fig::experiments::art_data_1_2_vp_time}]{\includegraphics[angle=270, scale=0.26]{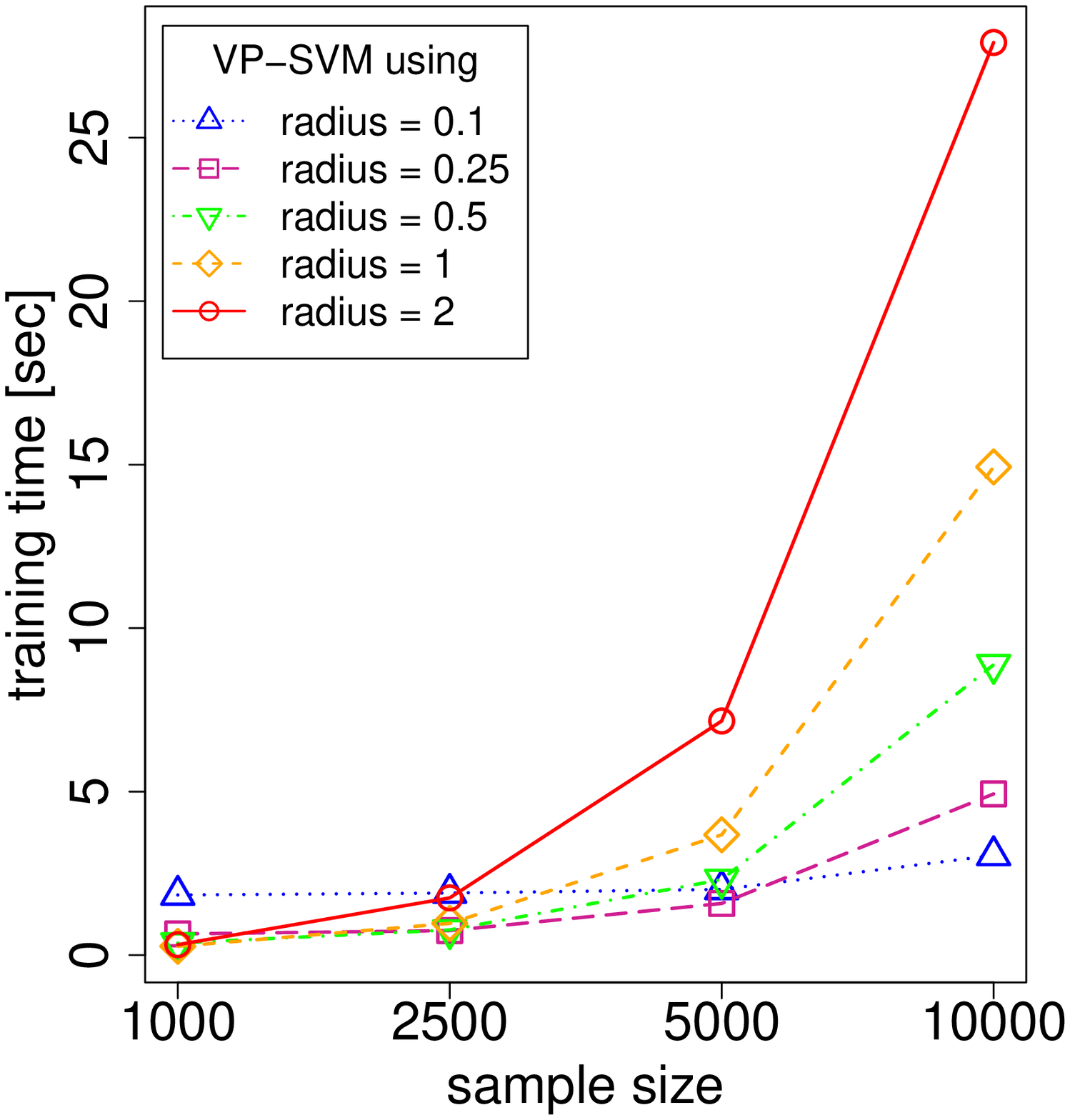}}
\hfill\subfigure[Average empirical risk of the various VP-SVMs\label{fig::experiments::art_data_1_2_vp_error}]{\includegraphics[angle=270, scale=0.26]{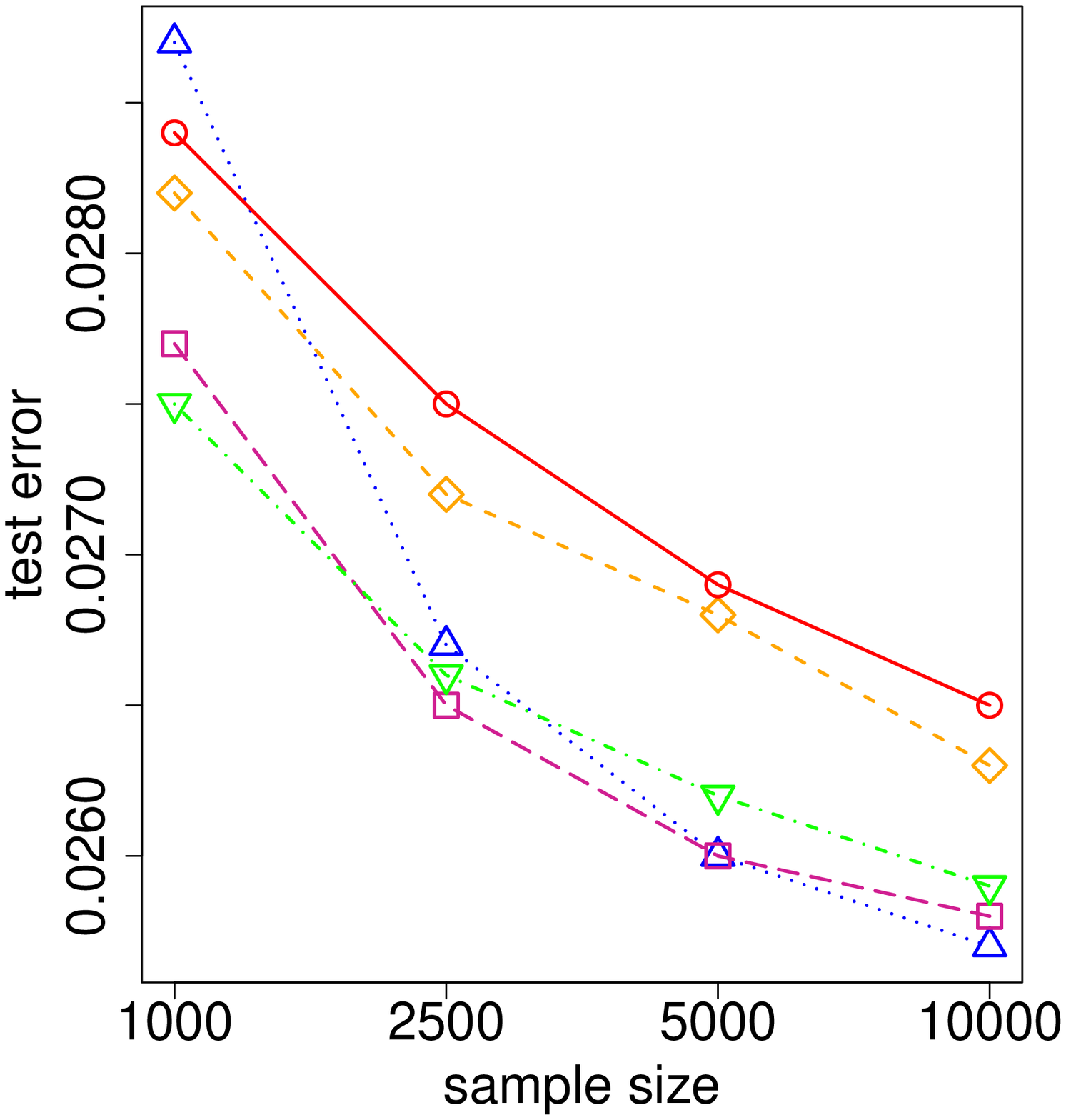}}
\hfill\subfigure[Average empirical $L_2$-error of the various VP-SVMs\label{fig::experiments::art_data_1_2_vp_L2error}]{\includegraphics[angle=270, scale=0.26]{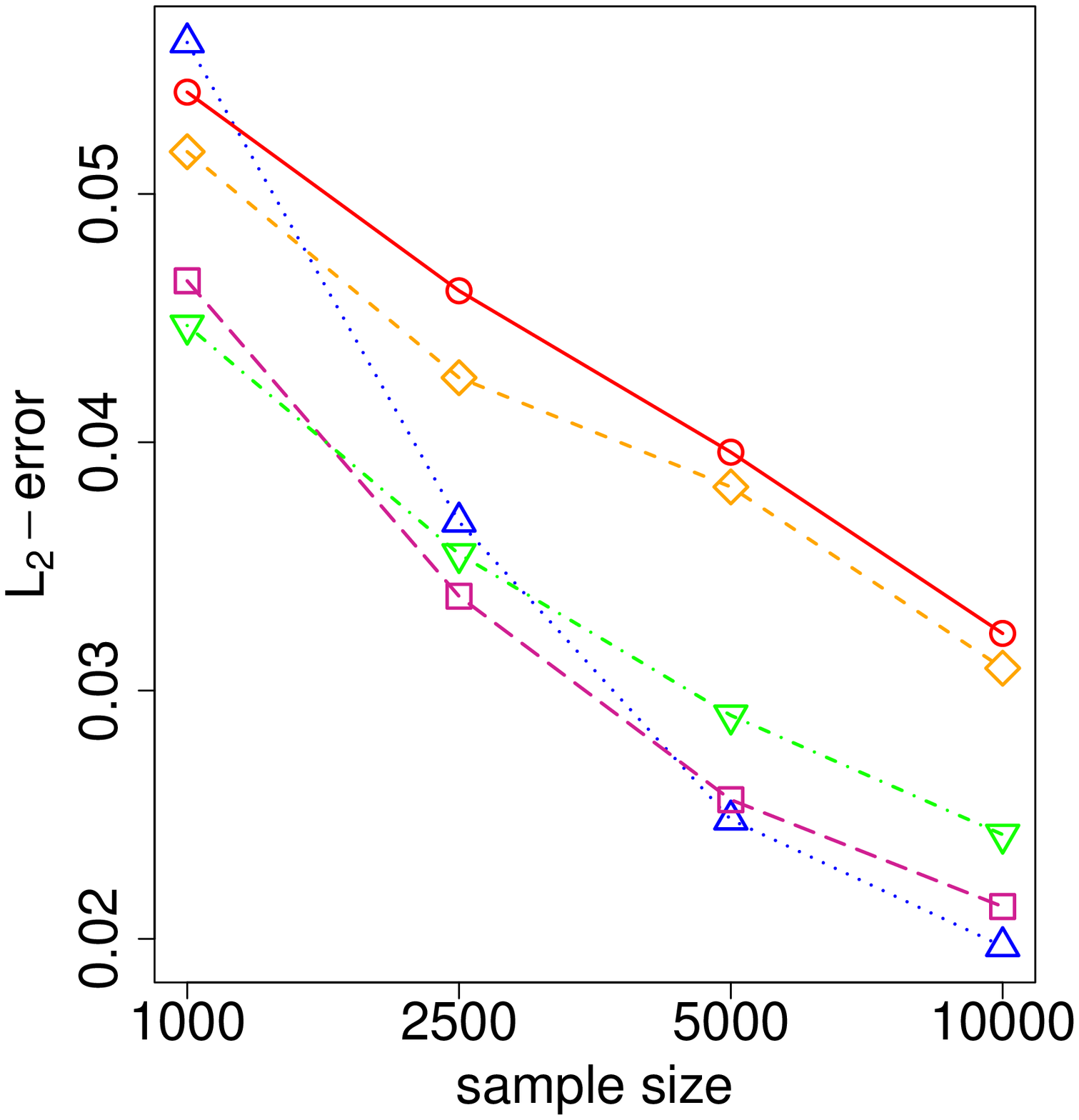}}\vspace{-0.5cm}\\
\subfigure[Average training time of \mbox{LS-,} VP-, and RC-SVMs\label{fig::experiments::art_data_1_2_time}]{\includegraphics[angle=270, scale=0.26]{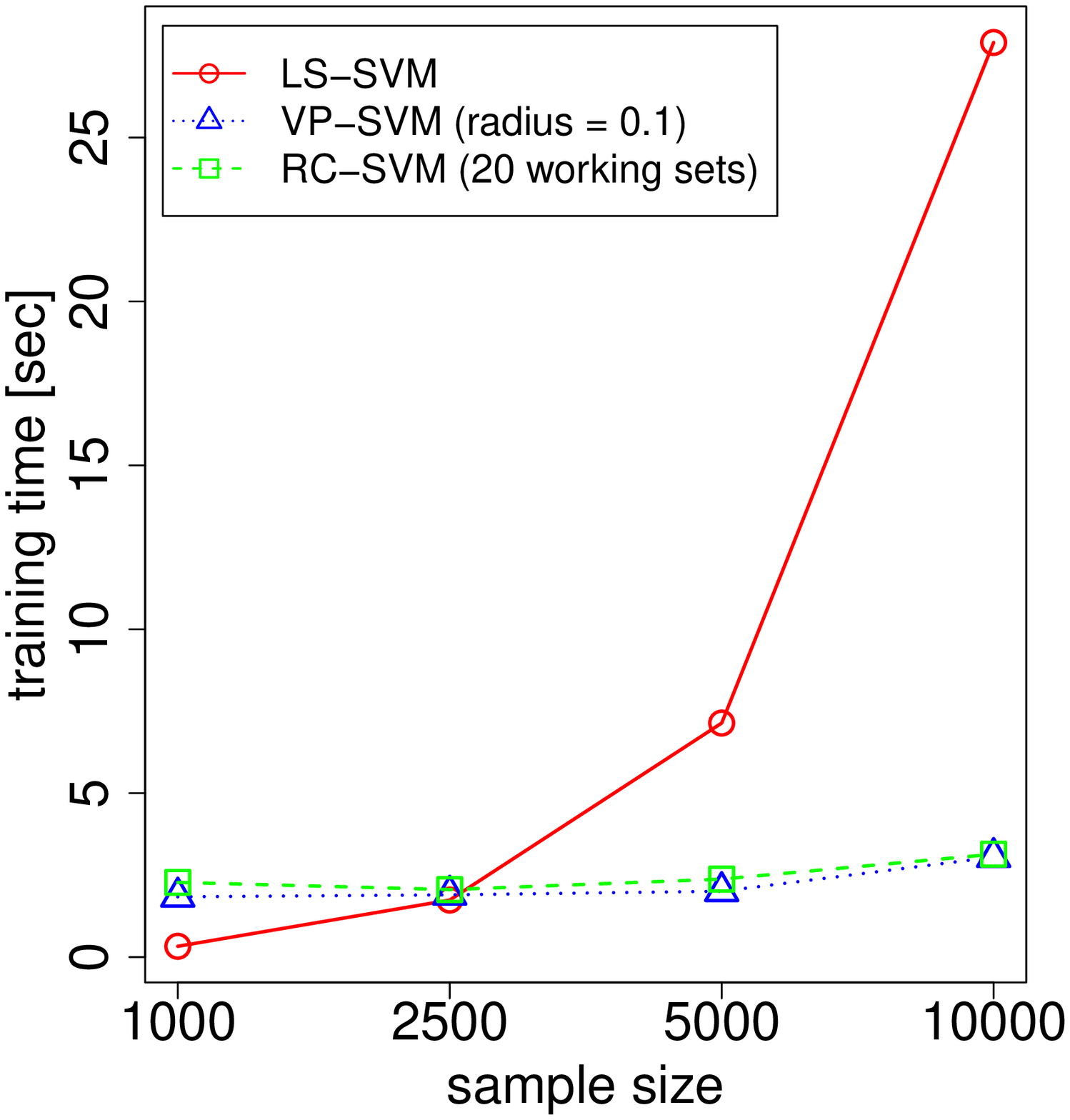}}
\hfill\subfigure[Average empirical risk of \mbox{LS-,} VP-, and RC-SVMs\label{fig::experiments::art_data_1_2_error}]{\includegraphics[angle=270, scale=0.26]{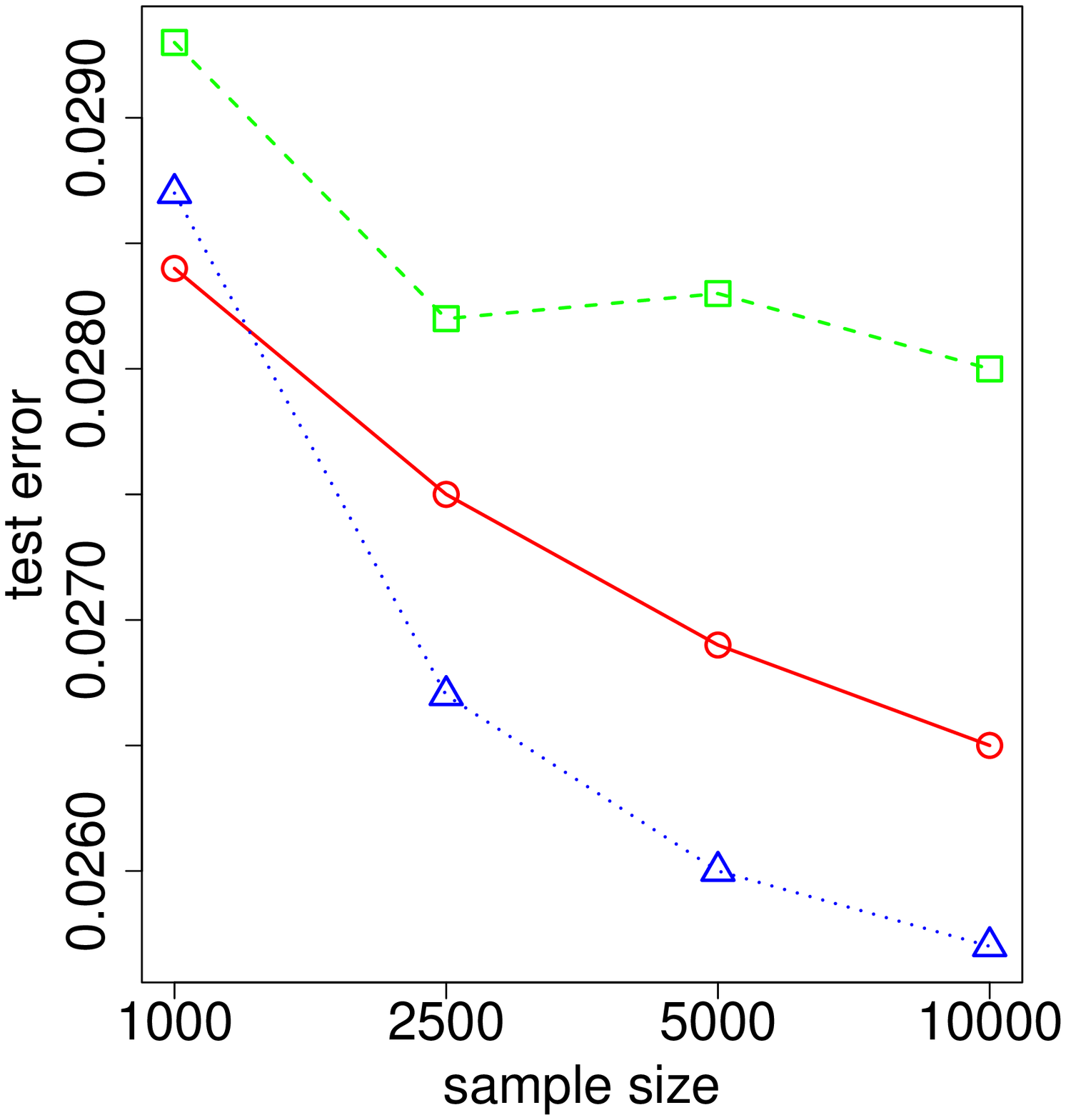}}
\hfill\subfigure[Average empirical $L_2$-error of \mbox{LS-,} VP-, and RC-SVMs\label{fig::experiments::art_data_1_2_L2error}]{\includegraphics[angle=270, scale=0.26]{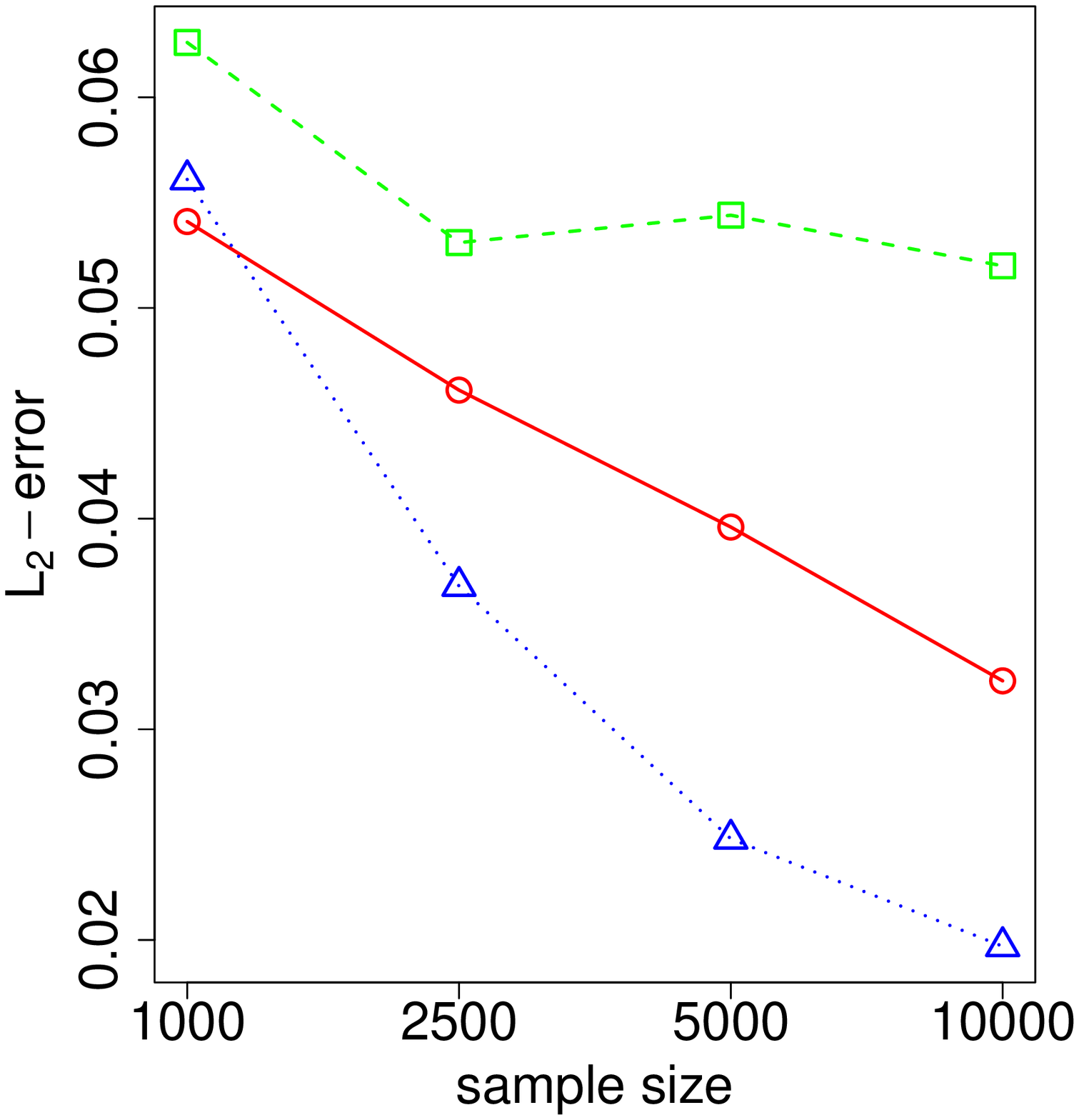}}
\caption{
Average training time and test error of LS-, VP-, and RC-SVMs for the artificial data 
Type \RM{1} depending on the training set size $n_\text{train}=1\,000,\ldots,10\,000$. Subfigures 
\subref{fig::experiments::art_data_1_2_rc_time}--\subref{fig::experiments::art_data_1_2_rc_L2error} 
show the results for RC-SVMs using different numbers of working sets and Subfigures 
\subref{fig::experiments::art_data_1_2_vp_time}--\subref{fig::experiments::art_data_1_2_vp_L2error} 
illustrate the results for VP-SVMs using various radii. At the bottom, Subfigures 
\subref{fig::experiments::art_data_1_2_time}--\subref{fig::experiments::art_data_1_2_L2error} 
contain the average training times and the average test errors of the LS-SVM, one VP-SVM and 
one RC-SVM. Here, the VP-SVM and the RC-SVM are those which train fastest for 
$n_\text{train}=10\,000$. Note that in the case at hand none of the considered RC-SVMs performs 
faster than the fastest VP-SVM for $n_\text{train}=10\,000$.} 
\label{fig::experiments::artificial_data_1_2}
\end{scriptsize}
\end{figure}

\begin{figure}[p]
\begin{scriptsize}\vspace{-0.6cm}
\subfigure[Average training time of the various RC-SVMs\label{fig::experiments::art_data_2_2_rc_time}]{\includegraphics[angle=270, scale=0.26]{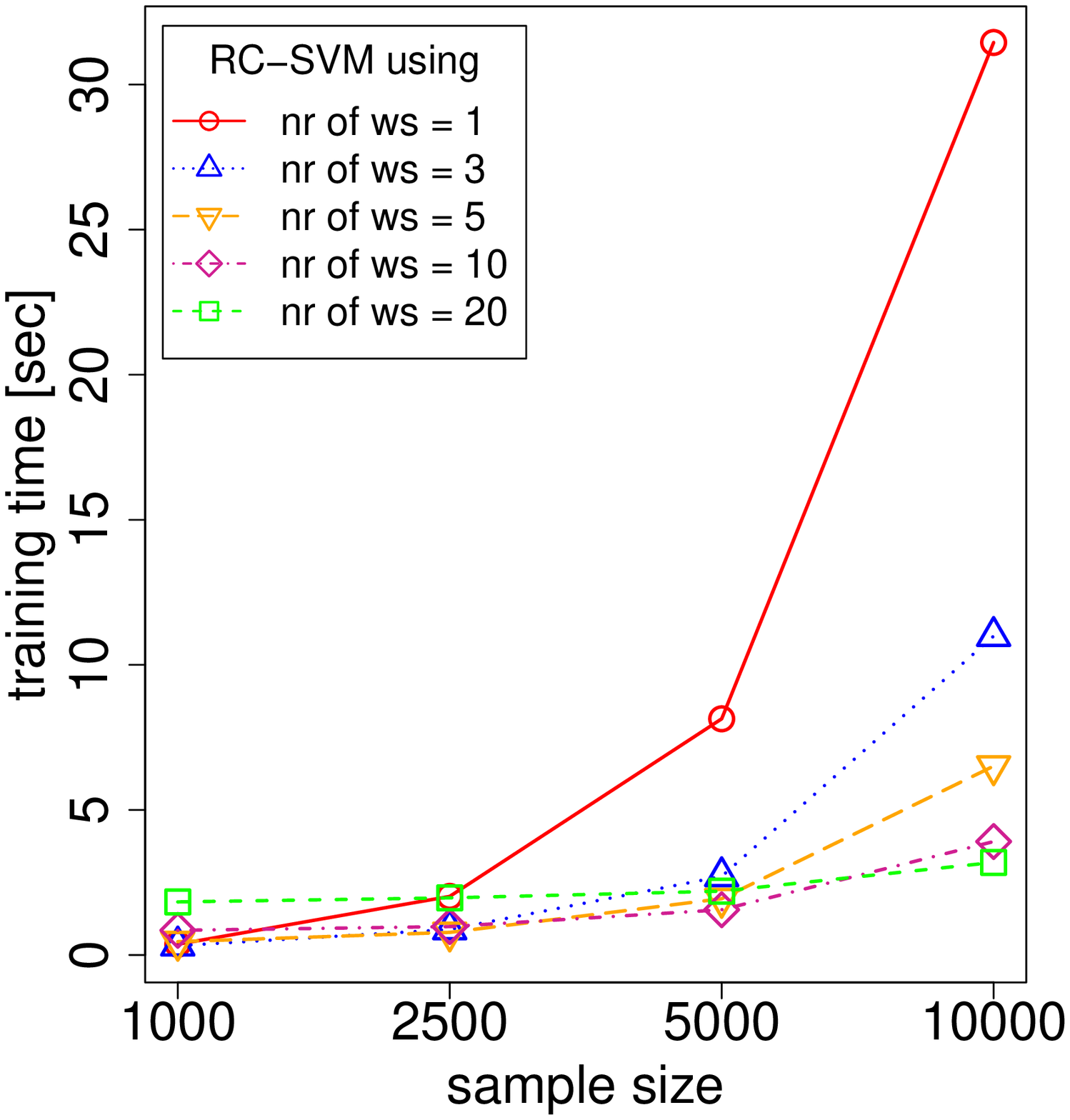}}
\hfill\subfigure[Average empirical risk of the various RC-SVMs\label{fig::experiments::art_data_2_2_rc_error}]{\includegraphics[angle=270, scale=0.26]{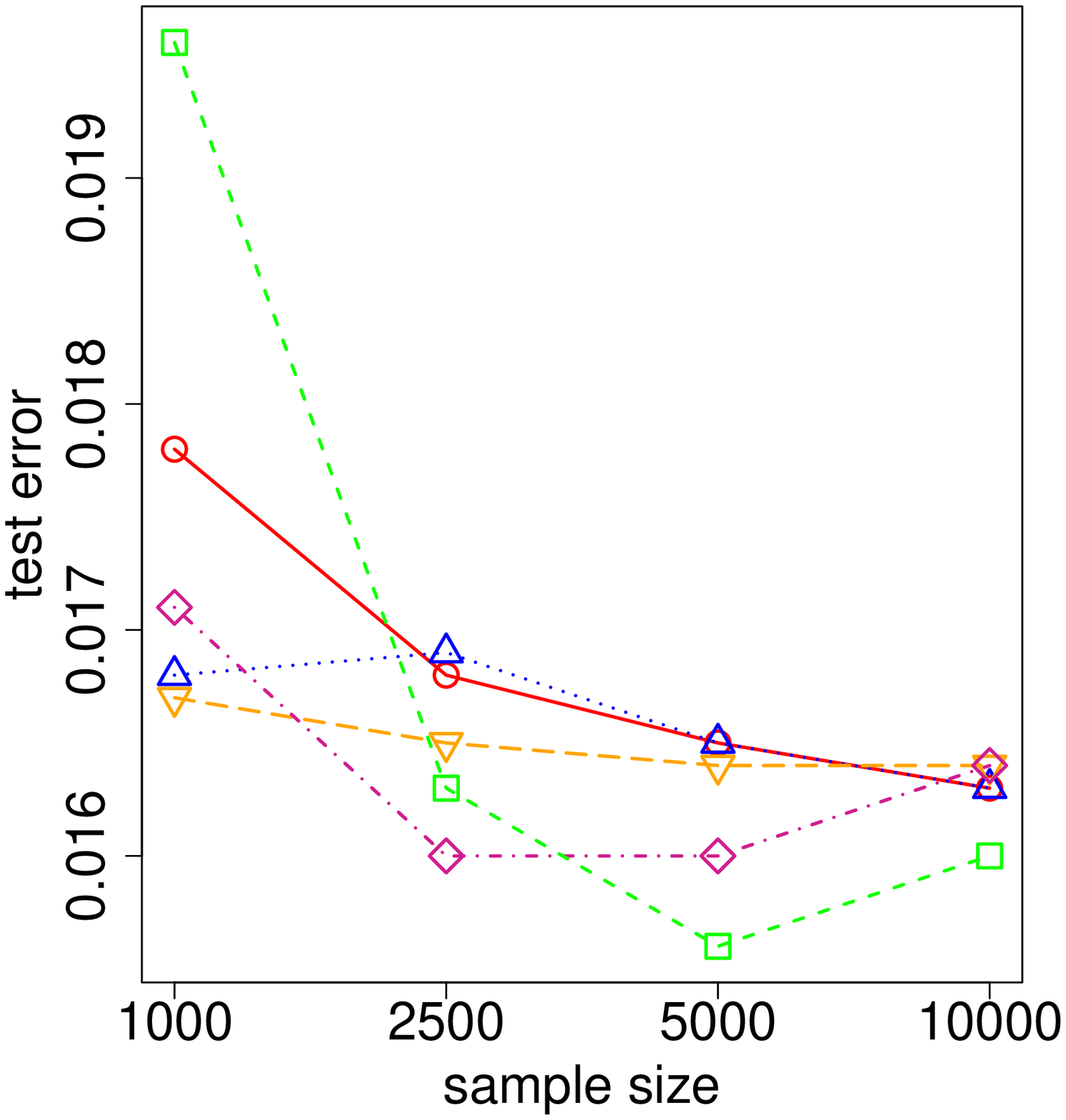}}
\hfill\subfigure[Average empirical $L_2$-error of the various RC-SVMs\label{fig::experiments::art_data_2_2_rc_L2error}]{\includegraphics[angle=270, scale=0.26]{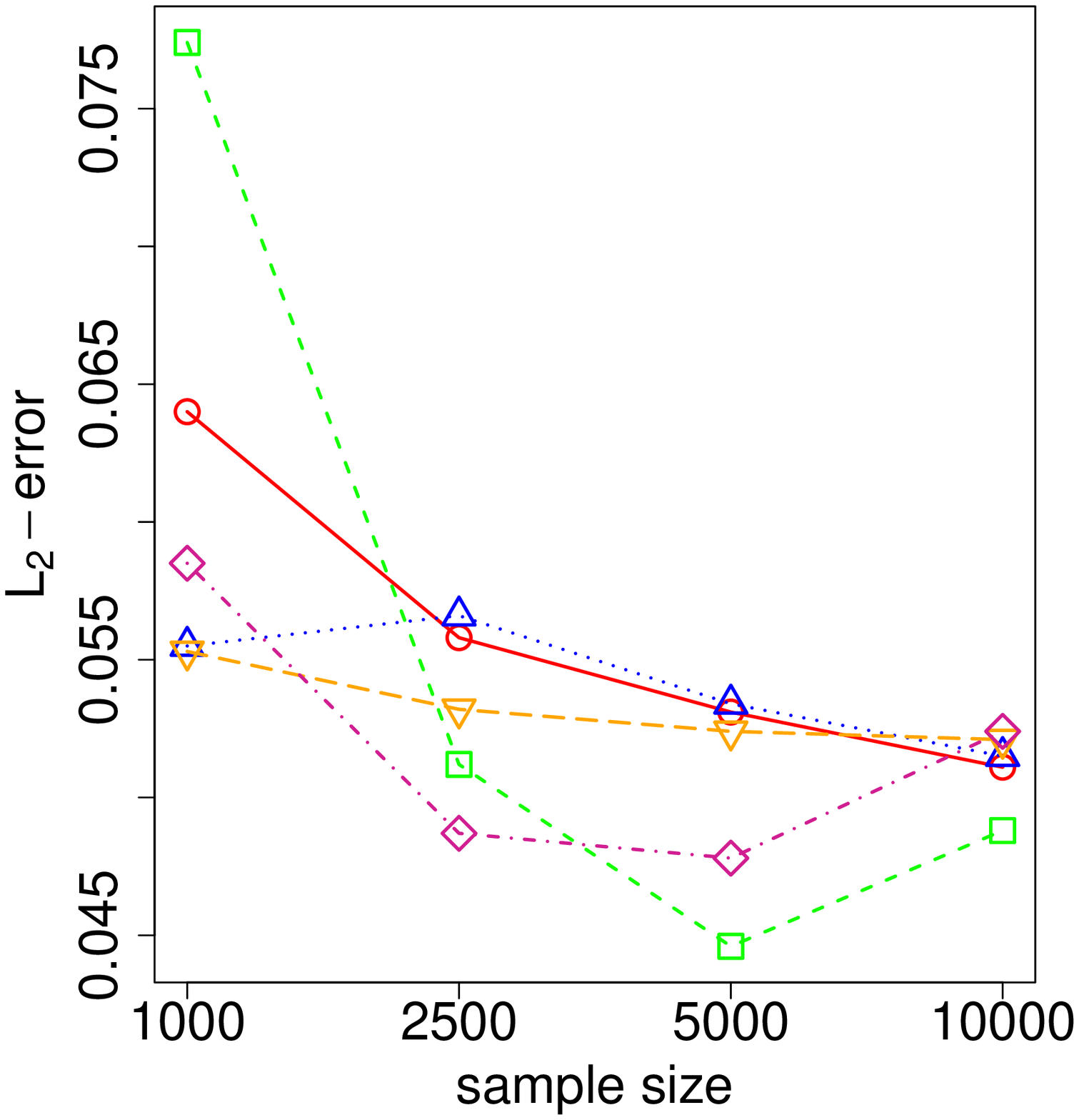}}\vspace{-0.5cm}\\
\subfigure[Average training time of the various VP-SVMs\label{fig::experiments::art_data_2_2_vp_time}]{\includegraphics[angle=270, scale=0.26]{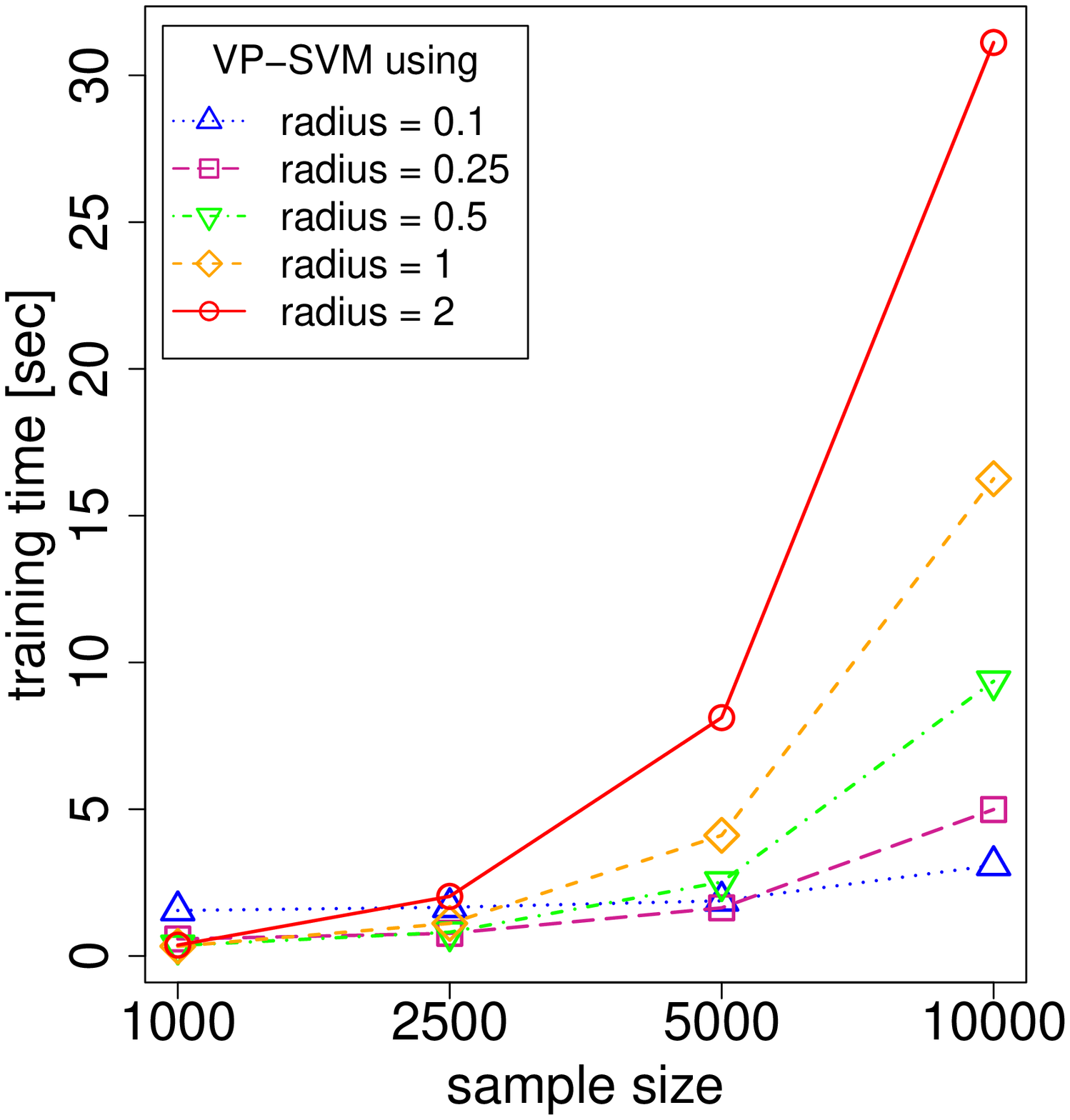}}
\hfill\subfigure[Average empirical risk of the various VP-SVMs\label{fig::experiments::art_data_2_2_vp_error}]{\includegraphics[angle=270, scale=0.26]{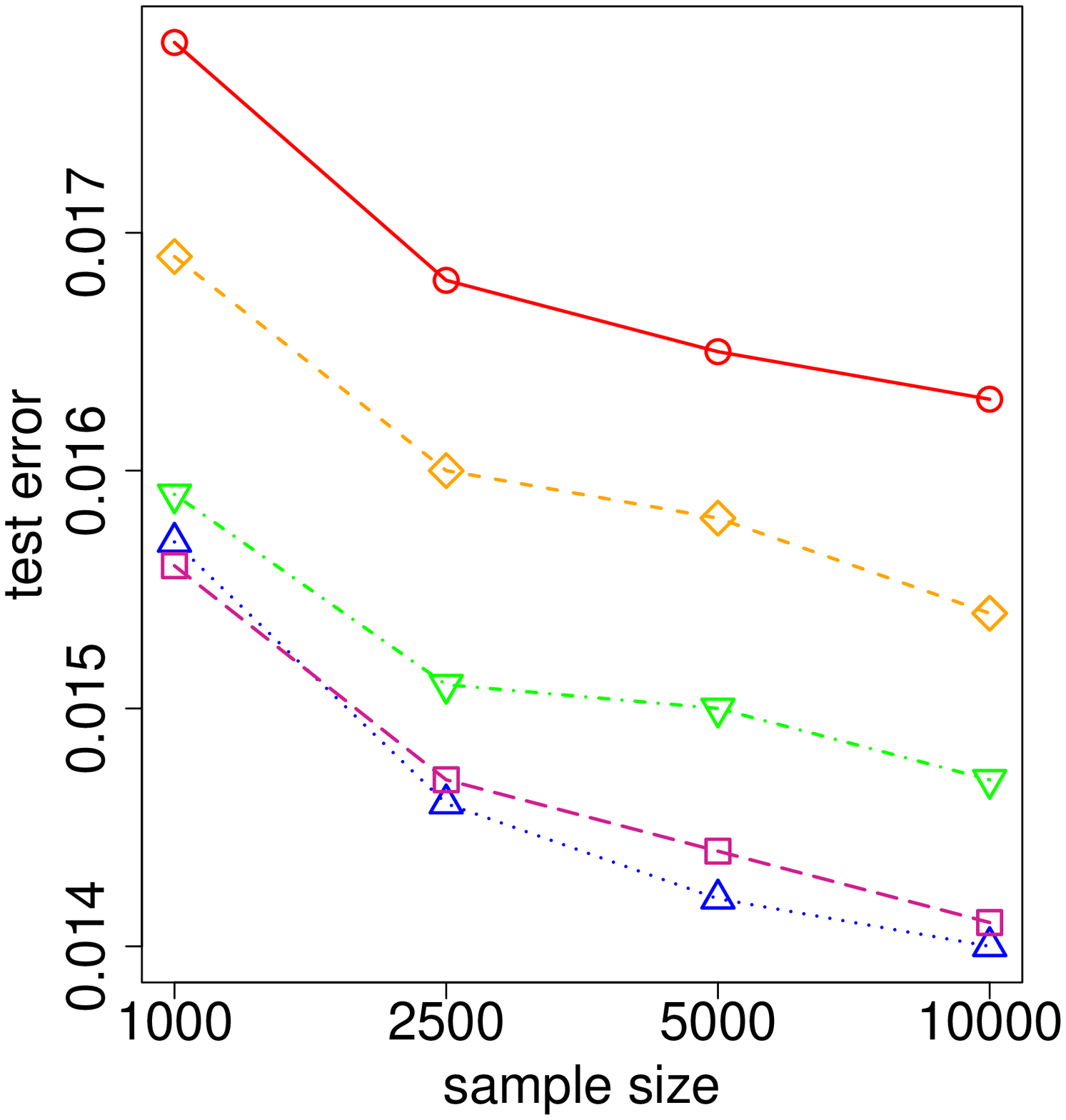}}
\hfill\subfigure[Average empirical $L_2$-error of the various VP-SVMs\label{fig::experiments::art_data_2_2_vp_L2error}]{\includegraphics[angle=270, scale=0.26]{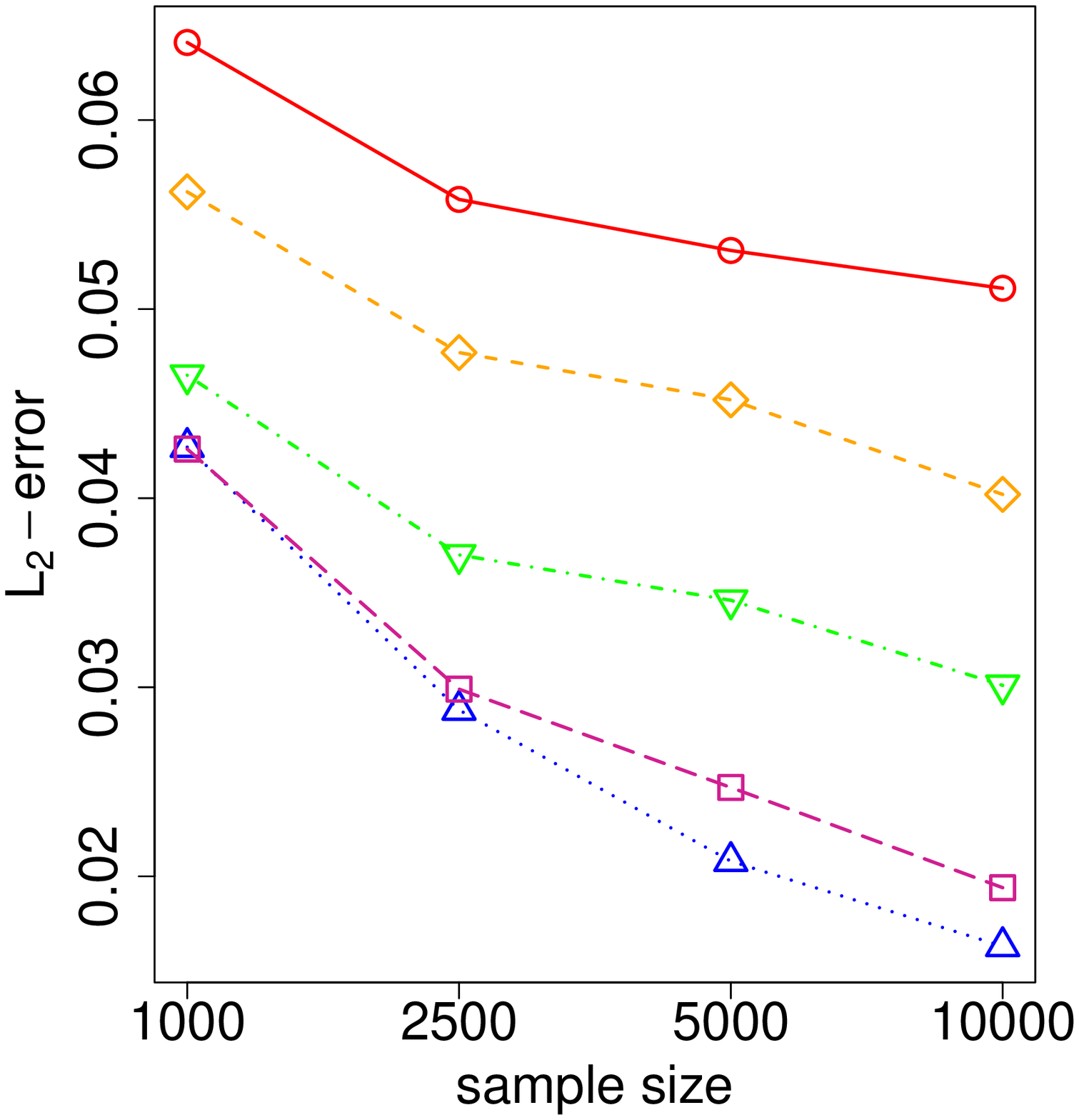}}\vspace{-0.5cm}\\
\subfigure[Average training time of \mbox{LS-,} VP-, and RC-SVMs\label{fig::experiments::art_data_2_2_time}]{\includegraphics[angle=270, scale=0.26]{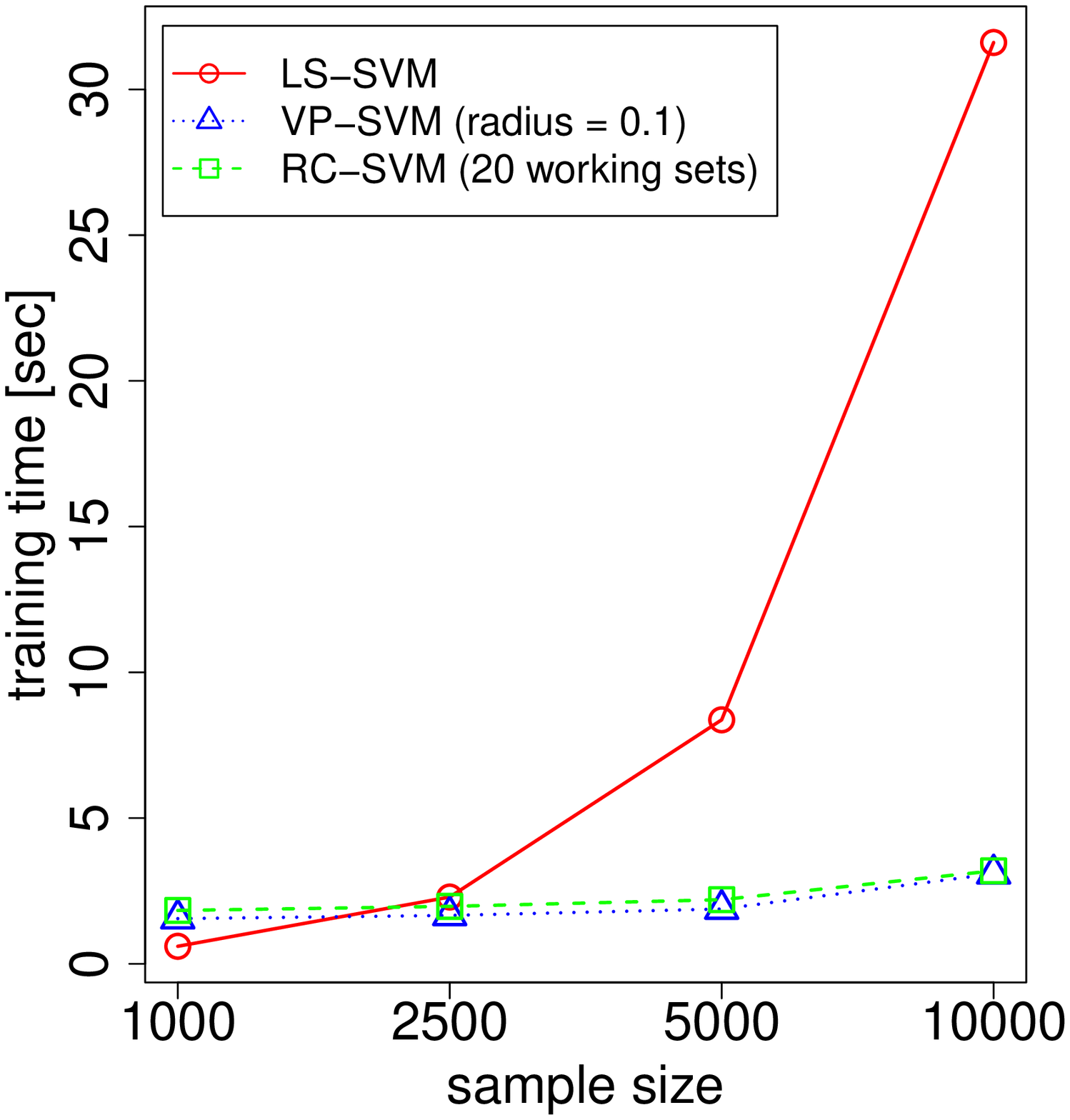}}
\hfill\subfigure[Average empirical risk of \mbox{LS-,} VP-, and RC-SVMs\label{fig::experiments::art_data_2_2_error}]{\includegraphics[angle=270, scale=0.26]{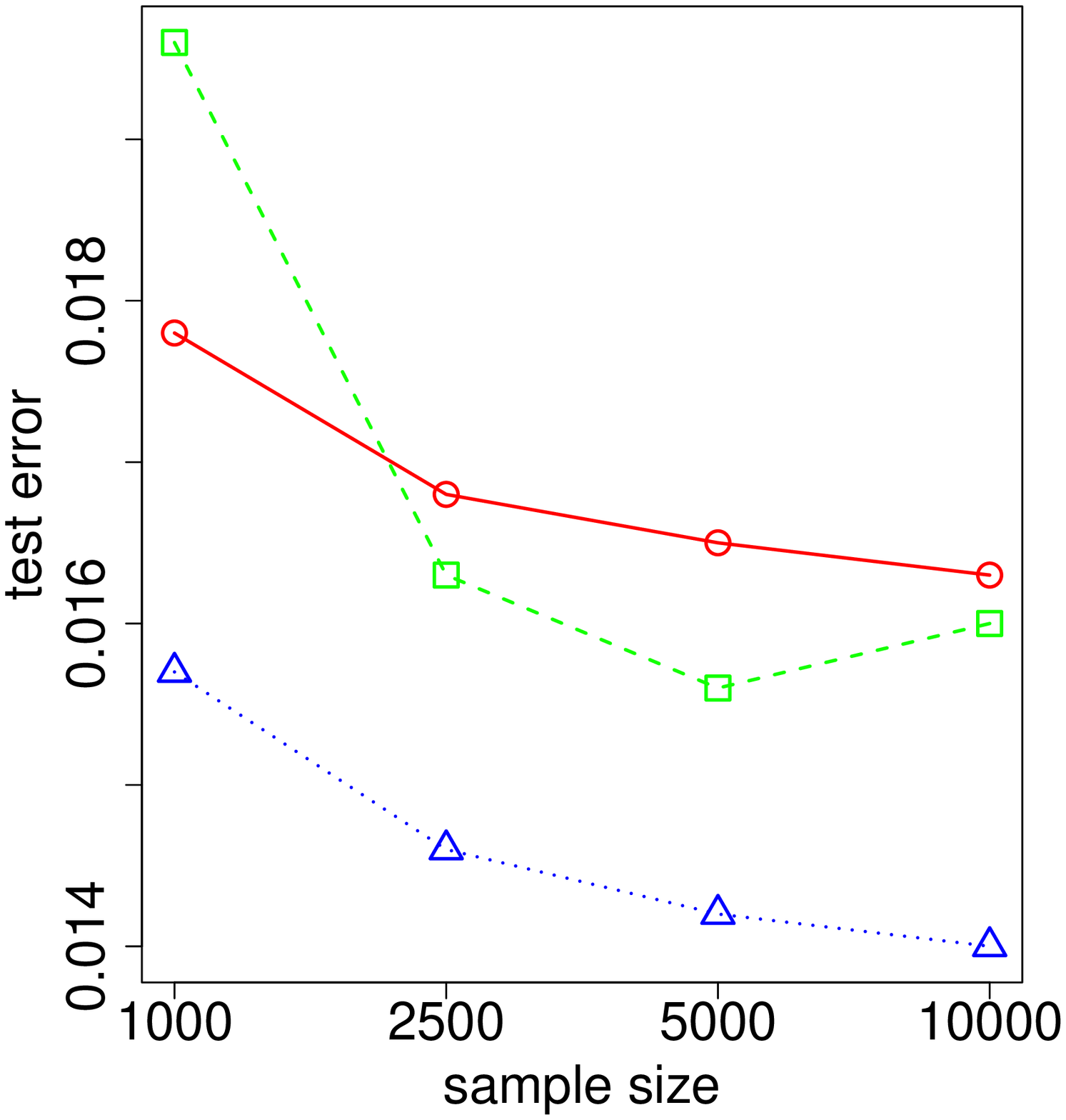}}
\hfill\subfigure[Average empirical $L_2$-error of \mbox{LS-,} VP-, and RC-SVMs\label{fig::experiments::art_data_2_2_L2error}]{\includegraphics[angle=270, scale=0.26]{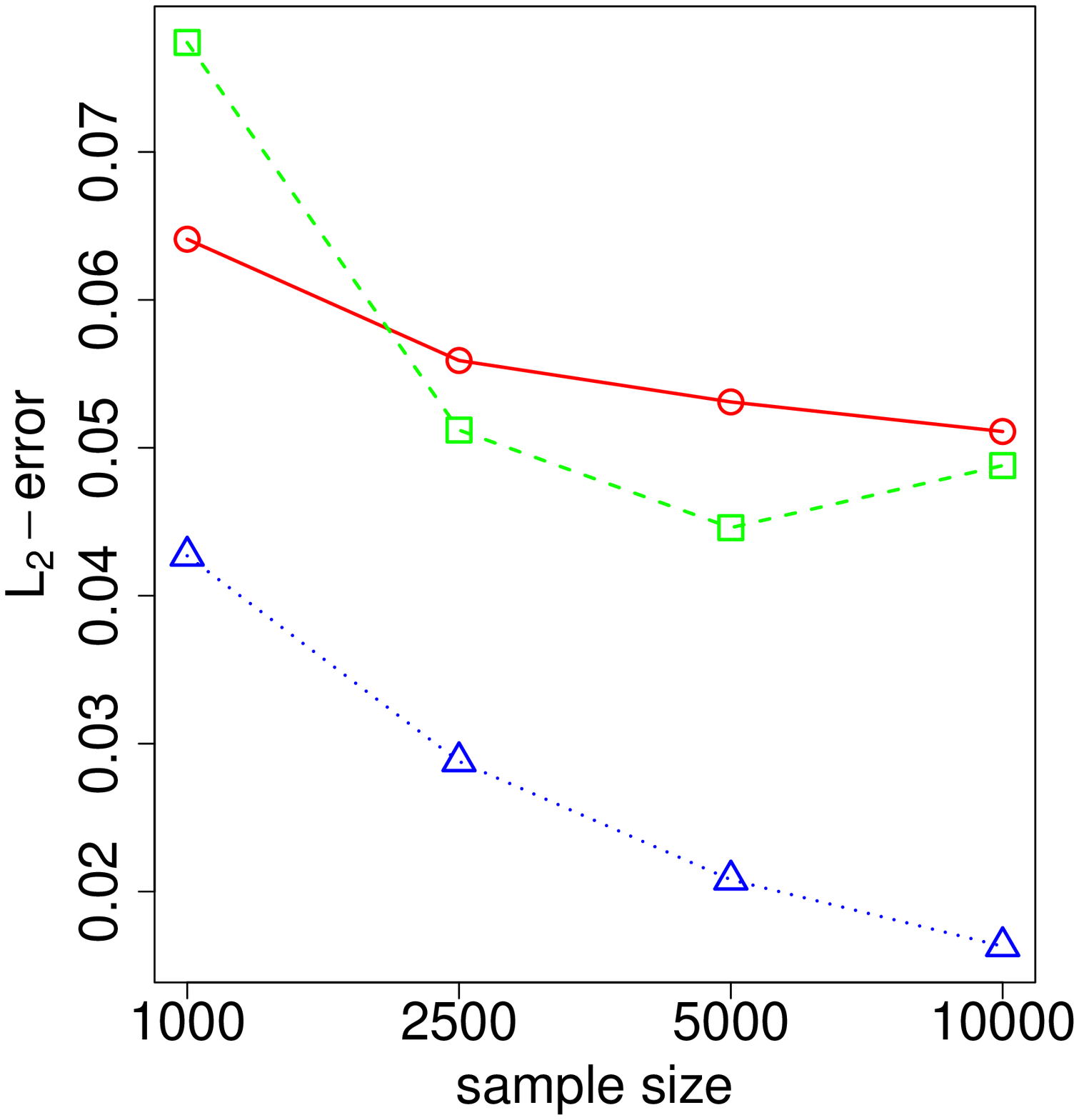}}
\caption{
Average training time and test error of LS-, VP-, and RC-SVMs for the artificial data 
Type \RM{2} depending on the training set size $n_\text{train}=1\,000,\ldots,10\,000$. Subfigures 
\subref{fig::experiments::art_data_2_2_rc_time}--\subref{fig::experiments::art_data_2_2_rc_L2error} 
show the results for RC-SVMs using different numbers of working sets and Subfigures 
\subref{fig::experiments::art_data_2_2_vp_time}--\subref{fig::experiments::art_data_2_2_vp_L2error} 
illustrate the results for VP-SVMs using various radii. At the bottom, Subfigures 
\subref{fig::experiments::art_data_2_2_time}--\subref{fig::experiments::art_data_2_2_L2error} 
contain the average training times and the average test errors of the LS-SVM, one VP-SVM and 
one RC-SVM. Here, the VP-SVM and the RC-SVM are those which train fastest for 
$n_\text{train}=10\,000$. Note that in the case at hand none of the considered RC-SVMs performs 
faster than the fastest VP-SVM for $n_\text{train}=10\,000$.} 
\label{fig::experiments::artificial_data_2_2}
\end{scriptsize}
\end{figure}

\begin{figure}[p]
\begin{scriptsize}\vspace{-0.6cm}
\subfigure[Average training time of the various RC-SVMs\label{fig::experiments::art_data_4_2_rc_time}]{\includegraphics[angle=270, scale=0.26]{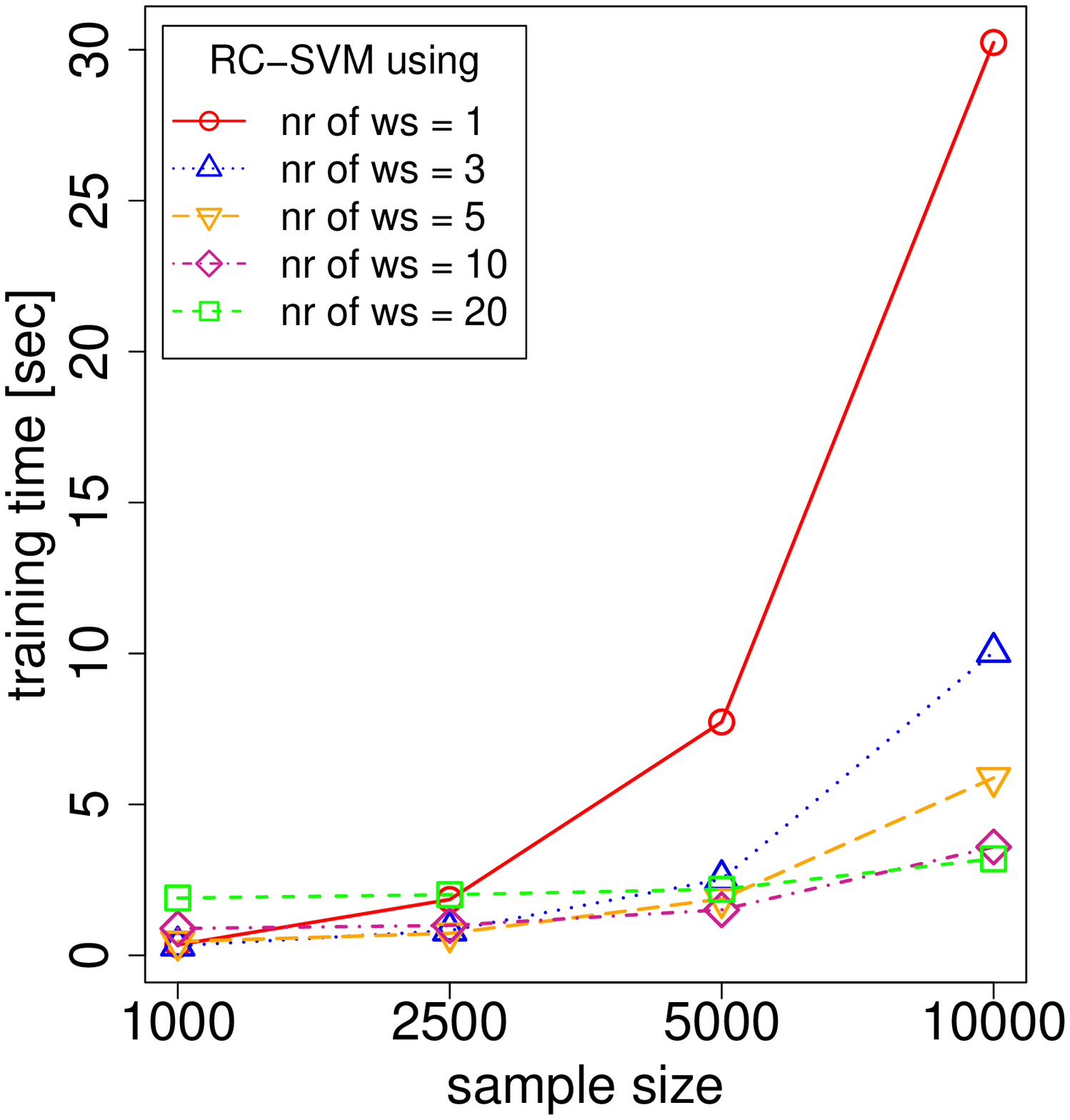}}
\hfill\subfigure[Average empirical risk of the various RC-SVMs\label{fig::experiments::art_data_4_2_rc_error}]{\includegraphics[angle=270, scale=0.26]{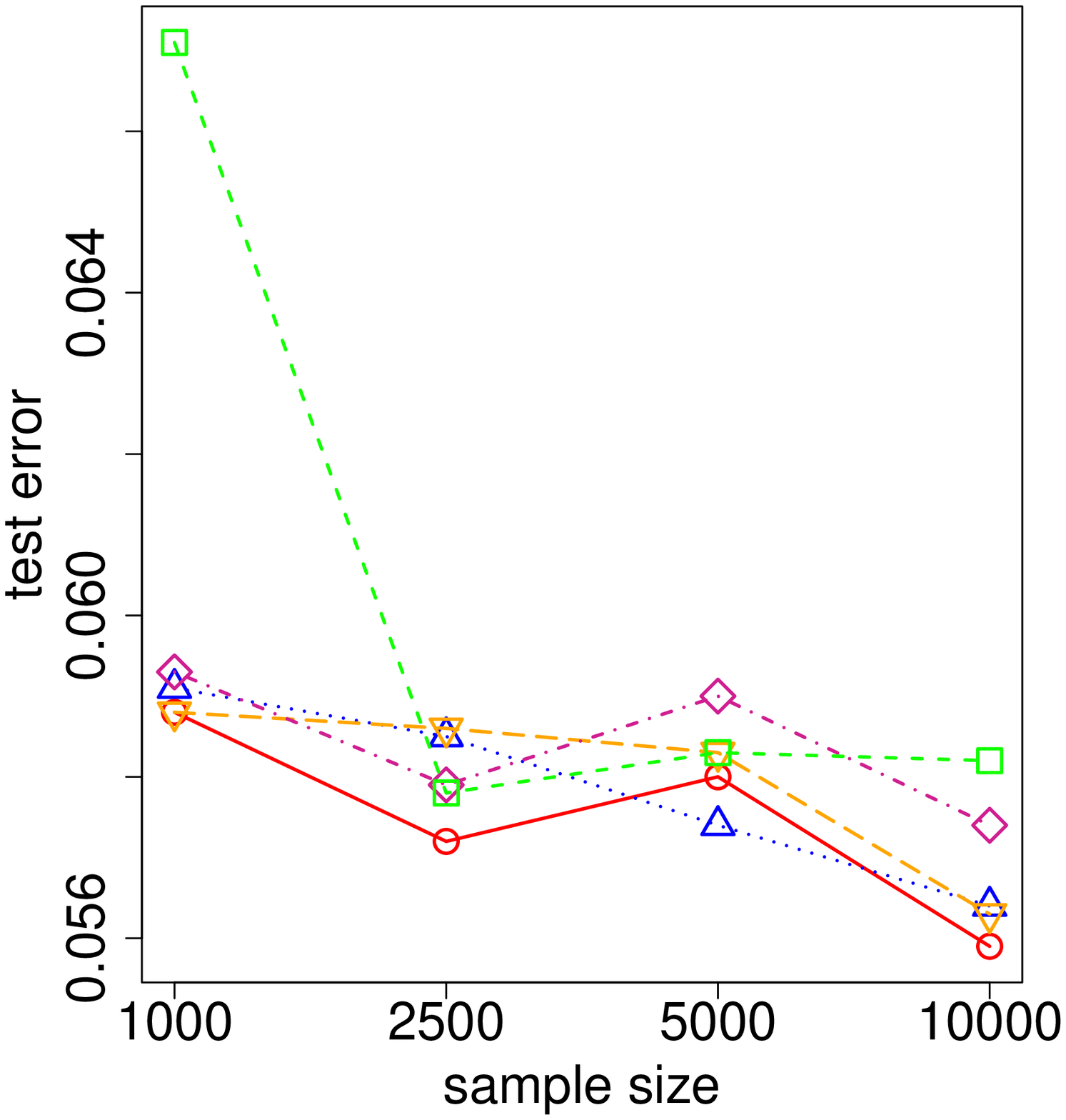}}
\hfill\subfigure[Average empirical $L_2$-error of the various RC-SVMs\label{fig::experiments::art_data_4_2_rc_L2error}]{\includegraphics[angle=270, scale=0.26]{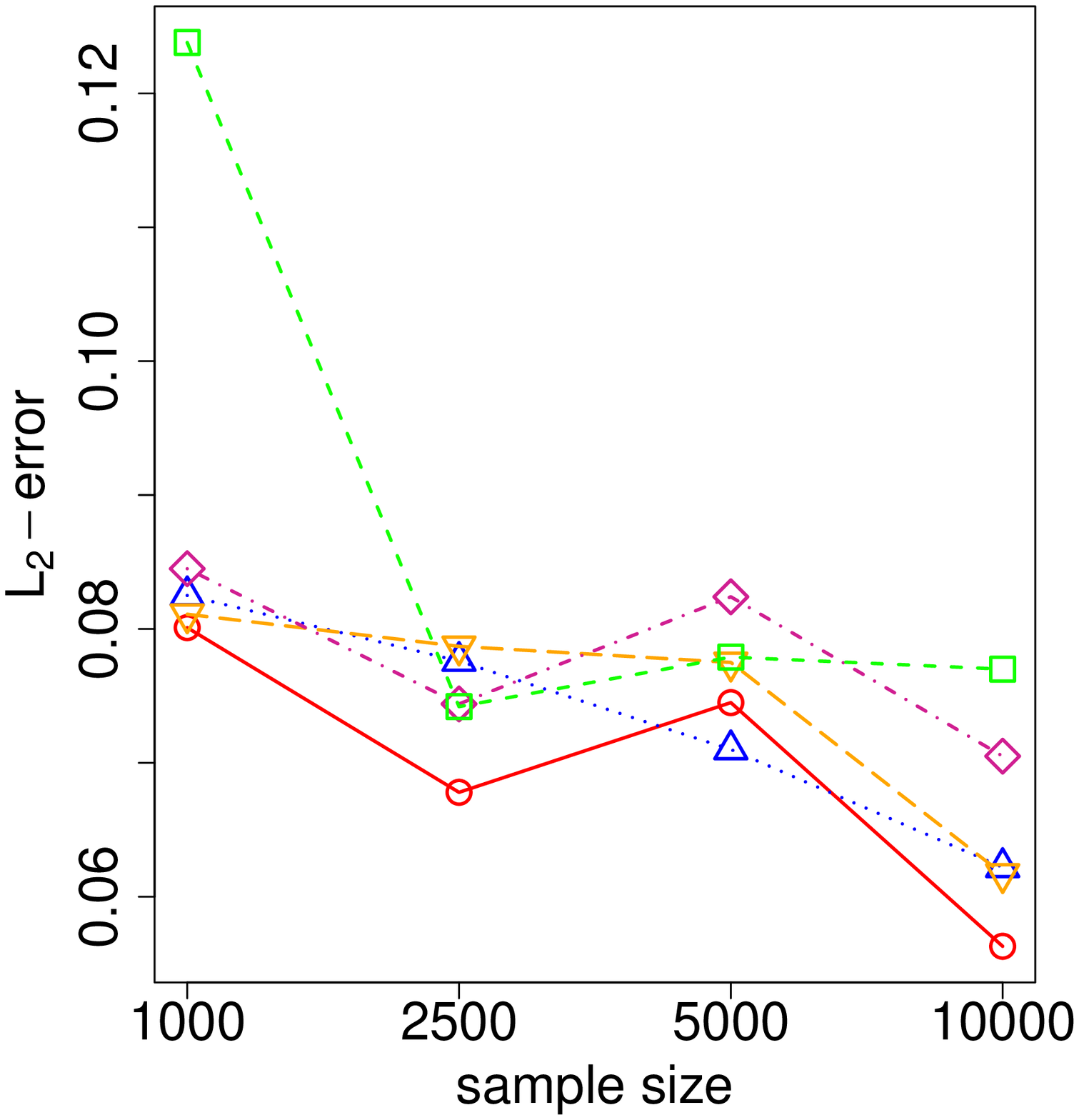}}\vspace{-0.5cm}\\
\subfigure[Average training time of the various VP-SVMs\label{fig::experiments::art_data_4_2_vp_time}]{\includegraphics[angle=270, scale=0.26]{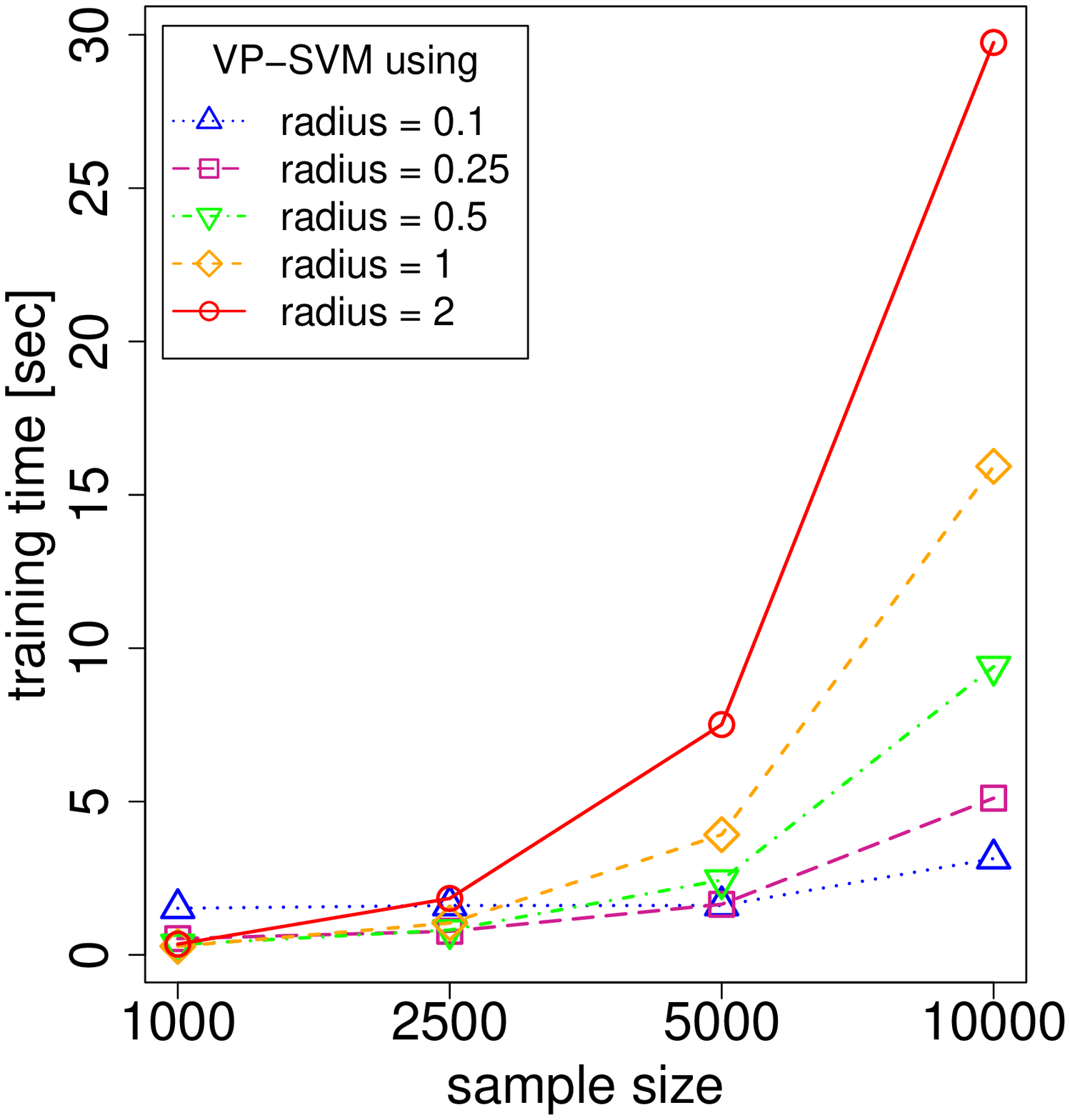}}
\hfill\subfigure[Average empirical risk of the various VP-SVMs\label{fig::experiments::art_data_4_2_vp_error}]{\includegraphics[angle=270, scale=0.26]{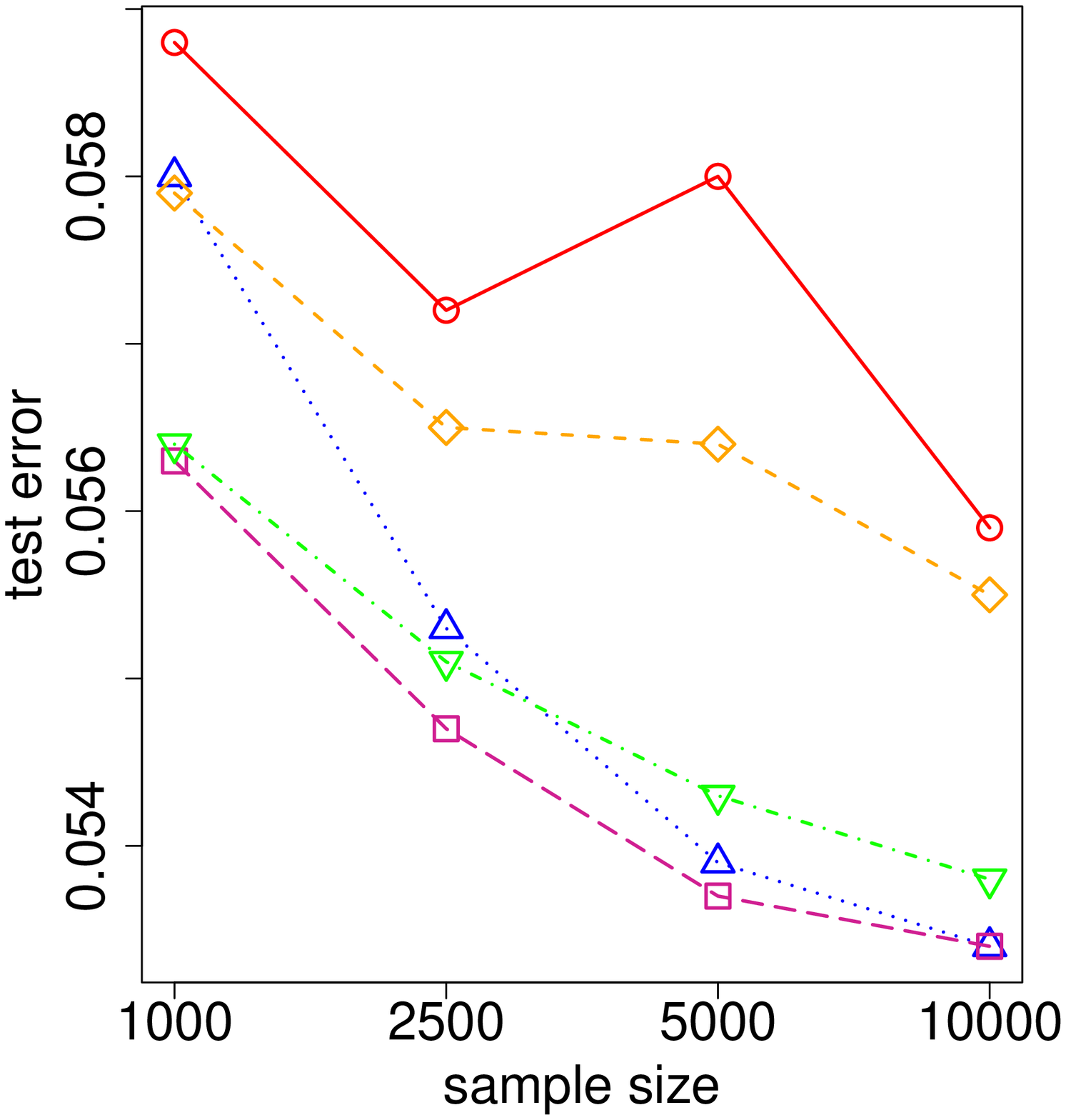}}
\hfill\subfigure[Average empirical $L_2$-error of the various VP-SVMs\label{fig::experiments::art_data_4_2_vp_L2error}]{\includegraphics[angle=270, scale=0.26]{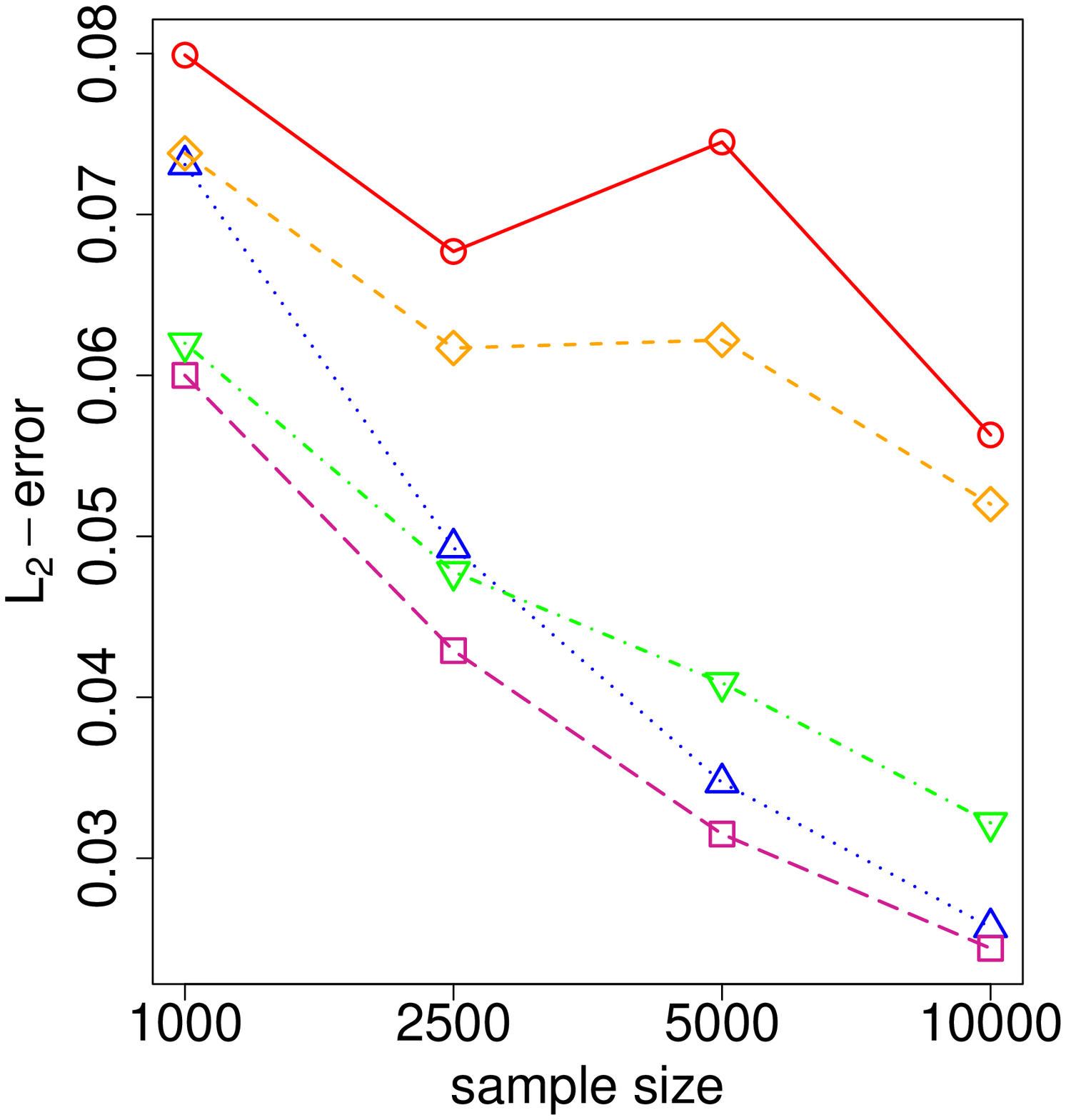}}\vspace{-0.5cm}\\
\subfigure[Average training time of \mbox{LS-,} VP-, and RC-SVMs\label{fig::experiments::art_data_4_2_time}]{\includegraphics[angle=270, scale=0.26]{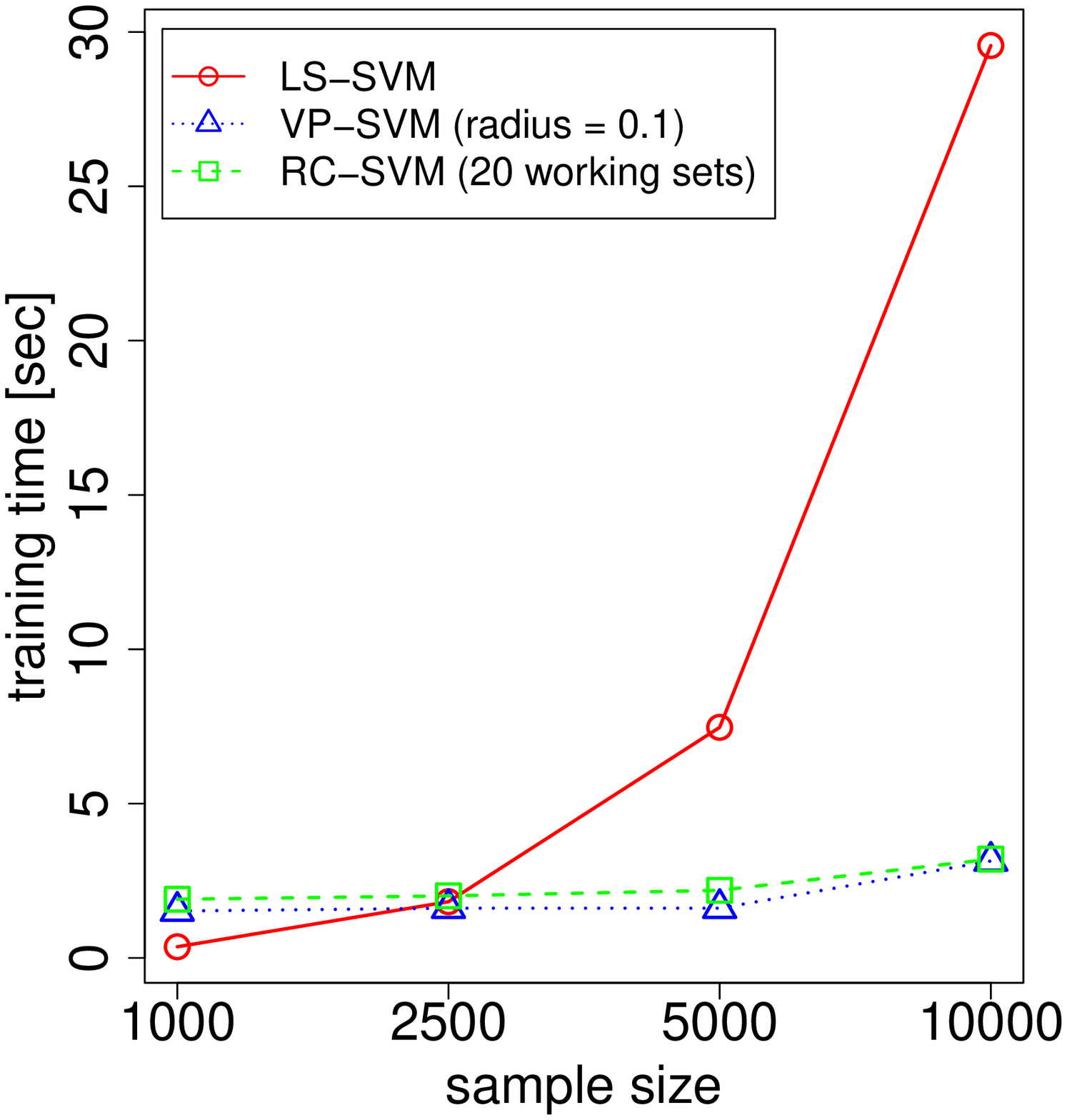}}
\hfill\subfigure[Average empirical risk of \mbox{LS-,} VP-, and RC-SVMs\label{fig::experiments::art_data_4_2_error}]{\includegraphics[angle=270, scale=0.26]{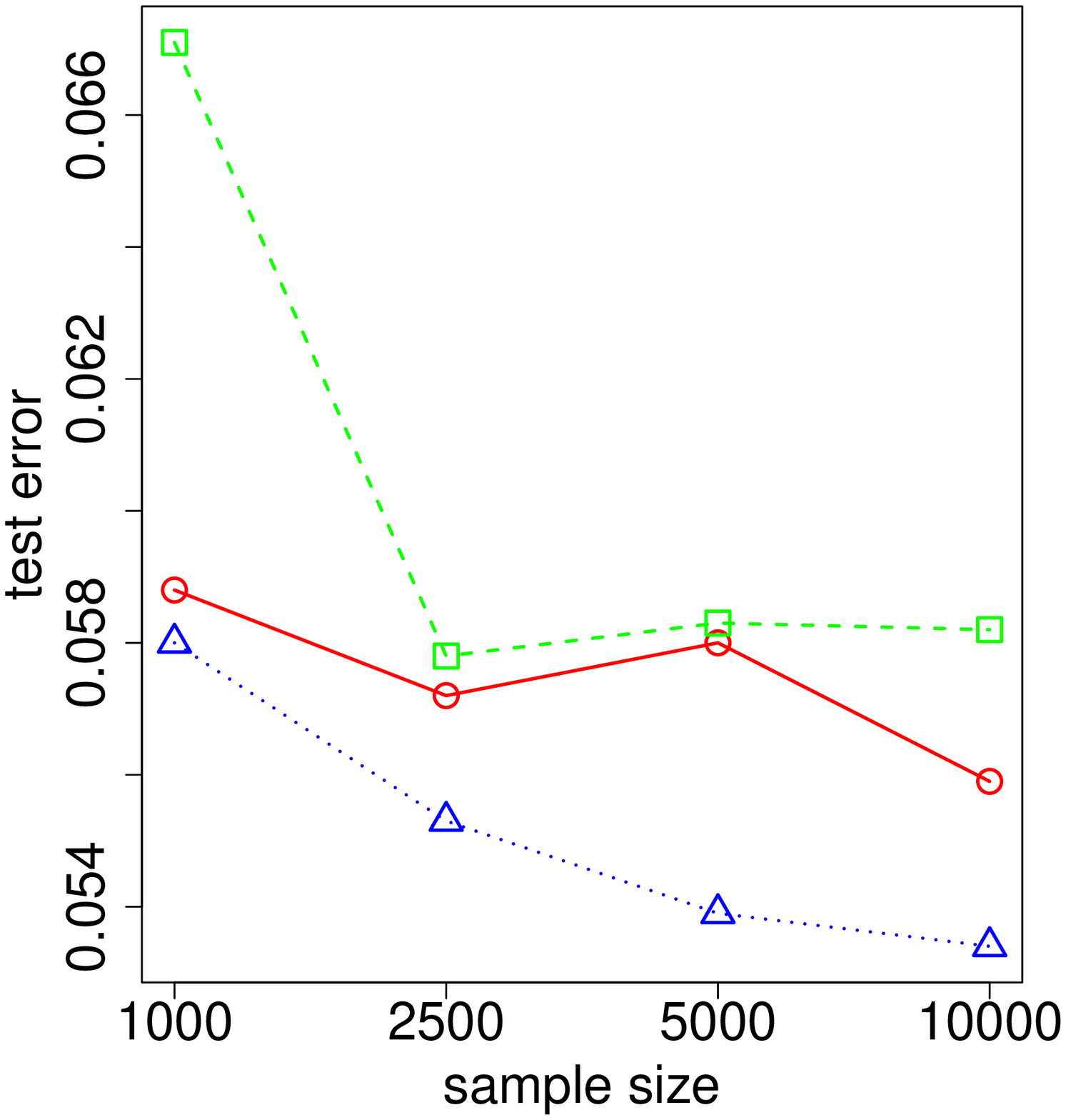}}
\hfill\subfigure[Average empirical $L_2$-error of \mbox{LS-,} VP-, and RC-SVMs\label{fig::experiments::art_data_4_2_L2error}]{\includegraphics[angle=270, scale=0.26]{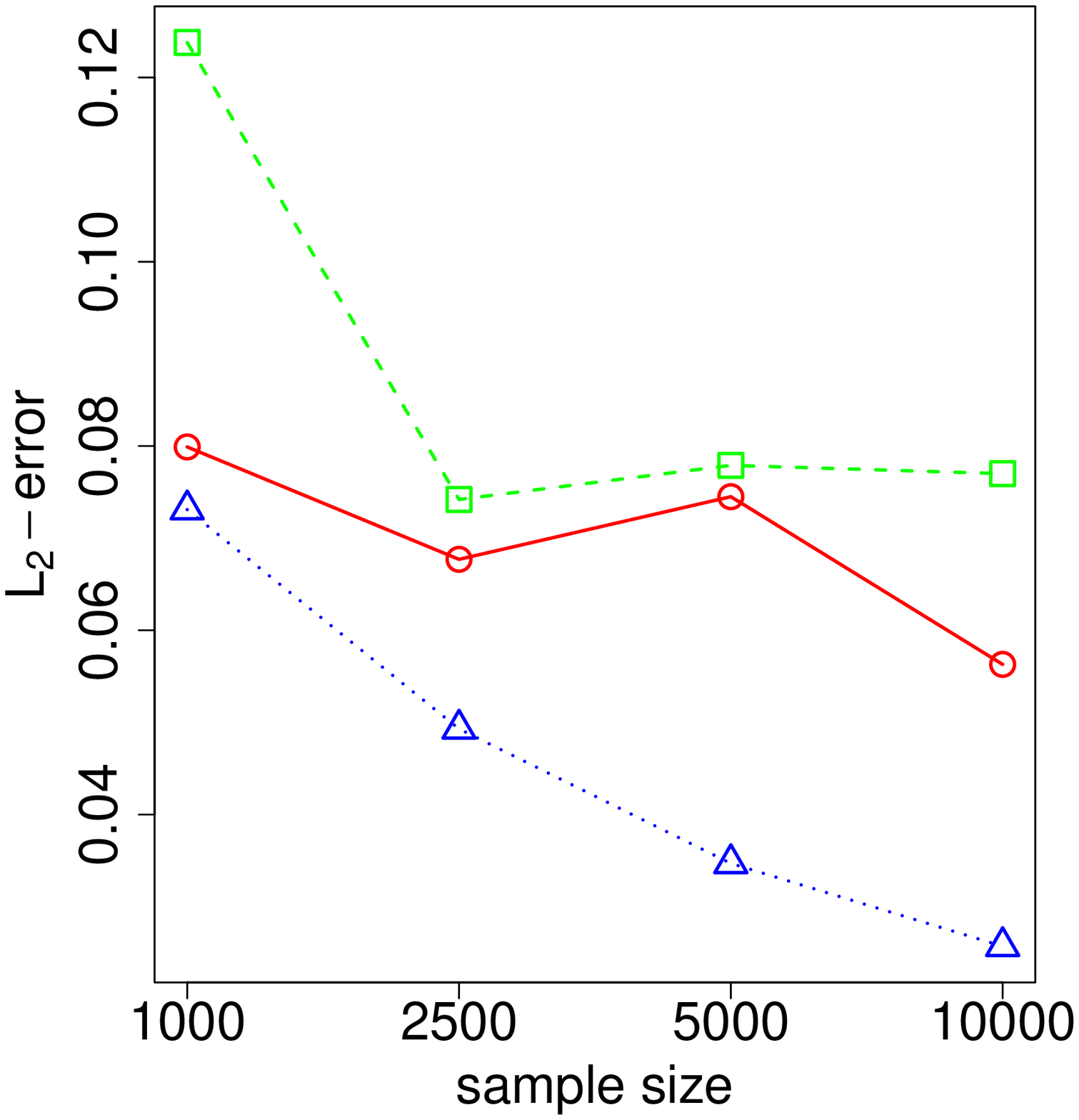}}
\caption{
Average training time and test error of LS-, VP-, and RC-SVMs for the artificial data 
Type \RM{3} depending on the training set size $n_\text{train}=1\,000,\ldots,10\,000$. Subfigures 
\subref{fig::experiments::art_data_4_2_rc_time}--\subref{fig::experiments::art_data_4_2_rc_L2error} 
show the results for RC-SVMs using different numbers of working sets and Subfigures 
\subref{fig::experiments::art_data_4_2_vp_time}--\subref{fig::experiments::art_data_4_2_vp_L2error} 
illustrate the results for VP-SVMs using various radii. At the bottom, Subfigures 
\subref{fig::experiments::art_data_4_2_time}--\subref{fig::experiments::art_data_4_2_L2error} 
contain the average training times and the average test errors of the LS-SVM, one VP-SVM and 
one RC-SVM. Here, the VP-SVM and the RC-SVM are those which train fastest for 
$n_\text{train}=10\,000$. Note that in the case at hand none of the considered RC-SVMs performs 
faster than the fastest VP-SVM for $n_\text{train}=10\,000$.} 
\label{fig::experiments::artificial_data_4_2}
\end{scriptsize}
\end{figure}

\begin{figure}[p]
\begin{scriptsize}\vspace{-0.6cm}
\subfigure[Average training time of the various RC-SVMs\label{fig::experiments::art_data_3_2_rc_time}]{\includegraphics[angle=270, scale=0.26]{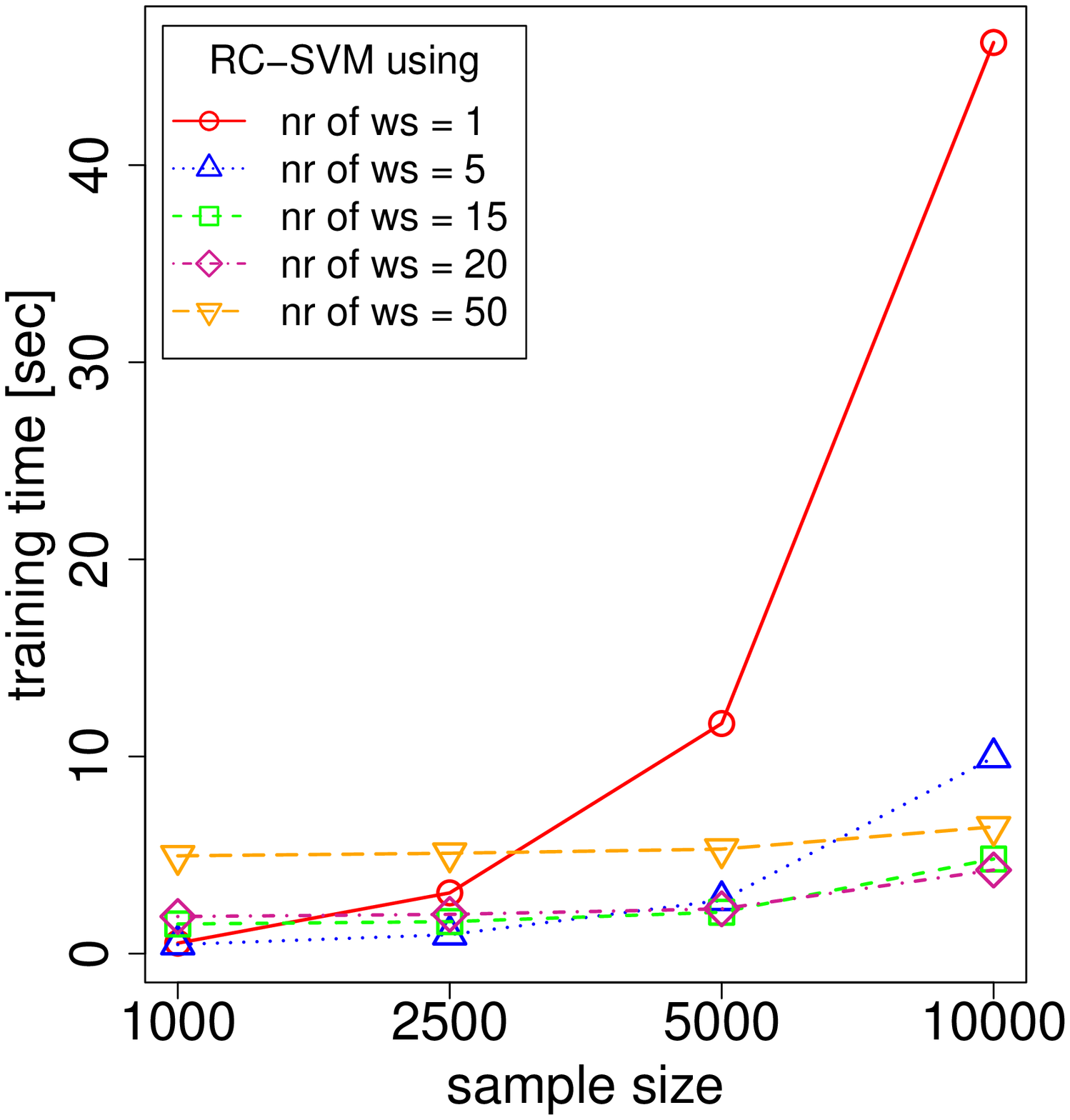}}
\hfill\subfigure[Average empirical risk of the various RC-SVMs\label{fig::experiments::art_data_3_2_rc_error}]{\includegraphics[angle=270, scale=0.26]{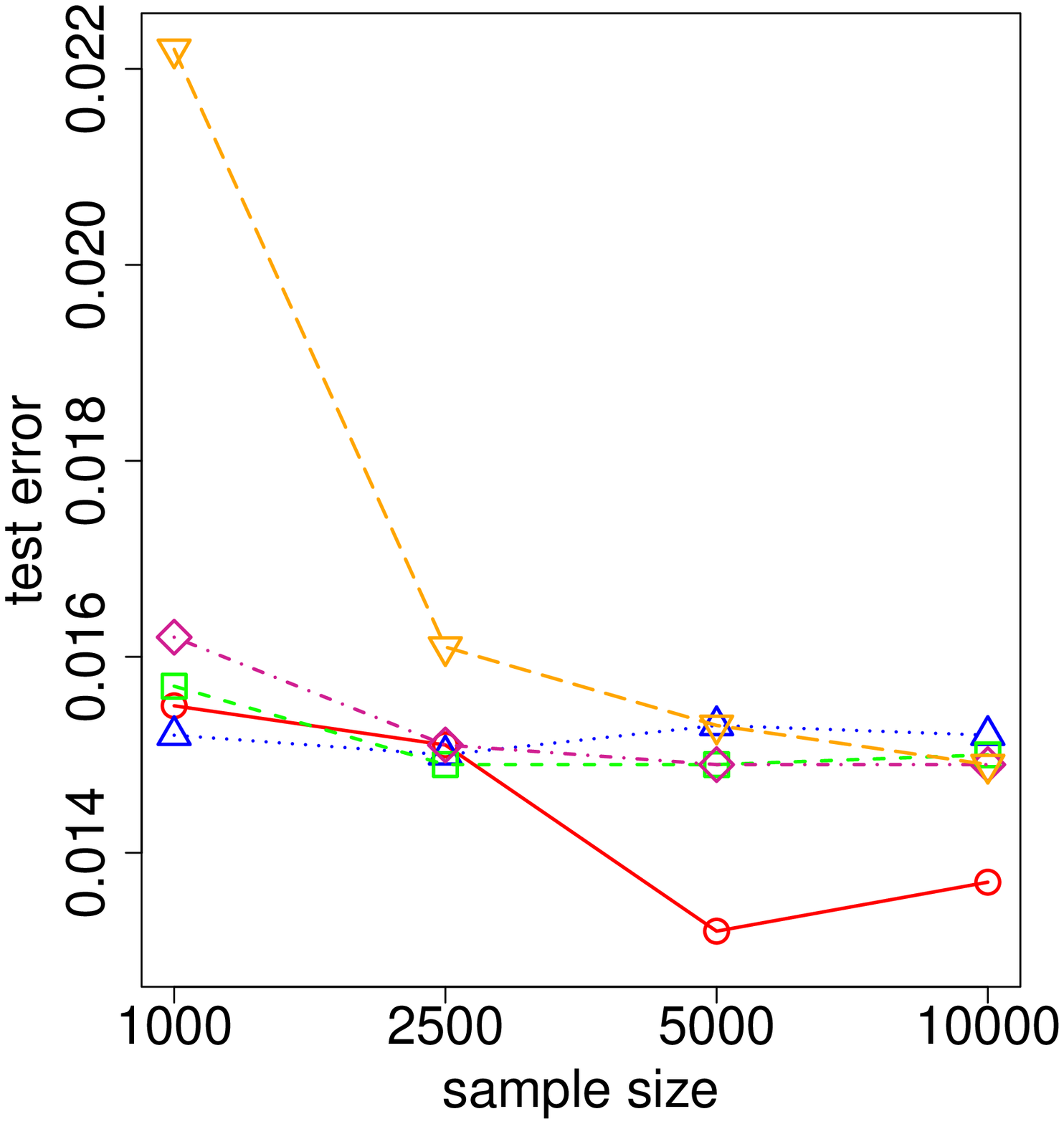}}
\hfill\subfigure[Average empirical $L_2$-error of the various RC-SVMs\label{fig::experiments::art_data_3_2_rc_L2error}]{\includegraphics[angle=270, scale=0.26]{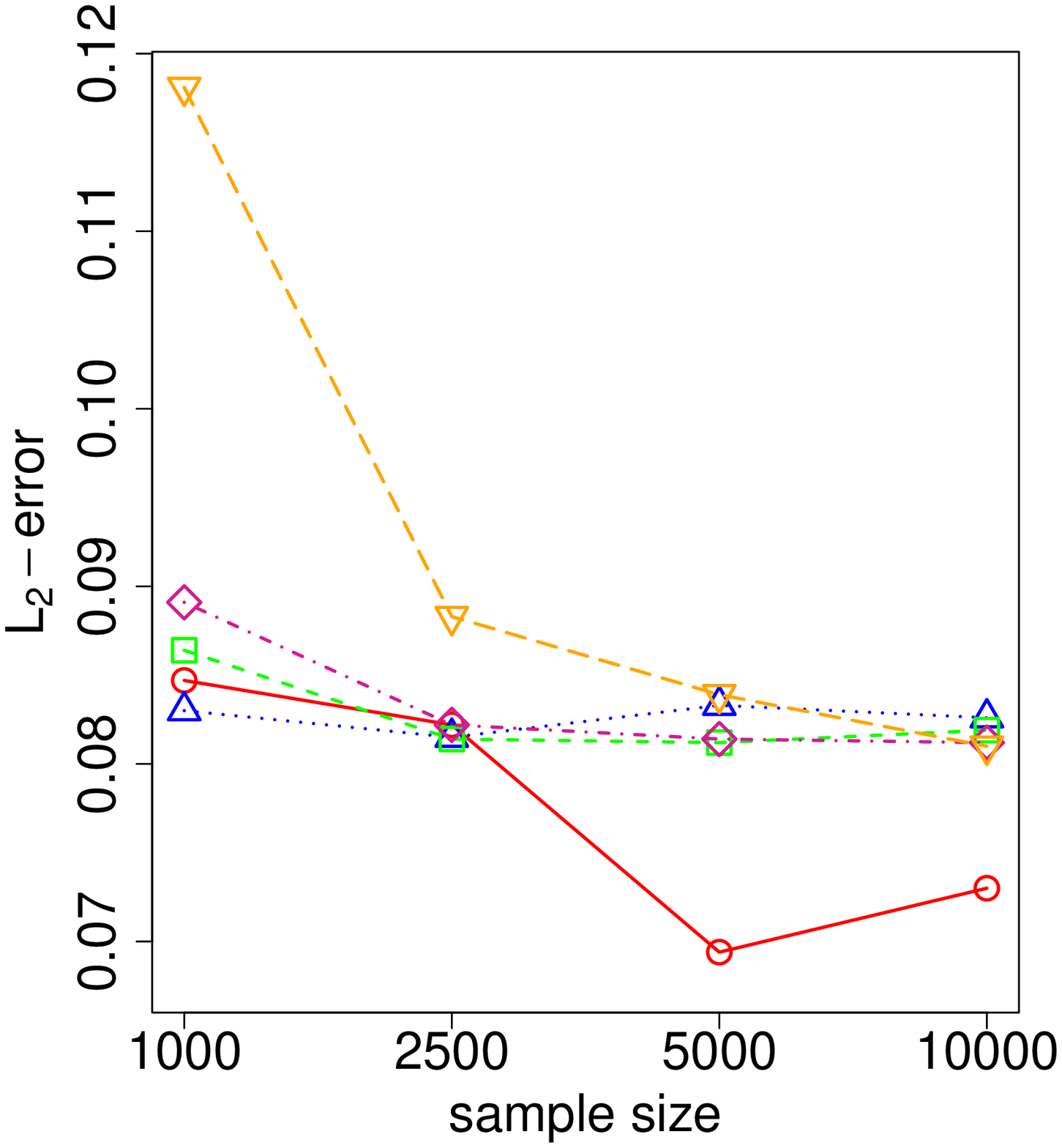}}\vspace{-0.5cm}\\
\subfigure[Average training time of the various VP-SVMs\label{fig::experiments::art_data_3_2_vp_time}]{\includegraphics[angle=270, scale=0.26]{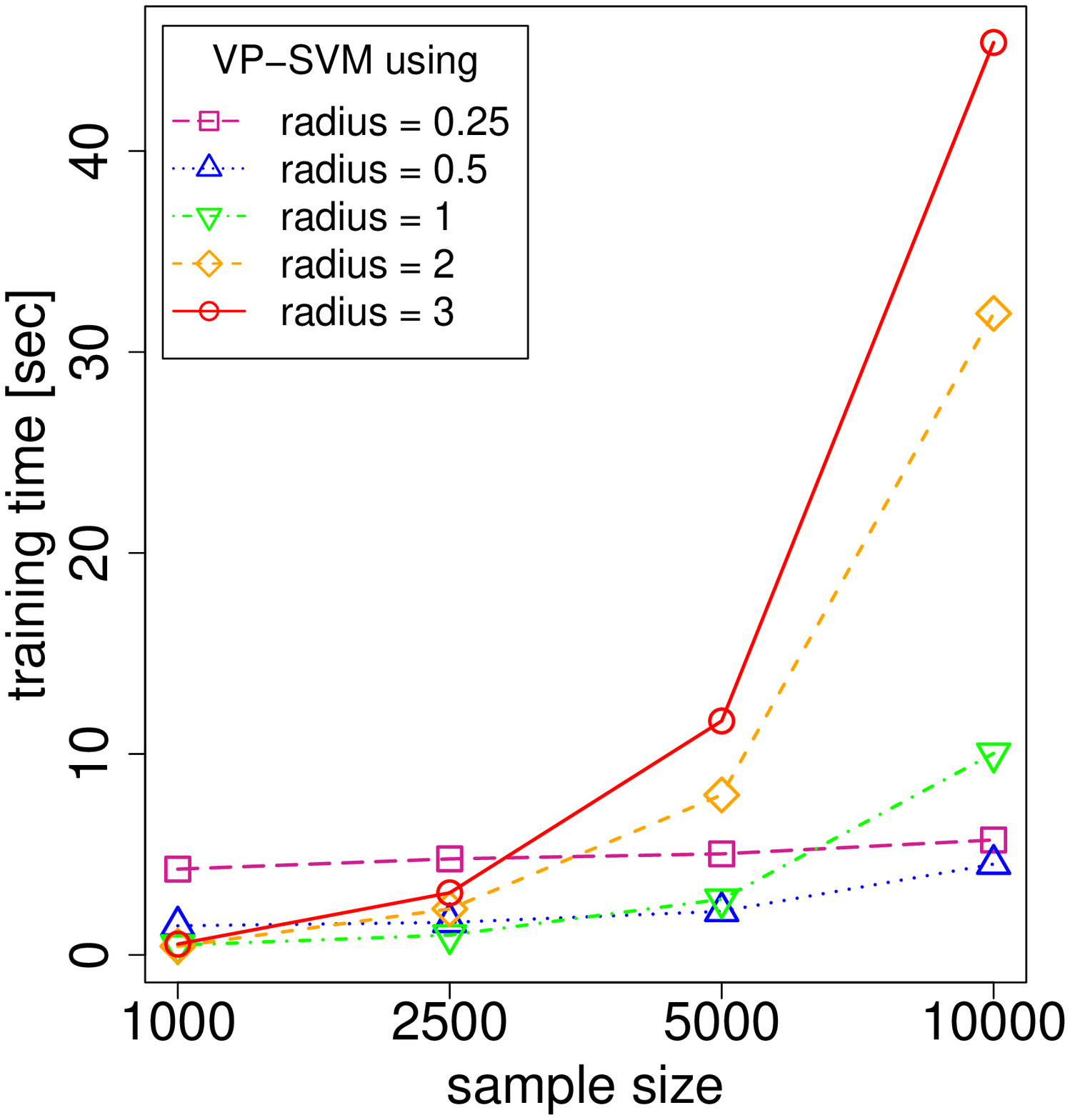}}
\hfill\subfigure[Average empirical risk of the various VP-SVMs\label{fig::experiments::art_data_3_2_vp_error}]{\includegraphics[angle=270, scale=0.26]{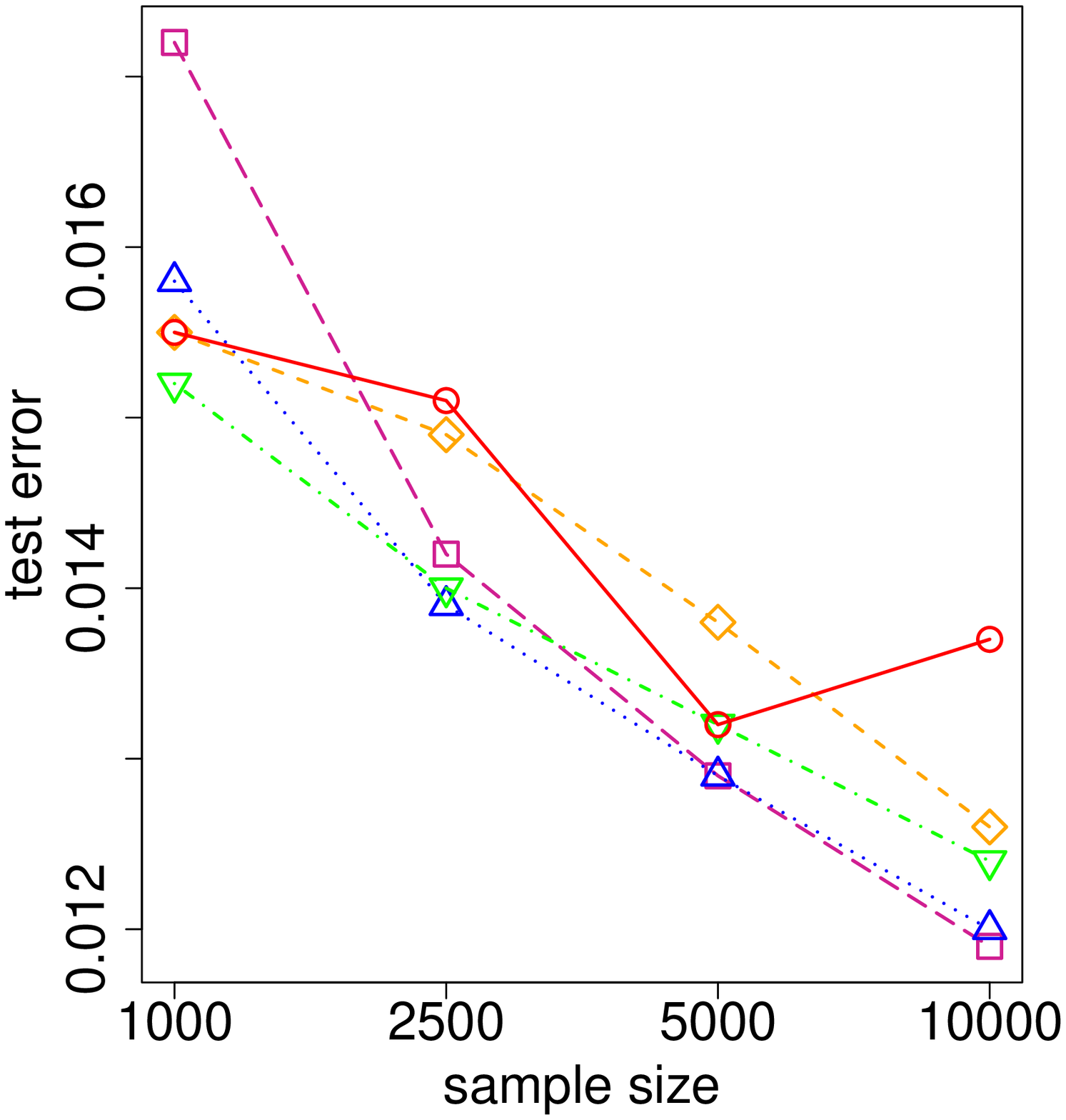}}
\hfill\subfigure[Average empirical $L_2$-error of the various VP-SVMs\label{fig::experiments::art_data_3_2_vp_L2error}]{\includegraphics[angle=270, scale=0.26]{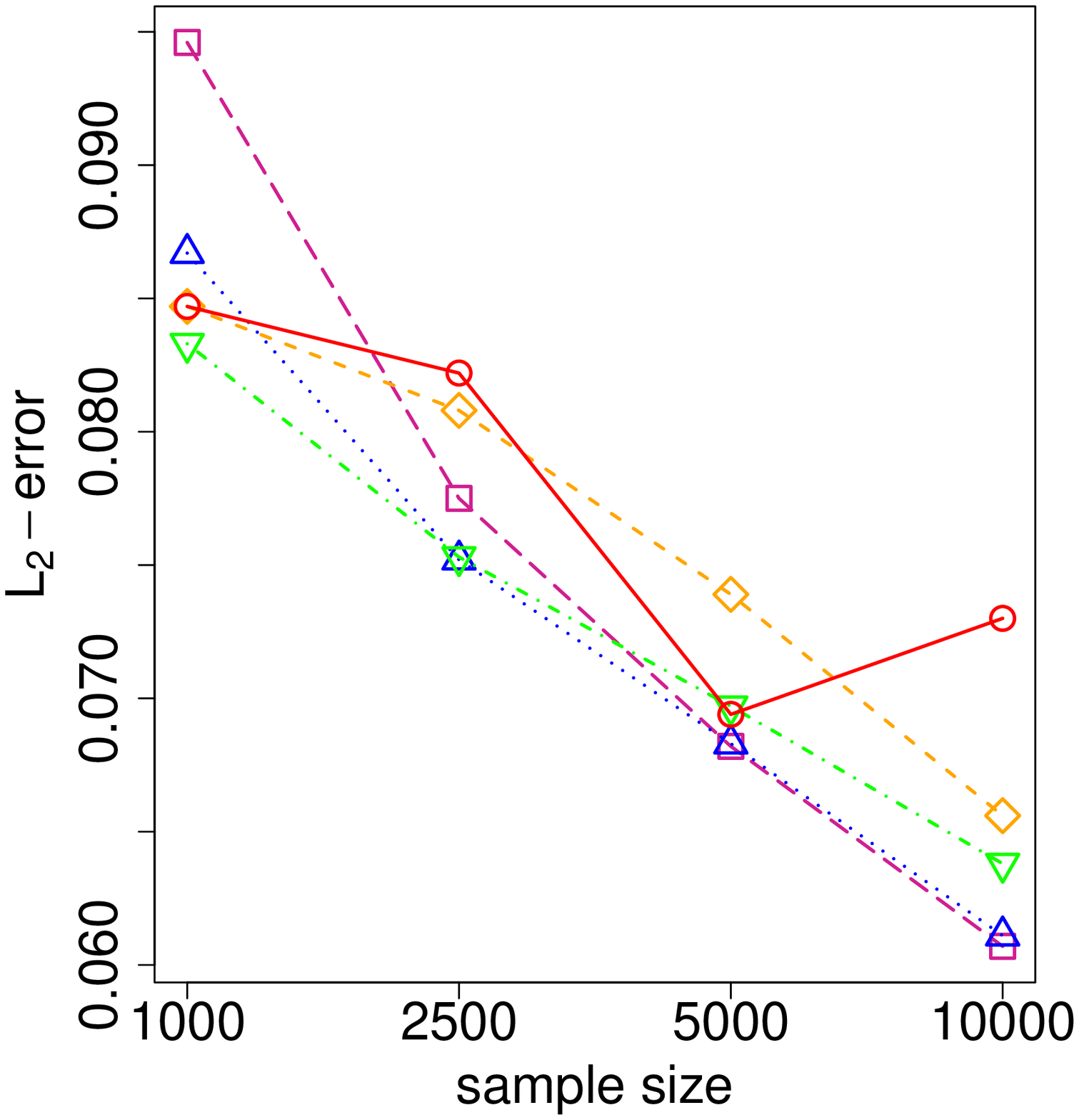}}\vspace{-0.5cm}\\
\subfigure[Average training time of \mbox{LS-,} VP-, and RC-SVMs\label{fig::experiments::art_data_3_2_time}]{\includegraphics[angle=270, scale=0.26]{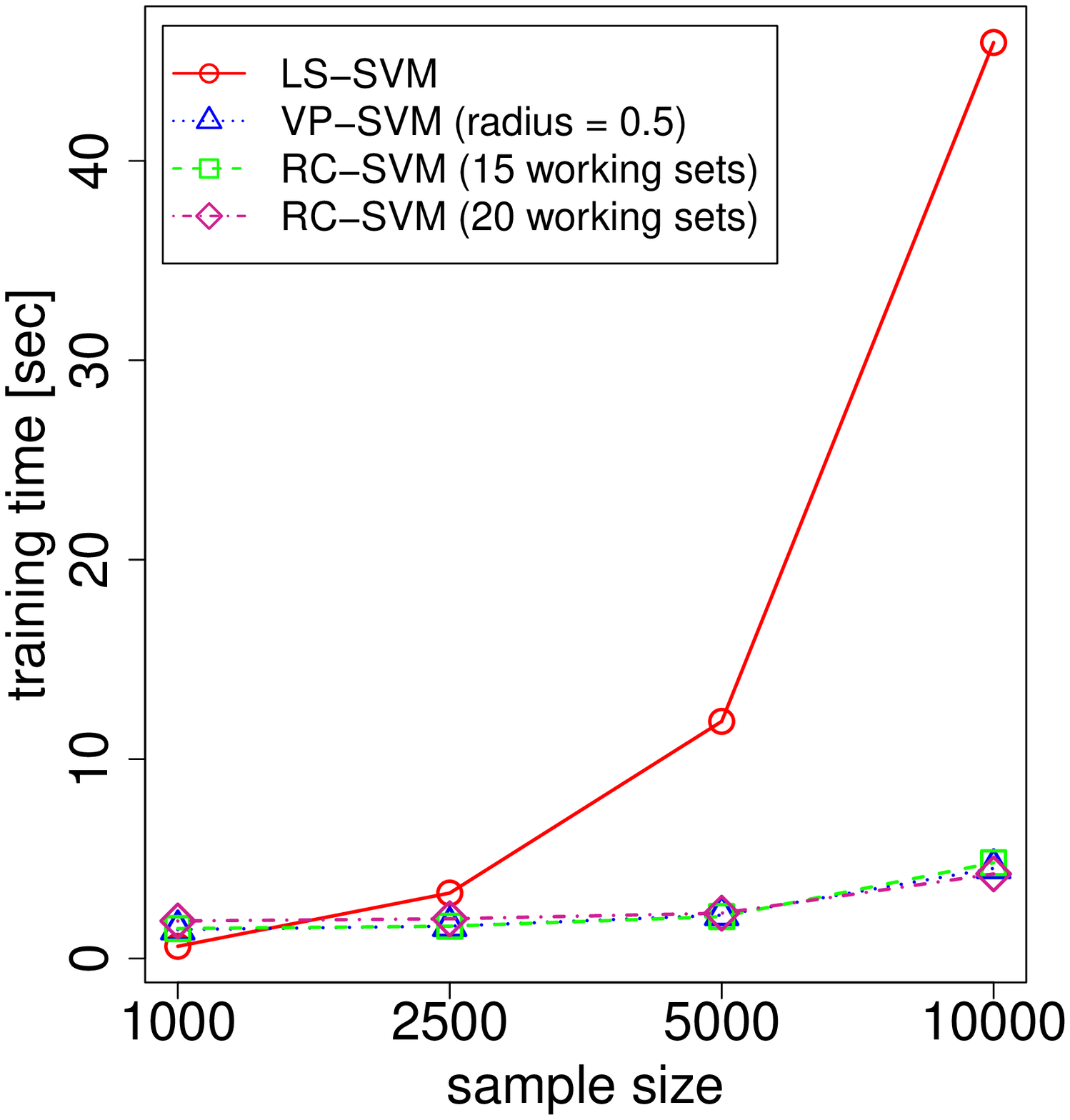}}
\hfill\subfigure[Average empirical risk of \mbox{LS-,} VP-, and RC-SVMs\label{fig::experiments::art_data_3_2_error}]{\includegraphics[angle=270, scale=0.26]{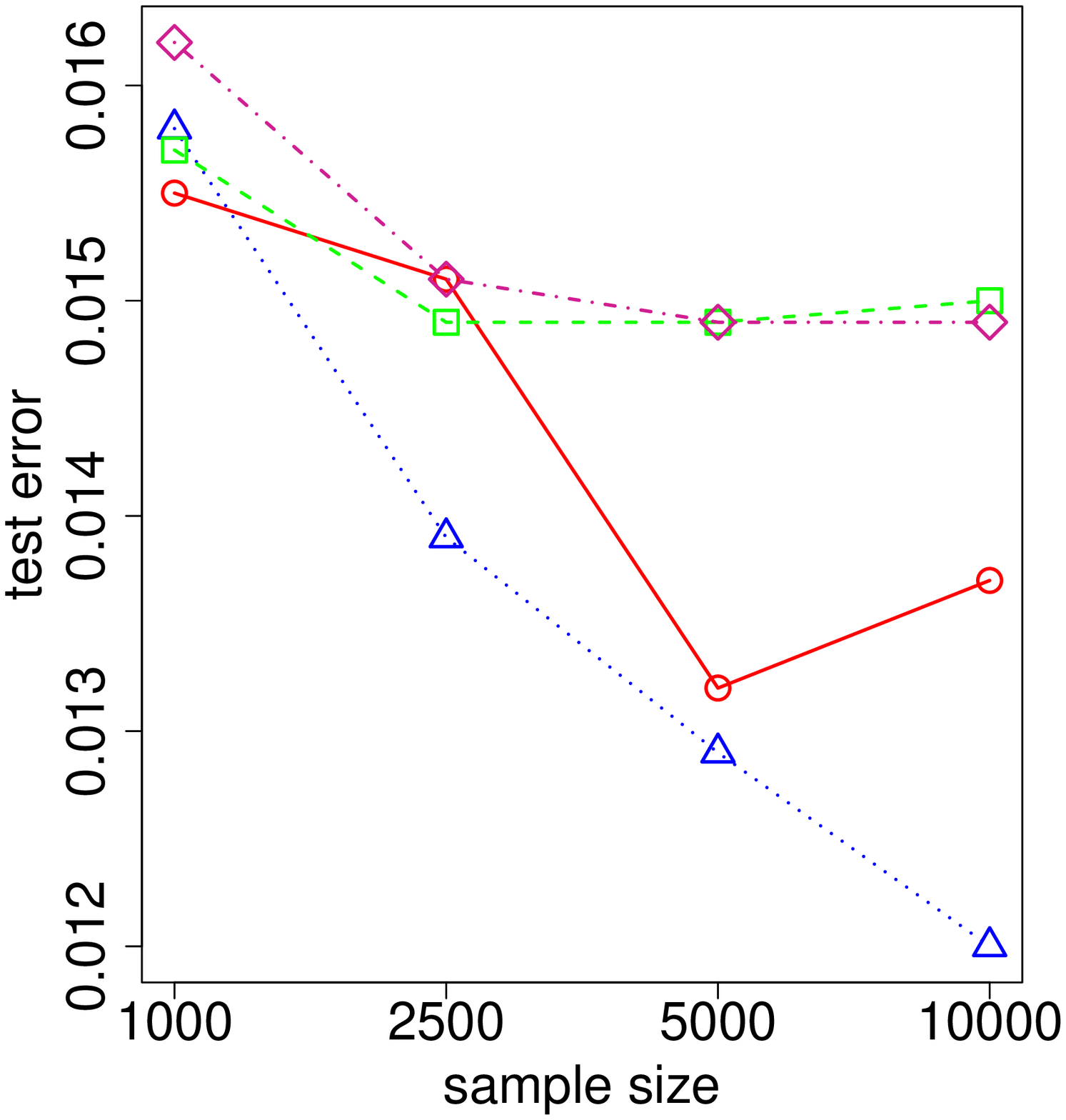}}
\hfill\subfigure[Average empirical $L_2$-error of \mbox{LS-,} VP-, and RC-SVMs\label{fig::experiments::art_data_3_2_L2error}]{\includegraphics[angle=270, scale=0.26]{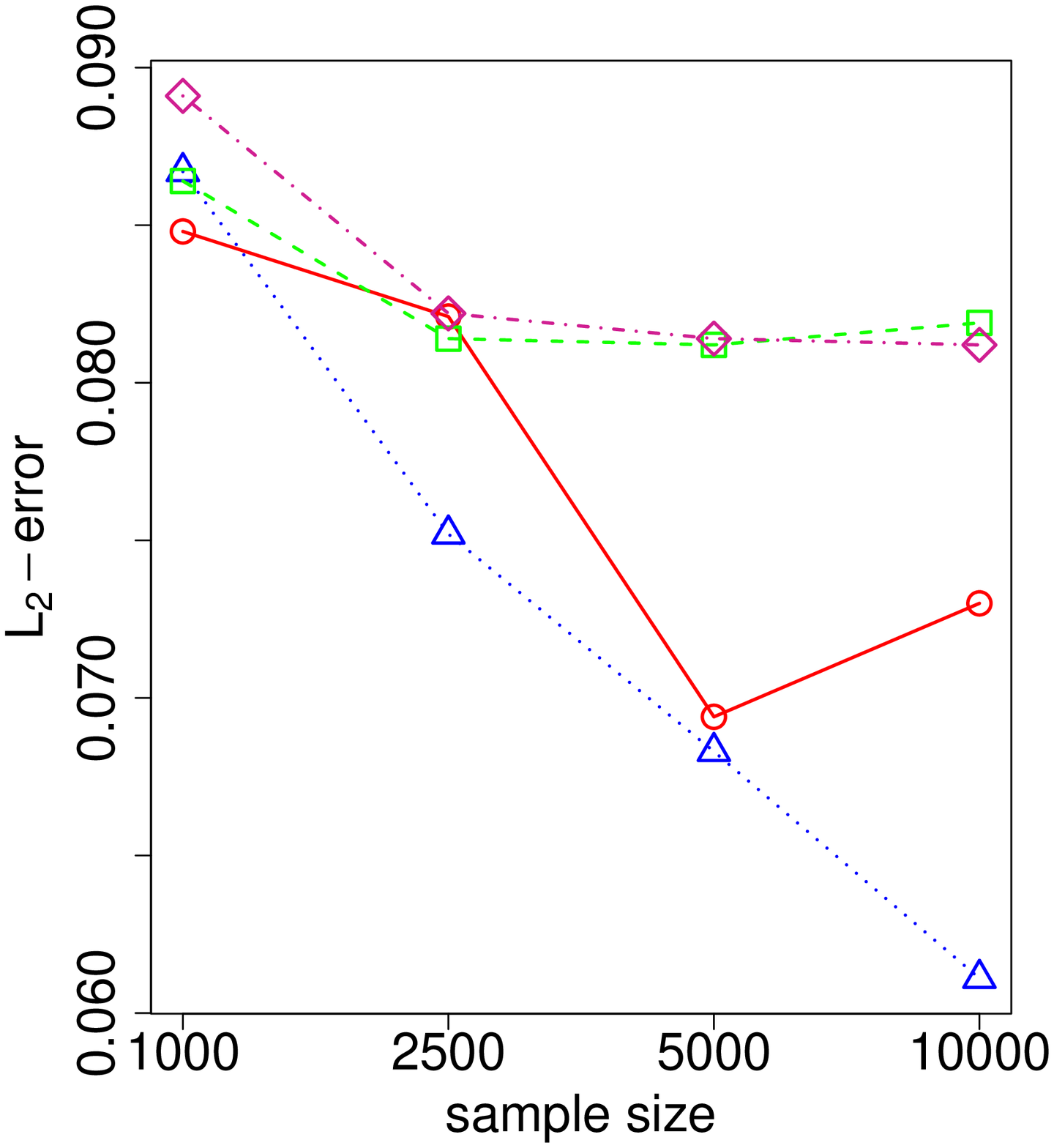}}
\caption{
Average training time and test error of LS-, VP-, and RC-SVMs for the artificial data 
Type \RM{4} depending on the training set size $n_\text{train}=1\,000,\ldots,10\,000$. Subfigures 
\subref{fig::experiments::art_data_3_2_rc_time}--\subref{fig::experiments::art_data_3_2_rc_L2error} 
show the results for RC-SVMs using different numbers of working sets and Subfigures 
\subref{fig::experiments::art_data_3_2_vp_time}--\subref{fig::experiments::art_data_3_2_vp_L2error} 
illustrate the results for VP-SVMs using various radii. At the bottom, Subfigures 
\subref{fig::experiments::art_data_3_2_time}--\subref{fig::experiments::art_data_3_2_L2error} 
contain the average training times and the average test errors of the LS-SVM, one VP-SVM and 
two RC-SVMs. Here, the VP-SVM is the one which trains fastest for $n_\text{train}=10\,000$
and the two RC-SVMs are those which achieve for $n_\text{train}=10\,000$ roughly the same 
training time as the chosen VP-SVM.}
\label{fig::experiments::artificial_data_3_2}
\end{scriptsize}
\end{figure}

\begin{figure}[p]
\begin{scriptsize}\vspace{-0.6cm}
\subfigure[Average training time of the various RC-SVMs\label{fig::experiments::art_data_5_2_rc_time}]{\includegraphics[angle=270, scale=0.26]{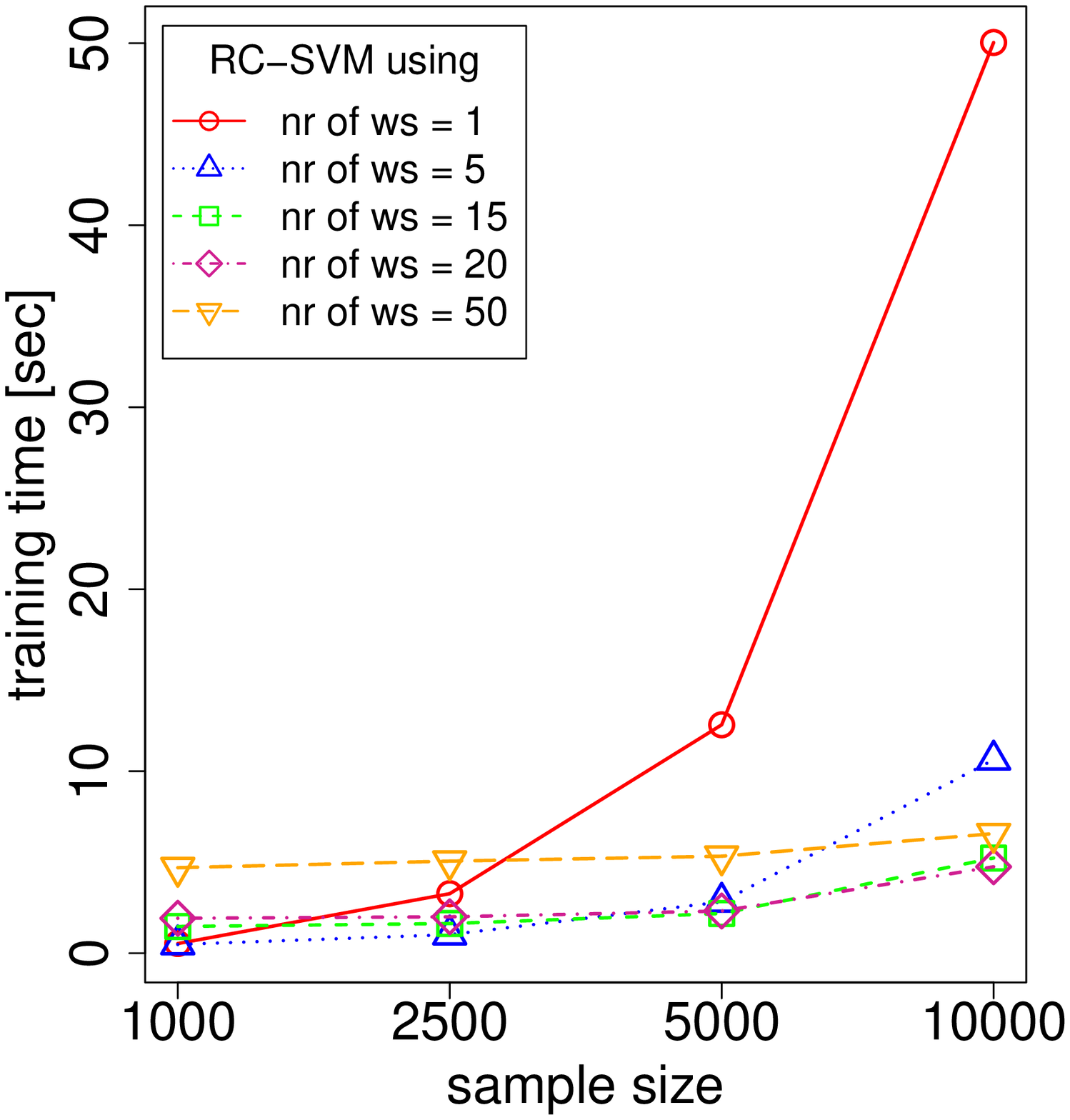}}
\hfill\subfigure[Average empirical risk of the various RC-SVMs\label{fig::experiments::art_data_5_2_rc_error}]{\includegraphics[angle=270, scale=0.26]{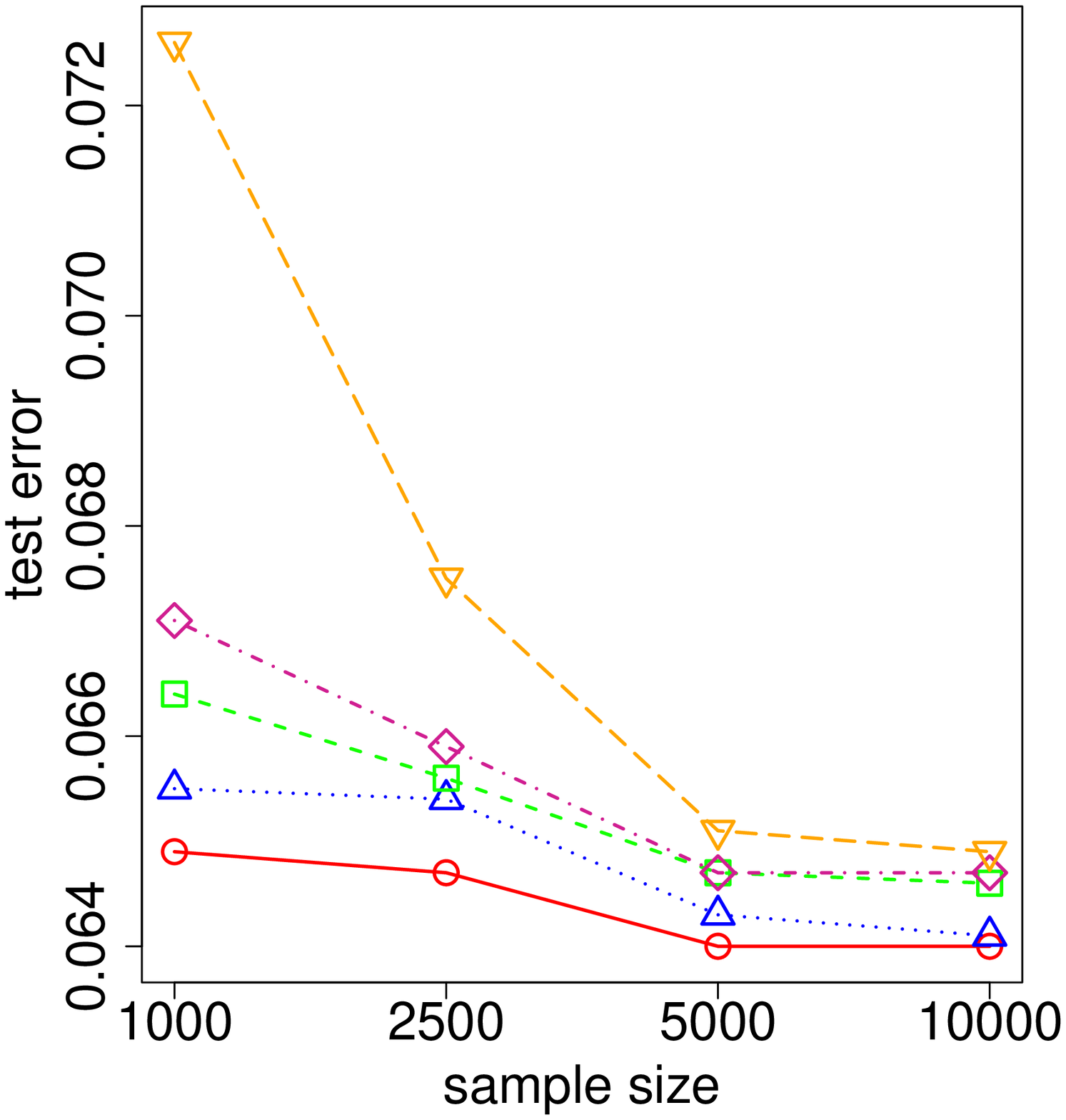}}
\hfill\subfigure[Average empirical $L_2$-error of the various RC-SVMs\label{fig::experiments::art_data_5_2_rc_L2error}]{\includegraphics[angle=270, scale=0.26]{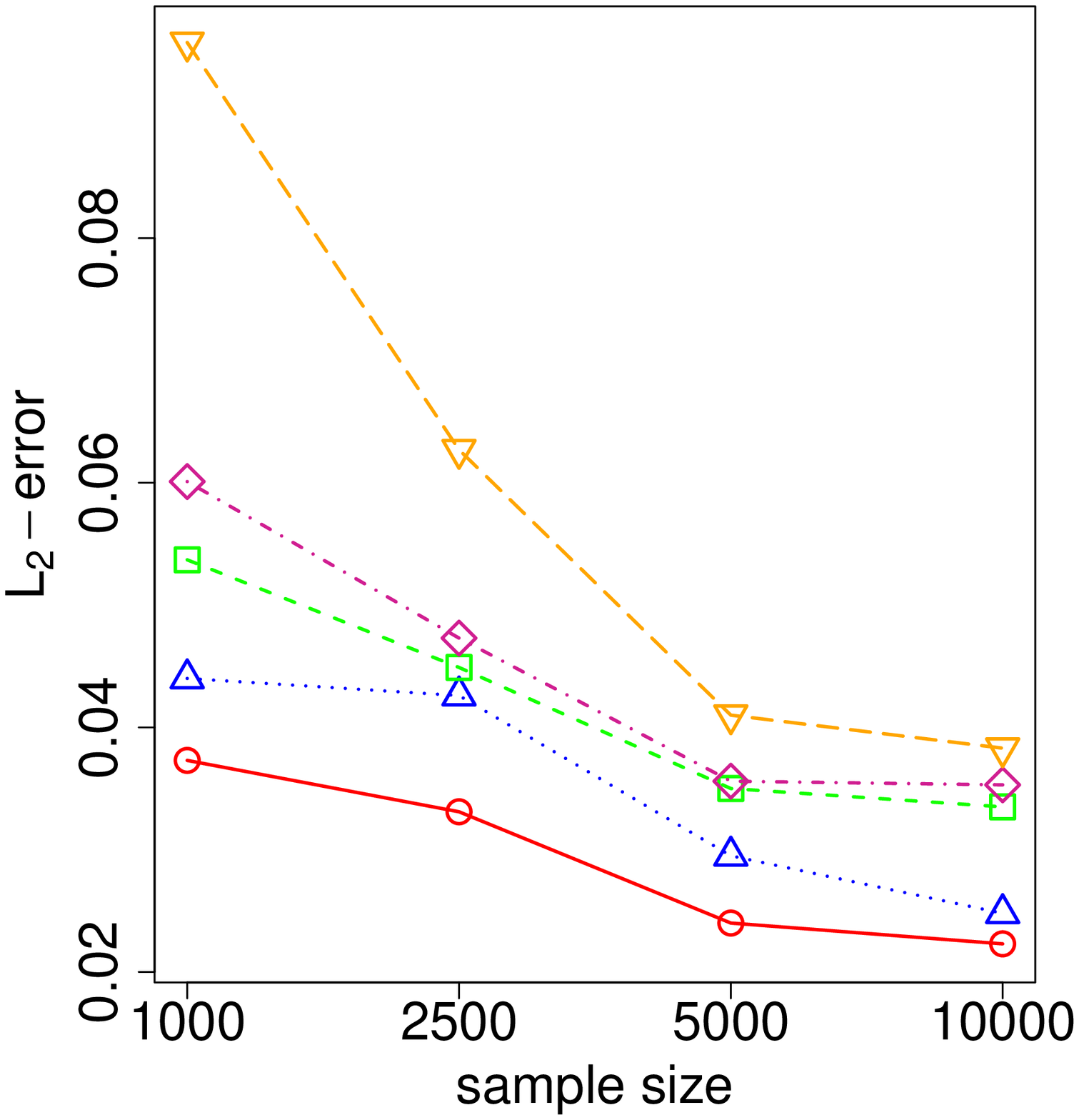}}\vspace{-0.5cm}\\
\subfigure[Average training time of the various VP-SVMs\label{fig::experiments::art_data_5_2_vp_time}]{\includegraphics[angle=270, scale=0.26]{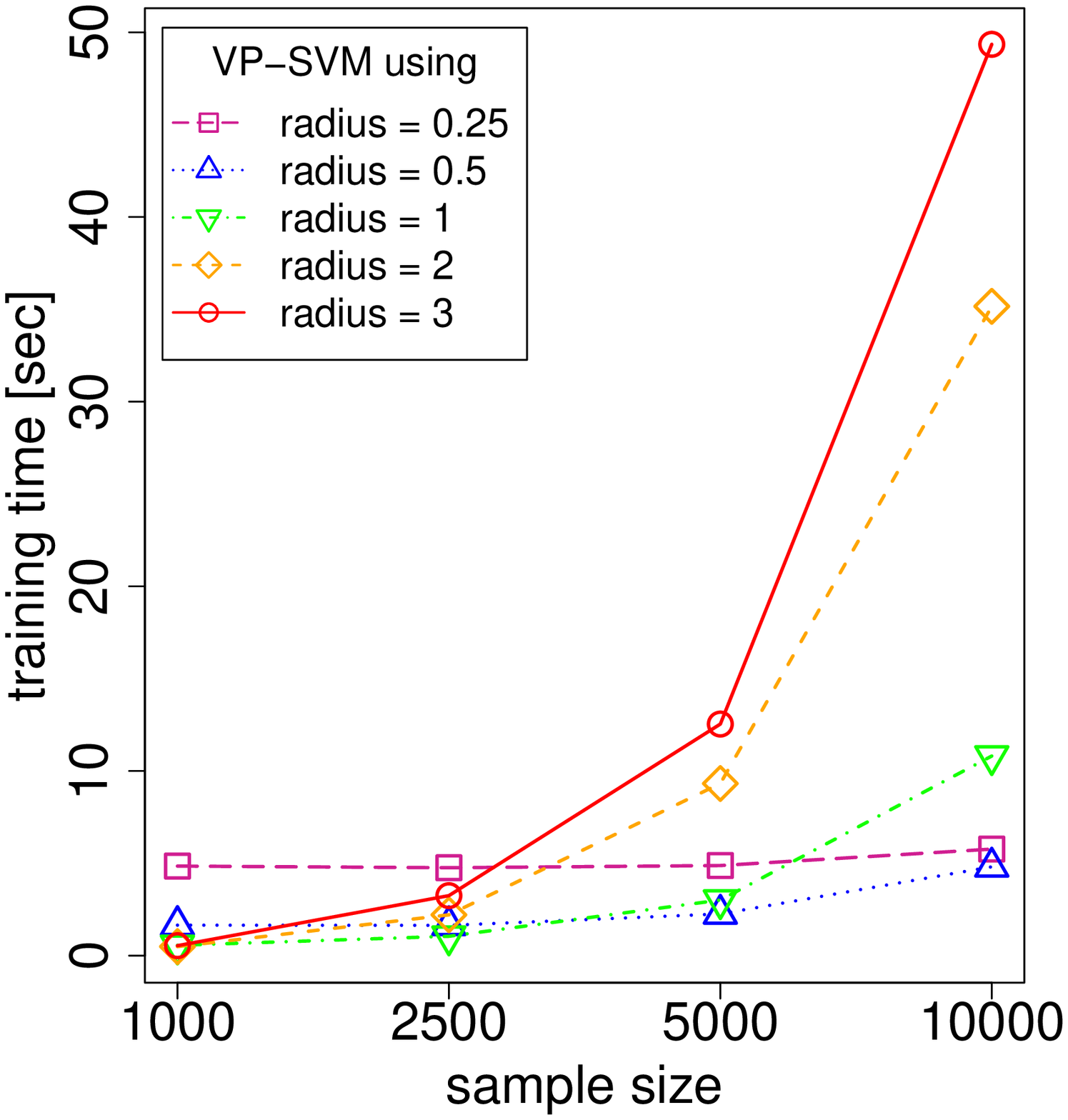}}
\hfill\subfigure[Average empirical risk of the various VP-SVMs\label{fig::experiments::art_data_5_2_vp_error}]{\includegraphics[angle=270, scale=0.26]{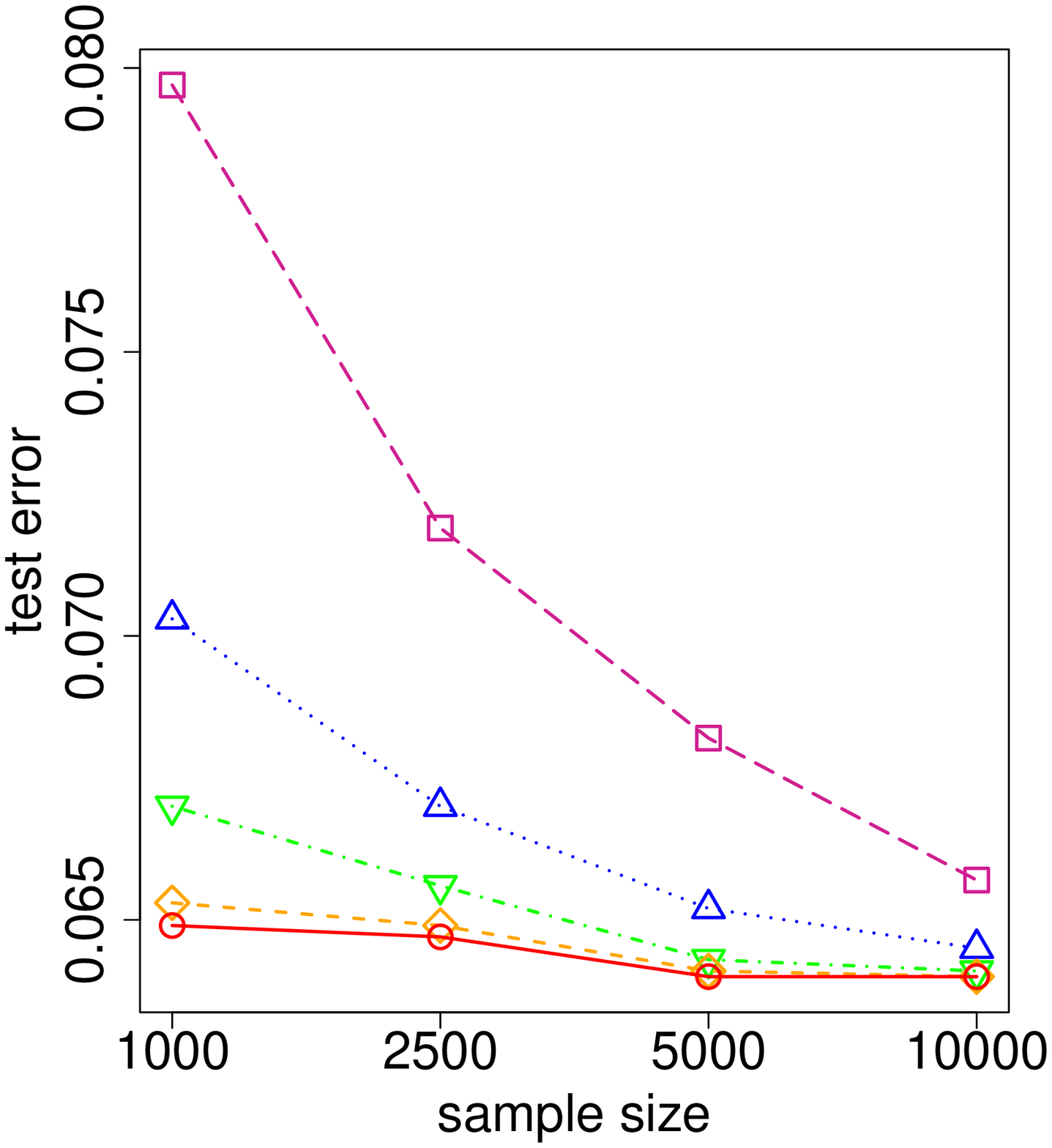}}
\hfill\subfigure[Average empirical $L_2$-error of the various VP-SVMs\label{fig::experiments::art_data_5_2_vp_L2error}]{\includegraphics[angle=270, scale=0.26]{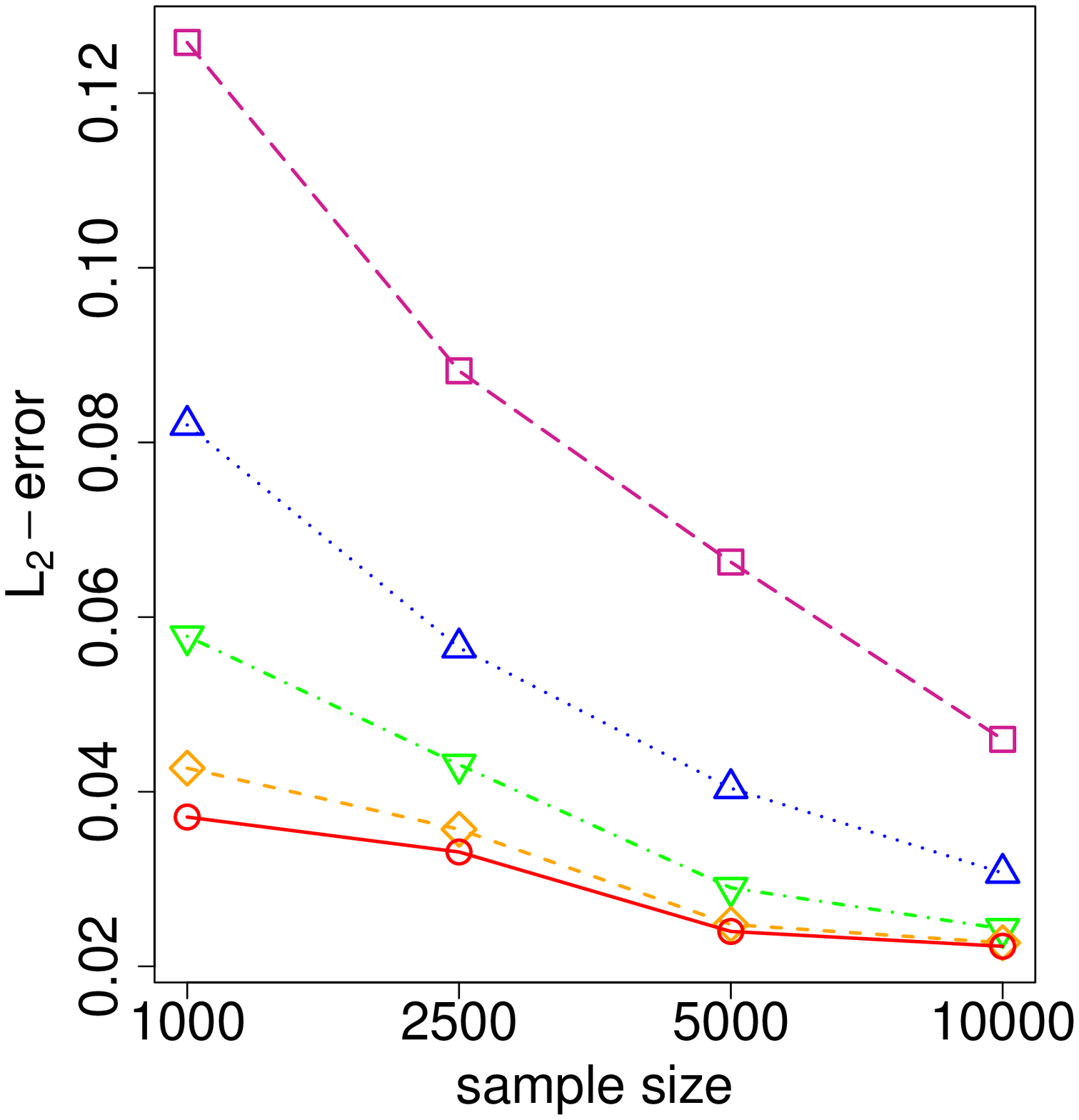}}\vspace{-0.5cm}\\
\subfigure[Average training time of \mbox{LS-,} VP-, and RC-SVMs\label{fig::experiments::art_data_5_2_time}]{\includegraphics[angle=270, scale=0.26]{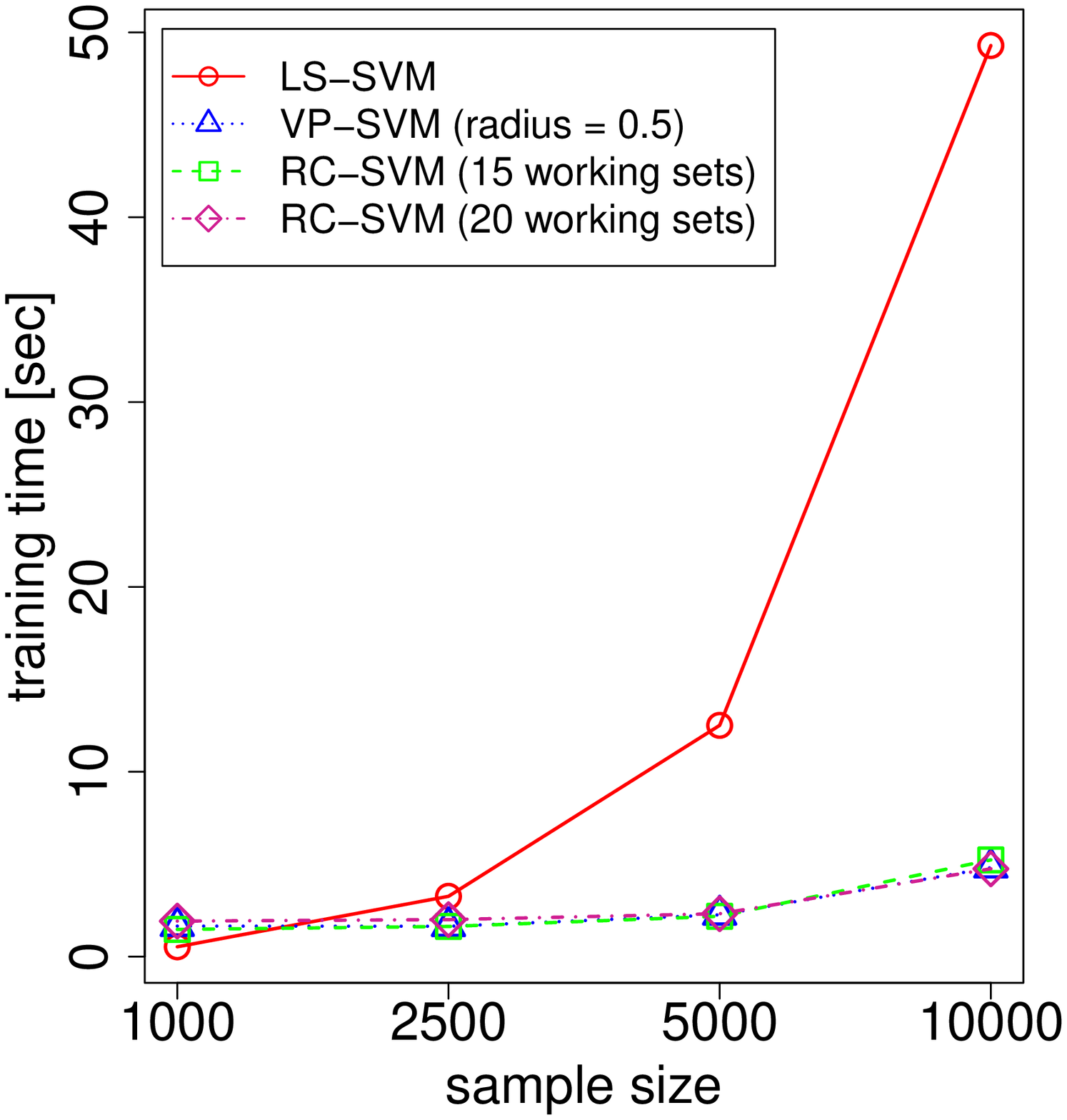}}
\hfill\subfigure[Average empirical risk of \mbox{LS-,} VP-, and RC-SVMs\label{fig::experiments::art_data_5_2_error}]{\includegraphics[angle=270, scale=0.26]{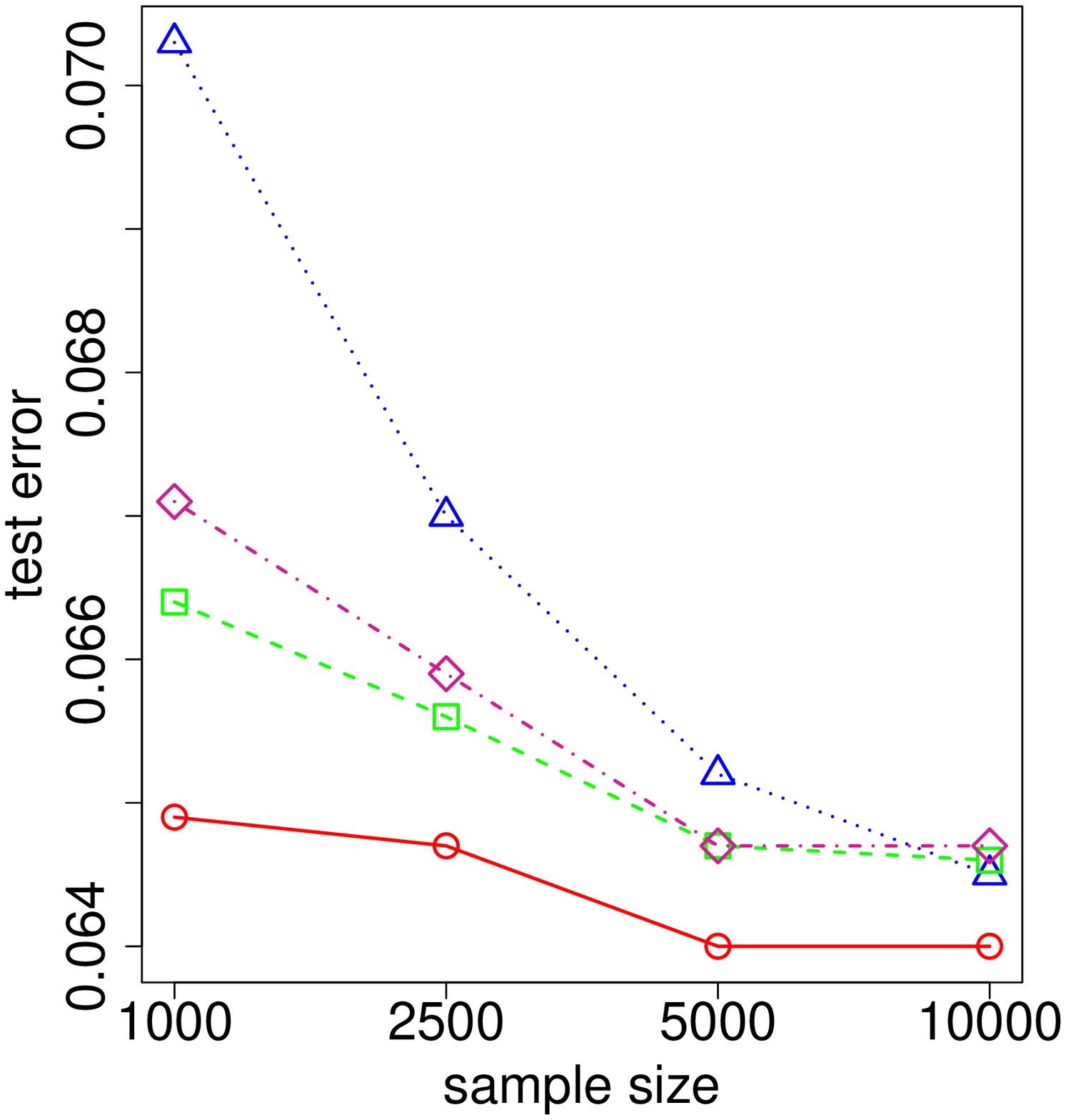}}
\hfill\subfigure[Average empirical $L_2$-error of \mbox{LS-,} VP-, and RC-SVMs\label{fig::experiments::art_data_5_2_L2error}]{\includegraphics[angle=270, scale=0.26]{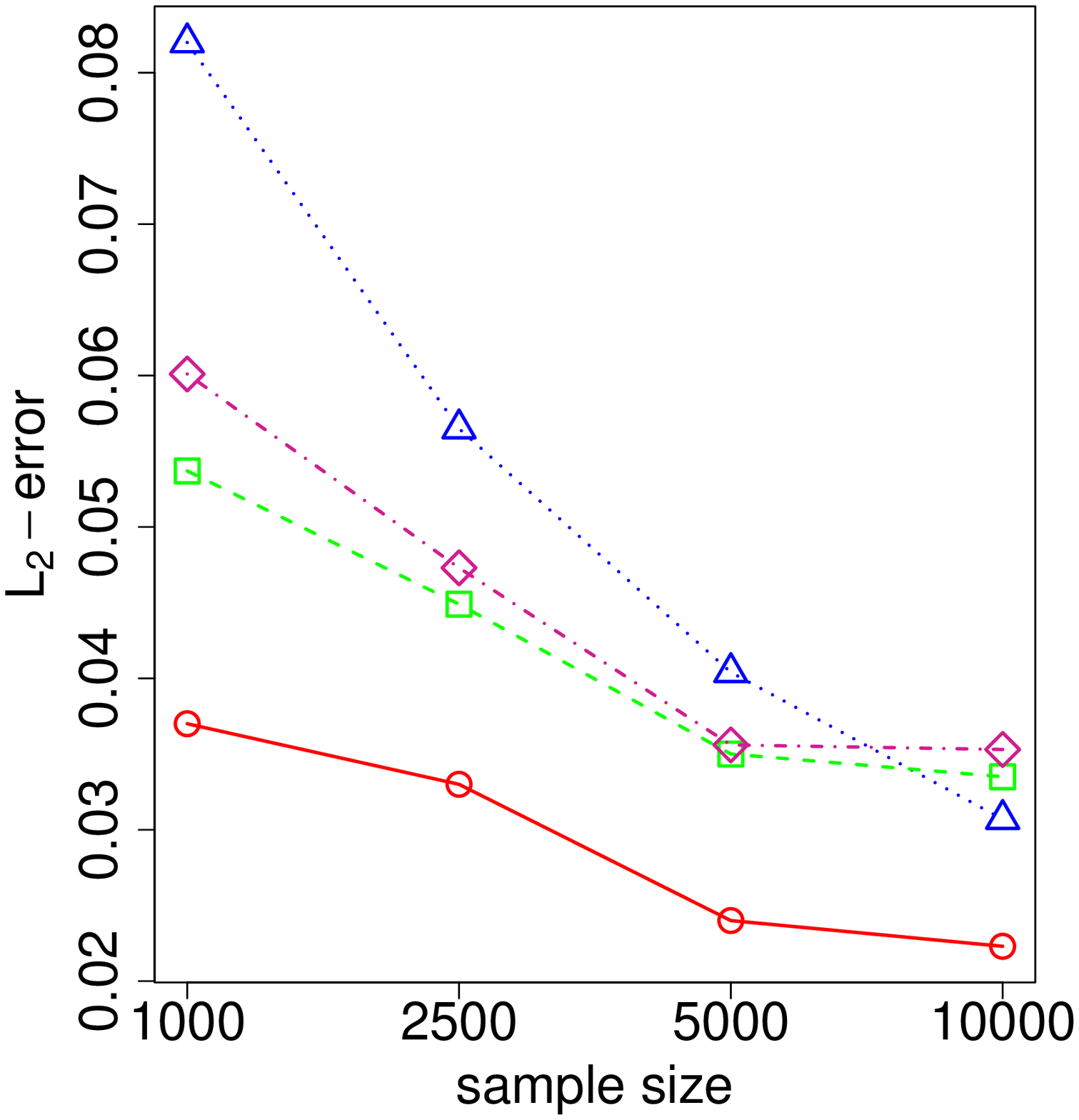}}
\caption{
Average training time and test error of LS-, VP-, and RC-SVMs for the artificial data 
Type \RM{5} depending on the training set size $n_\text{train}=1\,000,\ldots,10\,000$. Subfigures 
\subref{fig::experiments::art_data_5_2_rc_time}--\subref{fig::experiments::art_data_5_2_rc_L2error} 
show the results for RC-SVMs using different numbers of working sets and Subfigures 
\subref{fig::experiments::art_data_5_2_vp_time}--\subref{fig::experiments::art_data_5_2_vp_L2error} 
illustrate the results for VP-SVMs using various radii. At the bottom, Subfigures 
\subref{fig::experiments::art_data_5_2_time}--\subref{fig::experiments::art_data_5_2_L2error} 
contain the average training times and the average test errors of the LS-SVM, one VP-SVM and 
two RC-SVMs. Here, the VP-SVM is the one which trains fastest for $n_\text{train}=10\,000$
and the two RC-SVMs are those which achieve for $n_\text{train}=10\,000$ roughly the same 
training time as the chosen VP-SVM.}
\label{fig::experiments::artificial_data_5_2}
\end{scriptsize}
\end{figure}

\begin{figure}[t]
\subfigure{\includegraphics[angle=270, scale=0.28]{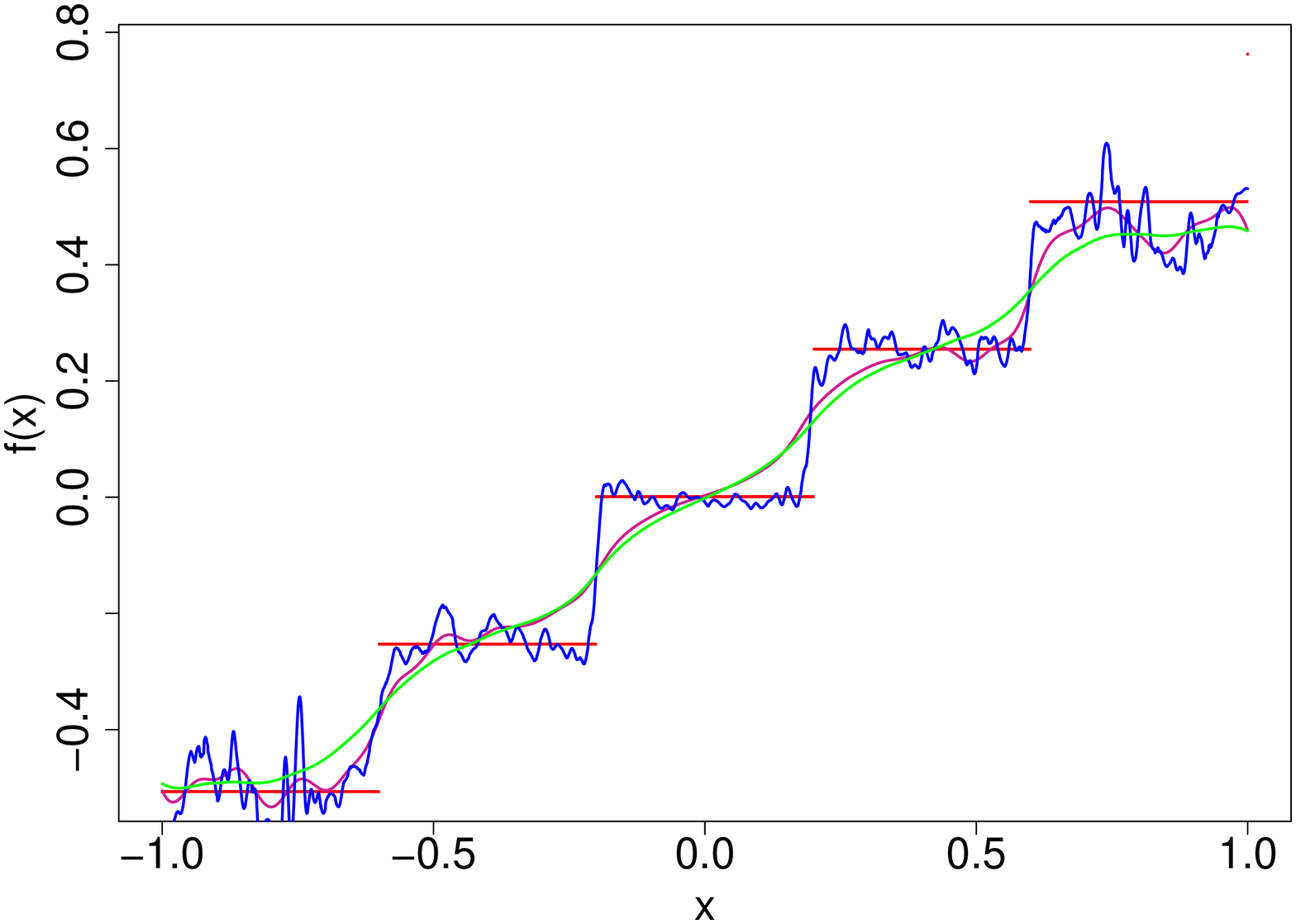}}
\hfill
\subfigure{\includegraphics[angle=270, scale=0.28]{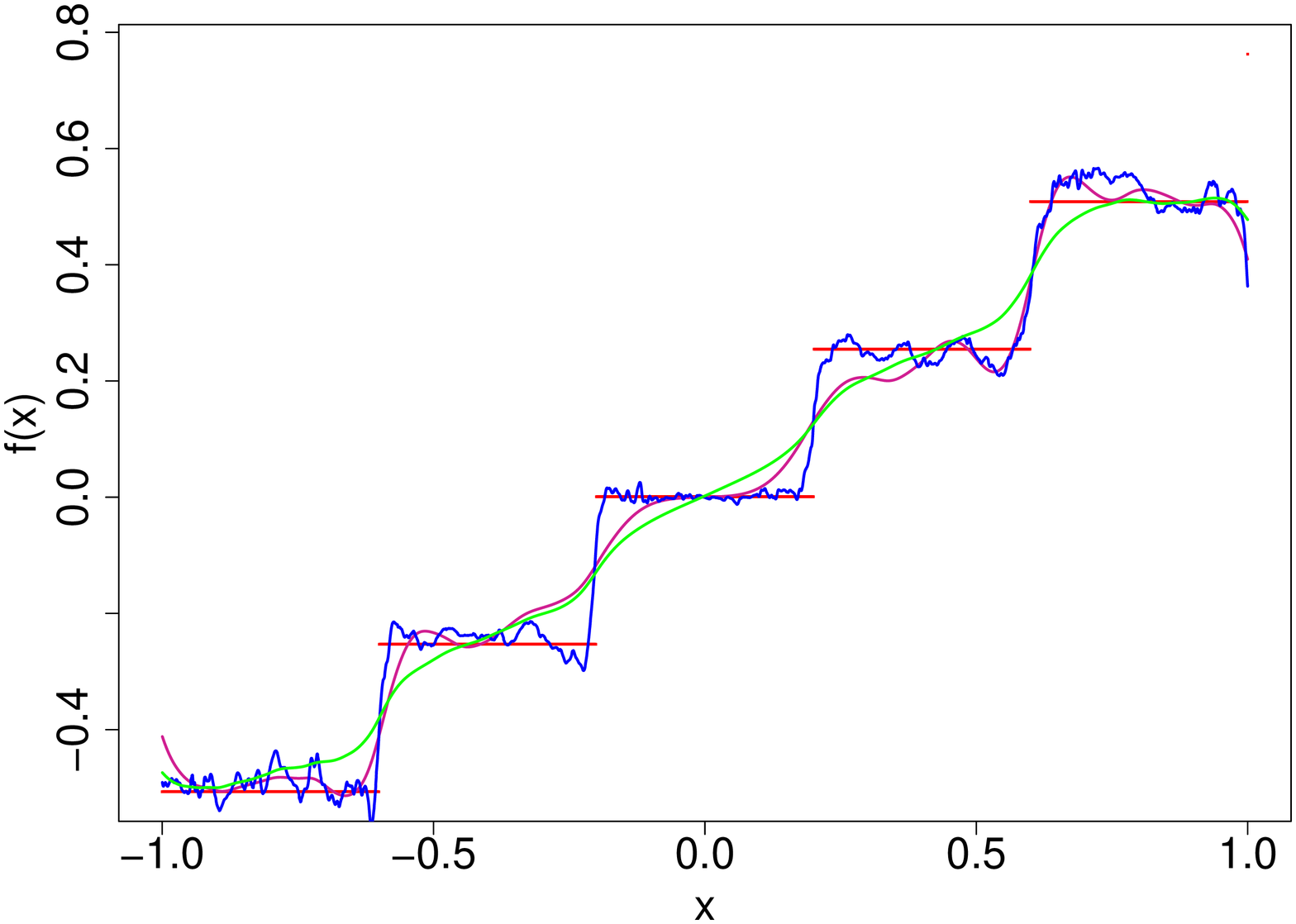}}
\begin{scriptsize}
\begin{minipage}{0.49\textwidth}
\begin{center}
\begin{tabular}{cl}
 \textcolor{rot}{\protect\rule[0.5ex]{0.7cm}{.4pt}} 
 & Bayes decision function \\
 \textcolor{violetred}{\protect\rule[0.5ex]{0.7cm}{.4pt}}
 & average empirical LS-SVM solution
\end{tabular}
\end{center} 
\end{minipage}\hfill
\begin{minipage}{0.49\textwidth}
\begin{center}
\begin{tabular}{cl}
 \textcolor{blau}{\protect\rule[0.5ex]{0.7cm}{.4pt}}
 & average empirical VP-SVM solution \\
 \textcolor{gruen}{\protect\rule[0.5ex]{0.7cm}{.4pt}}
 & average empirical RC-SVM solution
\end{tabular}
\end{center} 
\end{minipage}
\end{scriptsize}
\caption[Predictions for the artificial data sets of type 1]{
Predictions for the artificial data sets of Type \RM{1}, drawn from the step 
function in Figure \ref{fig::experiments::basic_fct_1} with noise depending on $x$. The left 
graphic shows the predictions for the data set of size $n_\text{train}=1\,000$ and the right 
graphic for the data set of size $n_\text{train}=10\,000$. Here, note that the VP-SVM solutions 
are not necessarily continuous, nevertheless we continuously connected its predicted values 
in the above plots.} 
\label{fig::experiments::prediction_artificial_data_1_2}
\end{figure}

\begin{figure}[ht!]
\subfigure{\includegraphics[angle=270, scale=0.28]{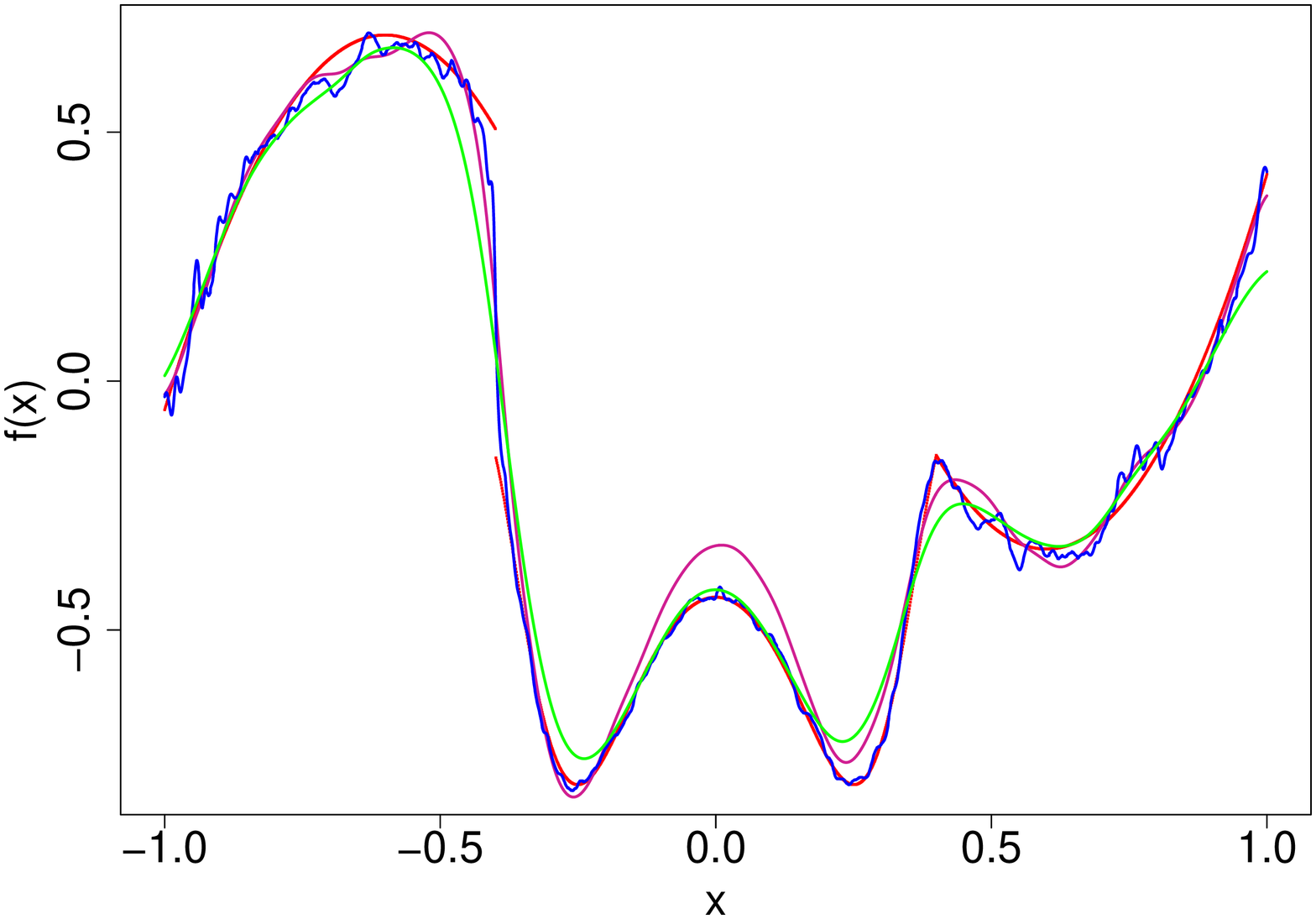}}
\hfill
\subfigure{\includegraphics[angle=270, scale=0.28]{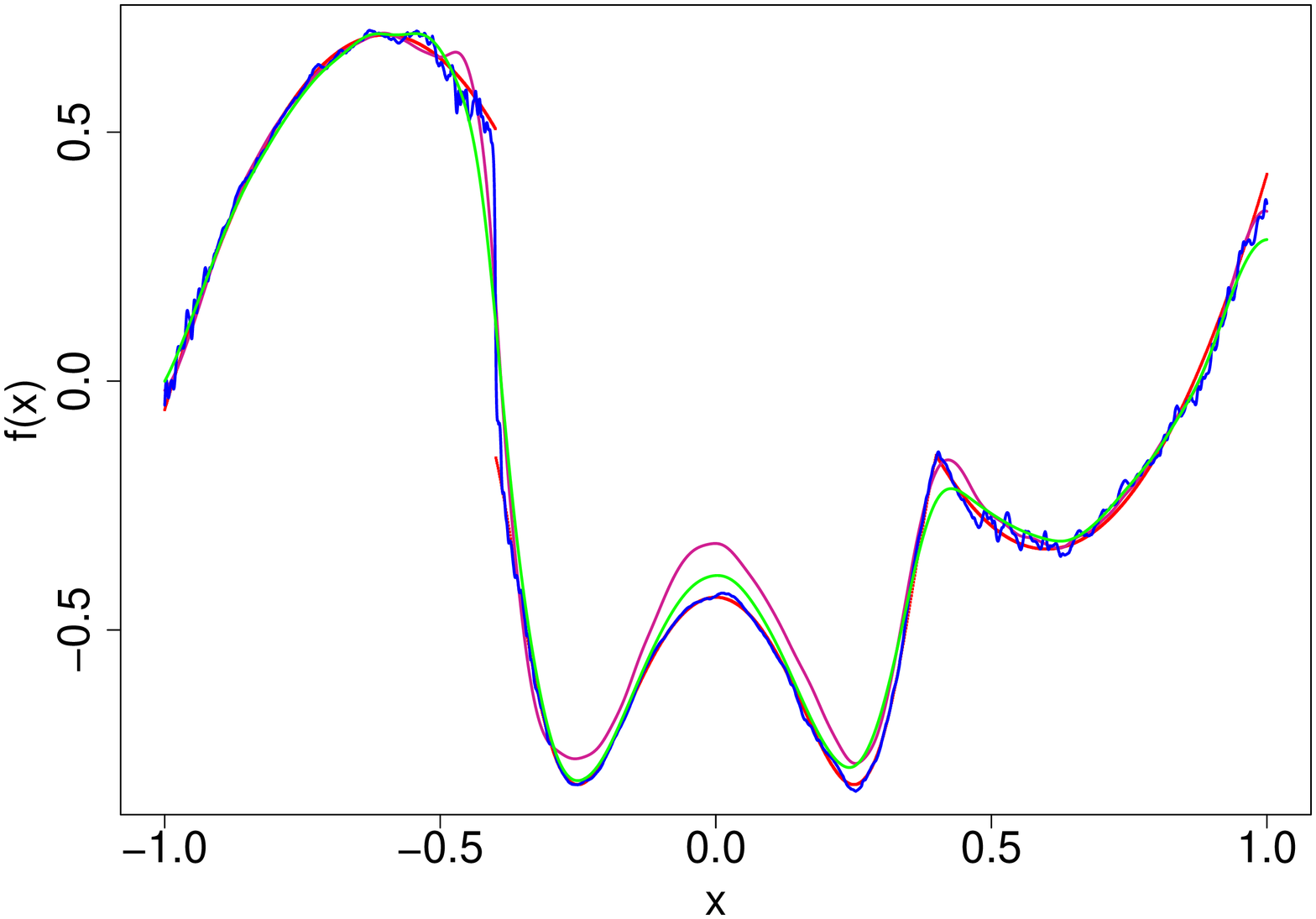}}
\begin{scriptsize}
\begin{minipage}{0.49\textwidth}
\begin{center}
\begin{tabular}{cl}
 \textcolor{rot}{\protect\rule[0.5ex]{0.7cm}{.4pt}} 
 & Bayes decision function \\
 \textcolor{violetred}{\protect\rule[0.5ex]{0.7cm}{.4pt}}
 & average empirical LS-SVM solution
\end{tabular}
\end{center} 
\end{minipage}\hfill
\begin{minipage}{0.49\textwidth}
\begin{center}
\begin{tabular}{cl}
 \textcolor{blau}{\protect\rule[0.5ex]{0.7cm}{.4pt}}
 & average empirical VP-SVM solution \\
 \textcolor{gruen}{\protect\rule[0.5ex]{0.7cm}{.4pt}}
 & average empirical RC-SVM solution
\end{tabular}
\end{center} 
\end{minipage}
\end{scriptsize}
\caption[Predictions for the artificial data sets of type 2]{Predictions for the artificial 
data sets of Type \RM{2}, drawn from the cracked function in Figure 
\ref{fig::experiments::basic_fct_2} with noise depending on $x$. The left 
graphic shows the predictions for the data set of size $n_\text{train}=1\,000$ and the right 
graphic for the data set of size $n_\text{train}=10\,000$.} 
\label{fig::experiments::prediction_artificial_data_2_2}
%
\begin{minipage}[t]{0.49\textwidth}
\hfill
\subfigure{\includegraphics[angle=270, scale=0.28]{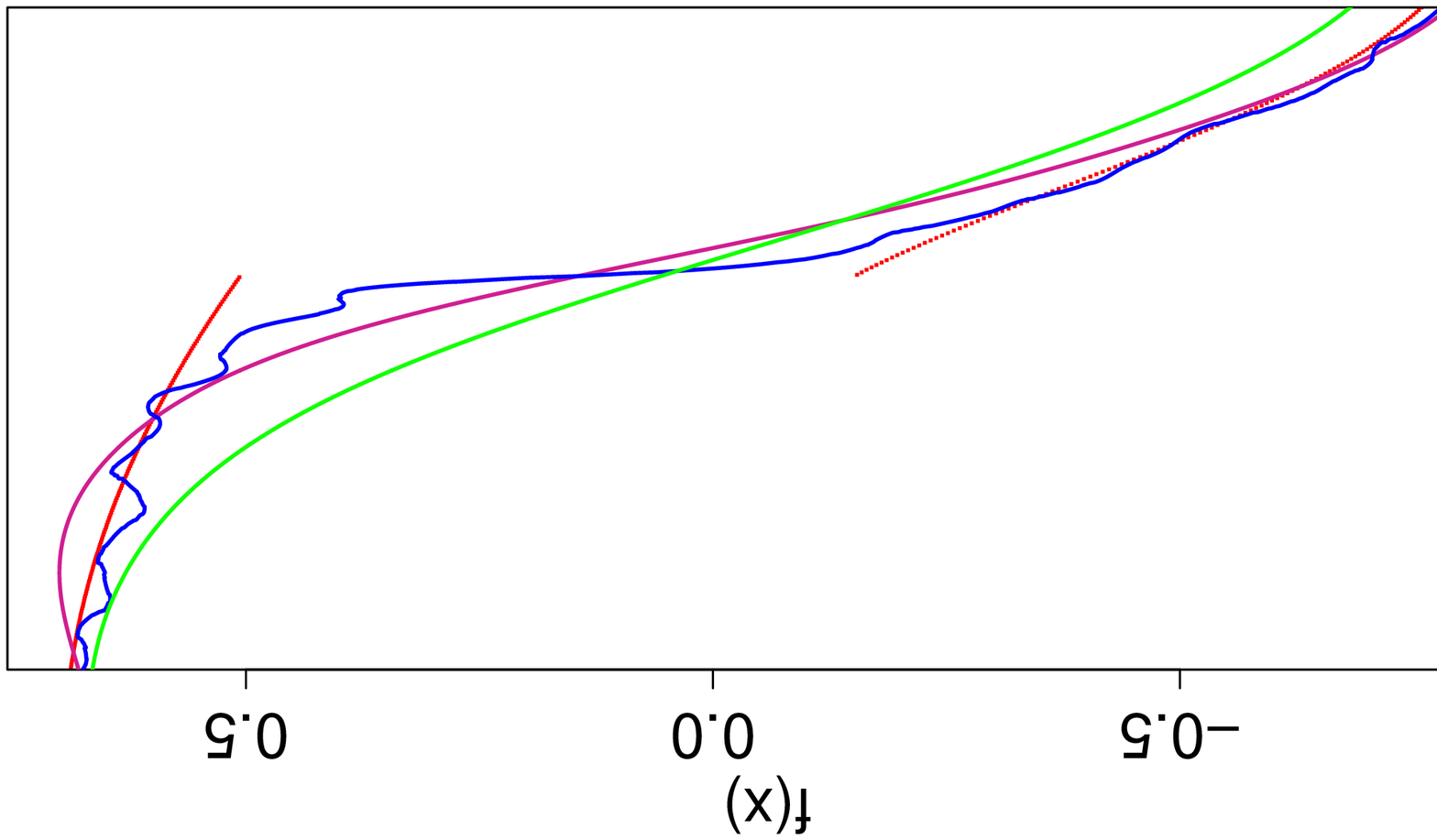}}
\hfill
\subfigure{\includegraphics[angle=270, scale=0.28]{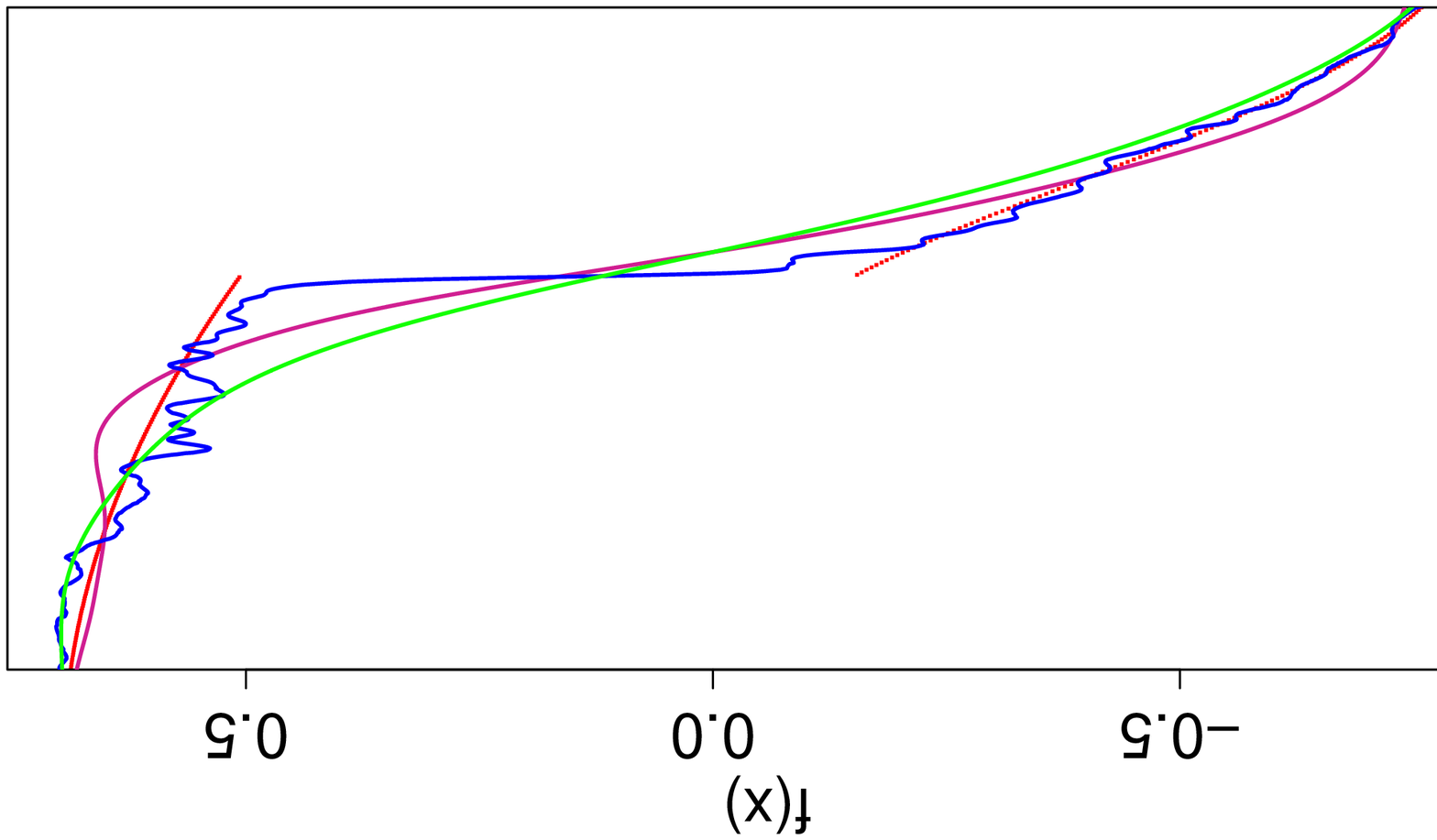}}
\hfill
\end{minipage}
\hfill
\begin{minipage}[t]{0.49\textwidth}
\subfigure{\includegraphics[angle=270, scale=0.28]{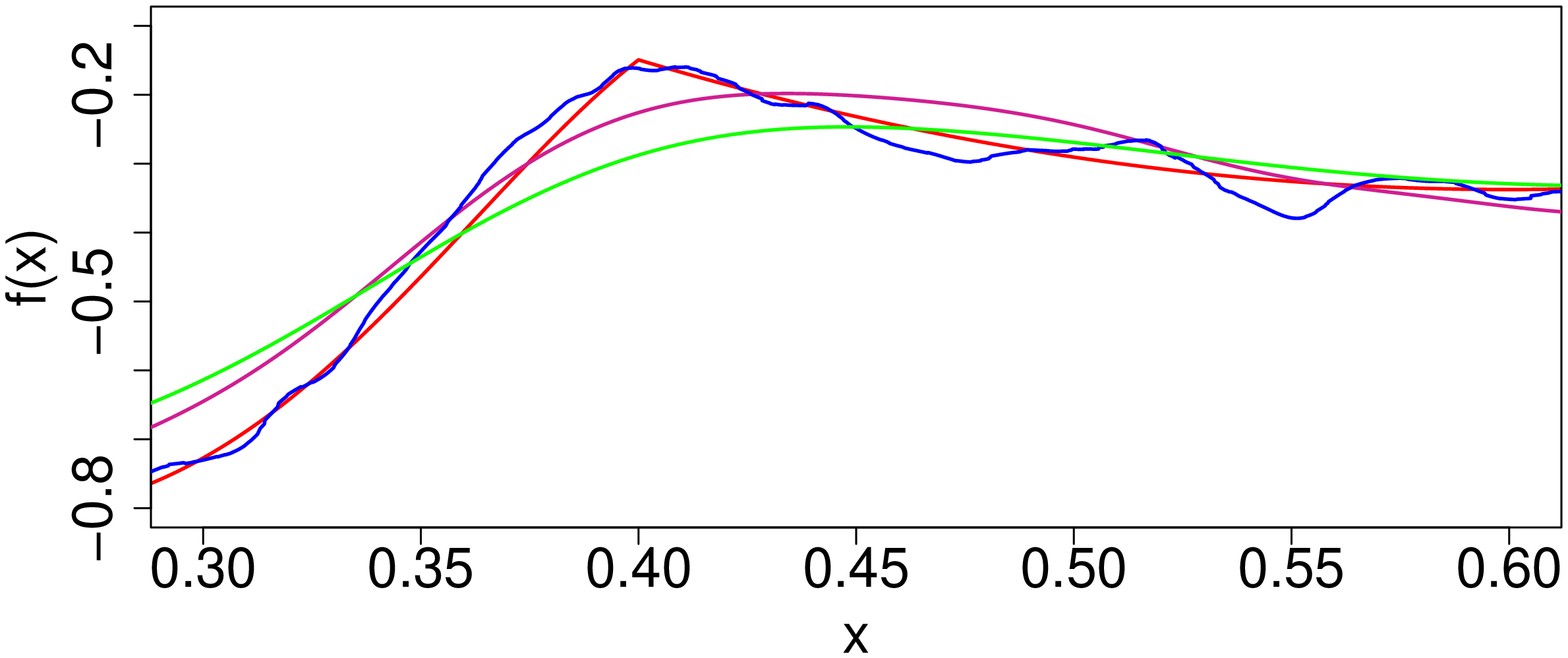}}
\\[-0.25cm]
\subfigure{\includegraphics[angle=270, scale=0.28]{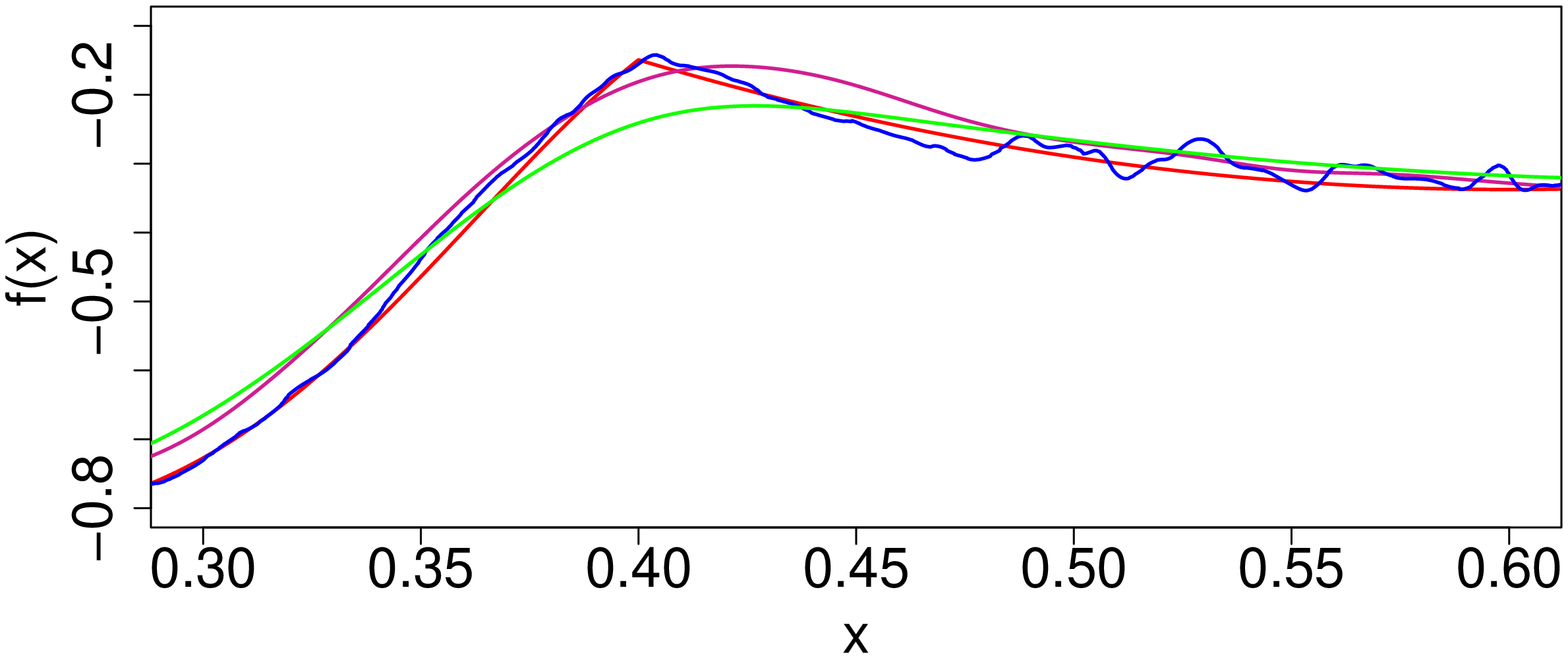}}
\end{minipage}\\
\begin{scriptsize}
\begin{minipage}{0.49\textwidth}
\begin{center}
\begin{tabular}{cl}
 \textcolor{rot}{\protect\rule[0.5ex]{0.7cm}{.4pt}} 
 & Bayes decision function \\
 \textcolor{violetred}{\protect\rule[0.5ex]{0.7cm}{.4pt}}
 & average empirical LS-SVM solution
\end{tabular}
\end{center} 
\end{minipage}\hfill
\begin{minipage}{0.49\textwidth}
\begin{center}
\begin{tabular}{cl}
 \textcolor{blau}{\protect\rule[0.5ex]{0.7cm}{.4pt}}
 & average empirical VP-SVM solution \\
 \textcolor{gruen}{\protect\rule[0.5ex]{0.7cm}{.4pt}}
 & average empirical RC-SVM solution
\end{tabular}
\end{center} 
\end{minipage}
\end{scriptsize}
\caption[Predictions for the artificial data sets of Type \RM{2}]{Predictions for 
the artificial data sets of Type \RM{2}. 
The left graphic shows the predictions for $x\in[-0.55,-0.4]$ and the data set of size 
$n_\text{train}=1\,000$, while the graphic on its right-hand side pictures the predictions 
for the same interval for $x$ and the data set of size $n_\text{train}=10\,000$.
The two graphics on the right-hand side illustrate the predictions for $x\in[0.3,0.6]$, 
the upper one 
for the data set of size $n_\text{train}=1\,000$ and the 
lower one for the data set of size $n_\text{train}=10\,000$.} 
\label{fig::experiments::prediction_artificial_data_2_2_zoom}
\end{figure}

\begin{figure}[ht]
\begin{center}
\subfigure{\includegraphics[angle=270, scale=0.28]{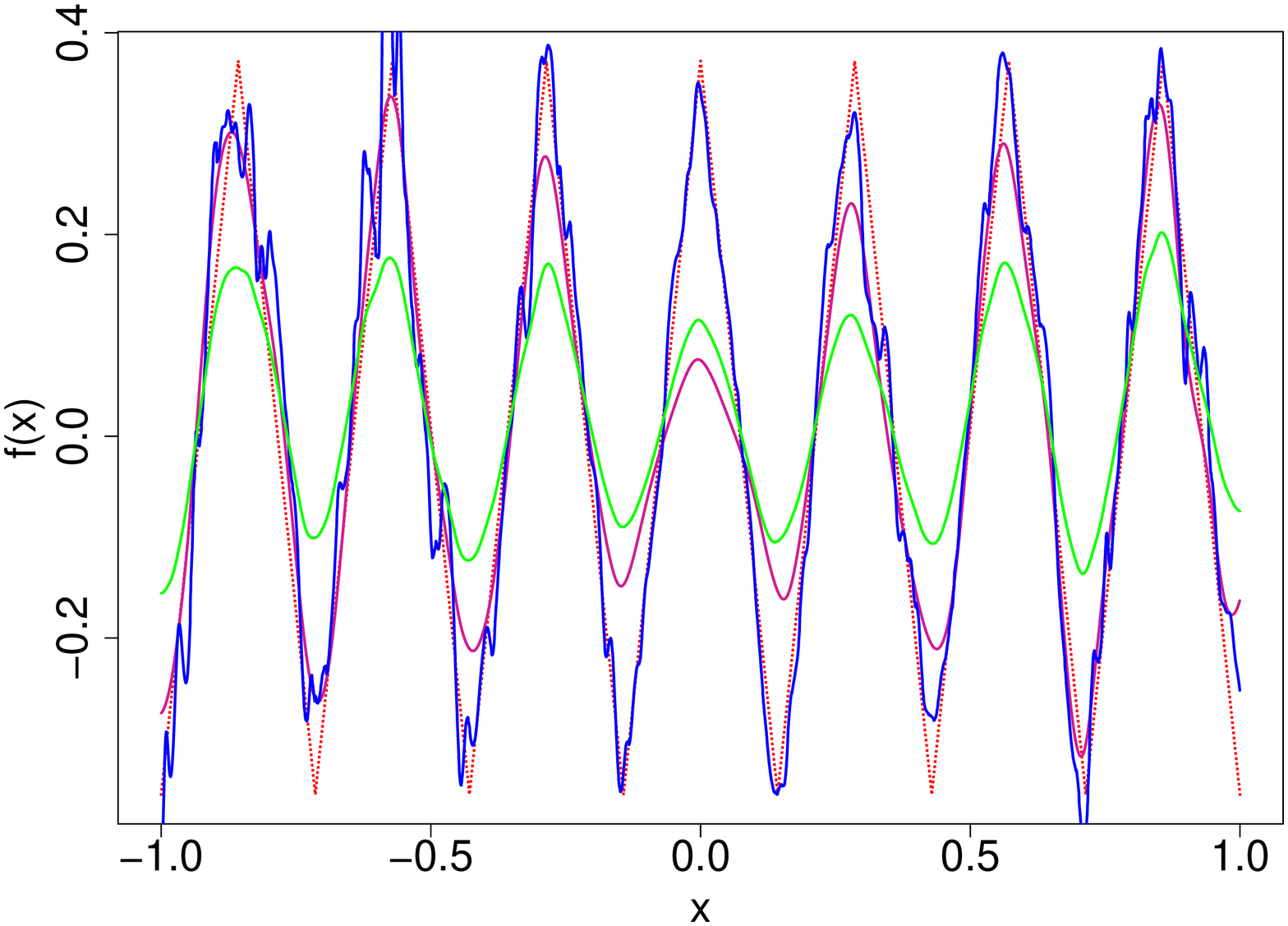}}
\hfill
\subfigure{\includegraphics[angle=270, scale=0.28]{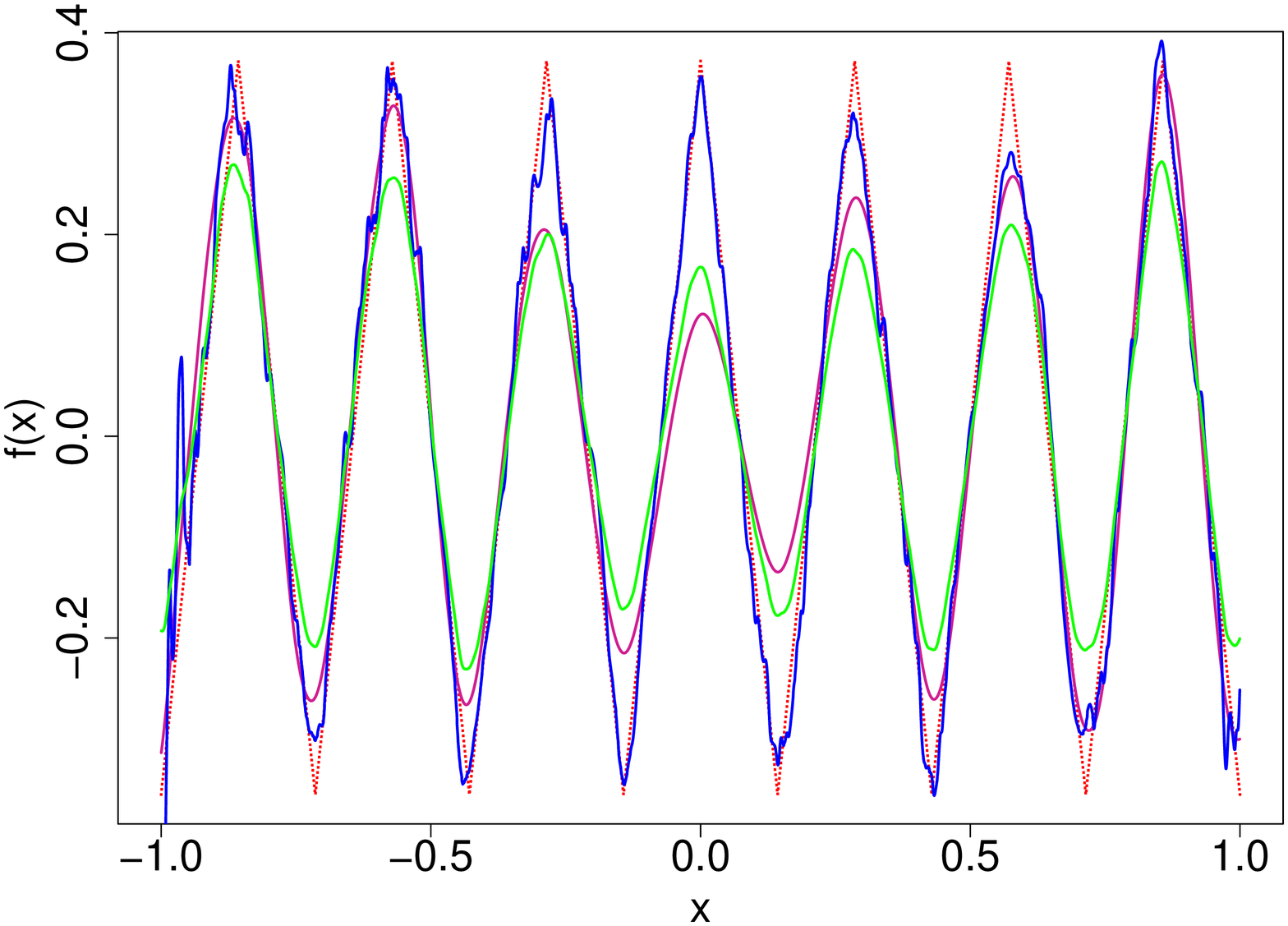}}
\end{center}
\begin{scriptsize}
\begin{minipage}{0.49\textwidth}
\begin{center}
\begin{tabular}{cl}
 \textcolor{rot}{\protect\rule[0.5ex]{0.7cm}{.4pt}} 
 & Bayes decision function \\
 \textcolor{violetred}{\protect\rule[0.5ex]{0.7cm}{.4pt}}
 & average empirical LS-SVM solution
\end{tabular}
\end{center} 
\end{minipage}\hfill
\begin{minipage}{0.49\textwidth}
\begin{center}
\begin{tabular}{cl}
 \textcolor{blau}{\protect\rule[0.5ex]{0.7cm}{.4pt}}
 & average empirical VP-SVM solution \\
 \textcolor{gruen}{\protect\rule[0.5ex]{0.7cm}{.4pt}}
 & average empirical RC-SVM solution
\end{tabular}
\end{center} 
\end{minipage}
\end{scriptsize}
\caption[Predictions for the artificial data sets of type 3]{Predictions for the artificial 
data sets of Type \RM{3}, drawn from the jagged function in Figure 
\ref{fig::experiments::basic_fct_3} with noise depending on $x$. The left 
graphic shows the predictions for the data set of size $n_\text{train}=1\,000$ and the right 
graphic for the data set of size $n_\text{train}=10\,000$.} 
\label{fig::experiments::prediction_artificial_data_4_2}
\end{figure}

\begin{figure}[p]
\subfigure[LS-SVM using $1\,000$ training samples]{\includegraphics[angle=270, scale=0.28]{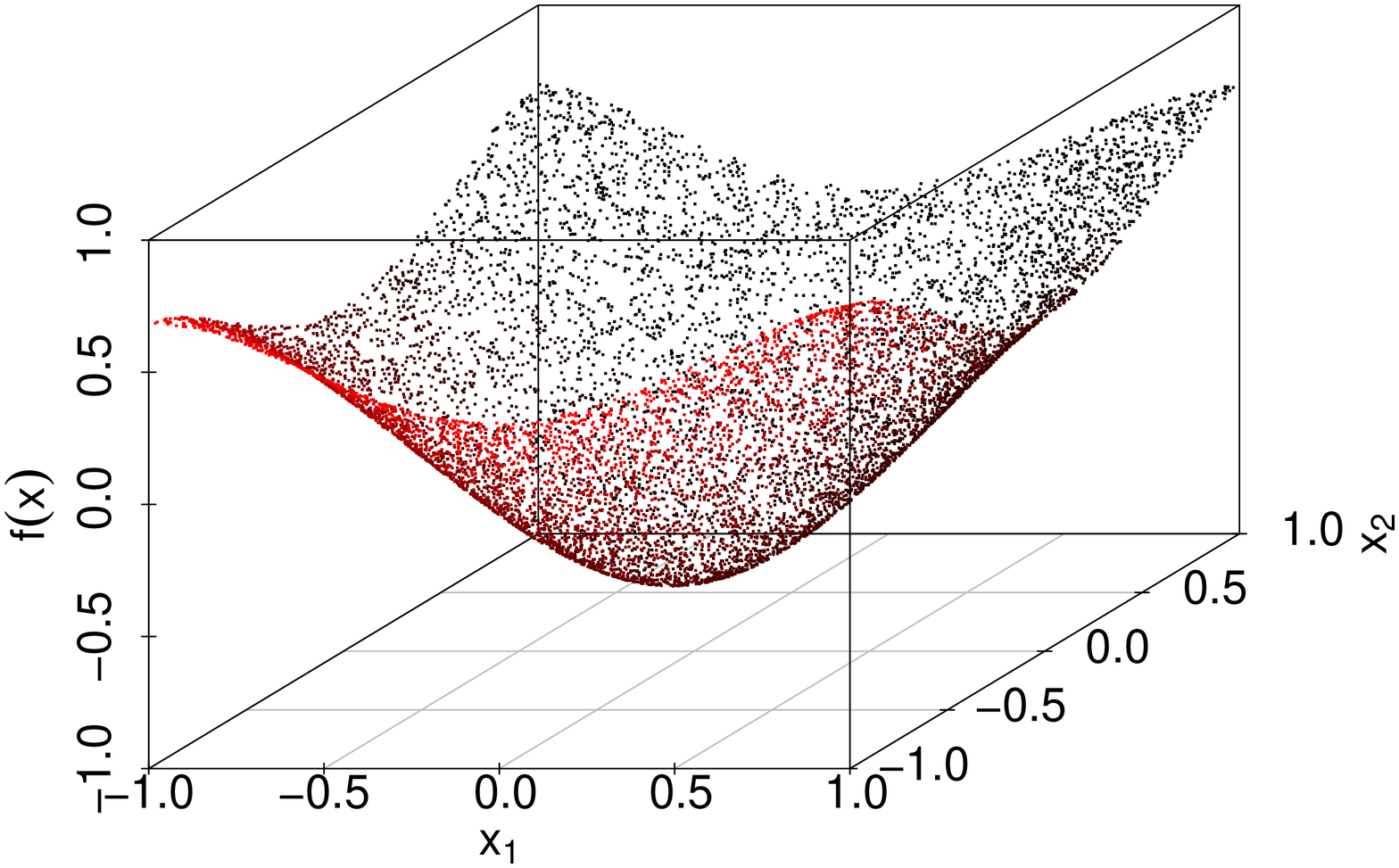}}\hfill
\subfigure[LS-SVM using $10\,000$ training samples]{\includegraphics[angle=270, scale=0.28]{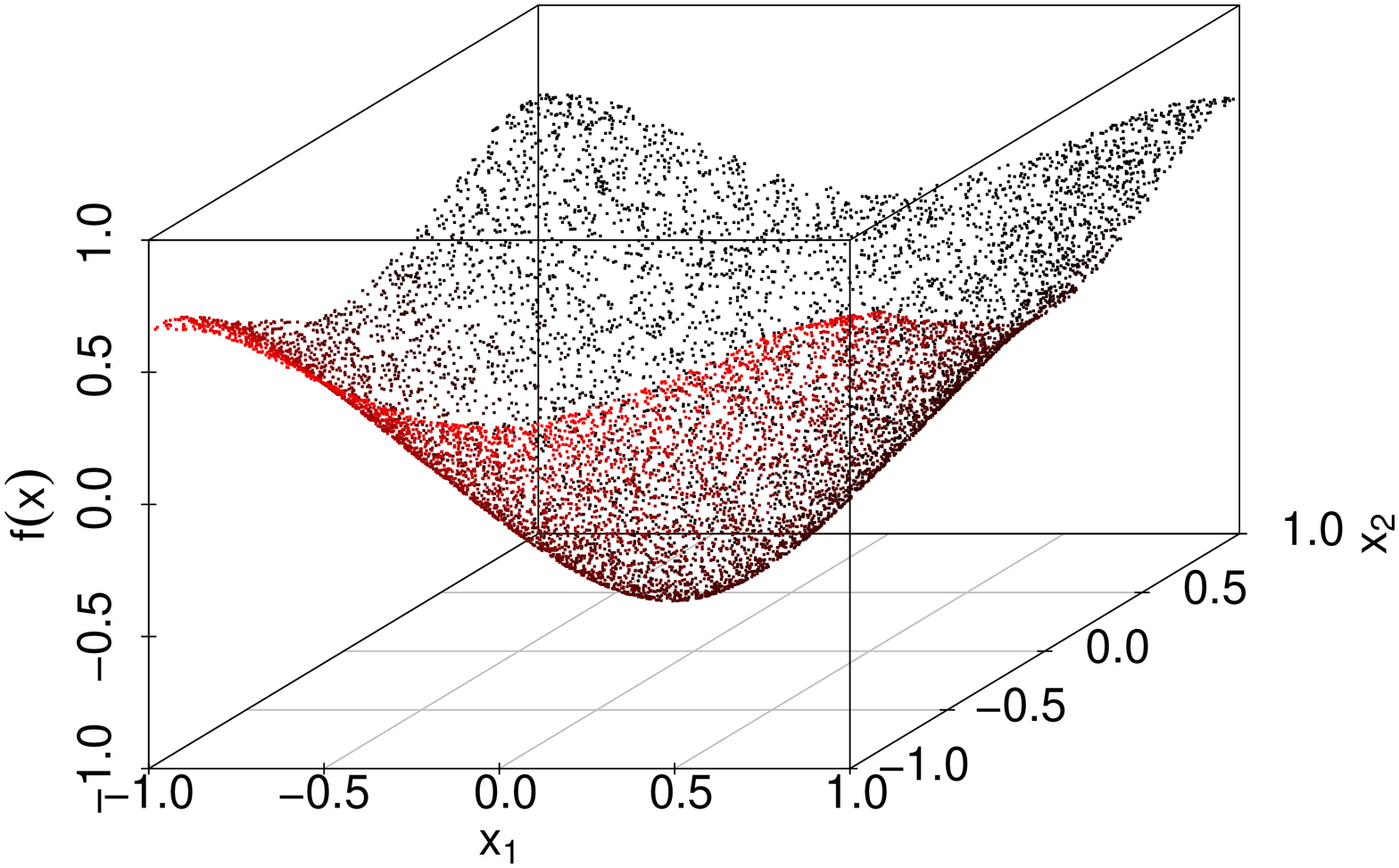}}\\
\subfigure[VP-SVM using radius $r=0.5$ and $1\,000$ training samples]{\includegraphics[angle=270, scale=0.28]{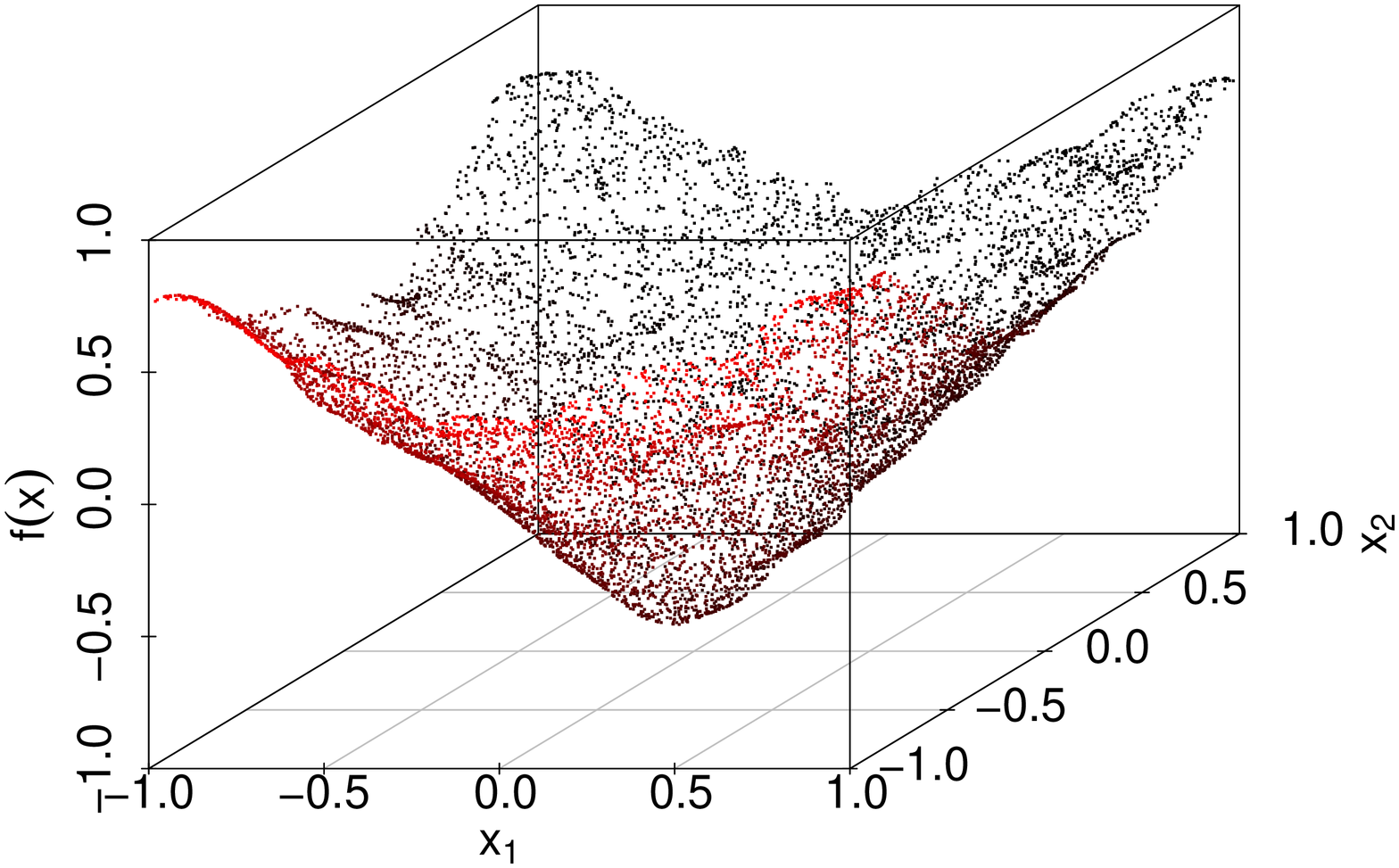}}\hfill
\subfigure[VP-SVM using radius $r=0.5$ and $10\,000$ training samples]{\includegraphics[angle=270, scale=0.28]{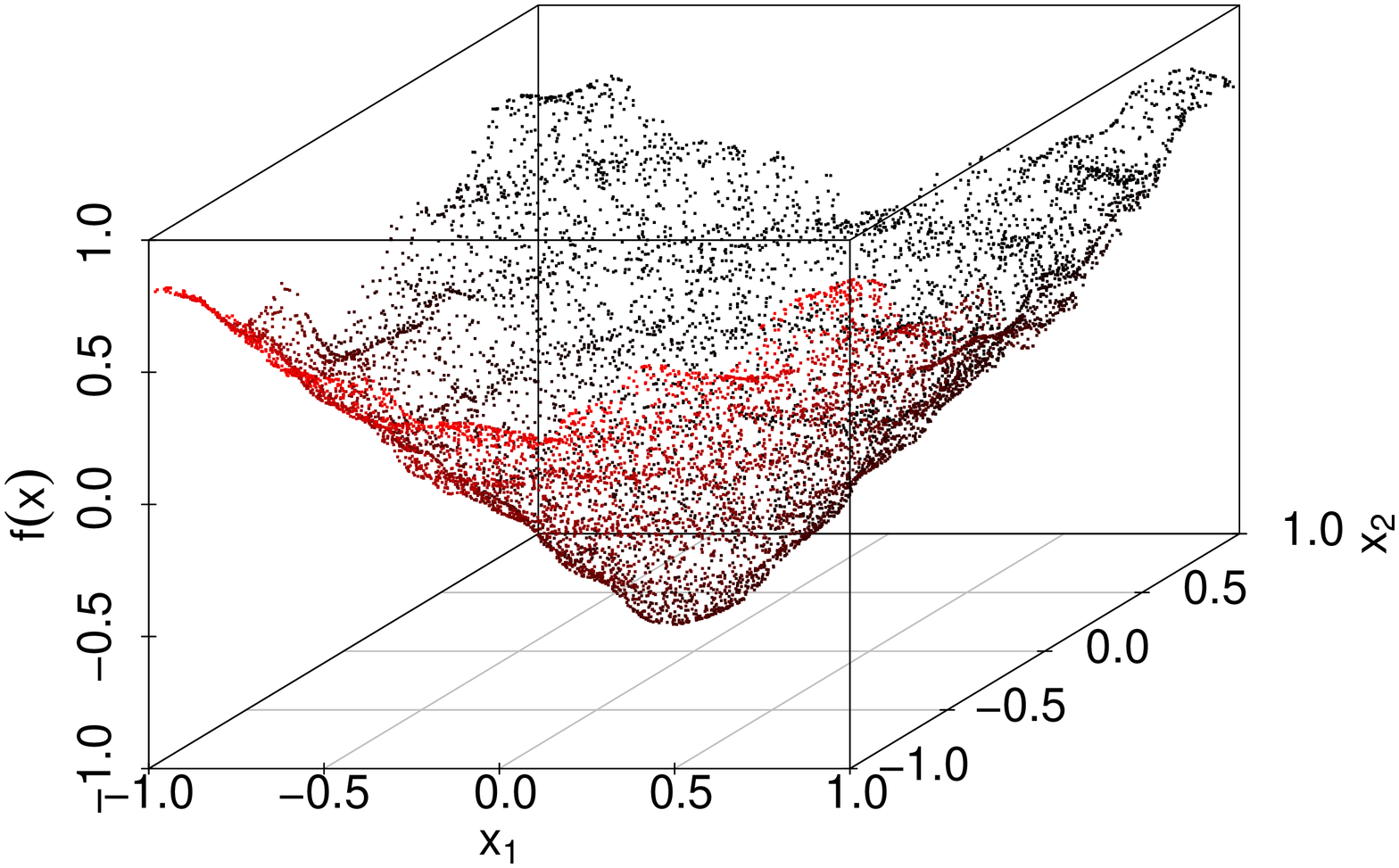}}\\
\subfigure[RC-SVM using $20$ working sets and $1\,000$ training samples]{\includegraphics[angle=270,scale=0.28]{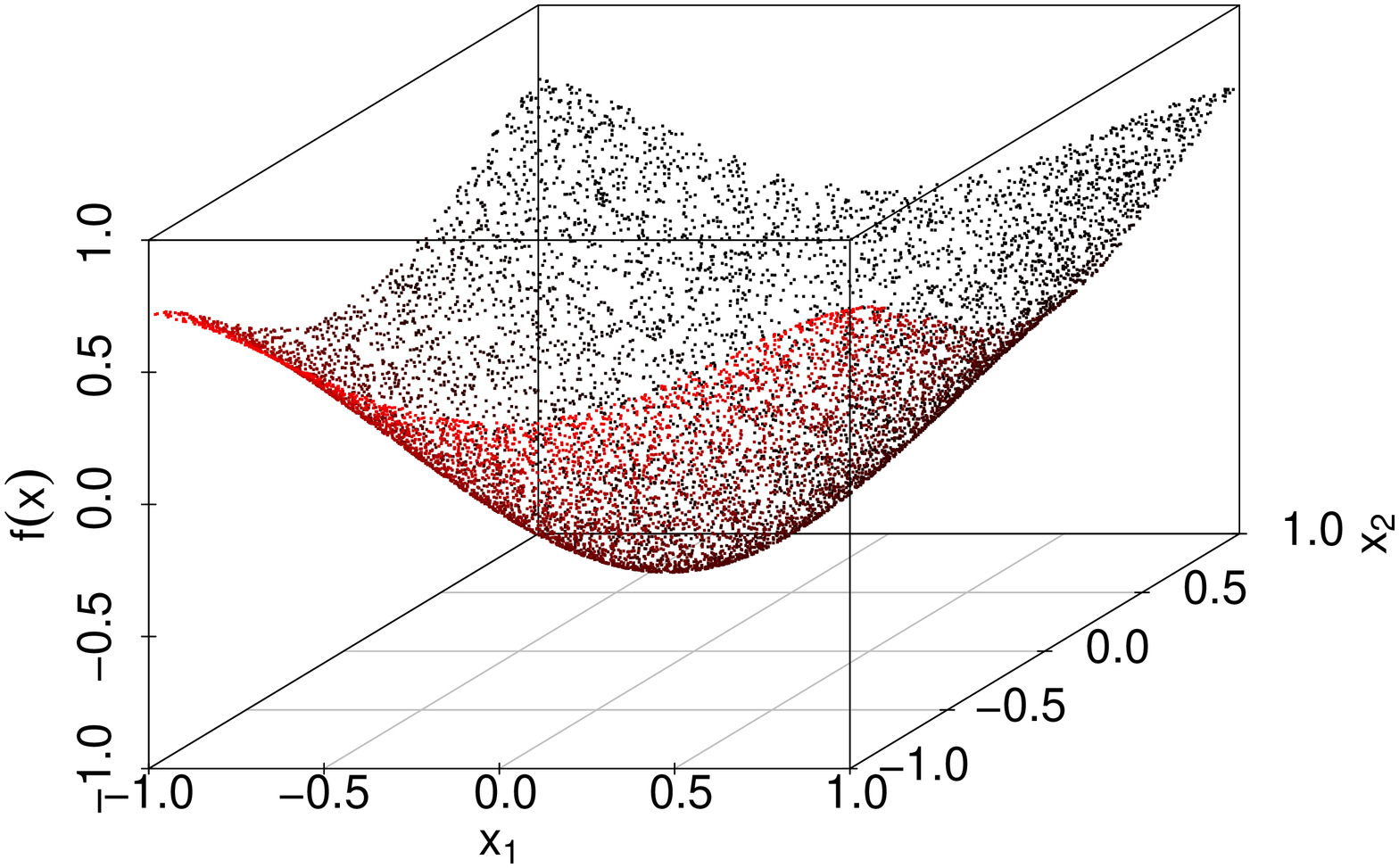}}\hfill
\subfigure[RC-SVM using $20$ working sets and $10\,000$ training samples]{\includegraphics[angle=270,scale=0.28]{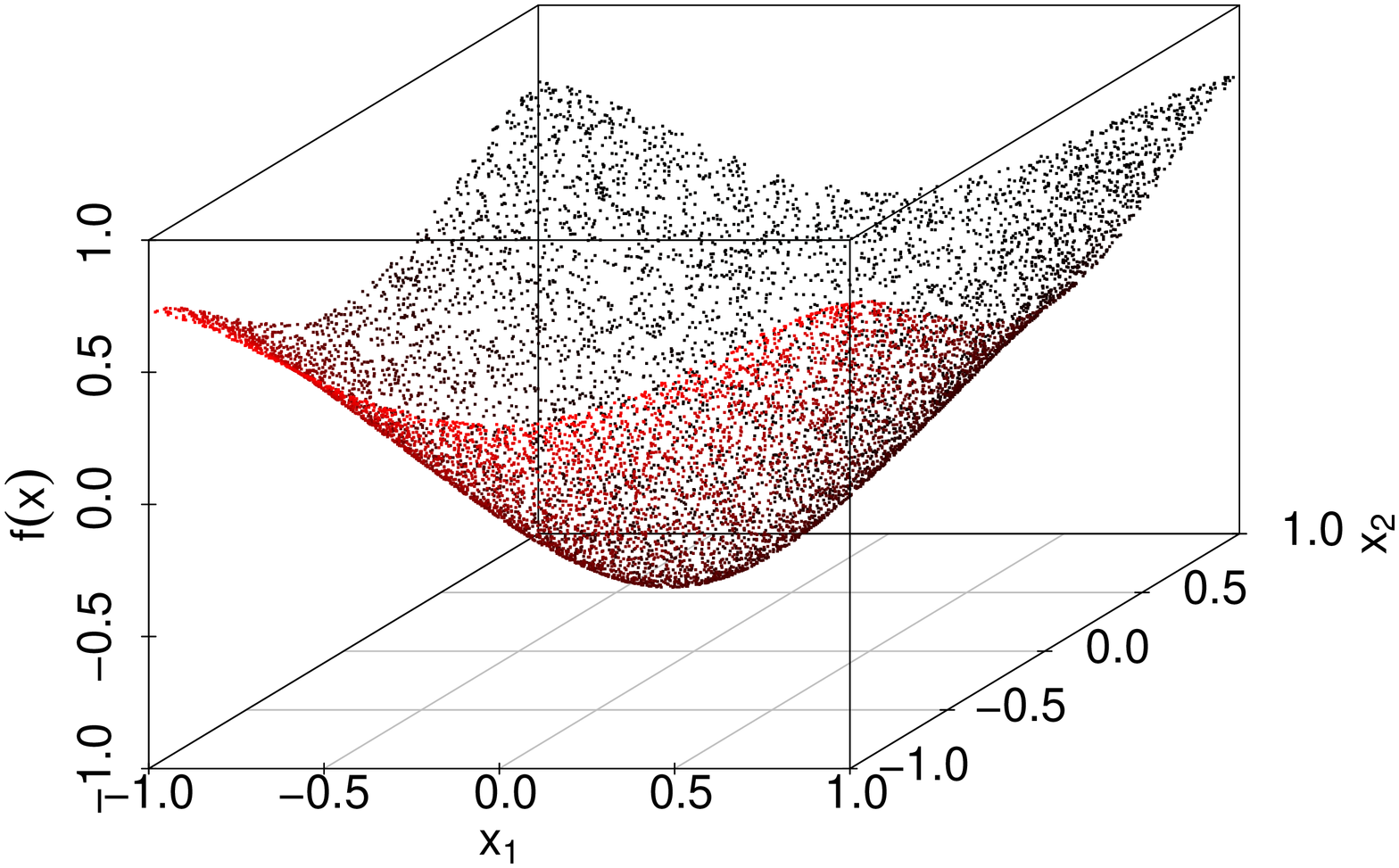}}
\caption[Predictions for the artificial data sets of type 4]{Predictions for the 
artificial data sets of Type \RM{4}, drawn from the circular step function in Figure 
\ref{fig::experiments::basic_fct_4} with noise independent of $x$.}
\label{fig::experiments::prediction_artificial_data_3_2}
\end{figure}

\begin{figure}[p]
\subfigure[LS-SVM using $1\,000$ training samples]{\includegraphics[angle=270, scale=0.28]{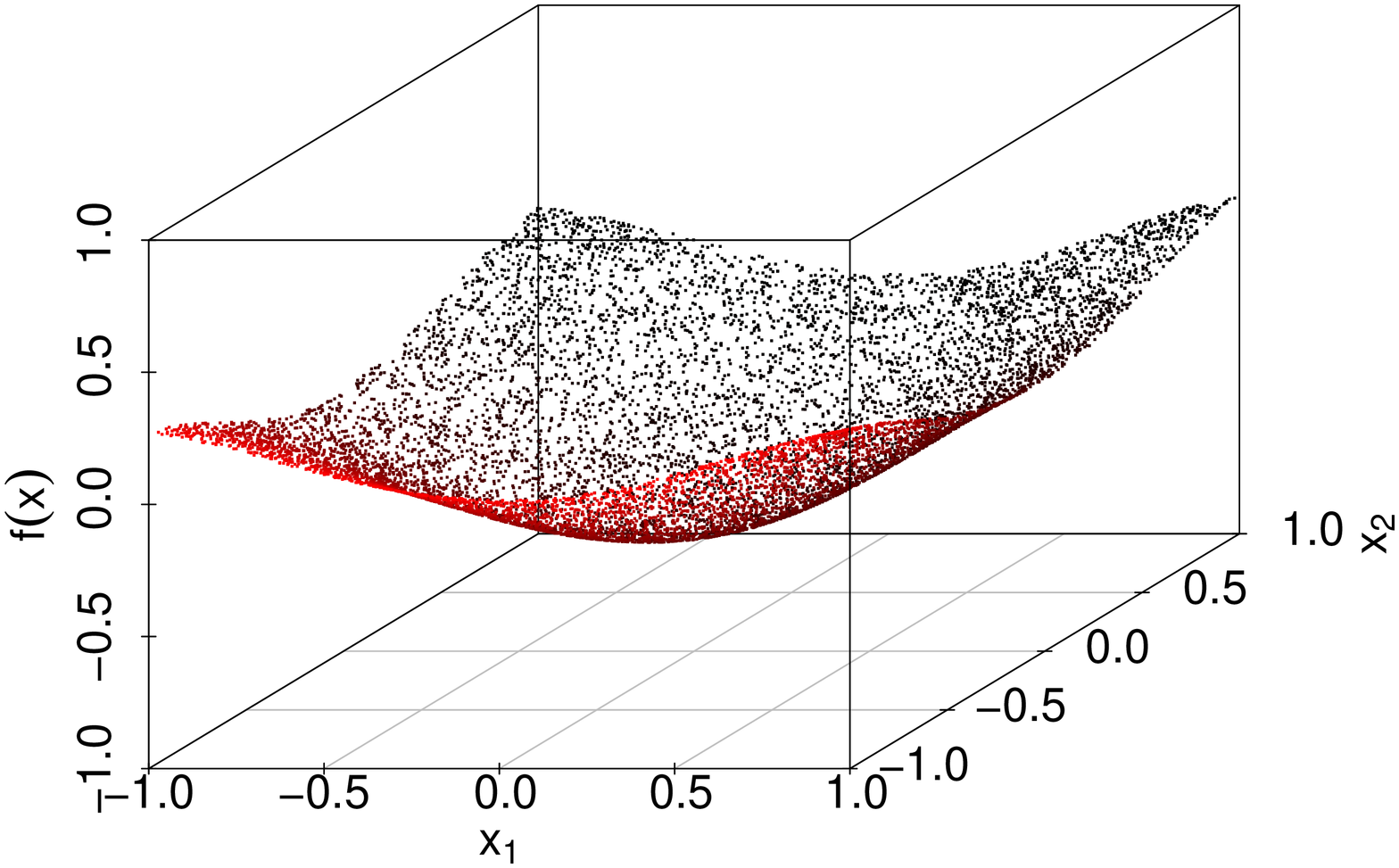}}\hfill
\subfigure[LS-SVM using $10\,000$ training samples]{\includegraphics[angle=270, scale=0.28]{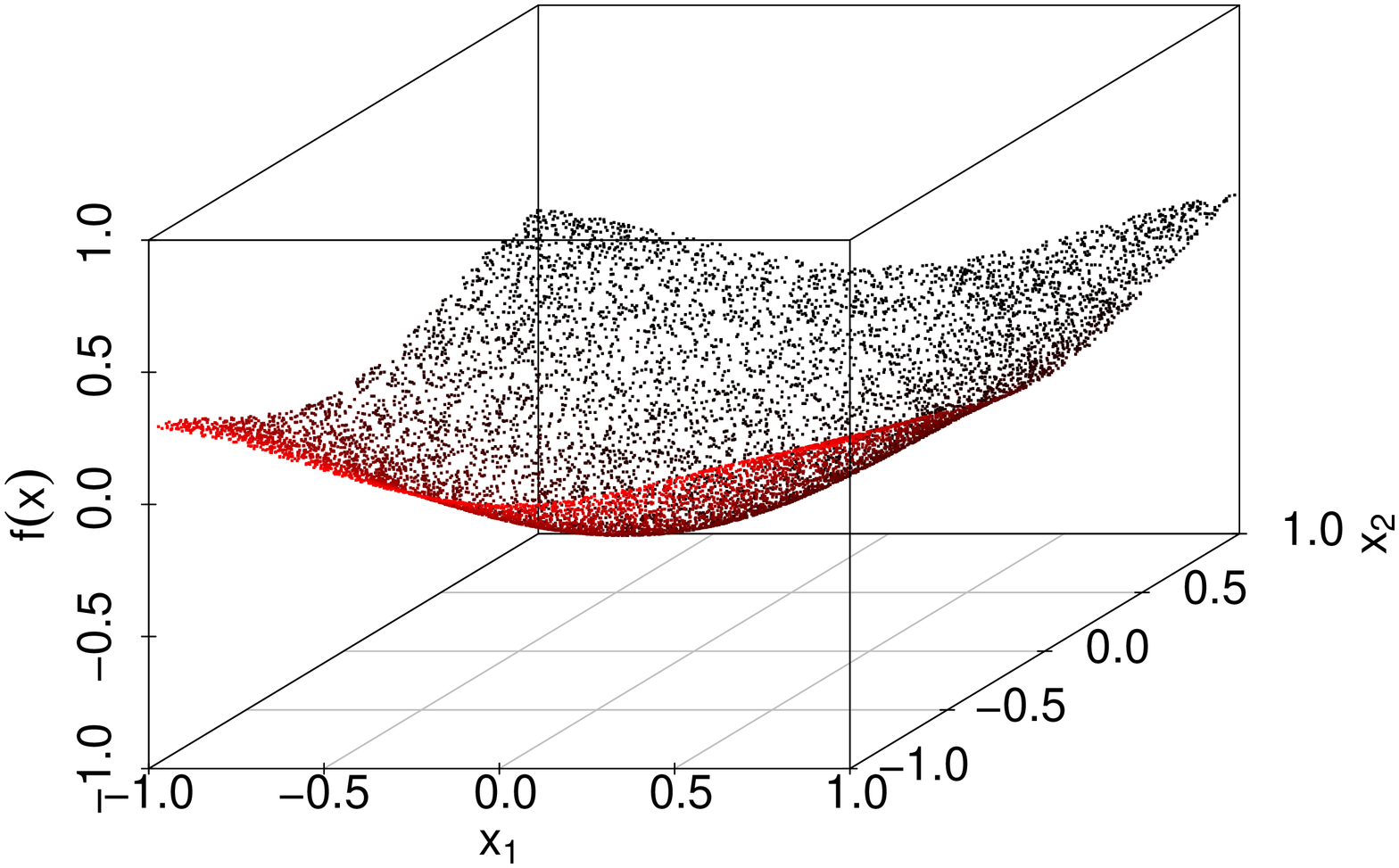}}\\
\subfigure[VP-SVM using radius $r=0.5$ and $1\,000$ training samples]{\includegraphics[angle=270, scale=0.28]{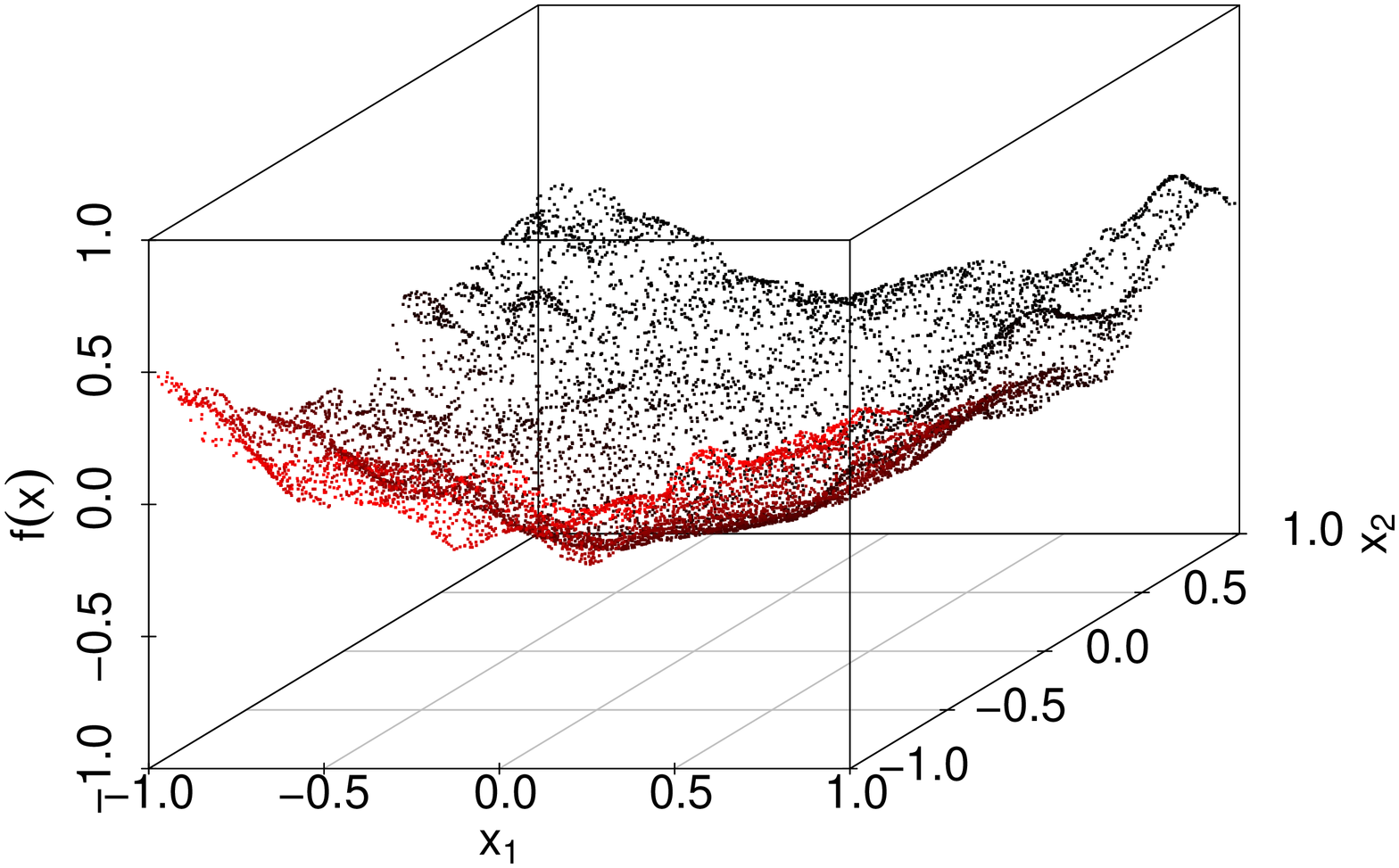}}\hfill
\subfigure[VP-SVM using radius $r=0.5$ and $10\,000$ training samples]{\includegraphics[angle=270, scale=0.28]{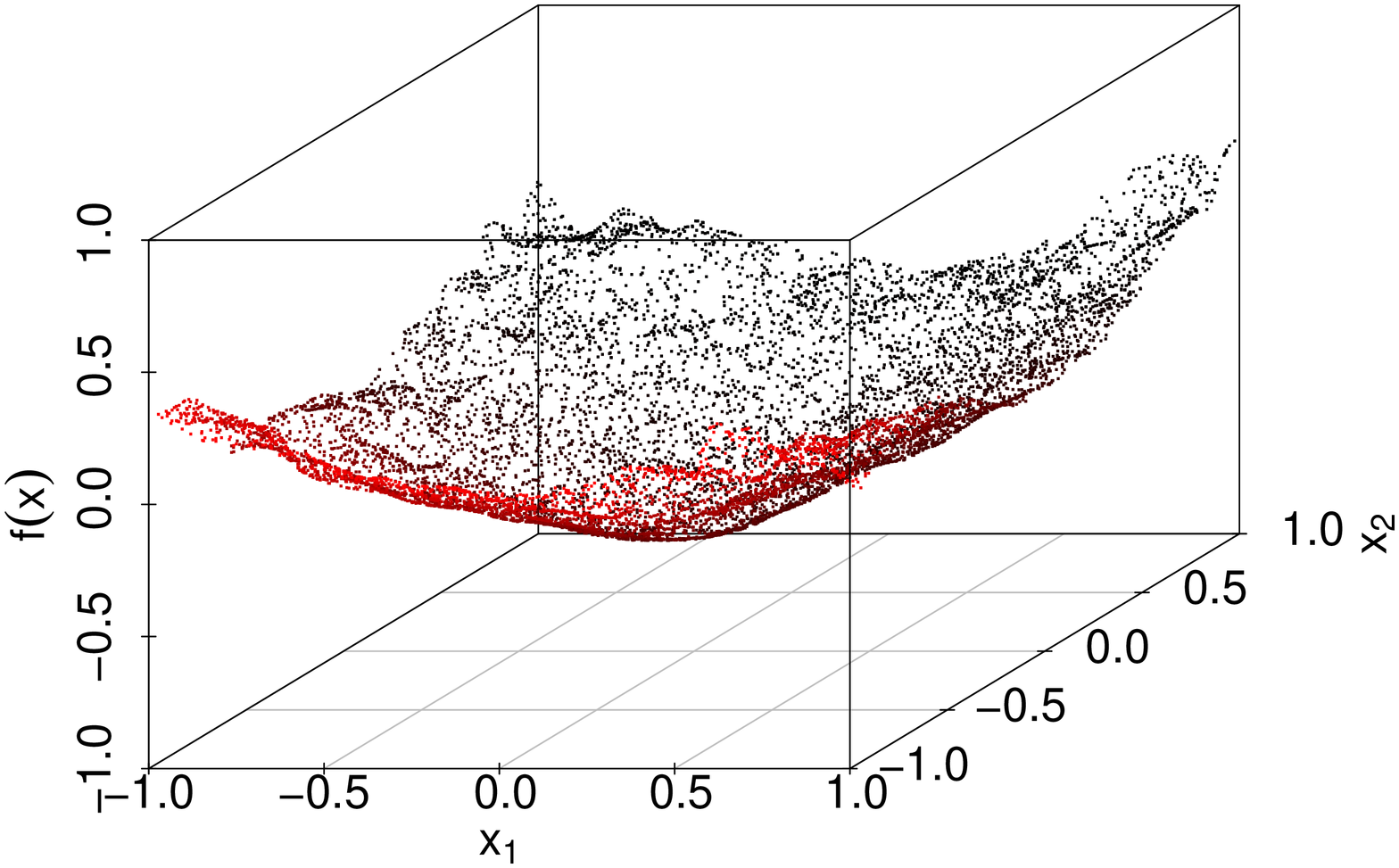}}\\
\subfigure[RC-SVM using $15$ working sets and $1\,000$ training samples]{\includegraphics[angle=270,scale=0.28]{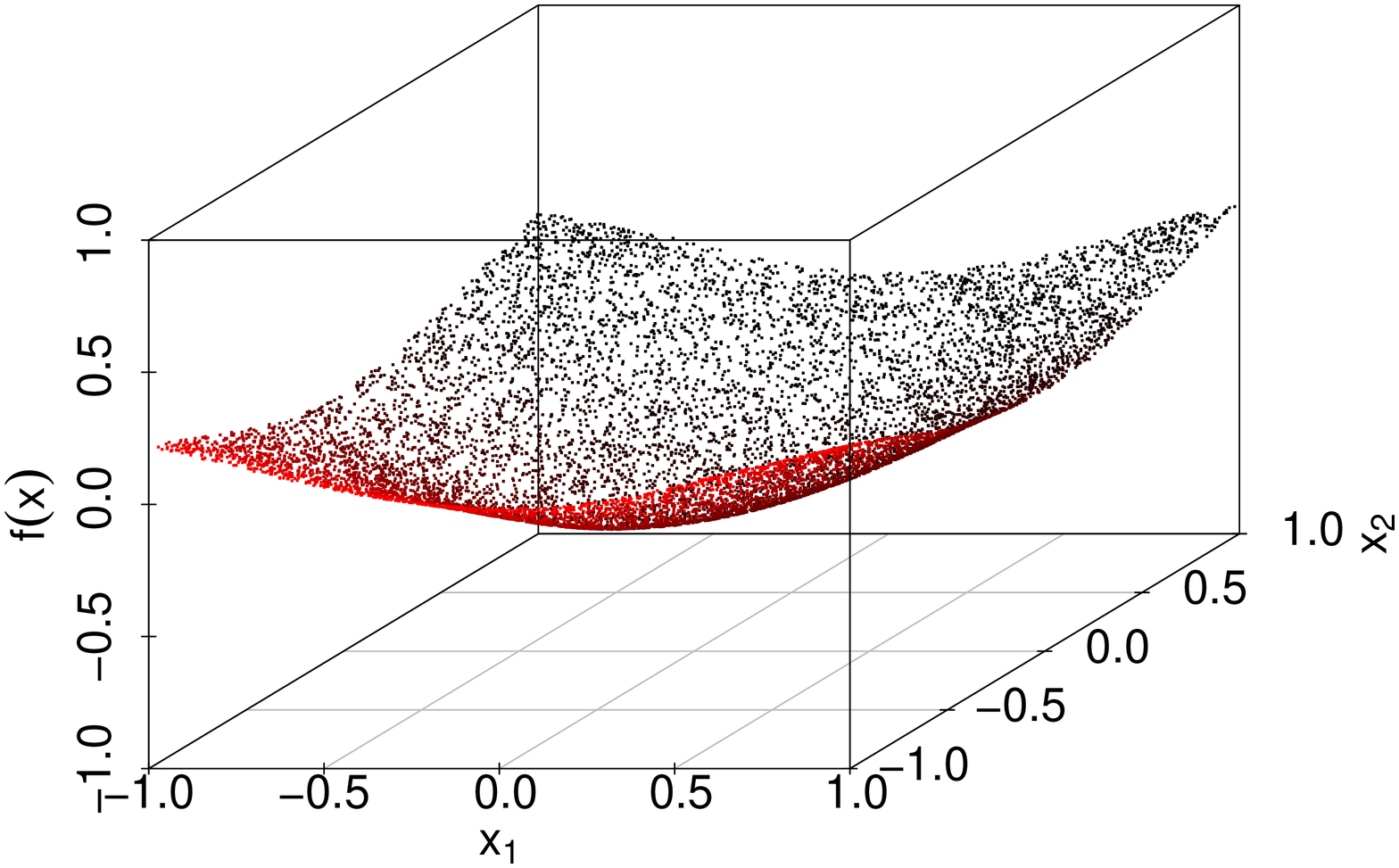}}\hfill
\subfigure[RC-SVM using $15$ working sets and $10\,000$ training samples]{\includegraphics[angle=270,scale=0.28]{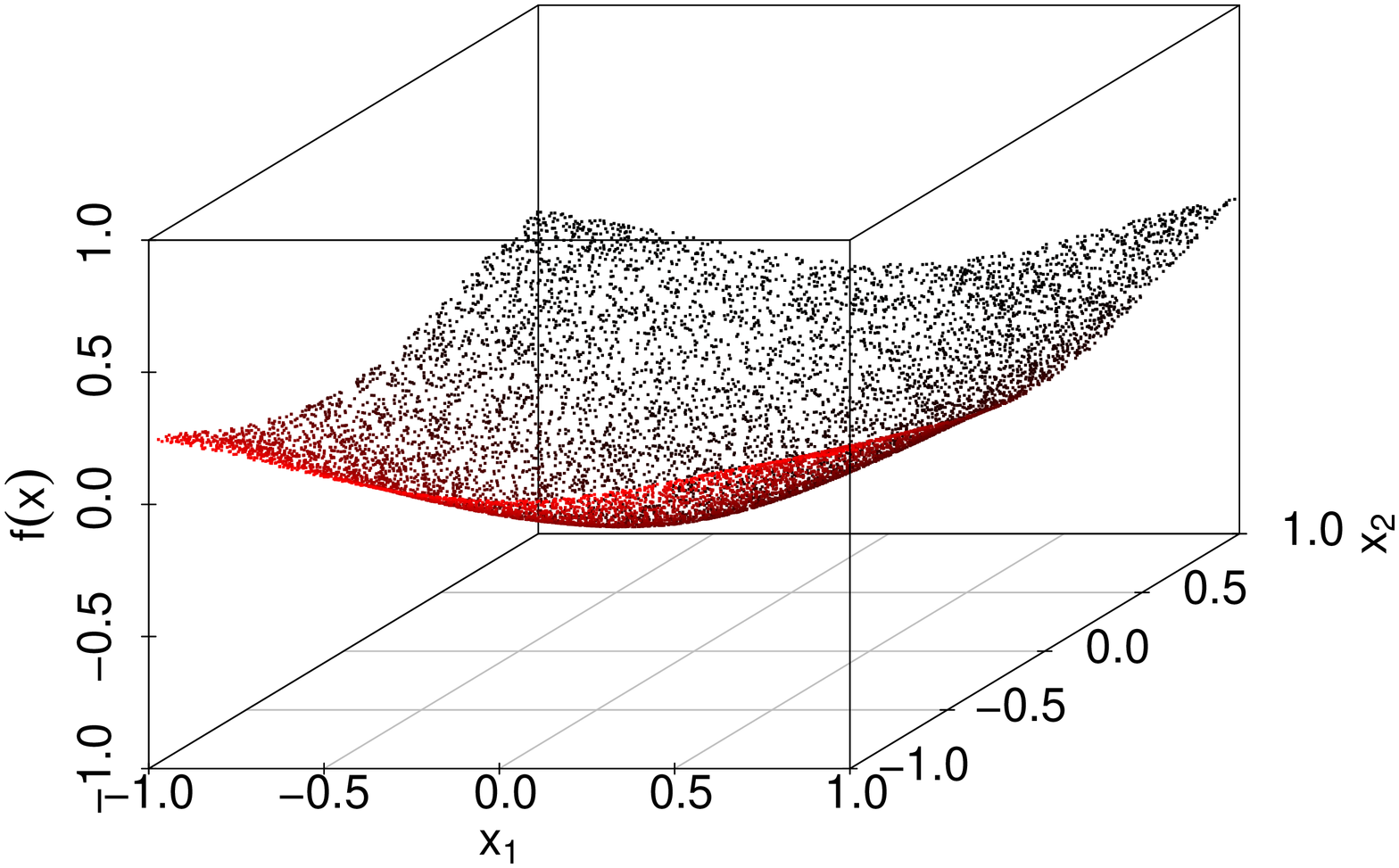}}
\caption[Predictions for the artificial data sets of type 5]{Predictions for the 
artificial data sets of Type \RM{5}, drawn from the 2-dimensional Euclidean norm in Figure 
\ref{fig::experiments::basic_fct_5} with noise independent of $x$.}
\label{fig::experiments::prediction_artificial_data_5_2}
\end{figure}

It remains to address  
the last question. To this end, we consider the results on the various artificial data sets, 
on the one hand, for the LS-SVM  and, on the other hand, for the VP-SVM performing fastest for 
$n_\text{train}=10\,000$.
Moreover, for the sake of comparability, we again add to this selection the two RC-SVMs 
training roughly as fast as the VP-SVM for the artificial data sets of size 
$n_\text{train}=10\,000$. 
However, for the artificial data sets of Type \RM{1}, \RM{2}, and \RM{3}, none of the executed 
RC-SVMs trained faster for 
$n_\text{train}= 10\,000$ than the VP-SVM with the smallest radius, so that we only 
consider one RC-SVM in these cases. The required average training times and average test 
errors of the selected SVMs are illustrated in Subfigures 
\subref{fig::experiments::art_data_1_2_time}--\subref{fig::experiments::art_data_1_2_L2error}
of Figures 
\ref{fig::experiments::artificial_data_1_2}--\ref{fig::experiments::artificial_data_5_2} 
and summarized in Tables 
\ref{tab::experiments::art_data_1_2}--\ref{tab::experiments::art_data_5_2}.
Here, we note that, for the artificial data of Type \RM{1}, \RM{2}, and \RM{3}, the VP-SVM 
using the smallest of the applied radii trains fastest for $n_\text{train}=10\,000$, while 
for the artificial data of Type \RM{4} and \RM{5} it is the VP-SVM using the second smallest 
radius.

Expectedly, we detect an evident improvement of the various average empirical SVM 
solutions using $10\,000$ training samples instead of $1\,000$ samples. 
Besides, the considered VP-SVM trains substantially faster than the standard LS-SVM with less 
than $11\%$ of the LS-SVM's training time for $n_\text{train}=10\,000$. 
Additionally, the VP-SVM's test errors are usually considerably lower than the test errors of 
the LS-SVM. Regarding the test errors of the RC-SVMs, we note that, in the majority of cases, 
they are higher than the VP-SVM's and the LS-SVM's test errors.

So far, we examined the behavior of LS-, VP-, and RC-SVMs in terms of training time and 
test error. Let us finally compare the three different kinds of SVMs w.r.t.~their optical 
appearance. To this end, the average empirical SVM solutions are plotted in Figures 
\ref{fig::experiments::prediction_artificial_data_1_2}--\ref{fig::experiments::prediction_artificial_data_5_2}
for the different artificial data sets of size $n_\text{train}=1\,000$ and $10\,000$. Here, 
note that, for the artificial data of Type \RM{4} and \RM{5}, we do not consider both RC-SVMs 
training roughly as fast as the selected VP-SVM but only the one of the both RC-SVMs with the
lower test error. 

The observation that, for the artificial data of Type \RM{1}, \RM{2}, and \RM{3}, the VP-SVMs 
perform best, 
is reinforced by the average empirical VP-SVM solutions illustrated in Figures 
\ref{fig::experiments::prediction_artificial_data_1_2}--\ref{fig::experiments::prediction_artificial_data_4_2}.
More precisely, Figure \ref{fig::experiments::prediction_artificial_data_1_2} shows that 
only the VP-SVMs exhaust the widths of the steps of $\fB$ almost completely. Moreover, 
in Figure \ref{fig::experiments::prediction_artificial_data_2_2} the smoothness interruptions 
of $\fB$ are again best illustrated by the VP-SVMs, which becomes even more evident in
Figure \ref{fig::experiments::prediction_artificial_data_2_2_zoom}. Besides, Figure 
\ref{fig::experiments::prediction_artificial_data_4_2} illustrates that the peaks of $\fB$ are 
best reproduced by the VP-SVMs. Considering the LS- and the RC-SVMs, we can not draw an 
universally valid conclusion, which one performs worse. In particular, Figures 
\ref{fig::experiments::prediction_artificial_data_1_2} and 
\ref{fig::experiments::prediction_artificial_data_2_2} show that, for the artificial data sets 
of Type \RM{1} and \RM{2}, both, the average empirical LS- and RC-SVM solutions, are not
very well suited to the Bayes decision function. 
Considering the data sets of Type \RM{3}, the 
LS-SVMs dominate the RC-SVMs in terms of the better test errors, though both kinds of SVMs 
do not reproduce the peaks of $\fB$, especially for small values of $|x|$.

It remains to optically analyze the results of the two-dimensional data sets in the following. 
For the artificial data sets of Type \RM{4}, the VP-SVM using $10\,000$ training samples 
achieves the best test error. Moreover, ensuing the optical impression, this VP-SVM is the 
only one of the considered SVM types that reflects the circular steps 
of the Bayes decision function as in Figure \ref{fig::experiments::basic_fct_4}, 
cf.~Figure \ref{fig::experiments::prediction_artificial_data_3_2}.
Finally, for the data sets of Type \RM{5}, it is always the LS-SVM which 
performs best in terms of the test errors, cf.~Table \ref{tab::experiments::art_data_5_2}. 
This observation is also substantiated optically. To be more precise, for 
$n_\text{train}=1\,000$, the uneven average empirical decision function induced by the VP-SVM 
(cf.~Figure \ref{fig::experiments::prediction_artificial_data_5_2}) shows that 
the RC-SVM even performs better than the VP-SVM. However, for $n_\text{train}=10\,000$, the 
VP-SVM results are substantially improved such that the RC-SVM is  
now outperformed by the VP-SVM.

Recapulatory, we realize that the VP-SVMs possess the most distinctive ability to 
handle smoothness interruptions of the Bayes decision function in most of our artificial data 
cases, especially if $n_\text{train}=10\,000$. 
For the sake of completeness, we point 
out that the worst performance was induced by the RC-SVMs in almost all cases, in particular for
a training sample size amounting to $10\,000$.

\subsection{Conclusions}

Finally, we summarize the essential findings of the previous subsections, where we considered 
standard LS-SVMs and two kinds of localized SVMs, namely VP-SVMs and RC-SVMs.
As just analyzed in Subsection \ref{subsec::exp::smoothness_interruptions}, VP-SVMs      
have the evident advantage that they manage smoothness interruptions of the Bayes decision 
function better than  LS- and RC-SVMs.

The real-world data sets demonstrated that the RC-SVMs perform considerably worse than 
the LS-SVMs and the VP-SVMs, while the performance of VP-SVMs using small radii is improved 
for increasing sample sizes. To be more precise, VP-SVMs outperform LS-SVMs or at most leads to 
a negligible worsening compared to LS-SVMs for a fraction of the training time and without memory constraints
on the large data sets.
%
For very small data sets, however,
LS-SVMs actually train faster than VP-SVMs and, hence, are preferable. 
What is more, for data sets of size $n_\text{train}\leq 2\,500$ all 
LS-SVMs require less than 9s to train, so that there are probably no reasons to apply 
a VP-SVM. 
Besides, really small training sample sizes involve considerably smaller working sets for a 
VP-SVM using a small radius, so that it is hard to find a well suited prediction.

Furthermore, despite a faster training procedure, a VP-SVM  using a sufficiently small radius 
induces considerably lower test errors for sample sizes $n_\text{train}>100\,000$ than a 
LS-SVM for training data 
sets that still enable computational feasibility.

\section{Proofs}\label{sec proofs}

This section is dedicated to prove the results of the previous sections. 
We begin with the proof of Lemma \ref{Lem relation r and m} relating the radius $r$ of 
a cover $B_{r}(z_1),\ldots,B_{r}(z_m)$ of $X$ defined by \eqref{Def. B_j} with the 
number $m$ of centers $z_1,\ldots,z_m$.

\begin{proof}[Proof of Lemma \ref{Lem relation r and m}]
First of all, let us recall the $m$-th entropy number of $X$ defined by 
\begin{align*}
 \e_m(X) := \inf\left\{ \e>0 : \exists z_1,\ldots,z_m\in X \text{ such that } 
                       X\subset \bigcup_{j=1}^m (z_j+\e B_{\ell_2^d}) \right\}\,.
\end{align*}
Since $X\subset c B_{\ell_2^d}$, the $m$-th entropy number of $X$ can be upper bounded by
\begin{align*}
 \e_m(X) \leq 2 \e_m(c B_{\ell_2^d}) \leq 2 c \e_m(B_{\ell_2^d})\,.
\end{align*}
Additionally, we know by \citep[Section 1.1]{ CaSt90} that
\begin{align*}
  m^{-\frac{1}{d}} \leq \e_m(B_{\ell_2^d}) \leq 4 m^{-\frac{1}{d}}\,,
\end{align*}
so that we can find a cover $(B_{j})_{j=1,\ldots,m}$ of $X\subset c B_{\ell_2^d}$
satisfying 
\begin{align*}
 r \leq 8 c m^{-\frac{1}{d}} \,.
\end{align*}
\end{proof}

\subsection{Proofs of Section \ref{general results}}

In Section \ref{general results} we presented a lemma that related the risk w.r.t.~the 
loss $L$ to the risk w.r.t.~the restricted loss $L_j$ and also transferred this 
result to the excess risk. Hereafter, the proof of this lemma can be found.

\begin{proof}[Proof of Lemma \ref{Lem RLP splitting}]
Simple transformations using $A\cup B=X$ and $A\cap B=\emptyset$ show
 \begin{align*}
   \RLP(f)
 &= \int_{X\times Y} L\left(x,y,\ind_A(x)f_A(x)+\ind_B(x)f_B(x)\right)\,d\P(x,y) \\
 & = \int_{X\times Y} \ind_{A}(x)  L(x,y,f_A(x)) 
     + \ind_{B}(x)  L(x,y,f_B(x))\, d\P(x,y) \\
 & = \RR_{L_A,\P}(f_A) + \RR_{L_B,\P}(f_B) \,.
\end{align*}
The second assertion follows immediately.
\end{proof}

To derive the new oracle inequality of Theorem \ref{main thm.} we first have to relate 
the entropy numbers of $H_j$, $j\in\{1,\ldots,m\}$, to those of $H$. To 
this end, we consider a similar concept to entropy numbers, namely covering numbers, 
cf.~\citep[Definition 9.3]{ GyKoKrWa02} or \citep[Definition 6.19]{ StCh08}.

\begin{definition}\label{Def. covering number}
 Let $(T,d)$ be a metric space and $\e>0$. A subset $S\subset T$ is called an $\e$-net 
of $T$ if for all $t\in T$ there exists an $s\in S$ such that $d(s,t)\leq\e$. 
Furthermore, we define the $\e$-covering number of $T$ by 
\begin{align*}
 \NN(T,d,\e):=\inf\left\{ n\geq 1: \exists s_1,\ldots,s_n\in T \text{ such that }
 T\subset \bigcup_{i=1}^n B_d(s_i,\e)\right\}\,, 
\end{align*}
where $\inf\emptyset:=\infty$ and $B_d(s,\e):=\{t\in T : d(t,s)\leq\e\}$.
\end{definition}

Note that an upper bound on entropy numbers involves a bound on covering numbers. To 
be more precise, for a metric space $(T,d)$ and constants $a>0$ and $q>0$, the implication 
\begin{align}\label{entropy -> covering number}
 e_i(T,d)\leq a i^{-1/q}\,,\qquad i\geq 1
 \qquad\qquad\Longrightarrow\qquad\qquad
 \ln \NN(T,d,\e)\leq \ln(4)\left(\frac{a}{\e}\right)^q \,,\qquad\forall\ \e>0
\end{align}
holds by \citep[Lemma 6.21]{ StCh08}. Additionally, \citep[Exercise 6.8]{ StCh08} yields 
the opposite implication, namely
\begin{align}\label{covering -> entropy number}
 \ln \NN(T,d,\e) < \left(\frac{a}{\e}\right)^q \,,\qquad \e>0
 \qquad\qquad\Longrightarrow\qquad\qquad
 e_i(T,d)\leq 3^{1/q} a i^{-1/q}\,,\qquad \forall\ i\geq 1\,.
\end{align}
Recall that we pursue the target to estimate 
$e_i(\mathrm{id} : H \to L_2(\P_X))$. In fact, the equivalence of 
entropy and covering numbers enables us to estimate the covering number 
$\NN(B_H,\|\,\cdot\,\|_{L_2(\P_X)},\e)$ of $H$ instead.

\begin{lemma}\label{Lem covering number}
Let $\nu$ be a distribution on $X$ and $A,B\subset X$ with $A\cap B=\emptyset$.
Moreover, let $H_A$ and $H_B$ be RKHSs on $A$ and $B$
that are embedded into $L_2(\nu_{|A})$ and $L_2(\nu_{|B})$, respectively.
Let the extended RKHSs $\hat{H}_A$ and $\hat{H}_B$ be defined as in Lemma 
\ref{Lemma extended RKHS} and denote their direct 
sum by $H$ as in \eqref{RKHS as direct sum}, where the norm is given by 
\eqref{norm RKHS as direct sum} with $\lb_A,\lb_B>0$.
Then, for the $\e$-covering number of $H$ w.r.t.~$\|\,\cdot\,\|_{L_2(\nu)}$, 
we have
\begin{align*}
 \NN(B_{H},\|\,\cdot\,\|_{L_2(\nu)},\e)
 \leq \NN\left(\lb_A^{-1/2}B_{\hat{H}_A},\|\,\cdot\,\|_{L_2(\nu_{|A})},\e_A\right)\cdot
      \NN\left(\lb_B^{-1/2}B_{\hat{H}_B},\|\,\cdot\,\|_{L_2(\nu_{|B})},\e_B\right)\,, 
\end{align*}
where $\e_A,\e_B>0$  and $\e:=\sqrt{\e_A^2+\e_B^2}$.
\end{lemma} 

\begin{proof}
First of all, we assume that there exist $a,b\in\N$ and functions 
$\hat{f}_{1},\ldots,\hat{f}_{a}\in \lb_A^{-\frac{1}{2}}B_{\hat{H}_A}$ 
and $\hat{h}_{1},\ldots,\hat{h}_{b}\in \lb_B^{-\frac{1}{2}}B_{\hat{H}_B}$ 
such that 
$\{\hat{f}_{1},\ldots,\hat{f}_{a}\}$ is an $\e_A$-cover of 
$\lb_A^{-\frac{1}{2}}B_{\hat{H}_A}$ w.r.t.~$\|\,\cdot\,\|_{L_2(\nu_{|A})}$, 
$\{\hat{h}_{1},\ldots,\hat{h}_{b}\}$ is an $\e_B$-cover of 
$\lb_B^{-\frac{1}{2}}B_{\hat{H}_B}$ w.r.t.~$\|\,\cdot\,\|_{L_2(\nu_{|B})}$,
\begin{align*}
a = \NN(\lb_A^{-\frac{1}{2}}B_{\hat{H}_A},\|\,\cdot\,\|_{L_2(\nu_{|A})},\e_A)
\qquad\text{ and }\qquad
b = \NN(\lb_B^{-\frac{1}{2}}B_{\hat{H}_B},\|\,\cdot\,\|_{L_2(\nu_{|B})},\e_B)\,.
\end{align*}
That is, for every function $\hat{g}_A\in\lb_A^{-\frac{1}{2}}B_{\hat{H}_A}$, there 
exists an $i_A\in\{1,\ldots,a\}$ such that
\begin{align}\label{e-net H_A}
 \left\|\hat{g}_A-\hat{f}_{i_A}\right\|_{L_2(\nu_{|A})}\leq \e_A \,,
\end{align}
and for every function $\hat{g}_B\in\lb_B^{-\frac{1}{2}}B_{\hat{H}_B}$, there 
exists an $i_B\in\{1,\ldots,b\}$ such that
\begin{align}\label{e-net H_B}
 \left\|\hat{g}_B-\hat{h}_{i_B}\right\|_{L_2(\nu_{|B})}\leq \e_B \,.
\end{align}
Let us now consider an arbitrary function $g\in B_{H}$. Then there exists an 
$\hat{g}_A\in \lb_A^{-\frac{1}{2}}B_{\hat{H}_A}$ and an 
$\hat{g}_B\in \lb_B^{-\frac{1}{2}}B_{\hat{H}_B}$ such that $g=\hat{g}_A+\hat{g}_B$. 
Together with \eqref{e-net H_A} and \eqref{e-net H_B}, this implies
\begin{align*}
  \left\|g-\left(\hat{f}_{i_A}+\hat{h}_{i_B}\right)\right\|_{L_2\left(\nu\right)}^2
  & = \left\|\left(\hat{g}_A-\hat{f}_{i_A}\right)+\left(\hat{g}_B-\hat{h}_{i_B}\right)
    \right\|_{L_2\left(\nu\right)}^2 \\
  & = \left\|\hat{g}_A-\hat{f}_{i_A}\right\|_{L_2(\nu_{|A})}^2
    + \left\|\hat{g}_B-\hat{h}_{i_B}\right\|_{L_2(\nu_{|B})}^2 \\
  & \leq \e_A^2+\e_B^2 \\
  & =: \e^2 \, .
\end{align*}
With this, we know that
\begin{align*}
 \left\{ \hat{f}_{i_A}+\hat{h}_{i_B} \;:\;
         \hat{f}_{i_A} \in \{\hat{f}_{1},\ldots,\hat{f}_{a}\} \text{ and }
         \hat{h}_{i_B}\in \{\hat{h}_{1},\ldots,\hat{h}_{b}\} \right\}
\end{align*}
is an $\e$-net of $H$ w.r.t.~$\|\,\cdot\,\|_{L_2\left(\nu\right)}$. Concerning the 
$\e$-covering number of $H$, this finally implies
\begin{align*}
 \NN(B_{H},\|\,\cdot\,\|_{L_2(\nu)},\e) \!
  \leq\! a\!\cdot\! b\!
  = \!\NN\!\left(\lb_A^{-1/2}B_{\hat{H}_A},\|\,\cdot\,\|_{L_2(\nu_{|A})},\e_A\right)\!\cdot
     \NN\!\left(\lb_B^{-1/2}B_{\hat{H}_B},\|\,\cdot\,\|_{L_2(\nu_{|B})},\e_B\right)\!\!\,.
\end{align*}
\end{proof}

Based on Lemma \ref{Lem covering number}, the following theorem relates entropy numbers 
of $H_A$ and $H_B$ to those of $H$.

\begin{theorem}\label{entropy estim}
Let $\P_X$ be a distribution on $X$ and $A_1,\ldots,A_m\subset X$ be pairwise disjoint. 
Moreover, we assume (H) with weights $\lb_1,\ldots,\lb_m>0$.
In addition, assume that there exist constants $p\in(0,1)$ and $a_j>0$, 
$j\in \{1,\ldots,m\}$, such that for every $j\in \{1,\ldots,m\}$ 
\begin{align}
  e_i(\mathrm{id} : H_j\to L_2(\TrP{A_j})) \leq  a_j\, i^{-\frac{1}{2p}}\,,
  \qquad  \qquad i\geq 1\,. \label{ass entropy estim}
\end{align}
Then we have
\begin{align*}
 e_i(\mathrm{id} : H\to L_2(\P_X))
 \leq 2 \sqrt{m} \left(3\ln(4)\sum_{j=1}^m\lb_{j}^{-p} a_j^{2p}\right)^{\frac{1}{2p}} 
        i^{-\frac{1}{2p}} \,, \qquad\qquad i\geq1\,,
\end{align*}
and, for the average entropy numbers,
\begin{align*}
 \E_{D_X\sim\P_X^n} e_i(\mathrm{id} : H\to L_2(\D_X))
 \leq c_p \sqrt{m} \left(\sum_{j=1}^m\lb_{j}^{-p} a_j^{2p}\right)^{\frac{1}{2p}} 
        i^{-\frac{1}{2p}} \,, \qquad\qquad i,n\geq1\,.
\end{align*}
\end{theorem}

\begin{proof}[Proof of Theorem \ref{entropy estim}]
First of all, 
note that the restriction operator $\I : B_{\hat{H}_j} \to B_{H_j}$ with $\I\hat{f}=f$ 
is an isometric isomorphism. Together with 
\citep[(A.36)]{ StCh08} 
and assumption \eqref{ass entropy estim}, this yields
\begin{align*}
 e_i(\lb_j^{-\frac{1}{2}}B_{\hat{H}_j},L_2(\TrP{A_j})) 
 & = 2 \lb_{j}^{-\frac{1}{2}} e_i(B_{\hat{H}_j}, L_2(\TrP{A_j})) \\
 & \leq 2 \lb_{j}^{-\frac{1}{2}} \|\I : B_{\hat{H}_j} \to B_{H_j}\|
      e_i(B_{H_j}, L_2(\TrP{A_j})) \\
 & \leq 2 \lb_{j}^{-\frac{1}{2}} a_j i^{-\frac{1}{2p}}\,.
\end{align*}
Furthermore, we know by \eqref{entropy -> covering number} 
that
\begin{align*}
 \ln \NN\left(\lb_j^{-\frac{1}{2}}B_{\hat{H}_j},\|\,\cdot\,\|_{L_2(\TrP{A_j})},\e\right) 
 \leq \ln(4) \left(2 \lb_{j}^{-\frac{1}{2}} a_j \right)^{2p} \e^{-2p}
\end{align*}
holds for all $\e>0$. With this and $\e_j:=\frac{\e}{\sqrt{m}}$ for every 
$j\in\{1,\ldots,m\}$, Lemma \ref{Lem covering number} implies
\begin{align*}
 \ln \NN(B_{H},\|\,\cdot\,\|_{L_2(\P_X)},\e) 
 & \leq \ln\left(\prod_{j=1}^m \NN\left(\lb_j^{-\frac{1}{2}}B_{\hat{H}_j},
        \|\,\cdot\,\|_{L_2(\TrP{A_j})},\e_j\right)\right) \\
 & = \sum_{j=1}^m \ln\NN\left(\lb_j^{-\frac{1}{2}}B_{\hat{H}_j},
     \|\,\cdot\,\|_{L_2(\TrP{A_j})},\frac{\e}{\sqrt{m}}\right) \\
 & \leq \sum_{j=1}^m\ln(4) \left(2\lb_{j}^{-\frac{1}{2}}
        a_j\right)^{2p} \left(\frac{\sqrt{m}}{\e}\right)^{2p} \\
 & = \left(2 \ln(4)^{\frac{1}{2p}} \sqrt{m} \left(\sum_{j=1}^m \lb_{j}^{-p} a_j^{2p}
     \right)^{\frac{1}{2p}}\right)^{2p} \e^{-2p}\,.
\end{align*}
Using \eqref{covering -> entropy number}, 
the latter bound for the covering number of $B_{H}$ finally implies the following 
entropy estimate 
\begin{align*}
 e_i(\mathrm{id} : H \to L_2(\P_X))
 & \leq 3^\frac{1}{2p} \left(2 \ln(4)^{\frac{1}{2p}} \sqrt{m} \left(\sum_{j=1}^m
        \lb_{j}^{-p} a_j^{2p}\right)^{\frac{1}{2p}}\right) i^{-\frac{1}{2p}} \\
 & \leq 2 \left(3\ln(4)\right)^{\frac{1}{2p}}\sqrt{m} \left(\sum_{j=1}^m
        \lb_{j}^{-p} a_j^{2p}\right)^{\frac{1}{2p}} i^{-\frac{1}{2p}} \,.
\end{align*}
The second assertion immediately follows by \citep[Corollary 7.31]{StCh08}.
\end{proof}

Applying Theorem \ref{entropy estim}, we now prove Theorem \ref{main thm.} and thus 
an oracle inequality for VP-SVMs using an ordinary type of losses. 

\begin{proof}[Proof of Theorem \ref{main thm.}]
Since $H_1,\ldots,H_m$ are seperable RKHS of mesurable kernels $k_1,\ldots,k_m$, $H$ is a
seperable RKHS and its kernel $k$ is measurable, too. Furthermore, Theorem 
\ref{entropy estim} yields 
\begin{align*}
 \E_{D_X\sim\P_X^n} e_i(\mathrm{id} : H\to L_2(\D_X))
 \leq  c_p \sqrt{m} \left(\sum_{j=1}^m\lb_{j}^{-p} a_j^{2p}\right)^{\frac{1}{2p}} 
        i^{-\frac{1}{2p}} \,, \qquad\qquad i,n\geq1\,.
\end{align*}
That is, we can apply \citep[Theorem 7.23]{StCh08} for a regularization parameter 
$\tilde{\lb}=1$ and, for all fixed $\t>0$ and $\lb_j>0$, $j\in \{1,\ldots,m\}$, we obtain
\begin{align*}
 & \sum_{j=1}^m\lb_j\|f_{\D_{j},\lb_j}\|^2_{\hat{H}_j}+\RLPJ(\fcl_{\D,\bs\lb})-\RBJ \\
 & = \|f_{\D,\bs\lb}\|^2_{H}+\RLPJ(\fcl_{\D,\bs\lb})-\RBJ \\  
 & \leq 9\left(\|f_0\|^2_{H}+\RLPJ(f_0)-\RBJ\right) 
        + C \left(a^{2p}n^{-1}\right)^{\frac{1}{2-p-\vt+\vt p}} 
        + 3\left(\frac{72 V\t}{n}\right)^{\frac{1}{2-\vt}} + \frac{15B_0\t}{n} \\
 & \leq 9\!\left(\sum_{j=1}^m\!\lb_j\|\ind_{A_j}f_0\|^2_{\hat{H}_j}\!+\!\RLPJ(f_0)-\!\RBJ\right) 
       \! +\! C \!\left(a^{2p}n^{-1}\right)^{\frac{1}{2-p-\vt+\vt p}} 
       \! + \!3\!\left(\frac{72 V\t}{n}\right)^{\frac{1}{2-\vt}}\! + \!\frac{15B_0\t}{n} 
\end{align*}
with probability $\P^n$ not less than $1-3e^{-\t}$, where $C>0$ is the constant of 
\cite[Theorem 7.23]{StCh08} only depending on $p$, $M$, $V$, $\vt$, and $B$. Moreover, 
\begin{align*}
 a := \max\left\{c_p \sqrt{m} \left(\sum_{j=1}^m\lb_{j}^{-p} a_j^{2p}\right)^{\frac{1}{2p}} , B\right\}\,,
\end{align*}
where we need $a\geq B$ since it is a condition of \citep[Theorem 7.23]{StCh08}.
\end{proof}

\subsection{Proofs Related to the Entropy Estimates of Section \ref{Sec entropy estim}}

In this subsection, just as in Section \ref{Sec entropy estim}, we focus on Gaussian RBF 
kernels and the associated RKHSs. To be more precise, we derive a bound for the entropy 
numbers of $H_\g(A)$, where $\g>0$ and $A\subset\Rd$ with $\mathring{A}\neq\emptyset$.

\begin{proof}[Proof of Theorem \ref{entropy number H_g}]
First of all, we consider the commutative diagram
\begin{align*}
  \begin{xy}
  \xymatrix{
    H_\g(A) \ar[rrr]^{\mathrm{id}} \ar[dd]_{\I_{B}^{-1}\circ \I_{A}} 
    & & & L_2(\TrP{A})   \\ \\
    H_\g(B)   \ar[rrr]_{\mathrm{id}} 
    & & & \ell_\infty(B) \ar[uu]_{\mathrm{id}}}
  \end{xy}
\end{align*}
where the extension operator $\I_{A} : H_\g(A)\to H_\g(\Rd)$ and the 
restriction operator $\I^{-1}_{B} : H_\g(\Rd) \to H_\g(B)$ given by
\citep[Corollary 4.43]{ StCh08} are isometric isomorphisms, so that
$\|\I_{B}^{-1}\circ \I_{A} : H_\g(A) \to H_\g(B)\|=1$.
Furthermore, for $f\in\ell_\infty(B)$, we have
\begin{align*}
 \|f\|_{L_2(\TrP{A})}
 = \left(\int_X \ind_{A}|f(x)|^2d\P_X(x)\right)^\frac{1}{2} 
 \leq \|f\|_\infty\left(\int_X \ind_{A}d\P_X(x)\right)^\frac{1}{2} 
 = \sqrt{\P_X(A)}\, \|f\|_\infty\,,
\end{align*}
i.e.~$\|\mathrm{id} : \ell_\infty(B) \to L_2(\TrP{A})\|\leq 
\sqrt{\P_X(A)}$. 
Together with \citep[(A.38) and (A.39)]{ StCh08} as well as \citep[Theorem 6.27]{ StCh08},
we obtain for all $i\geq 1$
\begin{align*}
 & e_i(\mathrm{id} : H_\g(A) \to L_2(\TrP{A})) \\
 & \leq \|\I_{B}^{-1}\circ \I_{A} : H_\g(A) \to H_\g(B)\|\cdot
        e_i(\mathrm{id} : H_\g(B) \to \ell_\infty(B)) \cdot
        \|\mathrm{id} : \ell_\infty(B) \to L_2(\TrP{A})\| \\
 & \leq \sqrt{\P_X(A)}\, c_{m,d} r^m\g^{-m}i^{-\frac{m}{d}}\, ,
\end{align*}
where $m\geq1$ is an arbitrary integer and $c_{m,d}$ a positive constant. 
For $p\in(0,1)$, the choice $m=\left\lceil\frac{d}{2p}\right\rceil$ finally yields
\begin{align*}
 e_i(\mathrm{id} : H_\g(A) \to L_2(\TrP{A})) 
 \leq \sqrt{\P_X(A)}\, c_{m,d} r^m\g^{-m}i^{-\frac{m}{d}} 
 \leq c_p \sqrt{\P_X(A)}\, r^\frac{d+2p}{2p}\g^{-\frac{d+2p}{2p}} i^{-\frac{1}{2p}} \,.
\end{align*}
\end{proof}

\subsection{Proofs Related to the Least Squares VP-SVMs}

In this subsection, we prove the results that are linked with the least squares loss, 
i.e.~the results of Section \ref{Sec LS loss}. Before we elaborate on the oracle 
inequality for VP-SVMs using the least squares loss as well as RKHSs of Gaussian kernels, 
we have to examine the excess risk
\begin{align}\label{eq excess risk}
 \RLPJT(f_0)-\RBJT
 = \|f_0-\fB\|^2_{L_2\left(\TrP{A_T}\right)} \,.
\end{align}
Let us begin by writing for fixed $\g_j>0$
\begin{align}\label{Def. K}
 K_{j}: \Rd\to\R\,,\qquad x\mapsto 
\sum_{\ell=1}^{s}{s\choose \ell}(-1)^{1-\ell}\left(
 \frac{2}{\ell^2\g_{j}^2\pi}\right)^\frac{d}{2}\exp\left(-\frac{2\|x\|_2^2}{\ell^2\g_{j}^2}\right)\,,
\end{align} 
and choosing $f_0 := \sum_{j=1}^{m} \ind_{A_{j}}\cdot ( K_{j} * \fB )$. 
Then \eqref{eq excess risk} can be estimated with the help of the following theorem, 
which is together with its proof basically a modification of 
\citep[Theorem 2.2]{ EbSt13}. Indeed, the 
proofs proceed mainly identically.
Note that we use the notation
\begin{align*}
 \g_{\max} := \max\{\g_1,\ldots,\g_m\} \qquad\text{and}\qquad
 \g_{\min} := \min\{\g_1,\ldots,\g_m\}
\end{align*}
in the following theorem and the associated proof.

\begin{theorem}\label{estim excess risk}
Let us fix some $q\in[1,\infty)$. Assume that $\nu$ is a finite measure on $\Rd$ with 
$\supp{\nu}=:X\subset\Rd$ and let $(A_{j})_{j=1,\ldots,m}$ be a partition of $X$. 
Furthermore, 
let $f:\Rd\to\R$ be such that $f\in B_{q,\infty}^{\a}(\nu)$ for some $\a\geq 1$.
For the functions $K_{j} : \mathbb{R}^d\to\mathbb{R}$, $j\in \{1,\ldots,m\}$, defined by 
\eqref{Def. K},
where $s:=\lfloor\a\rfloor+1$ and $\g_1,\ldots,\g_m>0$, we then have
\begin{align*}
 \|\sum_{j=1}^m \ind_{A_j} \cdot (K_j * f)-f\|_{L_q\left(\nu\right)}^{q} 
 \leq C_{\a,q}\,\left(\frac{\g_{\max}}{\g_{\min}}\right)^d \g_{\max}^{q\a} \,,
\end{align*}
where $C_{\a,q}:= \|f\|_{B_{q,\infty}^{\a}(\nu)}^q \left(\frac{d}{2}\right)^{
\frac{q\a}{2}} \pi^{-\frac{1}{4}} \Gamma\left(q\a+\frac{1}{2}\right)^{\frac{1}{2}}$.
\end{theorem}

\begin{proof}
In the following, we write $J:=\{1,\ldots,m\}$.
To show 
\begin{align*}
 \Bigl\|\sum_{j\in J} \ind_{A_j} \cdot (K_j * f)-f\Bigr\|_{L_q\left(\nu\right)}^{q} 
 \leq \|f\|_{B_{q,\infty}^{\a}(\nu)}^q \left(\frac{d}{2}\right)^{\frac{q\a}{2}} 
      \pi^{-\frac{1}{4}} \Gamma\left(q\a+\frac{1}{2}\right)^{\frac{1}{2}} 
      \left(\frac{\g_{\max}}{\g_{\min}}\right)^d \g_{\max}^{q\a}\,,
\end{align*}
we have to proceed in a similar way as in the proof of 
\citep[Theorem 2.2]{ EbSt13}.
First of all, we use the translation invariance of the Lebesgue measure and 
$\exp\left(-\|u\|_2^2\right)=\exp\left(-\|-u\|_2^2\right)$ ($u\in\Rd$) to obtain, 
for $x\in X$ and $j\in J$, 
\begin{align*}
 K_j * f\left(x\right) 
 & = \int_{\Rd}{\sum_{\ell =1}^{s}{s \choose \ell } \left(-1\right)^{1-\ell } 
     \frac{1}{\ell ^d}\left(\frac{2}{\g_{j}^2\pi}\right)^\frac{d}{2}
     \exp\left(-\frac{2\|x-t\|_2^2}{\ell^2\g_{j}^2}\right) f\left(t\right)} \, dt \\
 & = \int_{\Rd}\left(\frac{2}{\g_{j}^2\pi}\right)^\frac{d}{2} 
     \exp\left(-\frac{2\|h\|_2^2}{\g_{j}^2}\right) \left(\sum_{\ell =1}^{s}
     {s \choose \ell } \left(-1\right)^{1-\ell } f\left(x+\ell h\right)\right) \, dh  \,.
\end{align*} 
With this we can derive, for $q\geq 1$,
\begin{align*}
 & \left\|\sum_{j\in J} \ind_{A_j} \cdot (K_j * f)-f\right\|_{L_q\left(\nu\right)
   }^{q} \\
 & = \int_{\Rd}\left|\sum_{j\in J}\ind_{A_j}(x)\left(K_j * f\right)\left(x\right)
     -f\left(x\right)\right|^q \, d\nu\left(x\right)\\
 & \leq \int_{\Rd}\left(\sum_{j\in J}\ind_{A_j}(x)\left|K_j * f\left(x\right)
     -f\left(x\right)\right|\right)^q \, d\nu\left(x\right)\\
 & = \int_{\Rd}\sum_{j\in J}\ind_{A_j}(x)\left|K_j * f\left(x\right)
     -f\left(x\right)\right|^q \, d\nu\left(x\right)\\
 & = \sum_{j\in J}\int_{\Rd}\ind_{A_j}(x)\left|K_j * f\left(x\right)
     -f\left(x\right)\right|^q \, d\nu\left(x\right)\\
 & = \sum_{j\in J}\int_{\Rd}\ind_{A_j}(x)\left|\int_{\Rd}\!\!\left(\frac{2}{\g_{j}^2\pi}
     \right)^{\!\frac{d}{2}} \exp\!\left(-\frac{2\|h\|_2^2}{\g_{j}^2}\right) \!\!
     \left(\sum_{\ell =0}^{s}\!{s \choose \ell } 
     \left(-1\right)^{2s+1-\ell } \!f\left(x+\ell h\right)\right)  dh \right|^q \;
     d\nu\left(x\right) \\
 & = \sum_{j\in J}\int_{\Rd}\ind_{A_j}(x)\left|\int_{\Rd}\left(-1\right)^{s+1}\left(
     \frac{2}{\g_{j}^2\pi}
     \right)^\frac{d}{2} \exp\left(-\frac{2\|h\|_2^2}{\g_{j}^2}\right) 
     \triangle_h^s\left(f,x\right) \, dh \right|^q \, d\nu\left(x\right) \\ 
 & \leq \sum_{j\in J}\int_{\Rd}\ind_{A_j}(x)\left(\int_{\Rd}\left(\frac{2}{\g_{j}^2\pi}
     \right)^\frac{d}{2} \exp\left(-\frac{2\|h\|_2^2}{\g_{j}^2}\right) 
     \left|\triangle_h^s\left(f,x\right)\right|\, dh \right)^q\,d\nu\left(x\right) \,.
\end{align*}
Then H\"{o}lder's inequality and $\int_{\Rd}\exp\left(-2\g_{j}^{-2}\|h\|_2^2\right)\, dh
= \left(\frac{\g_j^2\pi}{2}\right)^{d/2}$ yield, for $q>1$,
\begin{align*}
 & \left\|\sum_{j\in J} \ind_{A_j} \cdot (K_j * f)-f\right\|_{L_q\left(\nu\right)
   }^{q} \\
 & \leq \sum_{j\in J}\int_{\Rd}\ind_{A_j}(x)\left(\left(\int_{\Rd}\left(\frac{2}{
        \g_{j}^2\pi}\right)^\frac{d}{2}\exp\left(-\frac{2\|h\|_2^2}{\g_{j}^2}\right)dh
        \right)^{\frac{q-1}{q}} \right.\\
 & \quad \left.\left(\int_{\Rd}\left(\frac{2}{\g_{j}^2\pi}
        \right)^\frac{d}{2} \exp\left(-\frac{2\|h\|_2^2}{\g_{j}^2}\right)
        \left|\triangle_h^s\left(f,x\right)\right|^qdh\right)^{\frac{1}{q}\,} \right)^{q}
        d\nu\left(x\right) \\
 & = \sum_{j\in J}\int_{\Rd}\ind_{A_j}(x)\int_{\Rd}\left(\frac{2}{\g_{j}^2\pi}
     \right)^\frac{d}{2} \exp\left(-\frac{2\|h\|_2^2}{\g_{j}^2}\right) 
     \left|\triangle_h^s\left(f,x\right)\right|^q \, dh \, d\nu\left(x\right) \\
 & = \sum_{j\in J}\int_{\Rd}\left(\frac{2}{\g_{j}^2\pi}\right)^\frac{d}{2} 
     \exp\left(-\frac{2\|h\|_2^2}{\g_{j}^2}\right) \int_{\Rd}\ind_{A_j}(x)
     \left|\triangle_h^s\left(f,x\right)\right|^q \, d\nu\left(x\right) \, dh \\
 & \leq \int_{\Rd}\left(\frac{2}{\pi\g_{\min}^2}\right)^\frac{d}{2} \exp\left(
        -\frac{2\|h\|_2^2}{\g_{\max}^2}\right) \int_{\Rd}\sum_{j\in J}
        \ind_{A_j}(x)\left|\triangle_h^s\left(f,x\right)\right|^q\, 
        d\, \nu\left(x\right)\, dh \\
 & = \int_{\Rd}\left(\frac{2}{\pi\g_{\min}^2}\right)^\frac{d}{2} 
     \exp\left(-\frac{2\|h\|_2^2}{\g_{\max}^2}\right)
     \left\|\triangle_h^s\left(f,\cdot\right)\right\|_{L_q(\nu)}^q \, dh \\
 & \leq \int_{\Rd}\left(\frac{2}{\pi\g_{\min}^2}\right)^\frac{d}{2} 
	\exp\left(-\frac{2\|h\|_2^2}{\g_{\max}^2}\right) 
	\omega_{s,L_q(\nu)}^q\left(f,\left\|h\right\|_2\right) \, dh \,. 
\end{align*}
Moreover, for $q=1$, we have
\begin{align*}
 & \left\|\sum_{j\in J}\ind_{A_j}\cdot(K_j * f)-f\right\|_{L_1\left(\nu\right)} \\
 & \leq \sum_{j\in J}\int_{\Rd}\ind_{A_j}(x)\int_{\Rd}\left(\frac{2}{\g_{j}^2\pi}
        \right)^\frac{d}{2} \exp\left(-\frac{2\|h\|_2^2}{\g_{j}^2}\right) 
        \left|\triangle_h^s\left(f,x\right)\right| \, dh \, d\nu\left(x\right) \\
 & \leq \int_{\Rd}\left(\frac{2}{\pi\g_{\min}^2}\right)^\frac{d}{2} \exp\left(
        -\frac{2\|h\|_2^2}{\g_{\max}^2}\right) \int_{\Rd}\sum_{j\in J}
        \ind_{A_j}(x)\left|\triangle_h^s\left(f,x\right)\right|\, 
        d\, \nu\left(x\right)\, dh \\
 & \leq \int_{\Rd}\left(\frac{2}{\pi\g_{\min}^2}\right)^\frac{d}{2} 
	\exp\left(-\frac{2\|h\|_2^2}{\g_{\max}^2}\right) 
	\omega_{s,L_1(\nu)}\left(f,\left\|h\right\|_2\right) \, dh \,. 
\end{align*}
Consequently, we can proceed in the same way for all $q\geq 1$. 
To this end, note that the assumption $f\in B_{q,\infty}^{\a}(\nu)$ implies 
$\omega_{s,L_q(\nu)}\left(f,t\right)\leq \|f\|_{B_{q,\infty}^{\a}(\nu)}\,t^\a$ for $t>0$. 
The latter together with H\"{o}lder's inequality yields
\begin{align*}
 & \left\|\sum_{j\in J}\ind_{A_j}\cdot(K_j * f)-f\right\|_{L_q\left(\nu\right)
   }^{q} \\
 & \leq \int_{\Rd}\left(\frac{2}{\pi\g_{\min}^2}\right)^\frac{d}{2} 
	\exp\left(-\frac{2\|h\|_2^2}{\g_{\max}^2}\right) 
	\omega_{s,L_q(\nu)}^q\left(f,\left\|h\right\|_2\right) \, dh \\
 & \leq \|f\|_{B_{q,\infty}^{\a}(\nu)}^q \left(\frac{2}{\pi\g_{\min}^2}\right)^\frac{d}{2} \int_{\Rd}
        \left\|h\right\|_2^{q\a}\exp\left(-\frac{2\|h\|_2^2}{\g_{\max}^2}
        \right)\, dh \\
 & \leq \|f\|_{B_{q,\infty}^{\a}(\nu)}^q \left(\frac{2}{\pi\g_{\min}^2}\right)^\frac{d}{2}
        \left(\int_{\Rd} \exp\left(-\frac{2\|h\|_2^2}{\g_{\max}^2}\right)\, 
        dh\right)^{\frac{1}{2}} \left(\int_{\Rd}\left\|h\right\|_2^{2q\a}\exp\left(
        -\frac{2\|h\|_2^2}{\g_{\max}^2}\right)\, dh\right)^{\frac{1}{2}} \\
 & = \|f\|_{B_{q,\infty}^{\a}(\nu)}^q \left(\frac{2\g_{\max}^2}{\pi\g_{\min}^4}\right)^\frac{d}{4}
        \left(\int_{\Rd}\left\|h\right\|_2^{2q\a}\exp\left(-\frac{2\|h\|_2^2}
        {\g_{\max}^2}\right)\, dh\right)^{\frac{1}{2}} \,.
\end{align*}
Using the embedding constant $d^\frac{q\a-1}{2q\a}$ of $\ell_{2q\a}^d$ to $\ell_{2}^d$,
we obtain
\begin{align*}
 \int_{\Rd}\left\|h\right\|_2^{2q\a}\exp\left(-\frac{2\left\|h\right\|_2^2}{\g_{\max}^2}
 \right)\, dh
 & \leq d^{q\a-1}\sum_{\ell =1}^{d}\int_{\Rd}h_\ell ^{2q\a}\prod_{l=1}^{d}
        \exp\left(-\frac{2h_l^2}{\g_{\max}^2}\right) \,d\left(h_1,\ldots,h_d\right) \\
 & = d^{q\a-1}\sum_{\ell =1}^{d}\left(\frac{\g_{\max}^2\pi}{2}\right)^{\frac{d-1}{2}}
     \int_{\R}h_\ell ^{2q\a}\exp\left(-\frac{2h_\ell ^2}{\g_{\max}^2}\right) \, dh_\ell  \\
 & = 2d^{q\a}\left(\frac{\g_{\max}^2\pi}{2}\right)^{\frac{d-1}{2}}\int_{0}^{\infty}
     t^{2q\a}\exp\left(-\frac{2t^2}{\g_{\max}^2}\right) \, dt \,.
\end{align*}
for $\g>0$. 
With the substitution $t=(\frac{1}{2}\g_{\max}^2u)^{\frac{1}{2}}$, the functional 
equation $\G(t+1)=t\,\G(t)$ of the Gamma function $\G$, and 
$\G\left(\frac{1}{2}\right)=\sqrt{\pi}$ we further have
\begin{align*}
 \int_{0}^{\infty}t^{2q\a}\exp\left(-\frac{2t^2}{\g_{\max}^2}\right) \, dt
 & = \frac{1}{2}\frac{\g_{\max}}{\sqrt{2}}\left(\frac{\g_{\max}^2}{2}\right)^{q\a} 
     \int_{0}^{\infty}u^{\left(q\a+\frac{1}{2}\right)-1}\exp\left(-u\right) \, du \\
 & = \frac{1}{2}\frac{\g_{\max}}{\sqrt{2}}\left(\frac{\g_{\max}^2}{2}\right)^{q\a} 
     \Gamma\left(q\a+\frac{1}{2}\right)\,.
\end{align*}
Altogether, we finally obtain
\begin{align*}
 & \left\|\sum_{j\in J}\ind_{A_j}\cdot(K_j * f)-f\right\|_{L_q\left(\nu\right)}^{q} \\
 & \leq \|f\|_{B_{q,\infty}^{\a}(\nu)}^q \left(\frac{2\g_{\max}^2}{\pi\g_{\min}^4}\right)^\frac{d}{4}
        \left(\int_{\Rd}\left\|h\right\|_2^{2q\a}\exp\left(-\frac{2\|h\|_2^2}
        {\g_{\max}^2}\right)\, dh\right)^{\frac{1}{2}} \\
 & \leq \|f\|_{B_{q,\infty}^{\a}(\nu)}^q \left(\frac{2\g_{\max}^2}{\pi\g_{\min}^4}\right)^\frac{d}{4}
        \left(\left(\frac{d}{2}\right)^{q\a} \left(\frac{\pi^{d-1}}{2^d}
        \right)^{\frac{1}{2}}\g_{\max}^{2q\a+d}
        \Gamma\left(q\a+\frac{1}{2}\right)\right)^{\frac{1}{2}} \\
 & = \|f\|_{B_{q,\infty}^{\a}(\nu)}^q \left(\frac{d}{2}\right)^{\frac{q\a}{2}}\pi^{-\frac{1}{4}}
       \Gamma\left(q\a+\frac{1}{2}\right)^{\frac{1}{2}}\left(\frac{\g_{\max}}
       {\g_{\min}}\right)^d \g_{\max}^{q\a} \,.
\end{align*}
\end{proof}

Based on Theorems \ref{main thm.} and \ref{estim excess risk}, we can now show 
Theorem \ref{LS oracle ineq}.

\begin{proof}[Proof of Theorem \ref{LS oracle ineq}]
First, we have to choose a function $f_0\in H$. To this end, we define functions
$K_j : \mathbb{R}^d\to\mathbb{R}$, $j\in\{1,\ldots,m\}$, by \eqref{Def. K}, 
where $s:=\lfloor\a\rfloor+1$ and $\g_j>0$. Then we define $f_0$ 
by convolving each $K_j$ with the Bayes decision function $\fB$, that is 
\begin{align*}
 f_0(x):=\sum_{j\in J_T}\ind_{A_j}(x) \cdot (K_j*f^*_{L,\P})(x)
\,, \qquad\qquad x\in\Rd\ .
\end{align*}
Now, to show that $f_0$ is indeed a suitable function to bound the approximation error,
we first need to ensure that $f_0$ is contained in $H$. In addition, we need to derive 
bounds for both, the regularization term and the excess risk of $f_0$.
To this end, we apply \citep[Theorem 2.3]{ EbSt13} and obtain, for every 
$j\in J_T$, 
\begin{align*}
 \left(K_j*\fB\right)_{|A_j} \in H_{\g_j}(A_{j})
\end{align*}
with
\begin{align*}
 \|\ind_{A_{j}}f_0\|_{\hat{H}_{\g_j}(A_{j})}
 & = \left\|\ind_{A_{j}}(K_j * \fB)\right\|_{\hat{H}_{\g_j}(A_{j})}  \\
 & = \left\|\left(K_j * \fB\right)_{|A_j}\right\|_{H_{\g_j}(A_{j})} \\
 & \leq (\g_j\sqrt{\pi})^{-\frac{d}{2}}(2^s-1)\|\fB\|_{L_2(\R^d)}\,.
\end{align*}
This implies
\begin{align*}
 f_0=\sum_{j\in J_T}\underbrace{\ind_{A_j}(K_j*\fB}_{\in \hat{H}_{\g_j}(A_{j})})
 \in H_{J_T}\,.
\end{align*} 
Besides, note that $0\in \hat{H}_{\g_j}(A_j)$ for every $j\in\{1,\ldots,m\}$ such that 
$f_0$ can be written as $f_0=\sum_{j=1}^m f_j$, where 
\begin{align*}
 f_j := \begin{cases}
         \ind_{A_j}(K_j*\fB) \,, & j\in J_T \,, \\
         0 \,,			 & j\notin J_T \,.
        \end{cases}
\end{align*}
Obviously, the latter implies $f_0\in H$. 
Furthermore, for $A_T:=\bigcup_{j\in J_T}A_{j}$, 
\eqref{eq excess risk} and Theorem \ref{estim excess risk} yield
\begin{align*}
 \RLPJT(f_0)-\RBJT
 & = \|f_0-\fB\|^2_{L_2(\TrP{A_T})} \\
 & = \|\sum_{j\in J_T}\ind_{A_{j}}(K_j * \fB)-\fB
     \|^2_{L_2(\TrP{A_T})} \\
 & \leq C_{\a,2}\, \left(\frac{\max_{j\in J_T}\g_{j}}{\min_{j\in J_T}\g_{j}}\right)^d 
        \max_{j\in J_T}\g_{j}^{2\a}\,,
\end{align*}
where $C_{\a,2}$ is a constant only depending on $\a$, $d$, and 
$\|\fB\|_{B_{2,\infty}^{\a}(\TrP{A_T})}$. To utilize Theorem \ref{main thm.}, it remains 
to examine the constants $B,\ V,\ \vt$, and $B_0$. 
Since we consider the least squares loss, which can be clipped at $M$ with $Y=[-M,M]$, 
the supremum bound \eqref{supbound} holds for $B=4M^2$ and the variance bound 
\eqref{varbound} for $V=16M^2$ and $\vt=1$ \citep[cf.][Example 7.3]{ StCh08}.
Next, we derive a bound for $\|L\circ f_0\|_\infty$ using \citep[Theorem 2.3]{ EbSt13} 
which provides, for every $x\in X$, the supremum bound
\begin{align}\label{sup f_0}
 |f_0(x)|\! =\! \left|\sum_{j\in J_T}\ind_{A_{j}}(x)\cdot(K_j * \fB)(x)\right|
 \leq\! \sum_{j\in J_T}\ind_{A_{j}}(x)\left|K_j * \fB(x)\right|
 \!\leq \!(2^s-1)\left\|\fB\right\|_{L_\infty(\R^d)}.
\end{align}
The latter implies
\begin{align*}
 \|L_{J_T}\circ f_0\|_\infty
 & = \sup_{(x,y)\in X\times Y} |L(y,f_0(x))| \\
 & \leq \sup_{(x,y)\in X\times Y} \left(M^2 + 2 M |f_0(x)| + |f_0(x)|^2\right) \\
 & \leq 4^s \max\left\{M^2, \|\fB\|_{L_\infty(\R^d)}^2\right\} \,,
\end{align*}
i.e.~$B_0:=4^s \max\{M^2, \|\fB\|_{L_\infty(\R^d)}^2\}$. 
Moreover, since Theorem \ref{entropy number H_g} provides 
\linebreak$e_i(\mathrm{id} : H_{\g_j}(A_{j}) \to L_2(\TrP{A_j})) \leq  a_j i^{-\frac{1}{2p}}$
for $i\geq 1$ with 
$a_j = \tilde{c}_p \sqrt{\P_X(A_{j})}\, r^\frac{d+2p}{2p} \g_j^{-\frac{d+2p}{2p}}$, we 
have 
\begin{align*}
 & \Biggl(\max\Biggl\{ c_p \sqrt{m} \Biggl(\sum_{j=1}^m\lb_{j}^{-p} a_j^{2p}\Biggr)^{\frac{1}{2p}},B\Biggr\}\Biggr)^{2p} \\
 & = \Biggl(\max\Biggl\{ c_p \tilde{c}_p \sqrt{m} r^\frac{d+2p}{2p} \Biggl(\sum_{j=1}^m \left(
     \lb_{j}^{-1} \g_j^{-\frac{d+2p}{p}} \P_X(A_{j})\right)^{p} \Biggr)^{\frac{1}{2p}},B\Biggr\} \Biggr)^{2p} \\
 & \leq \Biggl(\max\Biggl\{ c_p \tilde{c}_p m^\frac{1}{2p} r^\frac{d+2p}{2p} \Biggl(\sum_{j=1}^m 
     \lb_{j}^{-1} \g_j^{-\frac{d+2p}{p}} \P_X(A_{j})\Biggr)^{\frac{1}{2}},B\Biggr\} \Biggr)^{2p} \\
 & \leq \Biggl(\max\Biggl\{ c_p \tilde{c}_p 8^\frac{d}{2p} r \Biggl(\sum_{j=1}^m 
     \lb_{j}^{-1} \g_j^{-\frac{d+2p}{p}} \P_X(A_{j})\Biggr)^{\frac{1}{2}},B\Biggr\}\Biggr)^{2p}  \\
 & \leq C_p r^{2p} \Biggl(\sum_{j=1}^m 
     \lb_{j}^{-1} \g_j^{-\frac{d+2p}{p}} \P_X(A_{j})\Biggr)^p + B^{2p}  \\
 & =: a^{2p}\,,
\end{align*}
where we used the concavity of the function $t\mapsto t^p$ for $t\geq 0$, 
$m r^d\leq 8^d$ by \eqref{ex. Ueberdeckung}, and $C_p := c_p^{2p} \tilde{c}_p^{2p} 8^d$.
Finally, applying Theorem \ref{main thm.} yields
\begin{align*}
  & \RLPJT(\fcl_{\D,\bs\lb,\bs\g})-\RBJT \\
  & \leq \sum_{j=1}^m\lb_j\|f_{\D_{j},\lb_j,\g_j}\|^2_{\hat{H}_{\g_j}(A_{j})}
         + \RLPJT(\fcl_{\D,\bs\lb,\bs\g})-\RBJT \\ 
  & \leq 9 \left(\sum_{j=1}^m\lb_j\|\ind_{A_j}f_0\|^2_{\hat{H}_{\g_j}(A_{j})}+\RLPJT(f_0)-\RBJT
          \right) \\
  & \quad + C \left(a^{2p}n^{-1}\right)^{\frac{1}{2-p-\vt+\vt p}} 
          + 3\left(\frac{72 V\t}{n}\right)^{\frac{1}{2-\vt}} + \frac{15B_0\t}{n}  \\
  & \leq  9\left(\sum_{j\in J_T}\lb_j(\g_j\sqrt{\pi})^{-d}(2^s-1)^2
          \|\fB\|^2_{L_2(\R^d)} + C_{\a,2}\,\left(\frac{\max_{j\in J_T}\g_{j}}{
          \min_{j\in J_T}\g_{j}}\right)^d \max_{j\in J_T}\g_{j}^{2\a} \right) \\
  & \quad + C  C_p r^{2p} \left( \sum_{j=1}^m \lb_{j}^{-1} \g_j^{-\frac{d+2p}{p}} 
          \P_X(A_{j}) \right)^p n^{-1} + CB^{2p}n^{-1} + \frac{3456 M^2\t}{n} \\
  & \quad + 15\cdot 4^s \max\{M^2, \|\fB\|_{L_\infty(\R^d)}^2\}\frac{\t}{n} \\
  & \leq  9 (2^s-1)^2 \pi^{-\frac{d}{2}} \|\fB\|^2_{L_2(\R^d)} \sum_{j\in J_T} 
          \lb_j \g_j^{-d} + 9 C_{\a,2}\,\left(\frac{\max_{j\in J_T}\g_{j}}{
          \min_{j\in J_T}\g_{j}}\right)^d \max_{j\in J_T}\g_{j}^{2\a} \\
  & \quad + C   C_p r^{2p}\left( \sum_{j=1}^m \lb_{j}^{-1} \g_j^{-\frac{d+2p}{p}} 
          \P_X(A_{j}) \right)^p n^{-1} + 16^p C M^{4p} n^{-1} \\
  & \quad + \left(3456 M^2 + 15\cdot 4^s \max\{M^2, \|\fB\|_{L_\infty(\R^d)}^2\}\right) 
          \frac{\t}{n} 
\end{align*}
with probability $\P^n$ not less than $1-3e^{-\t}$. Finally, for $\hat{\t}\geq 1$, a 
variable transformation implies
\begin{align*}
 & \sum_{j=1}^m\lb_j\|f_{\D_{j},\lb_j,\g_j}\|^2_{\hat{H}_{\g_j}(A_{j})}
         + \RLPJT(\fcl_{\D,\bs\lb,\bs\g})-\RBJT \\
 & \leq C_{M,\a,p}\!\left( \sum_{j\in J_T}\!\! \lb_j \g_j^{-d} \!
        + \! \left(\frac{\max_{j\in J_T}\g_{j}}{\min_{j\in J_T}\g_{j}}\right)^d 
        \!\max_{j\in J_T}\g_{j}^{2\a} \!+\! r^{2p}\! \left( \sum_{j=1}^m\! \lb_{j}^{-1} 
        \g_j^{-\frac{d+2p}{p}}\P_X(A_{j})\! \right)^p\!\! n^{-1} \!+\! \hat{\t} n^{-1}\! \right)
\end{align*}
with probability $\P^n$ not less than $1-e^{-\hat{\t}}$, where the constant 
$C_{M,\a,p}$ is defined by
\begin{align*}
 C_{M,\a,p} := \max\Biggl\{\Biggr. 9 (2^s-1)^2 \pi^{-\frac{d}{2}} \|\fB\|^2_{L_2(\R^d)} \,,\,
                 9 \|\fB\|_{B_{2,\infty}^{\a}(\TrP{A_T})}^2 \left(\frac{d}{2}\right)^\a
		\pi^{-\frac{1}{4}} \Gamma\left(2\a+\frac{1}{2}\right)^{\frac{1}{2}}\,,\\
		 \ \,\,  8^d C  c_p^{2p} \tilde{c}_p^{2p} \,,\,
		16^p C M^{4p}+\left(3456 M^2 + 15\cdot 4^s \max\{M^2, 
		\|\fB\|_{L_\infty(\R^d)}^2\}\right)(1+\ln(3)) \Biggl.\Biggr\}\,.
\end{align*}
\end{proof}

Next, using the just proven oracle inequality presented in Theorem \ref{LS oracle ineq}, 
we show the learning rates of Theorem \ref{Thm LS rate} in only a few steps.

\begin{proof}[Proof of Theorem \ref{Thm LS rate}]
First of all, we define sequences $\tilde{\lb}_{n} := c_2 n^{-1}$ and 
$\tilde{\g}_{n}:= c_3 n^{-\frac{1}{2\a+d}}$ to simplify the presentation. Then Theorem 
\ref{LS oracle ineq}, $\sum_{j=1}^{m_n} \P_X(A_{j})=1$, and
$|J_T|\leq m_n \leq 8^d r_n^{-d}$ 
together with $\lb_{n,j}=r_n^d \tilde{\lb}_{n}$ and 
$\g_{n,j}=\tilde{\g}_{n}$ for all $j\in\{1,\ldots,m_n\}$ yield
\begin{align*}
 & \RLPJT(\fcl_{\D,\bs\lb_n,\bs\g_n})-\RBJT \\
 & \leq C_{M,\a,p}\!\left( \sum_{j\in J_T}\!\! \lb_{n,j} \g_{n,j}^{-d}\! +\! 
        \left(\frac{\max_{j\in J_T}\g_{n,j}}{\min_{j\in J_T} \g_{n,j}}\right)^{\!d} 
        \!\!\max_{j\in J_T}\g_{n,j}^{2\a}
        \!+\! r_n^{2p} \!\left( \sum_{j=1}^{m_n}\!
        \lb_{n,j}^{-1} \g_{n,j}^{-\frac{d+2p}{p}} \P_X(A_{j})\! \right)^{\!p}\!\! n^{-1} 
        \!+\! \frac{\t}{n} \right) \\
 &    = C_{M,\a,p}\left( |J_T| r_n^d \tilde{\lb}_{n} \tilde{\g}_{n}^{-d} + \tilde{\g}_{n}^{2\a}
        + r_n^{(2-d)p} \tilde{\lb}_{n}^{-p} \tilde{\g}_{n}^{-(d+2p)}
        \left( \sum_{j=1}^{m_n} \P_X(A_{j}) \right)^p n^{-1} + \t n^{-1} \right) \\
 & \leq 8^{d} C_{M,\a,p}\left( \tilde{\lb}_{n} \tilde{\g}_{n}^{-d} + \tilde{\g}_{n}^{2\a}
        + \tilde{\lb}_{n}^{-p} \tilde{\g}_{n}^{-(d+2p)} r_n^{(2-d)p} n^{-1} + \t n^{-1} \right) \,.
\end{align*}
Using the choices $\tilde{\lb}_{n} = c_2 n^{-1}$, $\tilde{\g}_{n} = c_3 n^{-\frac{1}{2\a+d}}$, as well 
as $r_n = c_1 n^{-\frac{1}{\b d}}$ finally implies
\begin{align*}
 & \RLPJT(\fcl_{\D,\bs\lb_n,\bs\g_n})-\RBJT \\
 & \leq 8^{d} C_{M,\a,p}\left( \tilde{\lb}_{n} \tilde{\g}_{n}^{-d} + \tilde{\g}_{n}^{2\a}
        + \tilde{\lb}_{n}^{-p} \tilde{\g}_{n}^{-(d+2p)} r_n^{(2-d)p} n^{-1} + \t n^{-1} \right) \\
 & \leq \hat{C}_{M,\a,p}\left( n^{-1}n^\frac{d}{2\a+d} + n^{-\frac{2\a}{2\a+d}} 
        + n^{p}n^\frac{d+2p}{2\a+d} n^{-\frac{(2-d)p}{\b d}} n^{-1} + \t n^{-1} \right) \\
 & = \hat{C}_{M,\a,p}\left( n^{-\frac{2\a}{2\a+d}} + n^{-\frac{2\a}{2\a+d}} 
       + n^{-\frac{2\a}{2\a+d}+\left(1+\frac{2}{2\a+d}+\frac{1}{\b}-\frac{2}{\b d}\right)p}
       + \t n^{-1} \right) \\
 & \leq C \t n^{-\frac{2\a}{2\a+d}+\x} 
\end{align*}
with probability $\P^n$ not less than $1-e^{-\t}$, where $C>0$ is a constant and 
\linebreak$\x\geq\left(1+\frac{2}{2\a+d}+\frac{1}{\b}-\frac{2}{\b d} \right)p>0$.
\end{proof}

\begin{proof}[Proof of Corollary \ref{rate for smoothness von T^{+d}}]
 For simplicity of notation, we write $\bs\lb$, $\lb_{j}$, $\bs\g$, and $\g_{j}$
instead of $\bs\lb_n$, $\lb_{n,j}$, $\bs\g_n$, and $\g_{n,j}$.
 Since $\bigcup_{j\in J_T}A_j\subset T^{+\d}$ for all $n\geq n_\d$, the assumption
$\fB\in B_{2,\infty}^{\a}(\TrP{T^{+\d}})$ implies
\begin{align*}
 \fB\in B_{2,\infty}^{\a}(\TrP{\bigcup_{j\in J_T}A_{j}})\,.
\end{align*}
With this, Theorems \ref{LS oracle ineq} and \ref{Thm LS rate} 
immediately yield 
\begin{align*}
  & \RLPT(\fcl_{\D,\bs\lb,\bs\g})-\RBT \\
  & \leq \sum_{j=1}^m\lb_j\|f_{\D_{j},\lb_j,\g_j}\|^2_{\hat{H}_{\g_j}(A_{j})}
         + \RLPT(\fcl_{\D,\bs\lb,\bs\g})-\RBT \\ 
  & \leq \sum_{j=1}^m\lb_j\|f_{\D_{j},\lb_j,\g_j}\|^2_{\hat{H}_{\g_j}(A_{j})}
         + \RLPJT(\fcl_{\D,\bs\lb,\bs\g})-\RBJT \\ 
  & \leq C_{M,\a,p}\left( \sum_{j\in J_T}\!\! \lb_j \g_j^{-d} 
        +\! \left(\frac{\max_{j\in J_T}\g_{j}}{\min_{j\in J_T}\g_{j}}\right)^d 
        \!\max_{j\in J_T}\g_{j}^{2\a} + r^{2p}\! \left( \sum_{j=1}^m\! \lb_{j}^{-1} 
        \g_j^{-\frac{d+2p}{p}}\P_X(A_{j})\! \right)^p\! n^{-1} + \frac{\t}{n} \right) \\
  & \leq C \t n^{-\frac{2\a}{2\a+d}+\x} 
\end{align*}
with probability $\P^n$ not less than $1-e^{-\t}$, where 
$\x\geq\left(1+\frac{2}{2\a+d}+\frac{1}{\b}-\frac{2}{\b d} \right)p>0$. 
Moreover, the constants $C_{M,\a,p}>0$ and $C>0$ coincide with those of Theorems 
\ref{LS oracle ineq} and \ref{Thm LS rate}.
\end{proof}

It remains to prove Theorem \ref{LS oracle ineq data-dep param}. However, we previously 
have to consider the following technical lemma.

\begin{lemma}\label{Lem data-dep param}
Let $d\geq 1$ and $r_n:=cn^{-\frac{1}{\b d}}$ with $\b>1$ and a constant $c>0$. 
We fix finite 
subsets $\Lb_n\subset(0,r_n^d]$ and $\G_n\subset(0,r_n]$ such that $\Lb_n$ is an 
$(r_n^d\e_n)$-net of $(0,r_n^d]$ and $\G_n$ is an $\d_n$-net of $(0,r_n]$ with 
$0<\e_n\leq \,n^{-1}$, $\d_n>0$, $r_n^d\in\Lb_n$, and $r_n\in\G_n$. 
Moreover, let $J\subset\{1,\ldots,m_n\}$ be an arbitrary non-empty index set and 
$|J|\leq m_n\leq 8^d r_n^{-d}$. 
Then, for all $0<\a<\frac{\b-1}{2}d$, 
$n\geq1$, and all $p\in(0,1)$ with 
$p\leq \frac{\b d-2\a-d}{2\a+d+2}$, we have
\begin{align*}
 & \inf_{(\lb_j,\g_j)_{j=1}^{m_n}\in(\Lb_n\times\G_n)^{m_n}}
 \!\!\left( \sum_{j \in J}\! \lb_j \g_j^{-d}\! +\! \left(\frac{\max_{j\in J}\g_{j}}{
 \min_{j\in J}\g_{j}}\right)^{\!d}\! \!\max_{j\in J}\g_{j}^{2\a} \!+\! r_n^{2p}\! \left( 
 \sum_{j=1}^{m_n}\! \lb_{j}^{-1} \g_j^{-\frac{d+2p}{p}} \P_X(A_{j})\! \right)^{\!p}\!\! n^{-1}\!
 \right) \\
 & \leq C\left(n^{-\frac{2\a}{2\a+d}+\xi}+\d_n^{2\a}\right)\, ,
\end{align*}
where $\xi:=\bigl(\frac{2\a(2\a+d+2)}{(2\a+d)\left((2\a+d)(1+p)+2p\right)}
+\max\bigl\{\frac{d-2}{\b d},0\bigr\}\bigr)p$ and $C>0$ is a constant independent of $n$, $\Lb_n$, $\e_n$, 
$\G_n$, and $\d_n$.
\end{lemma}

\begin{proof}
 Without loss of generality, we may assume that $\Lb_n$ and $\G_n$ are of the form 
$\Lb_n=\left\{\lb^{(1)},\ldots,\lb^{(u)}\right\}$ and 
$\G_n=\left\{\g^{(1)},\ldots,\g^{(v)}\right\}$ with $\lb^{(u)}=r_n^d$ and $\g^{(v)}=r_n$ 
as well as $\lb^{(i-1)}< \lb^{(i)}$ and 
$\g^{(\ell-1)}<\g^{(\ell)}$ for all $i=2,\ldots,u$ and $\ell=2,\ldots,v$.
With $\lb^{(0)}:=0$ and $\g^{(0)}:=0$  it is easy to see that
\begin{align}
 \lambda^{(i)}-\lambda^{(i-1)}\leq 2 r_n^d \e_n  \qquad\text{and}\qquad 
 \gamma^{(\ell)}-\gamma^{(\ell-1)}\leq 2\delta_n  \label{AbstaendeNetze}
\end{align}
hold for all $i=1,\ldots,u$ and $\ell=1,\ldots,v$. Furthermore, 
define 
$\lb^*:= n^{-\frac{2\a+d}{(2\a+d)(1+p)+2p}}$ and 
$\g^*:= c n^{-\frac{1}{(2\a+d)(1+p)+2p}}$. Then there exist indices 
$i\in\left\{1,\ldots,u\right\}$ and $\ell\in\left\{1,\ldots,v\right\}$ with 
$\lb^{(i-1)}\leq r_n^d\lb^*\leq\lb^{(i)}$ and 
$\g^{(\ell-1)}\leq \g^*\leq\g^{(\ell)}$. Together with (\ref{AbstaendeNetze}), this yields 
\begin{align}
 r_n^d \lb^*\leq\lb^{(i)}\leq r_n^d \lambda^*+2r_n^d\e_n  \qquad\text{and}\qquad 
 \g^*\leq\g^{(\ell)}\leq\g^*+2\delta_n \,. \label{(lb^*,g^*)}
\end{align}
Moreover, the definition of $\lb^*$ implies $\e_n\leq \lb^*$ and the 
one of $\g^*$ implies $\g^*\leq r_n$ for $\a<\frac{\b-1}{2}d$ and $p\in(0,p^*]$, where 
$p^*:= \frac{\b d-2\a-d}{2\a+d+2}$. Additionally, it is easy 
to check that
\begin{align}\label{estim. for (lb^*,g^*)}
 \lb^*\left(\g^*\right)^{-d}+\left(\g^*\right)^{2\a}+ \left(\lb^*\right)^{-p}
 \left(\g^*\right)^{-(d+2p)}r_n^{(2-d)p}n^{-1}
 \leq \hat{c} n^{-\frac{2\a}{(2\a+d)(1+p)+2p}+\max\left\{\frac{d-2}{\b d},0\right\}p}\,,
\end{align}
where $\hat{c}$ is a positive constant. Using \eqref{(lb^*,g^*)}, the bound
$|J|\leq m_n \leq 8^d r_n^{-d}$, 
and \eqref{estim. for (lb^*,g^*)}, we obtain
\begin{align*}
 &\inf_{(\lb_j,\g_j)_{j=1}^{m_n}\in(\Lb_n\times\G_n)^{m_n}}
 \!\!\left( \sum_{j \in J} \!\lb_j \g_j^{-d} \!+\! \left(\frac{\max_{j\in J}\g_{j}}{
 \min_{j\in J}\g_{j}}\right)^{\!d}\!\! \max_{j\in J}\g_{j}^{2\a} \!+\! r_n^{2p} \!\left( 
 \sum_{j=1}^{m_n} \!\lb_{j}^{-1} \g_j^{-\frac{d+2p}{p}} \P_X(A_{j}) \!\right)^{\!p}\!\! n^{-1}\!
 \right) \\
 & \leq \sum_{j \in J} \lb^{(i)} \left(\g^{(\ell)}\right)^{-d} 
        + \left(\g^{(\ell)}\right)^{2\a} + \left( \sum_{j=1}^{m_n} 
        \left(\lb^{(i)}\right)^{-1} \left(\g^{(\ell)}\right)^{-\frac{d+2p}{p}} \P_X(A_{j}) 
        \right)^p r_n^{2p} n^{-1}  \\
 & \leq |J| \lb^{(i)} \left(\g^{(\ell)}\right)^{-d} 
        + \left(\g^{(\ell)}\right)^{2\a} + \left(\lb^{(i)}\right)^{-p}
        \left(\g^{(\ell)}\right)^{-(d+2p)} r_n^{2p} n^{-1}  \\
 & \leq |J| \left(r_n^d\lb^*+2r_n^d\e_n\right)\left(\g^*\right)^{-d}
        +\left(\g^*+2\d_n\right)^{2\a} 
        + \left(r_n^d\lb^*\right)^{-p}\left(\g^*\right)^{-(d+2p)} r_n^{2p} n^{-1} \\
 & \leq 8^d\cdot 3\lb^*\left(\g^*\right)^{-d}
        +\left(\g^*+2\d_n\right)^{2\a} 
        + \left(\lb^*\right)^{-p}\left(\g^*\right)^{-(d+2p)}r_n^{(2-d)p}n^{-1}\\
 & \leq \tilde{c}\left(\lb^*\left(\g^*\right)^{-d}+\left(\g^*\right)^{2\a}+ \left(\lb^*
        \right)^{-p}\left(\g^*\right)^{-(d+2p)}r_n^{(2-d)p}n^{-1} \right) 
        + \tilde{c}\d_n^{2\a} \\
 & \leq \tilde{c}\hat{c} n^{-\frac{2\a}{(2\a+d)(1+p)+2p}+\max\left\{\frac{d-2}{\b d},0\right\}p} + \tilde{c}\d_n^{2\a}\\
 & \leq C\left(n^{-\frac{2\a}{2\a+d}+\xi}+\d_n^{2\a}\right)
\end{align*}
with $\xi:=\bigl(\frac{2\a(2\a+d+2)}{(2\a+d)\left((2\a+d)(1+p)+2p\right)}
+\max\bigl\{\frac{d-2}{\b d},0\bigr\}\bigr)p$ and constants $\tilde{c}>0$ and $C>0$ 
independent of $n$, $\Lb_n$, $\e_n $, $\G_n$, and $\d_n$.
\end{proof}

In the end, we show Theorem \ref{LS oracle ineq data-dep param} using Theorem 
\ref{LS oracle ineq} as well as Lemma \ref{Lem data-dep param}.

\begin{proof}[Proof of Theorem \ref{LS oracle ineq data-dep param}]
Let $l$ be defined by $l:=\left\lfloor \frac{n}{2} \right\rfloor +1$, 
i.e.~$l\geq \frac{n}{2}$. 
With this, Theorem \ref{LS oracle ineq} yields with probability $\P^l$ 
not less than $1-\left|\Lb_n\times\G_n\right|^{m_n}e^{-\t}$ that
\begin{align}
& \RLPJT(\fcl_{\D_1,\bs\lb,\bs\g}) - \RBJT \nonumber \\
& \leq \frac{c_1}{2}\! \left( \sum_{j \in J_T}\! \lb_j \g_j^{-d} 
       + \left(\frac{\max_{j\in J_T}\g_{j}}{\min_{j\in J_T}\g_{j}}\right)^d 
       \max_{j\in J_T}\g_{j}^{2\a} + r_n^{2p} \left( \sum_{j=1}^{m_n} \!
       \lb_{j}^{-1} \g_j^{-\frac{d+2p}{p}} \P_X(A_{j}) \right)^p l^{-1} + \t l^{-1}
       \right) \nonumber\\
& \leq c_1 \!\left( \sum_{j \in J_T} \!\lb_j \g_j^{-d} 
       + \left(\frac{\max_{j\in J_T}\g_{j}}{\min_{j\in J_T}\g_{j}}\right)^d 
       \!\max_{j\in J_T}\g_{j}^{2\a} + r_n^{2p} \left( \sum_{j=1}^{m_n} \!
       \lb_{j}^{-1} \g_j^{-\frac{d+2p}{p}} \P_X(A_{j}) \right)^p\! n^{-1}+ \t n^{-1}\right) 
  \label{Absch.2}
\end{align}
for all $(\lb_j,\g_j)\in\Lb_n\times\G_n$, $j\in\{1,\ldots,m_n\}$, simultaneously, where 
$c_1>0$ is a constant independent of $n$, $\t$, $\bs\lb$, and $\bs\g$. Furthermore, the 
oracle inequality of \citep[Theorem 7.2]{ StCh08} for empirical risk minimization, 
$n-l\geq \frac{n}{2}-1 \geq \frac{n}{4}$, and $\t_n:=\t+\ln(1+|\Lb_n\times\G_n|^{m_n})$ yield 
\begin{align}
 & \RLPJT(\fcl_{\D_1,\bs\lb_{\D_2},\bs\g_{\D_2}}) - \RBJT \\
 & < 6\left(\inf_{(\lb_j,\g_j)_{j=1}^{m_n}\in(\Lb_n\times\G_n)^{m_n}}
     \RLPJT(\fcl_{\D_1,\bs\lb,\bs\g})-\RBJT\right) + 512 M^2\frac{\t_n}{n-l} \nonumber \\ 
 & < 6\left(\inf_{(\lb_j,\g_j)_{j=1}^{m_n}\in(\Lb_n\times\G_n)^{m_n}}
     \RLPJT(\fcl_{\D_1,\bs\lb,\bs\g}) - \RBJT\right) + 2048 M^2\frac{\t_n}{n} \label{Absch.1}
\end{align}
with probability $\mathrm{P}^{n-l}$ not less than $1-e^{-\t}$. With \eqref{Absch.2}, 
\eqref{Absch.1} and Lemma \ref{Lem data-dep param} we can conclude
\begin{align*}
 & \RLPJT(\fcl_{\D_1,\bs\lb_{\D_2},\bs\g_{\D_2}}) - \RBJT  \\
 & < 6\left(\inf_{(\lb_j,\g_j)_{j=1}^{m_n}\in(\Lb_n\times\G_n)^{m_n}}
     \RLPJT(\fcl_{\D_1,\bs\lb,\bs\g}) - \RBJT\right) + 2048 M^2\frac{\t_n}{n}\\
 & \leq 6 c_1 \left( \inf_{(\lb_j,\g_j)_{j=1}^{m_n}\in(\Lb_n\times\G_n)^{m_n}}
        \left( \sum_{j \in J_T} \lb_j \g_j^{-d} + \left(\frac{\max_{j\in J_T}\g_{j}}{
        \min_{j\in J_T}\g_{j}}\right)^d \max_{j\in J_T}\g_{j}^{2\a} \right.\right. \\
 & \phantom{6 c_1\qquad\ }
        \left.\left. + r_n^{2p} \left(  \sum_{j=1}^{m_n} \lb_{j}^{-1} 
        \g_j^{-\frac{d+2p}{p}} \P_X(A_{j}) \right)^p n^{-1} \right)  + \t n^{-1}\right)
        + 2048M^2\frac{\t_n}{n}\\
 & \leq 6 c_1 \left( C\left(n^{-\frac{2\a}{2\a+d}+\x}+\d_n^{2\a}\right)
        + \t n^{-1}\right) + 2048M^2\frac{\t_n}{n} \\
 & \leq 12 c_1 C n^{-\frac{2\a}{2\a+d}+\x} + \left(6c_1\t + 2048 M^2 \t_n\right)n^{-1}
\end{align*}
with probability $\P^n$ not less than $1-\left(1+|\Lb_n\times\G_n|^{m_n}\right)e^{-\t}$. 
Finally, a variable transformation yields
\begin{align*}
 & \RLPJT(\fcl_{\D_1,\bs\lb_{\D_2},\bs\g_{\D_2}}) - \RBJT  \\
 & < 12 c_1 C n^{-\frac{2\a}{2\a+d}+\x} 
     + \bigl(\bigr.6c_1 \left(\t+\ln\left(1+|\Lb_n\times\G_n|^{m_n}\right)\right) \\
 &\quad + 2048 M^2 \left(\t+2\ln\left(1+|\Lb_n\times\G_n|^{m_n}\right)\right)\bigl.\bigr)n^{-1} \\
 & \leq 12 c_1 C n^{-\frac{2\a}{2\a+d}+\x} + (6c_1+ 2048 M^2 ) 
        \left(\t+2 m_n \ln\left(1+|\Lb_n\times\G_n|\right)\right) n^{-1} \\
 & \leq 12 c_1 C n^{-\frac{2\a}{2\a+d}+\x} + (6c_1+ 2048 M^2 ) 
        \left(\t+2 \cdot 8^d r_n^{-d} \ln\left(1+|\Lb_n\times\G_n|\right)\right)n^{-1}\\
 & = 12 c_1 C n^{-\frac{2\a}{2\a+d}+\x} + (6c_1+ 2048 M^2 ) 
        \left(\t n^{-1} + 2 \cdot 8^d c^{-d} \ln\left(1+|\Lb_n\times\G_n|\right)
        n^{-\frac{\b-1}{\b}}\right)\\
& < \left(12 c_1 C  + (6c_1+ 2048 M^2 )\left(\t + 2 \cdot 8^d c^{-d}
       \ln\left(1+|\Lb_n\times\G_n|\right)\right)\right) n^{-\frac{2\a}{2\a+d}+\x}
\end{align*}
with probability $\P^n$ not less than $1-e^{-\t}$, where we used 
\begin{align*}
 \a < \frac{\b-1}{2}d \quad \Longleftrightarrow \quad 
 n^{-\frac{\b-1}{\b}} < n^{-\frac{2\a}{2\a+d}} 
\end{align*}
in the last step.

\end{proof}


\bibliography{Paper}
\newpage

\section*{Appendix A}

For the sake of completeness, we present in the following some tables containing the 
computational results achieved by the LS-, VP-, and RC-SVMs for all real and artificial data 
set types. 
Here, the training and test times, given in seconds, are averaged over all successful runs. 
Moreover, for the test and $L_2$-errors, we also stated the mean of all runs plus/minus the 
standard deviation. The same is true for the number of working sets (\# of ws), except for 
the LS-SVMs, where we always have one working set by construction. The last two columns
contain median, minimum, and maximum of the working set sizes appearing during the 
various runs.

\bigskip\bigskip

\begin{table}[h]
  \centering
      \resizebox{0.87\linewidth}{!}{
\begin{tabular}{cc||c|c||S[tabnumalign=centre]S[tabnumalign=centre]rlrlS[tabnumalign=centre]rcl}
	&& data set sizes & runs & \multicolumn{1}{c}{train time} & \multicolumn{1}{c}{test time} & \multicolumn{2}{c}{test error} & \multicolumn{2}{c}{\# of ws} & \multicolumn{1}{c}{ws size: median} & \multicolumn{3}{c}{ws size: range}\\ 
	&&&&&&&&&&&&&\\[-0.35cm]\hline\hline&&&&&&&&&&&&&\\[-0.25cm]  
	\multirow{7}{*}{\begin{turn}{90} LS-SVM \end{turn}}
	&  & \phantom{00}1\,000 \footnotesize{(10\,000)} & 100 & 1.38 & 0.48 & 0.7142 & \hspace{-0.3cm}\footnotesize{\textpm 0.0097} & 1.00 &  & \\ 
 && \phantom{00}2\,500 \footnotesize{(10\,000)} & 100 & 6.42 & 0.99 & 0.6323 & \hspace{-0.3cm}\footnotesize{\textpm 0.0074} & 1.00 &  & \\ 
 && \phantom{00}5\,000 \footnotesize{(10\,000)} & 100 & 30.19 & 1.55 & 0.5707 & \hspace{-0.3cm}\footnotesize{\textpm 0.0070} & 1.00 &  & \\ 
 && \phantom{0}10\,000 \footnotesize{(50\,000)} & 10 & 138.01 & 12.14 & 0.4909 & \hspace{-0.3cm}\footnotesize{\textpm 0.0089} & 1.00 &  & \\ 
 && \phantom{0}25\,000 \footnotesize{(50\,000)} & 10 & 922.90 & 34.19 & 0.3816 & \hspace{-0.3cm}\footnotesize{\textpm 0.0042} & 1.00 &  & \\
\end{tabular}
    }
    \caption{LS- and VP-SVM results relating to the \textsc{covtype} data sets}
    \label{tab::experiments::covtype_ls_vp}
\end{table}

\begin{sidewaystable}[t!]
  \centering
  \begin{minipage}[t]{0.49\linewidth} \vspace{0pt}
    \resizebox{\textwidth}{!}{
      \begin{tabular}{cc||c|c||S[tabnumalign=centre]S[tabnumalign=centre]rlrlS[tabnumalign=centre]rcl}
	&& data set sizes & runs & \multicolumn{1}{c}{train time} & \multicolumn{1}{c}{test time} & \multicolumn{2}{c}{test error} & \multicolumn{2}{c}{\# of ws} & \multicolumn{1}{c}{ws size: median} & \multicolumn{3}{c}{ws size: range}\\ 
	&&&&&&&&&&&&&\\[-0.35cm]\hline\hline&&&&&&&&&&&&&\\[-0.25cm] 
	\multirow{9}{*}{ \begin{turn}{90} RC-SVM  \end{turn}} &
	\multirow{9}{*}{ \hspace{-0.3cm} \begin{turn}{90} \# of ws = 1  \end{turn}}
	  & \phantom{00}1\,000 \footnotesize{(10\,000)} & 100 & 1.76 & 0.30 & 0.7159 & \hspace{-0.3cm}\footnotesize{\textpm 0.0101} & 1.00 & \hspace{-0.3cm}\footnotesize{\textpm 0.00} & 1000 & \multicolumn{3}{c}{$\{ 1000 \}$} \\ 
 && \phantom{00}2\,500 \footnotesize{(10\,000)} & 100 & 22.55 & 2.19 & 0.6324 & \hspace{-0.3cm}\footnotesize{\textpm 0.0073} & 1.00 & \hspace{-0.3cm}\footnotesize{\textpm 0.00} & 2500 & \multicolumn{3}{c}{$\{ 2500 \}$} \\ 
 && \phantom{00}5\,000 \footnotesize{(10\,000)} & 100 & 97.30 & 9.16 & 0.5707 & \hspace{-0.3cm}\footnotesize{\textpm 0.0070} & 1.00 & \hspace{-0.3cm}\footnotesize{\textpm 0.00} & 5000 & \multicolumn{3}{c}{$\{ 5000 \}$} \\ 
 && \phantom{0}10\,000 \footnotesize{(50\,000)} & 10 & 741.46 & 111.75 & 0.4909 & \hspace{-0.3cm}\footnotesize{\textpm 0.0089} & 1.00 & \hspace{-0.3cm}\footnotesize{\textpm 0.00} & 10000 & \multicolumn{3}{c}{$\{ 10000 \}$} \\ 
 && \phantom{0}25\,000 \footnotesize{(50\,000)} & 10 & 4629 & 291.75 & 0.3816 & \hspace{-0.3cm}\footnotesize{\textpm 0.0042} & 1.00 & \hspace{-0.3cm}\footnotesize{\textpm 0.00} & 25000 & \multicolumn{3}{c}{$\{ 25000 \}$} \\
\end{tabular}
    }
  \end{minipage}\hfill
  \begin{minipage}[t]{0.49\linewidth} \vspace{0pt}
    \resizebox{\textwidth}{!}{
      \begin{tabular}{cc||c|c||S[tabnumalign=centre]S[tabnumalign=centre]rlrlS[tabnumalign=centre]rcl}
	&& data set sizes & runs & \multicolumn{1}{c}{train time} & \multicolumn{1}{c}{test time} & \multicolumn{2}{c}{test error} & \multicolumn{2}{c}{\# of ws} & \multicolumn{1}{c}{ws size: median} & \multicolumn{3}{c}{ws size: range}\\ 
	&&&&&&&&&&&&&\\[-0.35cm]\hline\hline&&&&&&&&&&&&&\\[-0.25cm] 
	\multirow{9}{*}{ \begin{turn}{90} RC-SVM  \end{turn}} &
	\multirow{9}{*}{ \hspace{-0.3cm} \begin{turn}{90} \# of ws = 10  \end{turn}}
	  & \phantom{00}1\,000 \footnotesize{(10\,000)} & 100 & 0.99 & 0.32 & 0.8134 & \hspace{-0.3cm}\footnotesize{\textpm 0.0133} & 10.00 & \hspace{-0.3cm}\footnotesize{\textpm 0.00} & 100 & \multicolumn{3}{c}{$\{ 100 \}$} \\ 
 && \phantom{00}2\,500 \footnotesize{(10\,000)} & 100 & 2.73 & 5.29 & 0.7188 & \hspace{-0.3cm}\footnotesize{\textpm 0.0090} & 10.00 & \hspace{-0.3cm}\footnotesize{\textpm 0.00} & 250 & \multicolumn{3}{c}{$\{ 250 \}$} \\ 
 && \phantom{00}5\,000 \footnotesize{(10\,000)} & 100 & 6.61 & 15.24 & 0.6825 & \hspace{-0.3cm}\footnotesize{\textpm 0.0074} & 10.00 & \hspace{-0.3cm}\footnotesize{\textpm 0.00} & 500 & \multicolumn{3}{c}{$\{ 500 \}$} \\ 
 && \phantom{0}10\,000 \footnotesize{(50\,000)} & 10 & 31.94 & 69.23 & 0.6278 & \hspace{-0.3cm}\footnotesize{\textpm 0.0079} & 10.00 & \hspace{-0.3cm}\footnotesize{\textpm 0.00} & 1000 & \multicolumn{3}{c}{$\{ 1000 \}$} \\ 
 && \phantom{0}25\,000 \footnotesize{(50\,000)} & 10 & 259.18 & 304.44 & 0.5444 & \hspace{-0.3cm}\footnotesize{\textpm 0.0041} & 10.00 & \hspace{-0.3cm}\footnotesize{\textpm 0.00} & 2500 & \multicolumn{3}{c}{$\{ 2500 \}$} \\
\end{tabular}
    }
  \end{minipage}
  \caption{RC-SVM results relating to the \textsc{covtype} data sets}
  \label{tab::experiments::covtype_rc}
\end{sidewaystable} 

\begin{sidewaystable}[t!]
  \centering
  \begin{minipage}[t]{0.49\linewidth} \vspace{1cm} 
    \resizebox{\textwidth}{!}{
      \begin{tabular}{cc||c|c||S[tabnumalign=centre]S[tabnumalign=centre]rlrlS[tabnumalign=centre]rcl}
	&& data set sizes & runs & \multicolumn{1}{c}{train time} & \multicolumn{1}{c}{test time} & \multicolumn{2}{c}{test error} & \multicolumn{2}{c}{\# of ws} & \multicolumn{1}{c}{ws size: median} & \multicolumn{3}{c}{ws size: range}\\ 
	&&&&&&&&&&&&&\\[-0.35cm]\hline\hline&&&&&&&&&&&&&\\[-0.25cm] 
	\multirow{7}{*}{ \begin{turn}{90} LS-SVM  \end{turn}}
	&  & \phantom{00}1\,000 \footnotesize{(10\,000)} & 100 & 1.21 & 0.25 & 0.1916 & \hspace{-0.3cm}\footnotesize{\textpm 0.0055} & 1.00 &  & \\ 
 && \phantom{00}2\,500 \footnotesize{(10\,000)} & 100 & 8.25 & 0.57 & 0.1842 & \hspace{-0.3cm}\footnotesize{\textpm 0.0043} & 1.00 &  & \\ 
 && \phantom{00}5\,000 \footnotesize{(10\,000)} & 100 & 33.98 & 1.03 & 0.1672 & \hspace{-0.3cm}\footnotesize{\textpm 0.0032} & 1.00 &  & \\ 
 && \phantom{0}10\,000 \footnotesize{(50\,000)} & 10 & 129.26 & 11.76 & 0.1596 & \hspace{-0.3cm}\footnotesize{\textpm 0.0032} & 1.00 &  & \\ 
 && \phantom{0}25\,000 \footnotesize{(50\,000)} & 10 & 791.14 & 52.75 & 0.1453 & \hspace{-0.3cm}\footnotesize{\textpm 0.0020} & 1.00 &  & \\
\end{tabular}
    }
  \end{minipage}\hfill
  \begin{minipage}[t]{0.49\linewidth} \vspace{1cm} 
    \resizebox{\textwidth}{!}{
      \begin{tabular}{cc||c|c||S[tabnumalign=centre]S[tabnumalign=centre]rlrlS[tabnumalign=centre]rcl}
	&& data set sizes & runs & \multicolumn{1}{c}{train time} & \multicolumn{1}{c}{test time} & \multicolumn{2}{c}{test error} & \multicolumn{2}{c}{\# of ws} & \multicolumn{1}{c}{ws size: median} & \multicolumn{3}{c}{ws size: range}\\ 
	&&&&&&&&&&&&&\\[-0.35cm]\hline\hline&&&&&&&&&&&&&\\[-0.25cm] 
	\multirow{9}{*}{ \begin{turn}{90} RC-SVM  \end{turn}} &
	\multirow{9}{*}{ \hspace{-0.3cm} \begin{turn}{90} \# of ws = 3  \end{turn}}
	  & \phantom{00}1\,000 \footnotesize{(10\,000)} & 100 & 0.87 & 0.39 & 0.2036 & \hspace{-0.3cm}\footnotesize{\textpm 0.0050} & 3.00 & \hspace{-0.3cm}\footnotesize{\textpm 0.00} & 333 & $[ 333$ & \hspace{-0.5cm}, & \hspace{-0.5cm}$334 ]$ \\ 
 && \phantom{00}2\,500 \footnotesize{(10\,000)} & 100 & 2.65 & 0.63 & 0.1925 & \hspace{-0.3cm}\footnotesize{\textpm 0.0035} & 3.00 & \hspace{-0.3cm}\footnotesize{\textpm 0.00} & 833 & $[ 833$ & \hspace{-0.5cm}, & \hspace{-0.5cm}$834 ]$ \\ 
 && \phantom{00}5\,000 \footnotesize{(10\,000)} & 100 & 9.58 & 0.99 & 0.1758 & \hspace{-0.3cm}\footnotesize{\textpm 0.0027} & 3.00 & \hspace{-0.3cm}\footnotesize{\textpm 0.00} & 1667 & $[ 1666$ & \hspace{-0.5cm}, & \hspace{-0.5cm}$1667 ]$ \\ 
 && \phantom{0}10\,000 \footnotesize{(50\,000)} & 10 & 34.21 & 11.14 & 0.1660 & \hspace{-0.3cm}\footnotesize{\textpm 0.0025} & 3.00 & \hspace{-0.3cm}\footnotesize{\textpm 0.00} & 3333 & $[ 3333$ & \hspace{-0.5cm}, & \hspace{-0.5cm}$3334 ]$ \\
\end{tabular}
    }
  \end{minipage} 
  \caption{Experimental results relating to the \textsc{cod-rna} data sets}
  \label{tab::experiments::cod-rna}
\end{sidewaystable}

\begin{sidewaystable}[t!]
  \centering
  \begin{minipage}[t]{0.49\linewidth} \vspace{0pt}
    \resizebox{\textwidth}{!}{
      \begin{tabular}{cc||c|c||S[tabnumalign=centre]S[tabnumalign=centre]rlrlS[tabnumalign=centre]rcl}
	&& data set sizes & runs & \multicolumn{1}{c}{train time} & \multicolumn{1}{c}{test time} & \multicolumn{2}{c}{test error} & \multicolumn{2}{c}{\# of ws} & \multicolumn{1}{c}{ws size: median} & \multicolumn{3}{c}{ws size: range}\\ 
	&&&&&&&&&&&&&\\[-0.35cm]\hline\hline&&&&&&&&&&&&&\\[-0.25cm] 
	\multirow{7}{*}{ \begin{turn}{90} LS-SVM  \end{turn}}
	&  & \phantom{00}1\,000 \footnotesize{(10\,000)} & 100 & 1.13 & 0.41 & 0.1582 & \hspace{-0.3cm}\footnotesize{\textpm 0.0052} & 1.00 &  & \\ 
 && \phantom{00}2\,500 \footnotesize{(10\,000)} & 100 & 4.08 & 0.80 & 0.1099 & \hspace{-0.3cm}\footnotesize{\textpm 0.0028} & 1.00 &  & \\ 
 && \phantom{00}5\,000 \footnotesize{(10\,000)} & 100 & 18.37 & 1.31 & 0.1010 & \hspace{-0.3cm}\footnotesize{\textpm 0.0016} & 1.00 &  & \\ 
 && \phantom{0}10\,000 \footnotesize{(50\,000)} & 10 & 82.65 & 9.29 & 0.0728 & \hspace{-0.3cm}\footnotesize{\textpm 0.0006} & 1.00 &  & \\ 
 && \phantom{0}25\,000 \footnotesize{(50\,000)} & 10 & 526.48 & 24.21 & 0.0542 & \hspace{-0.3cm}\footnotesize{\textpm 0.0004} & 1.00 &  & \\
\end{tabular}
    }
  \end{minipage}\hfill
  \begin{minipage}[t]{0.49\linewidth} \vspace{0pt}
    \resizebox{\textwidth}{!}{
      \begin{tabular}{cc||c|c||S[tabnumalign=centre]S[tabnumalign=centre]rlrlS[tabnumalign=centre]rcl}
	&& data set sizes & runs & \multicolumn{1}{c}{train time} & \multicolumn{1}{c}{test time} & \multicolumn{2}{c}{test error} & \multicolumn{2}{c}{\# of ws} & \multicolumn{1}{c}{ws size: median} & \multicolumn{3}{c}{ws size: range}\\ 
	&&&&&&&&&&&&&\\[-0.35cm]\hline\hline&&&&&&&&&&&&&\\[-0.25cm] 
	\multirow{7}{*}{ \begin{turn}{90} RC-SVM  \end{turn}} &
	\multirow{7}{*}{ \hspace{-0.3cm} \begin{turn}{90} \# of ws = 1  \end{turn}}
	  & \phantom{00}1\,000 \footnotesize{(10\,000)} & 100 & 1.08 & 0.38 & 0.1582 & \hspace{-0.3cm}\footnotesize{\textpm 0.0052} & 1.00 & \hspace{-0.3cm}\footnotesize{\textpm 0.00} & 1000 & \multicolumn{3}{c}{$\{ 1000 \}$} \\ 
 && \phantom{00}2\,500 \footnotesize{(10\,000)} & 100 & 4.01 & 0.85 & 0.1099 & \hspace{-0.3cm}\footnotesize{\textpm 0.0028} & 1.00 & \hspace{-0.3cm}\footnotesize{\textpm 0.00} & 2500 & \multicolumn{3}{c}{$\{ 2500 \}$} \\ 
 && \phantom{00}5\,000 \footnotesize{(10\,000)} & 100 & 17.47 & 1.29 & 0.1010 & \hspace{-0.3cm}\footnotesize{\textpm 0.0016} & 1.00 & \hspace{-0.3cm}\footnotesize{\textpm 0.00} & 5000 & \multicolumn{3}{c}{$\{ 5000 \}$} \\ 
 && \phantom{0}10\,000 \footnotesize{(50\,000)} & 10 & 79.67 & 8.62 & 0.0728 & \hspace{-0.3cm}\footnotesize{\textpm 0.0006} & 1.00 & \hspace{-0.3cm}\footnotesize{\textpm 0.00} & 10000 & \multicolumn{3}{c}{$\{ 10000 \}$} \\ 
 && \phantom{0}25\,000 \footnotesize{(50\,000)} & 10 & 500.55 & 22.20 & 0.0542 & \hspace{-0.3cm}\footnotesize{\textpm 0.0004} & 1.00 & \hspace{-0.3cm}\footnotesize{\textpm 0.00} & 25000 & \multicolumn{3}{c}{$\{ 25000 \}$} \\
\end{tabular}
    }
  \end{minipage} 
  \caption{Experimental results relating to the \textsc{ijcnn1} data sets}
  \label{tab::experiments::ijcnn1}
\end{sidewaystable}


\begin{sidewaystable}[t]
  \centering
  \begin{minipage}[t]{0.49\linewidth} \vspace{0pt}
    \resizebox{\textwidth}{!}{
        \begin{tabular}{cc||c|c||S[tabnumalign=centre]S[tabnumalign=centre]rlrlrlS[tabnumalign=centre]rcl}
	&& data set sizes & runs & \multicolumn{1}{c}{train time} & \multicolumn{1}{c}{test time} & \multicolumn{2}{c}{test error} & \multicolumn{2}{c}{$L_2$-error} & \multicolumn{2}{c}{\# of ws} & \multicolumn{1}{c}{ws size: median} & \multicolumn{3}{c}{ws size: range}\\ 
	&&&&&&&&&&&&&&&\\[-0.35cm]\hline\hline&&&&&&&&&&&&&&&\\[-0.25cm]  
	\multirow{4}{*}{ \begin{turn}{90} LS-SVM  \end{turn}}
	&  & \phantom{0}1\,000 \footnotesize{(10\,000)} & 100 & 0.33 & 0.07 & 0.0284 & \hspace{-0.3cm}\footnotesize{\textpm 0.0003} & 0.0541 & \hspace{-0.3cm}\footnotesize{\textpm 0.0033} & 1.00 &  & \\ 
 && \phantom{0}2\,500 \footnotesize{(10\,000)} & 100 & 1.74 & 0.07 & 0.0275 & \hspace{-0.3cm}\footnotesize{\textpm 0.0002} & 0.0461 & \hspace{-0.3cm}\footnotesize{\textpm 0.0019} & 1.00 &  & \\ 
 && \phantom{0}5\,000 \footnotesize{(10\,000)} & 100 & 7.14 & 0.14 & 0.0269 & \hspace{-0.3cm}\footnotesize{\textpm 0.0002} & 0.0396 & \hspace{-0.3cm}\footnotesize{\textpm 0.0019} & 1.00 &  & \\ 
 && 10\,000 \footnotesize{(10\,000)} & 100 & 27.90 & 0.22 & 0.0265 & \hspace{-0.3cm}\footnotesize{\textpm 0.0001} & 0.0323 & \hspace{-0.3cm}\footnotesize{\textpm 0.0014} & 1.00 &  & \\ 
 
	\noalign{\bigskip}\noalign{\bigskip}
	&& data set sizes & runs & \multicolumn{1}{c}{train time} & \multicolumn{1}{c}{test time} & \multicolumn{2}{c}{test error} & \multicolumn{2}{c}{$L_2$-error} & \multicolumn{2}{c}{\# of ws} & \multicolumn{1}{c}{ws size: median} & \multicolumn{3}{c}{ws size: range}\\ 
	&&&&&&&&&&&&&&&\\[-0.35cm]\hline\hline&&&&&&&&&&&&&&&\\[-0.25cm] 
	\multirow{4}{*}{ \begin{turn}{90} VP-SVM  \end{turn}} &
	\multirow{4}{*}{ \hspace{-0.3cm} \begin{turn}{90} $r$ =  0.1  \end{turn}}
	  & \phantom{0}1\,000 \footnotesize{(10\,000)} & 100 & 1.84 & 0.28 & 0.0287 & \hspace{-0.3cm}\footnotesize{\textpm 0.0005} & 0.0561 & \hspace{-0.3cm}\footnotesize{\textpm 0.0041} & 15.70 & \hspace{-0.3cm}\footnotesize{\textpm 2.43} & 64 & $[ 27$ & \hspace{-0.5cm}, & \hspace{-0.5cm}$109 ]$ \\ 
 && \phantom{0}2\,500 \footnotesize{(10\,000)} & 100 & 1.90 & 0.34 & 0.0267 & \hspace{-0.3cm}\footnotesize{\textpm 0.0002} & 0.0368 & \hspace{-0.3cm}\footnotesize{\textpm 0.0027} & 15.77 & \hspace{-0.3cm}\footnotesize{\textpm 2.34} & 159 & $[ 59$ & \hspace{-0.5cm}, & \hspace{-0.5cm}$268 ]$ \\ 
 && \phantom{0}5\,000 \footnotesize{(10\,000)} & 100 & 2.01 & 0.40 & 0.0260 & \hspace{-0.3cm}\footnotesize{\textpm 0.0001} & 0.0248 & \hspace{-0.3cm}\footnotesize{\textpm 0.0018} & 16.04 & \hspace{-0.3cm}\footnotesize{\textpm 2.45} & 312 & $[ 125$ & \hspace{-0.5cm}, & \hspace{-0.5cm}$532 ]$ \\ 
 && 10\,000 \footnotesize{(10\,000)} & 100 & 3.05 & 0.55 & 0.0257 & \hspace{-0.3cm}\footnotesize{\textpm 0.0001} & 0.0197 & \hspace{-0.3cm}\footnotesize{\textpm 0.0010} & 15.55 & \hspace{-0.3cm}\footnotesize{\textpm 2.43} & 643 & $[ 245$ & \hspace{-0.5cm}, & \hspace{-0.5cm}$1054 ]$ \\ 

	&&&&&&&&&&&&&&&\\[-0.35cm]\hline&&&&&&&&&&&&&&&\\[-0.25cm] 
	\multirow{4}{*}{ \begin{turn}{90} VP-SVM  \end{turn}} &
	\multirow{4}{*}{ \hspace{-0.3cm} \begin{turn}{90} $r$ =  0.25  \end{turn}}
	  & \phantom{0}1\,000 \footnotesize{(10\,000)} & 100 & 0.65 & 0.16 & 0.0277 & \hspace{-0.3cm}\footnotesize{\textpm 0.0004} & 0.0465 & \hspace{-0.3cm}\footnotesize{\textpm 0.0037} & 6.17 & \hspace{-0.3cm}\footnotesize{\textpm 0.88} & 162 & $[ 66$ & \hspace{-0.5cm}, & \hspace{-0.5cm}$251 ]$ \\ 
 && \phantom{0}2\,500 \footnotesize{(10\,000)} & 100 & 0.75 & 0.18 & 0.0265 & \hspace{-0.3cm}\footnotesize{\textpm 0.0002} & 0.0338 & \hspace{-0.3cm}\footnotesize{\textpm 0.0024} & 6.12 & \hspace{-0.3cm}\footnotesize{\textpm 0.87} & 408 & $[ 152$ & \hspace{-0.5cm}, & \hspace{-0.5cm}$645 ]$ \\ 
 && \phantom{0}5\,000 \footnotesize{(10\,000)} & 100 & 1.58 & 0.24 & 0.0260 & \hspace{-0.3cm}\footnotesize{\textpm 0.0001} & 0.0256 & \hspace{-0.3cm}\footnotesize{\textpm 0.0016} & 6.17 & \hspace{-0.3cm}\footnotesize{\textpm 0.79} & 810 & $[ 305$ & \hspace{-0.5cm}, & \hspace{-0.5cm}$1291 ]$ \\ 
 && 10\,000 \footnotesize{(10\,000)} & 100 & 4.93 & 0.38 & 0.0258 & \hspace{-0.3cm}\footnotesize{\textpm 0.0001} & 0.0213 & \hspace{-0.3cm}\footnotesize{\textpm 0.0013} & 6.27 & \hspace{-0.3cm}\footnotesize{\textpm 0.83} & 1595 & $[ 600$ & \hspace{-0.5cm}, & \hspace{-0.5cm}$2550 ]$ \\ 

	&&&&&&&&&&&&&&&\\[-0.35cm]\hline&&&&&&&&&&&&&&&\\[-0.25cm] 
	\multirow{4}{*}{ \begin{turn}{90} VP-SVM  \end{turn}} &
	\multirow{4}{*}{ \hspace{-0.3cm} \begin{turn}{90} $r$ =  0.5  \end{turn}}
	  & \phantom{0}1\,000 \footnotesize{(10\,000)} & 100 & 0.37 & 0.11 & 0.0275 & \hspace{-0.3cm}\footnotesize{\textpm 0.0003} & 0.0447 & \hspace{-0.3cm}\footnotesize{\textpm 0.0032} & 3.49 & \hspace{-0.3cm}\footnotesize{\textpm 0.50} & 287 & $[ 121$ & \hspace{-0.5cm}, & \hspace{-0.5cm}$502 ]$ \\ 
 && \phantom{0}2\,500 \footnotesize{(10\,000)} & 100 & 0.77 & 0.14 & 0.0266 & \hspace{-0.3cm}\footnotesize{\textpm 0.0002} & 0.0355 & \hspace{-0.3cm}\footnotesize{\textpm 0.0025} & 3.44 & \hspace{-0.3cm}\footnotesize{\textpm 0.50} & 727 & $[ 329$ & \hspace{-0.5cm}, & \hspace{-0.5cm}$1202 ]$ \\ 
 && \phantom{0}5\,000 \footnotesize{(10\,000)} & 100 & 2.31 & 0.19 & 0.0262 & \hspace{-0.3cm}\footnotesize{\textpm 0.0001} & 0.0290 & \hspace{-0.3cm}\footnotesize{\textpm 0.0021} & 3.41 & \hspace{-0.3cm}\footnotesize{\textpm 0.49} & 1466 & $[ 627$ & \hspace{-0.5cm}, & \hspace{-0.5cm}$2511 ]$ \\ 
 && 10\,000 \footnotesize{(10\,000)} & 100 & 8.88 & 0.31 & 0.0259 & \hspace{-0.3cm}\footnotesize{\textpm 0.0001} & 0.0242 & \hspace{-0.3cm}\footnotesize{\textpm 0.0019} & 3.51 & \hspace{-0.3cm}\footnotesize{\textpm 0.50} & 2849 & $[ 1253$ & \hspace{-0.5cm}, & \hspace{-0.5cm}$5003 ]$ \\ 

	&&&&&&&&&&&&&&&\\[-0.35cm]\hline&&&&&&&&&&&&&&&\\[-0.25cm] 
	\multirow{4}{*}{ \begin{turn}{90} VP-SVM  \end{turn}} &
	\multirow{4}{*}{ \hspace{-0.3cm} \begin{turn}{90} $r$ =  1  \end{turn}}
	  & \phantom{0}1\,000 \footnotesize{(10\,000)} & 100 & 0.27 & 0.08 & 0.0282 & \hspace{-0.3cm}\footnotesize{\textpm 0.0003} & 0.0517 & \hspace{-0.3cm}\footnotesize{\textpm 0.0030} & 2.00 & \hspace{-0.3cm}\footnotesize{\textpm 0.00} & 500 & $[ 266$ & \hspace{-0.5cm}, & \hspace{-0.5cm}$734 ]$ \\ 
 && \phantom{0}2\,500 \footnotesize{(10\,000)} & 100 & 0.97 & 0.11 & 0.0272 & \hspace{-0.3cm}\footnotesize{\textpm 0.0002} & 0.0426 & \hspace{-0.3cm}\footnotesize{\textpm 0.0024} & 2.00 & \hspace{-0.3cm}\footnotesize{\textpm 0.00} & 1250 & $[ 658$ & \hspace{-0.5cm}, & \hspace{-0.5cm}$1842 ]$ \\ 
 && \phantom{0}5\,000 \footnotesize{(10\,000)} & 100 & 3.68 & 0.15 & 0.0268 & \hspace{-0.3cm}\footnotesize{\textpm 0.0002} & 0.0382 & \hspace{-0.3cm}\footnotesize{\textpm 0.0021} & 2.00 & \hspace{-0.3cm}\footnotesize{\textpm 0.00} & 2500 & $[ 1224$ & \hspace{-0.5cm}, & \hspace{-0.5cm}$3776 ]$ \\ 
 && 10\,000 \footnotesize{(10\,000)} & 100 & 14.93 & 0.44 & 0.0263 & \hspace{-0.3cm}\footnotesize{\textpm 0.0001} & 0.0309 & \hspace{-0.3cm}\footnotesize{\textpm 0.0020} & 2.00 & \hspace{-0.3cm}\footnotesize{\textpm 0.00} & 5000 & $[ 2501$ & \hspace{-0.5cm}, & \hspace{-0.5cm}$7499 ]$ \\ 

	&&&&&&&&&&&&&&&\\[-0.35cm]\hline&&&&&&&&&&&&&&&\\[-0.25cm] 
	\multirow{4}{*}{ \begin{turn}{90} VP-SVM  \end{turn}} &
	\multirow{4}{*}{ \hspace{-0.3cm} \begin{turn}{90} $r$ =  2  \end{turn}}
	  & \phantom{0}1\,000 \footnotesize{(10\,000)} & 100 & 0.32 & 0.07 & 0.0284 & \hspace{-0.3cm}\footnotesize{\textpm 0.0004} & 0.0541 & \hspace{-0.3cm}\footnotesize{\textpm 0.0033} & 1.00 & \hspace{-0.3cm}\footnotesize{\textpm 0.00} & 1000 & \multicolumn{3}{c}{$\{ 1000 \}$} \\ 
 && \phantom{0}2\,500 \footnotesize{(10\,000)} & 100 & 1.75 & 0.08 & 0.0275 & \hspace{-0.3cm}\footnotesize{\textpm 0.0002} & 0.0461 & \hspace{-0.3cm}\footnotesize{\textpm 0.0020} & 1.00 & \hspace{-0.3cm}\footnotesize{\textpm 0.00} & 2500 & \multicolumn{3}{c}{$\{ 2500 \}$} \\ 
 && \phantom{0}5\,000 \footnotesize{(10\,000)} & 100 & 7.16 & 0.15 & 0.0269 & \hspace{-0.3cm}\footnotesize{\textpm 0.0002} & 0.0396 & \hspace{-0.3cm}\footnotesize{\textpm 0.0019} & 1.00 & \hspace{-0.3cm}\footnotesize{\textpm 0.00} & 5000 & \multicolumn{3}{c}{$\{ 5000 \}$} \\ 
 && 10\,000 \footnotesize{(10\,000)} & 100 & 27.91 & 0.24 & 0.0265 & \hspace{-0.3cm}\footnotesize{\textpm 0.0001} & 0.0323 & \hspace{-0.3cm}\footnotesize{\textpm 0.0014} & 1.00 & \hspace{-0.3cm}\footnotesize{\textpm 0.00} & 10000 & \multicolumn{3}{c}{$\{ 10000 \}$} \\ 

	\noalign{\bigskip}\noalign{\bigskip}
	&& data set sizes & runs & \multicolumn{1}{c}{train time} & \multicolumn{1}{c}{test time} & \multicolumn{2}{c}{test error} & \multicolumn{2}{c}{$L_2$-error} & \multicolumn{2}{c}{\# of ws} & \multicolumn{1}{c}{ws size: median} & \multicolumn{3}{c}{ws size: range}\\ 
	&&&&&&&&&&&&&&&\\[-0.35cm]\hline\hline&&&&&&&&&&&&&&&\\[-0.25cm] 
	\multirow{4}{*}{ \begin{turn}{90} RC-SVM  \end{turn}} &
	\multirow{4}{*}{ \hspace{-0.3cm} \begin{turn}{90} \# ws = 1  \end{turn}}
	  & \phantom{0}1\,000 \footnotesize{(10\,000)} & 100 & 0.32 & 0.07 & 0.0284 & \hspace{-0.3cm}\footnotesize{\textpm 0.0003} & 0.0540 & \hspace{-0.3cm}\footnotesize{\textpm 0.0031} & 1.00 & \hspace{-0.3cm}\footnotesize{\textpm 0.00} & 1000 & \multicolumn{3}{c}{$\{ 1000 \}$} \\ 
 && \phantom{0}2\,500 \footnotesize{(10\,000)} & 100 & 1.76 & 0.08 & 0.0275 & \hspace{-0.3cm}\footnotesize{\textpm 0.0002} & 0.0461 & \hspace{-0.3cm}\footnotesize{\textpm 0.0020} & 1.00 & \hspace{-0.3cm}\footnotesize{\textpm 0.00} & 2500 & \multicolumn{3}{c}{$\{ 2500 \}$} \\ 
 && \phantom{0}5\,000 \footnotesize{(10\,000)} & 100 & 7.16 & 0.15 & 0.0269 & \hspace{-0.3cm}\footnotesize{\textpm 0.0002} & 0.0396 & \hspace{-0.3cm}\footnotesize{\textpm 0.0019} & 1.00 & \hspace{-0.3cm}\footnotesize{\textpm 0.00} & 5000 & \multicolumn{3}{c}{$\{ 5000 \}$} \\ 
 && 10\,000 \footnotesize{(10\,000)} & 100 & 28.05 & 0.24 & 0.0265 & \hspace{-0.3cm}\footnotesize{\textpm 0.0001} & 0.0323 & \hspace{-0.3cm}\footnotesize{\textpm 0.0014} & 1.00 & \hspace{-0.3cm}\footnotesize{\textpm 0.00} & 10000 & \multicolumn{3}{c}{$\{ 10000 \}$} \\ 

	&&&&&&&&&&&&&&&\\[-0.35cm]\hline&&&&&&&&&&&&&&&\\[-0.25cm] 
	\multirow{4}{*}{ \begin{turn}{90} RC-SVM  \end{turn}} &
	\multirow{4}{*}{ \hspace{-0.3cm} \begin{turn}{90} \# ws = 2  \end{turn}}
	  & \phantom{0}1\,000 \footnotesize{(10\,000)} & 100 & 0.26 & 0.09 & 0.0287 & \hspace{-0.3cm}\footnotesize{\textpm 0.0005} & 0.0571 & \hspace{-0.3cm}\footnotesize{\textpm 0.0042} & 2.00 & \hspace{-0.3cm}\footnotesize{\textpm 0.00} & 500 & \multicolumn{3}{c}{$\{ 500 \}$} \\ 
 && \phantom{0}2\,500 \footnotesize{(10\,000)} & 100 & 0.94 & 0.11 & 0.0276 & \hspace{-0.3cm}\footnotesize{\textpm 0.0002} & 0.0475 & \hspace{-0.3cm}\footnotesize{\textpm 0.0018} & 2.00 & \hspace{-0.3cm}\footnotesize{\textpm 0.00} & 1250 & \multicolumn{3}{c}{$\{ 1250 \}$} \\ 
 && \phantom{0}5\,000 \footnotesize{(10\,000)} & 100 & 3.46 & 0.14 & 0.0273 & \hspace{-0.3cm}\footnotesize{\textpm 0.0001} & 0.0444 & \hspace{-0.3cm}\footnotesize{\textpm 0.0014} & 2.00 & \hspace{-0.3cm}\footnotesize{\textpm 0.00} & 2500 & \multicolumn{3}{c}{$\{ 2500 \}$} \\ 
 && 10\,000 \footnotesize{(10\,000)} & 100 & 14.37 & 0.28 & 0.0268 & \hspace{-0.3cm}\footnotesize{\textpm 0.0001} & 0.0376 & \hspace{-0.3cm}\footnotesize{\textpm 0.0016} & 2.00 & \hspace{-0.3cm}\footnotesize{\textpm 0.00} & 5000 & \multicolumn{3}{c}{$\{ 5000 \}$} \\ 

	&&&&&&&&&&&&&&&\\[-0.35cm]\hline&&&&&&&&&&&&&&&\\[-0.25cm] 
	\multirow{4}{*}{ \begin{turn}{90} RC-SVM  \end{turn}} &
	\multirow{4}{*}{ \hspace{-0.3cm} \begin{turn}{90} \# ws = 3  \end{turn}}
	  & \phantom{0}1\,000 \footnotesize{(10\,000)} & 100 & 0.31 & 0.11 & 0.0288 & \hspace{-0.3cm}\footnotesize{\textpm 0.0004} & 0.0583 & \hspace{-0.3cm}\footnotesize{\textpm 0.0037} & 3.00 & \hspace{-0.3cm}\footnotesize{\textpm 0.00} & 333 & $[ 333$ & \hspace{-0.5cm}, & \hspace{-0.5cm}$334 ]$ \\ 
 && \phantom{0}2\,500 \footnotesize{(10\,000)} & 100 & 0.76 & 0.14 & 0.0277 & \hspace{-0.3cm}\footnotesize{\textpm 0.0002} & 0.0488 & \hspace{-0.3cm}\footnotesize{\textpm 0.0022} & 3.00 & \hspace{-0.3cm}\footnotesize{\textpm 0.00} & 833 & $[ 833$ & \hspace{-0.5cm}, & \hspace{-0.5cm}$834 ]$ \\ 
 && \phantom{0}5\,000 \footnotesize{(10\,000)} & 100 & 2.32 & 0.17 & 0.0274 & \hspace{-0.3cm}\footnotesize{\textpm 0.0001} & 0.0456 & \hspace{-0.3cm}\footnotesize{\textpm 0.0014} & 3.00 & \hspace{-0.3cm}\footnotesize{\textpm 0.00} & 1667 & $[ 1666$ & \hspace{-0.5cm}, & \hspace{-0.5cm}$1667 ]$ \\ 
 && 10\,000 \footnotesize{(10\,000)} & 100 & 9.58 & 0.27 & 0.0271 & \hspace{-0.3cm}\footnotesize{\textpm 0.0001} & 0.0417 & \hspace{-0.3cm}\footnotesize{\textpm 0.0015} & 3.00 & \hspace{-0.3cm}\footnotesize{\textpm 0.00} & 3333 & $[ 3333$ & \hspace{-0.5cm}, & \hspace{-0.5cm}$3334 ]$ \\ 

	&&&&&&&&&&&&&&&\\[-0.35cm]\hline&&&&&&&&&&&&&&&\\[-0.25cm] 
	\multirow{4}{*}{ \begin{turn}{90} RC-SVM  \end{turn}} &
	\multirow{4}{*}{ \hspace{-0.3cm} \begin{turn}{90} \# ws = 4  \end{turn}}
	  & \phantom{0}1\,000 \footnotesize{(10\,000)} & 100 & 0.43 & 0.12 & 0.0289 & \hspace{-0.3cm}\footnotesize{\textpm 0.0004} & 0.0588 & \hspace{-0.3cm}\footnotesize{\textpm 0.0037} & 4.00 & \hspace{-0.3cm}\footnotesize{\textpm 0.00} & 250 & \multicolumn{3}{c}{$\{ 250 \}$} \\ 
 && \phantom{0}2\,500 \footnotesize{(10\,000)} & 100 & 0.69 & 0.16 & 0.0278 & \hspace{-0.3cm}\footnotesize{\textpm 0.0003} & 0.0500 & \hspace{-0.3cm}\footnotesize{\textpm 0.0025} & 4.00 & \hspace{-0.3cm}\footnotesize{\textpm 0.00} & 625 & \multicolumn{3}{c}{$\{ 625 \}$} \\ 
 && \phantom{0}5\,000 \footnotesize{(10\,000)} & 100 & 1.90 & 0.20 & 0.0275 & \hspace{-0.3cm}\footnotesize{\textpm 0.0002} & 0.0469 & \hspace{-0.3cm}\footnotesize{\textpm 0.0016} & 4.00 & \hspace{-0.3cm}\footnotesize{\textpm 0.00} & 1250 & \multicolumn{3}{c}{$\{ 1250 \}$} \\ 
 && 10\,000 \footnotesize{(10\,000)} & 100 & 6.95 & 0.27 & 0.0272 & \hspace{-0.3cm}\footnotesize{\textpm 0.0001} & 0.0432 & \hspace{-0.3cm}\footnotesize{\textpm 0.0010} & 4.00 & \hspace{-0.3cm}\footnotesize{\textpm 0.00} & 2500 & \multicolumn{3}{c}{$\{ 2500 \}$} \\ 

	&&&&&&&&&&&&&&&\\[-0.35cm]\hline&&&&&&&&&&&&&&&\\[-0.25cm] 
	\multirow{4}{*}{ \begin{turn}{90} RC-SVM  \end{turn}} &
	\multirow{4}{*}{ \hspace{-0.3cm} \begin{turn}{90} \# ws = 5  \end{turn}}
	  & \phantom{0}1\,000 \footnotesize{(10\,000)} & 100 & 0.52 & 0.15 & 0.0287 & \hspace{-0.3cm}\footnotesize{\textpm 0.0004} & 0.0577 & \hspace{-0.3cm}\footnotesize{\textpm 0.0033} & 5.00 & \hspace{-0.3cm}\footnotesize{\textpm 0.00} & 200 & \multicolumn{3}{c}{$\{ 200 \}$} \\ 
 && \phantom{0}2\,500 \footnotesize{(10\,000)} & 100 & 0.67 & 0.18 & 0.0280 & \hspace{-0.3cm}\footnotesize{\textpm 0.0002} & 0.0515 & \hspace{-0.3cm}\footnotesize{\textpm 0.0022} & 5.00 & \hspace{-0.3cm}\footnotesize{\textpm 0.00} & 500 & \multicolumn{3}{c}{$\{ 500 \}$} \\ 
 && \phantom{0}5\,000 \footnotesize{(10\,000)} & 100 & 1.65 & 0.23 & 0.0276 & \hspace{-0.3cm}\footnotesize{\textpm 0.0002} & 0.0480 & \hspace{-0.3cm}\footnotesize{\textpm 0.0020} & 5.00 & \hspace{-0.3cm}\footnotesize{\textpm 0.00} & 1000 & \multicolumn{3}{c}{$\{ 1000 \}$} \\ 
 && 10\,000 \footnotesize{(10\,000)} & 100 & 5.52 & 0.30 & 0.0273 & \hspace{-0.3cm}\footnotesize{\textpm 0.0001} & 0.0438 & \hspace{-0.3cm}\footnotesize{\textpm 0.0012} & 5.00 & \hspace{-0.3cm}\footnotesize{\textpm 0.00} & 2000 & \multicolumn{3}{c}{$\{ 2000 \}$} \\ 

	&&&&&&&&&&&&&&&\\[-0.35cm]\hline&&&&&&&&&&&&&&&\\[-0.25cm] 
	\multirow{4}{*}{ \begin{turn}{90} RC-SVM  \end{turn}} &
	\multirow{4}{*}{ \hspace{-0.3cm} \begin{turn}{90} \# ws = 6  \end{turn}}
	  & \phantom{0}1\,000 \footnotesize{(10\,000)} & 100 & 0.64 & 0.16 & 0.0288 & \hspace{-0.3cm}\footnotesize{\textpm 0.0004} & 0.0586 & \hspace{-0.3cm}\footnotesize{\textpm 0.0031} & 6.00 & \hspace{-0.3cm}\footnotesize{\textpm 0.00} & 167 & $[ 166$ & \hspace{-0.5cm}, & \hspace{-0.5cm}$167 ]$ \\ 
 && \phantom{0}2\,500 \footnotesize{(10\,000)} & 100 & 0.69 & 0.19 & 0.0281 & \hspace{-0.3cm}\footnotesize{\textpm 0.0003} & 0.0525 & \hspace{-0.3cm}\footnotesize{\textpm 0.0028} & 6.00 & \hspace{-0.3cm}\footnotesize{\textpm 0.00} & 417 & $[ 416$ & \hspace{-0.5cm}, & \hspace{-0.5cm}$417 ]$ \\ 
 && \phantom{0}5\,000 \footnotesize{(10\,000)} & 100 & 1.52 & 0.25 & 0.0277 & \hspace{-0.3cm}\footnotesize{\textpm 0.0002} & 0.0490 & \hspace{-0.3cm}\footnotesize{\textpm 0.0022} & 6.00 & \hspace{-0.3cm}\footnotesize{\textpm 0.00} & 833 & $[ 833$ & \hspace{-0.5cm}, & \hspace{-0.5cm}$834 ]$ \\ 
 && 10\,000 \footnotesize{(10\,000)} & 100 & 4.67 & 0.32 & 0.0273 & \hspace{-0.3cm}\footnotesize{\textpm 0.0001} & 0.0446 & \hspace{-0.3cm}\footnotesize{\textpm 0.0011} & 6.00 & \hspace{-0.3cm}\footnotesize{\textpm 0.00} & 1667 & $[ 1666$ & \hspace{-0.5cm}, & \hspace{-0.5cm}$1667 ]$ \\ 

	&&&&&&&&&&&&&&&\\[-0.35cm]\hline&&&&&&&&&&&&&&&\\[-0.25cm] 
	\multirow{4}{*}{ \begin{turn}{90} RC-SVM  \end{turn}} &
	\multirow{4}{*}{ \hspace{-0.3cm} \begin{turn}{90} \# ws = 10  \end{turn}}
	  & \phantom{0}1\,000 \footnotesize{(10\,000)} & 100 & 1.05 & 0.22 & 0.0288 & \hspace{-0.3cm}\footnotesize{\textpm 0.0003} & 0.0584 & \hspace{-0.3cm}\footnotesize{\textpm 0.0029} & 10.00 & \hspace{-0.3cm}\footnotesize{\textpm 0.00} & 100 & \multicolumn{3}{c}{$\{ 100 \}$} \\ 
 && \phantom{0}2\,500 \footnotesize{(10\,000)} & 100 & 1.08 & 0.27 & 0.0282 & \hspace{-0.3cm}\footnotesize{\textpm 0.0003} & 0.0534 & \hspace{-0.3cm}\footnotesize{\textpm 0.0026} & 10.00 & \hspace{-0.3cm}\footnotesize{\textpm 0.00} & 250 & \multicolumn{3}{c}{$\{ 250 \}$} \\ 
 && \phantom{0}5\,000 \footnotesize{(10\,000)} & 100 & 1.39 & 0.34 & 0.0280 & \hspace{-0.3cm}\footnotesize{\textpm 0.0002} & 0.0522 & \hspace{-0.3cm}\footnotesize{\textpm 0.0023} & 10.00 & \hspace{-0.3cm}\footnotesize{\textpm 0.00} & 500 & \multicolumn{3}{c}{$\{ 500 \}$} \\ 
 && 10\,000 \footnotesize{(10\,000)} & 100 & 3.45 & 0.44 & 0.0275 & \hspace{-0.3cm}\footnotesize{\textpm 0.0001} & 0.0471 & \hspace{-0.3cm}\footnotesize{\textpm 0.0012} & 10.00 & \hspace{-0.3cm}\footnotesize{\textpm 0.00} & 1000 & \multicolumn{3}{c}{$\{ 1000 \}$} \\ 

	&&&&&&&&&&&&&&&\\[-0.35cm]\hline&&&&&&&&&&&&&&&\\[-0.25cm] 
	\multirow{4}{*}{ \begin{turn}{90} RC-SVM  \end{turn}} &
	\multirow{4}{*}{ \hspace{-0.3cm} \begin{turn}{90} \# ws = 20  \end{turn}}
	  & \phantom{0}1\,000 \footnotesize{(10\,000)} & 100 & 2.28 & 0.34 & 0.0293 & \hspace{-0.3cm}\footnotesize{\textpm 0.0004} & 0.0626 & \hspace{-0.3cm}\footnotesize{\textpm 0.0027} & 20.00 & \hspace{-0.3cm}\footnotesize{\textpm 0.00} & 50 & \multicolumn{3}{c}{$\{ 50 \}$} \\ 
 && \phantom{0}2\,500 \footnotesize{(10\,000)} & 100 & 2.06 & 0.43 & 0.0282 & \hspace{-0.3cm}\footnotesize{\textpm 0.0002} & 0.0531 & \hspace{-0.3cm}\footnotesize{\textpm 0.0022} & 20.00 & \hspace{-0.3cm}\footnotesize{\textpm 0.00} & 125 & \multicolumn{3}{c}{$\{ 125 \}$} \\ 
 && \phantom{0}5\,000 \footnotesize{(10\,000)} & 100 & 2.38 & 0.51 & 0.0283 & \hspace{-0.3cm}\footnotesize{\textpm 0.0002} & 0.0544 & \hspace{-0.3cm}\footnotesize{\textpm 0.0017} & 20.00 & \hspace{-0.3cm}\footnotesize{\textpm 0.00} & 250 & \multicolumn{3}{c}{$\{ 250 \}$} \\ 
 && 10\,000 \footnotesize{(10\,000)} & 100 & 3.14 & 0.64 & 0.0280 & \hspace{-0.3cm}\footnotesize{\textpm 0.0001} & 0.0520 & \hspace{-0.3cm}\footnotesize{\textpm 0.0014} & 20.00 & \hspace{-0.3cm}\footnotesize{\textpm 0.00} & 500 & \multicolumn{3}{c}{$\{ 500 \}$} \\ 

      \end{tabular}
    } 
    \caption{Experimental results relating to the artificial data of Type \RM{1}}
    \label{tab::experiments::art_data_1_2}
  \end{minipage} 
  \hfill
  \begin{minipage}[t]{0.49\linewidth} \vspace{0pt}
    \resizebox{\textwidth}{!}{
        \begin{tabular}{cc||c|c||S[tabnumalign=centre]S[tabnumalign=centre]rlrlrlS[tabnumalign=centre]rcl}
	&& data set sizes & runs & \multicolumn{1}{c}{train time} & \multicolumn{1}{c}{test time} & \multicolumn{2}{c}{test error} & \multicolumn{2}{c}{$L_2$-error} & \multicolumn{2}{c}{\# of ws} & \multicolumn{1}{c}{ws size: median} & \multicolumn{3}{c}{ws size: range}\\ 
	&&&&&&&&&&&&&&&\\[-0.35cm]\hline\hline&&&&&&&&&&&&&&&\\[-0.25cm]  
	\multirow{4}{*}{ \begin{turn}{90} LS-SVM  \end{turn}}
	&  & \phantom{0}1\,000 \footnotesize{(10\,000)} & 100 & 0.60 & 0.12 & 0.0178 & \hspace{-0.3cm}\footnotesize{\textpm 0.0006} & 0.0641 & \hspace{-0.3cm}\footnotesize{\textpm 0.0047} & 1.00 &  & \\ 
 && \phantom{0}2\,500 \footnotesize{(10\,000)} & 100 & 2.29 & 0.21 & 0.0168 & \hspace{-0.3cm}\footnotesize{\textpm 0.0004} & 0.0559 & \hspace{-0.3cm}\footnotesize{\textpm 0.0033} & 1.00 &  & \\ 
 && \phantom{0}5\,000 \footnotesize{(10\,000)} & 100 & 8.37 & 0.25 & 0.0165 & \hspace{-0.3cm}\footnotesize{\textpm 0.0003} & 0.0531 & \hspace{-0.3cm}\footnotesize{\textpm 0.0026} & 1.00 &  & \\ 
 && 10\,000 \footnotesize{(10\,000)} & 100 & 31.61 & 0.44 & 0.0163 & \hspace{-0.3cm}\footnotesize{\textpm 0.0003} & 0.0511 & \hspace{-0.3cm}\footnotesize{\textpm 0.0033} & 1.00 &  & \\ 
 
	\noalign{\bigskip}\noalign{\bigskip}
	&& data set sizes & runs & \multicolumn{1}{c}{train time} & \multicolumn{1}{c}{test time} & \multicolumn{2}{c}{test error} & \multicolumn{2}{c}{$L_2$-error} & \multicolumn{2}{c}{\# of ws} & \multicolumn{1}{c}{ws size: median} & \multicolumn{3}{c}{ws size: range}\\ 
	&&&&&&&&&&&&&&&\\[-0.35cm]\hline\hline&&&&&&&&&&&&&&&\\[-0.25cm] 
	\multirow{4}{*}{ \begin{turn}{90} VP-SVM  \end{turn}} &
	\multirow{4}{*}{ \hspace{-0.3cm} \begin{turn}{90} $r$ =  0.1  \end{turn}}
	  & \phantom{0}1\,000 \footnotesize{(10\,000)} & 100 & 1.55 & 0.29 & 0.0157 & \hspace{-0.3cm}\footnotesize{\textpm 0.0004} & 0.0427 & \hspace{-0.3cm}\footnotesize{\textpm 0.0040} & 16.08 & \hspace{-0.3cm}\footnotesize{\textpm 2.50} & 62 & $[ 22$ & \hspace{-0.5cm}, & \hspace{-0.5cm}$114 ]$ \\ 
 && \phantom{0}2\,500 \footnotesize{(10\,000)} & 100 & 1.66 & 0.36 & 0.0146 & \hspace{-0.3cm}\footnotesize{\textpm 0.0002} & 0.0288 & \hspace{-0.3cm}\footnotesize{\textpm 0.0028} & 15.79 & \hspace{-0.3cm}\footnotesize{\textpm 2.40} & 158 & $[ 51$ & \hspace{-0.5cm}, & \hspace{-0.5cm}$292 ]$ \\ 
 && \phantom{0}5\,000 \footnotesize{(10\,000)} & 100 & 1.88 & 0.43 & 0.0142 & \hspace{-0.3cm}\footnotesize{\textpm 0.0001} & 0.0208 & \hspace{-0.3cm}\footnotesize{\textpm 0.0016} & 15.95 & \hspace{-0.3cm}\footnotesize{\textpm 2.53} & 313 & $[ 105$ & \hspace{-0.5cm}, & \hspace{-0.5cm}$549 ]$ \\ 
 && 10\,000 \footnotesize{(10\,000)} & 100 & 3.09 & 0.59 & 0.0140 & \hspace{-0.3cm}\footnotesize{\textpm 0.0001} & 0.0163 & \hspace{-0.3cm}\footnotesize{\textpm 0.0013} & 15.84 & \hspace{-0.3cm}\footnotesize{\textpm 2.50} & 631 & $[ 231$ & \hspace{-0.5cm}, & \hspace{-0.5cm}$1043 ]$ \\ 

	&&&&&&&&&&&&&&&\\[-0.35cm]\hline&&&&&&&&&&&&&&&\\[-0.25cm] 
	\multirow{4}{*}{ \begin{turn}{90} VP-SVM  \end{turn}} &
	\multirow{4}{*}{ \hspace{-0.3cm} \begin{turn}{90} $r$ =  0.25  \end{turn}}
	  & \phantom{0}1\,000 \footnotesize{(10\,000)} & 100 & 0.58 & 0.17 & 0.0156 & \hspace{-0.3cm}\footnotesize{\textpm 0.0004} & 0.0426 & \hspace{-0.3cm}\footnotesize{\textpm 0.0045} & 6.21 & \hspace{-0.3cm}\footnotesize{\textpm 0.83} & 161 & $[ 60$ & \hspace{-0.5cm}, & \hspace{-0.5cm}$267 ]$ \\ 
 && \phantom{0}2\,500 \footnotesize{(10\,000)} & 100 & 0.77 & 0.22 & 0.0147 & \hspace{-0.3cm}\footnotesize{\textpm 0.0002} & 0.0299 & \hspace{-0.3cm}\footnotesize{\textpm 0.0030} & 6.29 & \hspace{-0.3cm}\footnotesize{\textpm 0.81} & 397 & $[ 133$ & \hspace{-0.5cm}, & \hspace{-0.5cm}$676 ]$ \\ 
 && \phantom{0}5\,000 \footnotesize{(10\,000)} & 100 & 1.64 & 0.30 & 0.0144 & \hspace{-0.3cm}\footnotesize{\textpm 0.0001} & 0.0247 & \hspace{-0.3cm}\footnotesize{\textpm 0.0023} & 6.22 & \hspace{-0.3cm}\footnotesize{\textpm 0.82} & 804 & $[ 285$ & \hspace{-0.5cm}, & \hspace{-0.5cm}$1299 ]$ \\ 
 && 10\,000 \footnotesize{(10\,000)} & 100 & 4.99 & 0.44 & 0.0141 & \hspace{-0.3cm}\footnotesize{\textpm 0.0001} & 0.0194 & \hspace{-0.3cm}\footnotesize{\textpm 0.0016} & 6.38 & \hspace{-0.3cm}\footnotesize{\textpm 0.80} & 1567 & $[ 597$ & \hspace{-0.5cm}, & \hspace{-0.5cm}$2535 ]$ \\ 

	&&&&&&&&&&&&&&&\\[-0.35cm]\hline&&&&&&&&&&&&&&&\\[-0.25cm] 
	\multirow{4}{*}{ \begin{turn}{90} VP-SVM  \end{turn}} &
	\multirow{4}{*}{ \hspace{-0.3cm} \begin{turn}{90} $r$ =  0.5  \end{turn}}
	  & \phantom{0}1\,000 \footnotesize{(10\,000)} & 100 & 0.36 & 0.12 & 0.0159 & \hspace{-0.3cm}\footnotesize{\textpm 0.0004} & 0.0465 & \hspace{-0.3cm}\footnotesize{\textpm 0.0050} & 3.47 & \hspace{-0.3cm}\footnotesize{\textpm 0.50} & 288 & $[ 122$ & \hspace{-0.5cm}, & \hspace{-0.5cm}$482 ]$ \\ 
 && \phantom{0}2\,500 \footnotesize{(10\,000)} & 100 & 0.81 & 0.18 & 0.0151 & \hspace{-0.3cm}\footnotesize{\textpm 0.0002} & 0.0370 & \hspace{-0.3cm}\footnotesize{\textpm 0.0031} & 3.52 & \hspace{-0.3cm}\footnotesize{\textpm 0.50} & 710 & $[ 297$ & \hspace{-0.5cm}, & \hspace{-0.5cm}$1248 ]$ \\ 
 && \phantom{0}5\,000 \footnotesize{(10\,000)} & 100 & 2.52 & 0.28 & 0.0150 & \hspace{-0.3cm}\footnotesize{\textpm 0.0002} & 0.0346 & \hspace{-0.3cm}\footnotesize{\textpm 0.0032} & 3.47 & \hspace{-0.3cm}\footnotesize{\textpm 0.50} & 1441 & $[ 604$ & \hspace{-0.5cm}, & \hspace{-0.5cm}$2508 ]$ \\ 
 && 10\,000 \footnotesize{(10\,000)} & 100 & 9.36 & 0.43 & 0.0147 & \hspace{-0.3cm}\footnotesize{\textpm 0.0002} & 0.0301 & \hspace{-0.3cm}\footnotesize{\textpm 0.0034} & 3.58 & \hspace{-0.3cm}\footnotesize{\textpm 0.50} & 2793 & $[ 1249$ & \hspace{-0.5cm}, & \hspace{-0.5cm}$4962 ]$ \\ 

	&&&&&&&&&&&&&&&\\[-0.35cm]\hline&&&&&&&&&&&&&&&\\[-0.25cm] 
	\multirow{4}{*}{ \begin{turn}{90} VP-SVM  \end{turn}} &
	\multirow{4}{*}{ \hspace{-0.3cm} \begin{turn}{90} $r$ =  1  \end{turn}}
	  & \phantom{0}1\,000 \footnotesize{(10\,000)} & 100 & 0.33 & 0.09 & 0.0169 & \hspace{-0.3cm}\footnotesize{\textpm 0.0005} & 0.0562 & \hspace{-0.3cm}\footnotesize{\textpm 0.0047} & 2.00 & \hspace{-0.3cm}\footnotesize{\textpm 0.00} & 500 & $[ 265$ & \hspace{-0.5cm}, & \hspace{-0.5cm}$735 ]$ \\ 
 && \phantom{0}2\,500 \footnotesize{(10\,000)} & 100 & 1.11 & 0.14 & 0.0160 & \hspace{-0.3cm}\footnotesize{\textpm 0.0004} & 0.0477 & \hspace{-0.3cm}\footnotesize{\textpm 0.0041} & 2.00 & \hspace{-0.3cm}\footnotesize{\textpm 0.00} & 1250 & $[ 613$ & \hspace{-0.5cm}, & \hspace{-0.5cm}$1887 ]$ \\ 
 && \phantom{0}5\,000 \footnotesize{(10\,000)} & 100 & 4.11 & 0.23 & 0.0158 & \hspace{-0.3cm}\footnotesize{\textpm 0.0004} & 0.0452 & \hspace{-0.3cm}\footnotesize{\textpm 0.0044} & 2.00 & \hspace{-0.3cm}\footnotesize{\textpm 0.00} & 2500 & $[ 1232$ & \hspace{-0.5cm}, & \hspace{-0.5cm}$3768 ]$ \\ 
 && 10\,000 \footnotesize{(10\,000)} & 100 & 16.26 & 0.42 & 0.0154 & \hspace{-0.3cm}\footnotesize{\textpm 0.0005} & 0.0402 & \hspace{-0.3cm}\footnotesize{\textpm 0.0064} & 2.00 & \hspace{-0.3cm}\footnotesize{\textpm 0.00} & 5000 & $[ 2533$ & \hspace{-0.5cm}, & \hspace{-0.5cm}$7467 ]$ \\ 

	&&&&&&&&&&&&&&&\\[-0.35cm]\hline&&&&&&&&&&&&&&&\\[-0.25cm] 
	\multirow{4}{*}{ \begin{turn}{90} VP-SVM  \end{turn}} &
	\multirow{4}{*}{ \hspace{-0.3cm} \begin{turn}{90} $r$ =  2  \end{turn}}
	  & \phantom{0}1\,000 \footnotesize{(10\,000)} & 100 & 0.37 & 0.08 & 0.0178 & \hspace{-0.3cm}\footnotesize{\textpm 0.0006} & 0.0641 & \hspace{-0.3cm}\footnotesize{\textpm 0.0045} & 1.00 & \hspace{-0.3cm}\footnotesize{\textpm 0.00} & 1000 & \multicolumn{3}{c}{$\{ 1000 \}$} \\ 
 && \phantom{0}2\,500 \footnotesize{(10\,000)} & 100 & 2.03 & 0.12 & 0.0168 & \hspace{-0.3cm}\footnotesize{\textpm 0.0003} & 0.0558 & \hspace{-0.3cm}\footnotesize{\textpm 0.0033} & 1.00 & \hspace{-0.3cm}\footnotesize{\textpm 0.00} & 2500 & \multicolumn{3}{c}{$\{ 2500 \}$} \\ 
 && \phantom{0}5\,000 \footnotesize{(10\,000)} & 100 & 8.12 & 0.20 & 0.0165 & \hspace{-0.3cm}\footnotesize{\textpm 0.0003} & 0.0531 & \hspace{-0.3cm}\footnotesize{\textpm 0.0026} & 1.00 & \hspace{-0.3cm}\footnotesize{\textpm 0.00} & 5000 & \multicolumn{3}{c}{$\{ 5000 \}$} \\ 
 && 10\,000 \footnotesize{(10\,000)} & 100 & 31.12 & 0.36 & 0.0163 & \hspace{-0.3cm}\footnotesize{\textpm 0.0003} & 0.0511 & \hspace{-0.3cm}\footnotesize{\textpm 0.0033} & 1.00 & \hspace{-0.3cm}\footnotesize{\textpm 0.00} & 10000 & \multicolumn{3}{c}{$\{ 10000 \}$} \\ 

	\noalign{\bigskip}\noalign{\bigskip}
	&& data set sizes & runs & \multicolumn{1}{c}{train time} & \multicolumn{1}{c}{test time} & \multicolumn{2}{c}{test error} & \multicolumn{2}{c}{$L_2$-error} & \multicolumn{2}{c}{\# of ws} & \multicolumn{1}{c}{ws size: median} & \multicolumn{3}{c}{ws size: range}\\ 
	&&&&&&&&&&&&&&&\\[-0.35cm]\hline\hline&&&&&&&&&&&&&&&\\[-0.25cm] 
	\multirow{4}{*}{ \begin{turn}{90} RC-SVM  \end{turn}} &
	\multirow{4}{*}{ \hspace{-0.3cm} \begin{turn}{90} \# ws = 1  \end{turn}}
	  & \phantom{0}1\,000 \footnotesize{(10\,000)} & 100 & 0.37 & 0.07 & 0.0178 & \hspace{-0.3cm}\footnotesize{\textpm 0.0006} & 0.0640 & \hspace{-0.3cm}\footnotesize{\textpm 0.0046} & 1.00 & \hspace{-0.3cm}\footnotesize{\textpm 0.00} & 1000 & \multicolumn{3}{c}{$\{ 1000 \}$} \\ 
 && \phantom{0}2\,500 \footnotesize{(10\,000)} & 100 & 2.02 & 0.13 & 0.0168 & \hspace{-0.3cm}\footnotesize{\textpm 0.0003} & 0.0558 & \hspace{-0.3cm}\footnotesize{\textpm 0.0033} & 1.00 & \hspace{-0.3cm}\footnotesize{\textpm 0.00} & 2500 & \multicolumn{3}{c}{$\{ 2500 \}$} \\ 
 && \phantom{0}5\,000 \footnotesize{(10\,000)} & 100 & 8.14 & 0.19 & 0.0165 & \hspace{-0.3cm}\footnotesize{\textpm 0.0003} & 0.0531 & \hspace{-0.3cm}\footnotesize{\textpm 0.0026} & 1.00 & \hspace{-0.3cm}\footnotesize{\textpm 0.00} & 5000 & \multicolumn{3}{c}{$\{ 5000 \}$} \\ 
 && 10\,000 \footnotesize{(10\,000)} & 100 & 31.45 & 0.36 & 0.0163 & \hspace{-0.3cm}\footnotesize{\textpm 0.0003} & 0.0511 & \hspace{-0.3cm}\footnotesize{\textpm 0.0033} & 1.00 & \hspace{-0.3cm}\footnotesize{\textpm 0.00} & 10000 & \multicolumn{3}{c}{$\{ 10000 \}$} \\ 

	&&&&&&&&&&&&&&&\\[-0.35cm]\hline&&&&&&&&&&&&&&&\\[-0.25cm] 
	\multirow{4}{*}{ \begin{turn}{90} RC-SVM  \end{turn}} &
	\multirow{4}{*}{ \hspace{-0.3cm} \begin{turn}{90} \# ws = 2  \end{turn}}
	  & \phantom{0}1\,000 \footnotesize{(10\,000)} & 100 & 0.30 & 0.11 & 0.0172 & \hspace{-0.3cm}\footnotesize{\textpm 0.0005} & 0.0595 & \hspace{-0.3cm}\footnotesize{\textpm 0.0046} & 2.00 & \hspace{-0.3cm}\footnotesize{\textpm 0.00} & 500 & \multicolumn{3}{c}{$\{ 500 \}$} \\ 
 && \phantom{0}2\,500 \footnotesize{(10\,000)} & 100 & 1.10 & 0.15 & 0.0169 & \hspace{-0.3cm}\footnotesize{\textpm 0.0004} & 0.0567 & \hspace{-0.3cm}\footnotesize{\textpm 0.0037} & 2.00 & \hspace{-0.3cm}\footnotesize{\textpm 0.00} & 1250 & \multicolumn{3}{c}{$\{ 1250 \}$} \\ 
 && \phantom{0}5\,000 \footnotesize{(10\,000)} & 100 & 4.06 & 0.25 & 0.0165 & \hspace{-0.3cm}\footnotesize{\textpm 0.0003} & 0.0528 & \hspace{-0.3cm}\footnotesize{\textpm 0.0029} & 2.00 & \hspace{-0.3cm}\footnotesize{\textpm 0.00} & 2500 & \multicolumn{3}{c}{$\{ 2500 \}$} \\ 
 && 10\,000 \footnotesize{(10\,000)} & 100 & 16.33 & 0.41 & 0.0163 & \hspace{-0.3cm}\footnotesize{\textpm 0.0002} & 0.0516 & \hspace{-0.3cm}\footnotesize{\textpm 0.0020} & 2.00 & \hspace{-0.3cm}\footnotesize{\textpm 0.00} & 5000 & \multicolumn{3}{c}{$\{ 5000 \}$} \\ 

	&&&&&&&&&&&&&&&\\[-0.35cm]\hline&&&&&&&&&&&&&&&\\[-0.25cm] 
	\multirow{4}{*}{ \begin{turn}{90} RC-SVM  \end{turn}} &
	\multirow{4}{*}{ \hspace{-0.3cm} \begin{turn}{90} \# ws = 3  \end{turn}}
	  & \phantom{0}1\,000 \footnotesize{(10\,000)} & 100 & 0.32 & 0.14 & 0.0168 & \hspace{-0.3cm}\footnotesize{\textpm 0.0004} & 0.0555 & \hspace{-0.3cm}\footnotesize{\textpm 0.0040} & 3.00 & \hspace{-0.3cm}\footnotesize{\textpm 0.00} & 333 & $[ 333$ & \hspace{-0.5cm}, & \hspace{-0.5cm}$334 ]$ \\ 
 && \phantom{0}2\,500 \footnotesize{(10\,000)} & 100 & 0.88 & 0.17 & 0.0169 & \hspace{-0.3cm}\footnotesize{\textpm 0.0004} & 0.0566 & \hspace{-0.3cm}\footnotesize{\textpm 0.0033} & 3.00 & \hspace{-0.3cm}\footnotesize{\textpm 0.00} & 833 & $[ 833$ & \hspace{-0.5cm}, & \hspace{-0.5cm}$834 ]$ \\ 
 && \phantom{0}5\,000 \footnotesize{(10\,000)} & 100 & 2.71 & 0.28 & 0.0165 & \hspace{-0.3cm}\footnotesize{\textpm 0.0002} & 0.0534 & \hspace{-0.3cm}\footnotesize{\textpm 0.0022} & 3.00 & \hspace{-0.3cm}\footnotesize{\textpm 0.00} & 1667 & $[ 1666$ & \hspace{-0.5cm}, & \hspace{-0.5cm}$1667 ]$ \\ 
 && 10\,000 \footnotesize{(10\,000)} & 100 & 10.99 & 0.45 & 0.0163 & \hspace{-0.3cm}\footnotesize{\textpm 0.0002} & 0.0515 & \hspace{-0.3cm}\footnotesize{\textpm 0.0018} & 3.00 & \hspace{-0.3cm}\footnotesize{\textpm 0.00} & 3333 & $[ 3333$ & \hspace{-0.5cm}, & \hspace{-0.5cm}$3334 ]$ \\ 

	&&&&&&&&&&&&&&&\\[-0.35cm]\hline&&&&&&&&&&&&&&&\\[-0.25cm] 
	\multirow{4}{*}{ \begin{turn}{90} RC-SVM  \end{turn}} &
	\multirow{4}{*}{ \hspace{-0.3cm} \begin{turn}{90} \# ws = 4  \end{turn}}
	  & \phantom{0}1\,000 \footnotesize{(10\,000)} & 100 & 0.39 & 0.17 & 0.0166 & \hspace{-0.3cm}\footnotesize{\textpm 0.0004} & 0.0544 & \hspace{-0.3cm}\footnotesize{\textpm 0.0039} & 4.00 & \hspace{-0.3cm}\footnotesize{\textpm 0.00} & 250 & \multicolumn{3}{c}{$\{ 250 \}$} \\ 
 && \phantom{0}2\,500 \footnotesize{(10\,000)} & 100 & 0.81 & 0.21 & 0.0166 & \hspace{-0.3cm}\footnotesize{\textpm 0.0003} & 0.0537 & \hspace{-0.3cm}\footnotesize{\textpm 0.0032} & 4.00 & \hspace{-0.3cm}\footnotesize{\textpm 0.00} & 625 & \multicolumn{3}{c}{$\{ 625 \}$} \\ 
 && \phantom{0}5\,000 \footnotesize{(10\,000)} & 100 & 2.23 & 0.31 & 0.0165 & \hspace{-0.3cm}\footnotesize{\textpm 0.0003} & 0.0532 & \hspace{-0.3cm}\footnotesize{\textpm 0.0027} & 4.00 & \hspace{-0.3cm}\footnotesize{\textpm 0.00} & 1250 & \multicolumn{3}{c}{$\{ 1250 \}$} \\ 
 && 10\,000 \footnotesize{(10\,000)} & 100 & 8.16 & 0.49 & 0.0163 & \hspace{-0.3cm}\footnotesize{\textpm 0.0002} & 0.0516 & \hspace{-0.3cm}\footnotesize{\textpm 0.0020} & 4.00 & \hspace{-0.3cm}\footnotesize{\textpm 0.00} & 2500 & \multicolumn{3}{c}{$\{ 2500 \}$} \\ 

	&&&&&&&&&&&&&&&\\[-0.35cm]\hline&&&&&&&&&&&&&&&\\[-0.25cm] 
	\multirow{4}{*}{ \begin{turn}{90} RC-SVM  \end{turn}} &
	\multirow{4}{*}{ \hspace{-0.3cm} \begin{turn}{90} \# ws = 5  \end{turn}}
	  & \phantom{0}1\,000 \footnotesize{(10\,000)} & 100 & 0.46 & 0.19 & 0.0167 & \hspace{-0.3cm}\footnotesize{\textpm 0.0004} & 0.0553 & \hspace{-0.3cm}\footnotesize{\textpm 0.0041} & 5.00 & \hspace{-0.3cm}\footnotesize{\textpm 0.00} & 200 & \multicolumn{3}{c}{$\{ 200 \}$} \\ 
 && \phantom{0}2\,500 \footnotesize{(10\,000)} & 100 & 0.78 & 0.25 & 0.0165 & \hspace{-0.3cm}\footnotesize{\textpm 0.0004} & 0.0532 & \hspace{-0.3cm}\footnotesize{\textpm 0.0035} & 5.00 & \hspace{-0.3cm}\footnotesize{\textpm 0.00} & 500 & \multicolumn{3}{c}{$\{ 500 \}$} \\ 
 && \phantom{0}5\,000 \footnotesize{(10\,000)} & 100 & 1.94 & 0.32 & 0.0164 & \hspace{-0.3cm}\footnotesize{\textpm 0.0003} & 0.0524 & \hspace{-0.3cm}\footnotesize{\textpm 0.0029} & 5.00 & \hspace{-0.3cm}\footnotesize{\textpm 0.00} & 1000 & \multicolumn{3}{c}{$\{ 1000 \}$} \\ 
 && 10\,000 \footnotesize{(10\,000)} & 100 & 6.51 & 0.51 & 0.0164 & \hspace{-0.3cm}\footnotesize{\textpm 0.0002} & 0.0521 & \hspace{-0.3cm}\footnotesize{\textpm 0.0022} & 5.00 & \hspace{-0.3cm}\footnotesize{\textpm 0.00} & 2000 & \multicolumn{3}{c}{$\{ 2000 \}$} \\ 

	&&&&&&&&&&&&&&&\\[-0.35cm]\hline&&&&&&&&&&&&&&&\\[-0.25cm] 
	\multirow{4}{*}{ \begin{turn}{90} RC-SVM  \end{turn}} &
	\multirow{4}{*}{ \hspace{-0.3cm} \begin{turn}{90} \# ws = 6  \end{turn}}
	  & \phantom{0}1\,000 \footnotesize{(10\,000)} & 100 & 0.53 & 0.21 & 0.0168 & \hspace{-0.3cm}\footnotesize{\textpm 0.0004} & 0.0556 & \hspace{-0.3cm}\footnotesize{\textpm 0.0031} & 6.00 & \hspace{-0.3cm}\footnotesize{\textpm 0.00} & 167 & $[ 166$ & \hspace{-0.5cm}, & \hspace{-0.5cm}$167 ]$ \\ 
 && \phantom{0}2\,500 \footnotesize{(10\,000)} & 100 & 0.76 & 0.28 & 0.0164 & \hspace{-0.3cm}\footnotesize{\textpm 0.0003} & 0.0520 & \hspace{-0.3cm}\footnotesize{\textpm 0.0029} & 6.00 & \hspace{-0.3cm}\footnotesize{\textpm 0.00} & 417 & $[ 416$ & \hspace{-0.5cm}, & \hspace{-0.5cm}$417 ]$ \\ 
 && \phantom{0}5\,000 \footnotesize{(10\,000)} & 100 & 1.77 & 0.37 & 0.0164 & \hspace{-0.3cm}\footnotesize{\textpm 0.0003} & 0.0518 & \hspace{-0.3cm}\footnotesize{\textpm 0.0027} & 6.00 & \hspace{-0.3cm}\footnotesize{\textpm 0.00} & 833 & $[ 833$ & \hspace{-0.5cm}, & \hspace{-0.5cm}$834 ]$ \\ 
 && 10\,000 \footnotesize{(10\,000)} & 100 & 5.49 & 0.54 & 0.0165 & \hspace{-0.3cm}\footnotesize{\textpm 0.0002} & 0.0530 & \hspace{-0.3cm}\footnotesize{\textpm 0.0019} & 6.00 & \hspace{-0.3cm}\footnotesize{\textpm 0.00} & 1667 & $[ 1666$ & \hspace{-0.5cm}, & \hspace{-0.5cm}$1667 ]$ \\ 

	&&&&&&&&&&&&&&&\\[-0.35cm]\hline&&&&&&&&&&&&&&&\\[-0.25cm] 
	\multirow{4}{*}{ \begin{turn}{90} RC-SVM  \end{turn}} &
	\multirow{4}{*}{ \hspace{-0.3cm} \begin{turn}{90} \# ws = 10  \end{turn}}
	  & \phantom{0}1\,000 \footnotesize{(10\,000)} & 100 & 0.85 & 0.27 & 0.0171 & \hspace{-0.3cm}\footnotesize{\textpm 0.0003} & 0.0585 & \hspace{-0.3cm}\footnotesize{\textpm 0.0027} & 10.00 & \hspace{-0.3cm}\footnotesize{\textpm 0.00} & 100 & \multicolumn{3}{c}{$\{ 100 \}$} \\ 
 && \phantom{0}2\,500 \footnotesize{(10\,000)} & 100 & 0.98 & 0.40 & 0.0160 & \hspace{-0.3cm}\footnotesize{\textpm 0.0003} & 0.0487 & \hspace{-0.3cm}\footnotesize{\textpm 0.0027} & 10.00 & \hspace{-0.3cm}\footnotesize{\textpm 0.00} & 250 & \multicolumn{3}{c}{$\{ 250 \}$} \\ 
 && \phantom{0}5\,000 \footnotesize{(10\,000)} & 100 & 1.56 & 0.52 & 0.0160 & \hspace{-0.3cm}\footnotesize{\textpm 0.0002} & 0.0478 & \hspace{-0.3cm}\footnotesize{\textpm 0.0023} & 10.00 & \hspace{-0.3cm}\footnotesize{\textpm 0.00} & 500 & \multicolumn{3}{c}{$\{ 500 \}$} \\ 
 && 10\,000 \footnotesize{(10\,000)} & 100 & 3.91 & 0.63 & 0.0164 & \hspace{-0.3cm}\footnotesize{\textpm 0.0002} & 0.0524 & \hspace{-0.3cm}\footnotesize{\textpm 0.0017} & 10.00 & \hspace{-0.3cm}\footnotesize{\textpm 0.00} & 1000 & \multicolumn{3}{c}{$\{ 1000 \}$} \\ 

	&&&&&&&&&&&&&&&\\[-0.35cm]\hline&&&&&&&&&&&&&&&\\[-0.25cm] 
	\multirow{4}{*}{ \begin{turn}{90} RC-SVM  \end{turn}} &
	\multirow{4}{*}{ \hspace{-0.3cm} \begin{turn}{90} \# ws = 20  \end{turn}}
	  & \phantom{0}1\,000 \footnotesize{(10\,000)} & 100 & 1.83 & 0.37 & 0.0196 & \hspace{-0.3cm}\footnotesize{\textpm 0.0007} & 0.0774 & \hspace{-0.3cm}\footnotesize{\textpm 0.0042} & 20.00 & \hspace{-0.3cm}\footnotesize{\textpm 0.00} & 50 & \multicolumn{3}{c}{$\{ 50 \}$} \\ 
 && \phantom{0}2\,500 \footnotesize{(10\,000)} & 100 & 1.97 & 0.56 & 0.0163 & \hspace{-0.3cm}\footnotesize{\textpm 0.0002} & 0.0512 & \hspace{-0.3cm}\footnotesize{\textpm 0.0020} & 20.00 & \hspace{-0.3cm}\footnotesize{\textpm 0.00} & 125 & \multicolumn{3}{c}{$\{ 125 \}$} \\ 
 && \phantom{0}5\,000 \footnotesize{(10\,000)} & 100 & 2.20 & 0.79 & 0.0156 & \hspace{-0.3cm}\footnotesize{\textpm 0.0002} & 0.0446 & \hspace{-0.3cm}\footnotesize{\textpm 0.0018} & 20.00 & \hspace{-0.3cm}\footnotesize{\textpm 0.00} & 250 & \multicolumn{3}{c}{$\{ 250 \}$} \\ 
 && 10\,000 \footnotesize{(10\,000)} & 100 & 3.19 & 0.99 & 0.0160 & \hspace{-0.3cm}\footnotesize{\textpm 0.0002} & 0.0488 & \hspace{-0.3cm}\footnotesize{\textpm 0.0018} & 20.00 & \hspace{-0.3cm}\footnotesize{\textpm 0.00} & 500 & \multicolumn{3}{c}{$\{ 500 \}$} \\ 

      \end{tabular}
    } 
    \caption{Experimental results relating to the artificial data of Type \RM{2}}
    \label{tab::experiments::art_data_2_2}
  \end{minipage} 
\end{sidewaystable}

\begin{sidewaystable}[t!]
  \centering
  \begin{minipage}[t]{0.49\linewidth} \vspace{0pt}
    \resizebox{\textwidth}{!}{
        \begin{tabular}{cc||c|c||S[tabnumalign=centre]S[tabnumalign=centre]rlrlrlS[tabnumalign=centre]rcl}
	&& data set sizes & runs & \multicolumn{1}{c}{train time} & \multicolumn{1}{c}{test time} & \multicolumn{2}{c}{test error} & \multicolumn{2}{c}{$L_2$-error} & \multicolumn{2}{c}{\# of ws} & \multicolumn{1}{c}{ws size: median} & \multicolumn{3}{c}{ws size: range}\\ 
	&&&&&&&&&&&&&&&\\[-0.35cm]\hline\hline&&&&&&&&&&&&&&&\\[-0.25cm]  
	\multirow{4}{*}{ \begin{turn}{90} LS-SVM  \end{turn}}
	&  & \phantom{0}1\,000 \footnotesize{(10\,000)} & 100 & 0.36 & 0.05 & 0.0588 & \hspace{-0.3cm}\footnotesize{\textpm 0.0009} & 0.0799 & \hspace{-0.3cm}\footnotesize{\textpm 0.0056} & 1.00 &  & \\ 
 && \phantom{0}2\,500 \footnotesize{(10\,000)} & 100 & 1.82 & 0.07 & 0.0572 & \hspace{-0.3cm}\footnotesize{\textpm 0.0005} & 0.0677 & \hspace{-0.3cm}\footnotesize{\textpm 0.0039} & 1.00 &  & \\ 
 && \phantom{0}5\,000 \footnotesize{(10\,000)} & 100 & 7.47 & 0.11 & 0.0580 & \hspace{-0.3cm}\footnotesize{\textpm 0.0003} & 0.0745 & \hspace{-0.3cm}\footnotesize{\textpm 0.0019} & 1.00 &  & \\ 
 && 10\,000 \footnotesize{(10\,000)} & 100 & 29.56 & 0.24 & 0.0559 & \hspace{-0.3cm}\footnotesize{\textpm 0.0003} & 0.0563 & \hspace{-0.3cm}\footnotesize{\textpm 0.0028} & 1.00 &  & \\ 
 
	\noalign{\bigskip}\noalign{\bigskip}
	&& data set sizes & runs & \multicolumn{1}{c}{train time} & \multicolumn{1}{c}{test time} & \multicolumn{2}{c}{test error} & \multicolumn{2}{c}{$L_2$-error} & \multicolumn{2}{c}{\# of ws} & \multicolumn{1}{c}{ws size: median} & \multicolumn{3}{c}{ws size: range}\\ 
	&&&&&&&&&&&&&&&\\[-0.35cm]\hline\hline&&&&&&&&&&&&&&&\\[-0.25cm] 
	\multirow{4}{*}{ \begin{turn}{90} VP-SVM  \end{turn}} &
	\multirow{4}{*}{ \hspace{-0.3cm} \begin{turn}{90} $r$ =  0.1  \end{turn}}
	  & \phantom{0}1\,000 \footnotesize{(10\,000)} & 100 & 1.52 & 0.29 & 0.0580 & \hspace{-0.3cm}\footnotesize{\textpm 0.0009} & 0.0731 & \hspace{-0.3cm}\footnotesize{\textpm 0.0057} & 15.55 & \hspace{-0.3cm}\footnotesize{\textpm 2.36} & 64 & $[ 26$ & \hspace{-0.5cm}, & \hspace{-0.5cm}$109 ]$ \\ 
 && \phantom{0}2\,500 \footnotesize{(10\,000)} & 100 & 1.61 & 0.35 & 0.0553 & \hspace{-0.3cm}\footnotesize{\textpm 0.0004} & 0.0493 & \hspace{-0.3cm}\footnotesize{\textpm 0.0038} & 15.99 & \hspace{-0.3cm}\footnotesize{\textpm 2.39} & 156 & $[ 56$ & \hspace{-0.5cm}, & \hspace{-0.5cm}$279 ]$ \\ 
 && \phantom{0}5\,000 \footnotesize{(10\,000)} & 100 & 1.61 & 0.42 & 0.0539 & \hspace{-0.3cm}\footnotesize{\textpm 0.0002} & 0.0347 & \hspace{-0.3cm}\footnotesize{\textpm 0.0024} & 15.80 & \hspace{-0.3cm}\footnotesize{\textpm 2.45} & 316 & $[ 116$ & \hspace{-0.5cm}, & \hspace{-0.5cm}$531 ]$ \\ 
 && 10\,000 \footnotesize{(10\,000)} & 100 & 3.14 & 0.58 & 0.0534 & \hspace{-0.3cm}\footnotesize{\textpm 0.0001} & 0.0257 & \hspace{-0.3cm}\footnotesize{\textpm 0.0016} & 15.73 & \hspace{-0.3cm}\footnotesize{\textpm 2.58} & 636 & $[ 236$ & \hspace{-0.5cm}, & \hspace{-0.5cm}$1068 ]$ \\ 

	&&&&&&&&&&&&&&&\\[-0.35cm]\hline&&&&&&&&&&&&&&&\\[-0.25cm] 
	\multirow{4}{*}{ \begin{turn}{90} VP-SVM  \end{turn}} &
	\multirow{4}{*}{ \hspace{-0.3cm} \begin{turn}{90} $r$ =  0.25  \end{turn}}
	  & \phantom{0}1\,000 \footnotesize{(10\,000)} & 100 & 0.53 & 0.16 & 0.0563 & \hspace{-0.3cm}\footnotesize{\textpm 0.0007} & 0.0600 & \hspace{-0.3cm}\footnotesize{\textpm 0.0057} & 6.18 & \hspace{-0.3cm}\footnotesize{\textpm 0.87} & 162 & $[ 56$ & \hspace{-0.5cm}, & \hspace{-0.5cm}$259 ]$ \\ 
 && \phantom{0}2\,500 \footnotesize{(10\,000)} & 100 & 0.77 & 0.19 & 0.0547 & \hspace{-0.3cm}\footnotesize{\textpm 0.0003} & 0.0429 & \hspace{-0.3cm}\footnotesize{\textpm 0.0034} & 6.26 & \hspace{-0.3cm}\footnotesize{\textpm 0.81} & 399 & $[ 150$ & \hspace{-0.5cm}, & \hspace{-0.5cm}$658 ]$ \\ 
 && \phantom{0}5\,000 \footnotesize{(10\,000)} & 100 & 1.65 & 0.26 & 0.0537 & \hspace{-0.3cm}\footnotesize{\textpm 0.0001} & 0.0315 & \hspace{-0.3cm}\footnotesize{\textpm 0.0024} & 6.27 & \hspace{-0.3cm}\footnotesize{\textpm 0.81} & 797 & $[ 321$ & \hspace{-0.5cm}, & \hspace{-0.5cm}$1270 ]$ \\ 
 && 10\,000 \footnotesize{(10\,000)} & 100 & 5.11 & 0.39 & 0.0534 & \hspace{-0.3cm}\footnotesize{\textpm 0.0001} & 0.0244 & \hspace{-0.3cm}\footnotesize{\textpm 0.0021} & 6.34 & \hspace{-0.3cm}\footnotesize{\textpm 0.79} & 1577 & $[ 597$ & \hspace{-0.5cm}, & \hspace{-0.5cm}$2534 ]$ \\ 

	&&&&&&&&&&&&&&&\\[-0.35cm]\hline&&&&&&&&&&&&&&&\\[-0.25cm] 
	\multirow{4}{*}{ \begin{turn}{90} VP-SVM  \end{turn}} &
	\multirow{4}{*}{ \hspace{-0.3cm} \begin{turn}{90} $r$ =  0.5  \end{turn}}
	  & \phantom{0}1\,000 \footnotesize{(10\,000)} & 100 & 0.34 & 0.11 & 0.0564 & \hspace{-0.3cm}\footnotesize{\textpm 0.0007} & 0.0620 & \hspace{-0.3cm}\footnotesize{\textpm 0.0057} & 3.48 & \hspace{-0.3cm}\footnotesize{\textpm 0.50} & 287 & $[ 126$ & \hspace{-0.5cm}, & \hspace{-0.5cm}$501 ]$ \\ 
 && \phantom{0}2\,500 \footnotesize{(10\,000)} & 100 & 0.81 & 0.14 & 0.0551 & \hspace{-0.3cm}\footnotesize{\textpm 0.0004} & 0.0478 & \hspace{-0.3cm}\footnotesize{\textpm 0.0047} & 3.49 & \hspace{-0.3cm}\footnotesize{\textpm 0.50} & 716 & $[ 297$ & \hspace{-0.5cm}, & \hspace{-0.5cm}$1244 ]$ \\ 
 && \phantom{0}5\,000 \footnotesize{(10\,000)} & 100 & 2.44 & 0.22 & 0.0543 & \hspace{-0.3cm}\footnotesize{\textpm 0.0004} & 0.0409 & \hspace{-0.3cm}\footnotesize{\textpm 0.0048} & 3.50 & \hspace{-0.3cm}\footnotesize{\textpm 0.50} & 1429 & $[ 641$ & \hspace{-0.5cm}, & \hspace{-0.5cm}$2517 ]$ \\ 
 && 10\,000 \footnotesize{(10\,000)} & 100 & 9.40 & 0.30 & 0.0538 & \hspace{-0.3cm}\footnotesize{\textpm 0.0004} & 0.0322 & \hspace{-0.3cm}\footnotesize{\textpm 0.0054} & 3.54 & \hspace{-0.3cm}\footnotesize{\textpm 0.50} & 2825 & $[ 1209$ & \hspace{-0.5cm}, & \hspace{-0.5cm}$4993 ]$ \\ 

	&&&&&&&&&&&&&&&\\[-0.35cm]\hline&&&&&&&&&&&&&&&\\[-0.25cm] 
	\multirow{4}{*}{ \begin{turn}{90} VP-SVM  \end{turn}} &
	\multirow{4}{*}{ \hspace{-0.3cm} \begin{turn}{90} $r$ =  1  \end{turn}}
	  & \phantom{0}1\,000 \footnotesize{(10\,000)} & 100 & 0.28 & 0.08 & 0.0579 & \hspace{-0.3cm}\footnotesize{\textpm 0.0009} & 0.0738 & \hspace{-0.3cm}\footnotesize{\textpm 0.0062} & 2.00 & \hspace{-0.3cm}\footnotesize{\textpm 0.00} & 500 & $[ 248$ & \hspace{-0.5cm}, & \hspace{-0.5cm}$752 ]$ \\ 
 && \phantom{0}2\,500 \footnotesize{(10\,000)} & 100 & 1.04 & 0.11 & 0.0565 & \hspace{-0.3cm}\footnotesize{\textpm 0.0007} & 0.0617 & \hspace{-0.3cm}\footnotesize{\textpm 0.0059} & 2.00 & \hspace{-0.3cm}\footnotesize{\textpm 0.00} & 1250 & $[ 625$ & \hspace{-0.5cm}, & \hspace{-0.5cm}$1875 ]$ \\ 
 && \phantom{0}5\,000 \footnotesize{(10\,000)} & 100 & 3.92 & 0.15 & 0.0564 & \hspace{-0.3cm}\footnotesize{\textpm 0.0005} & 0.0622 & \hspace{-0.3cm}\footnotesize{\textpm 0.0044} & 2.00 & \hspace{-0.3cm}\footnotesize{\textpm 0.00} & 2500 & $[ 1281$ & \hspace{-0.5cm}, & \hspace{-0.5cm}$3719 ]$ \\ 
 && 10\,000 \footnotesize{(10\,000)} & 100 & 15.93 & 0.23 & 0.0555 & \hspace{-0.3cm}\footnotesize{\textpm 0.0004} & 0.0520 & \hspace{-0.3cm}\footnotesize{\textpm 0.0035} & 2.00 & \hspace{-0.3cm}\footnotesize{\textpm 0.00} & 5000 & $[ 2404$ & \hspace{-0.5cm}, & \hspace{-0.5cm}$7596 ]$ \\ 

	&&&&&&&&&&&&&&&\\[-0.35cm]\hline&&&&&&&&&&&&&&&\\[-0.25cm] 
	\multirow{4}{*}{ \begin{turn}{90} VP-SVM  \end{turn}} &
	\multirow{4}{*}{ \hspace{-0.3cm} \begin{turn}{90} $r$ =  2  \end{turn}}
	  & \phantom{0}1\,000 \footnotesize{(10\,000)} & 100 & 0.35 & 0.06 & 0.0588 & \hspace{-0.3cm}\footnotesize{\textpm 0.0009} & 0.0799 & \hspace{-0.3cm}\footnotesize{\textpm 0.0057} & 1.00 & \hspace{-0.3cm}\footnotesize{\textpm 0.00} & 1000 & \multicolumn{3}{c}{$\{ 1000 \}$} \\ 
 && \phantom{0}2\,500 \footnotesize{(10\,000)} & 100 & 1.84 & 0.08 & 0.0572 & \hspace{-0.3cm}\footnotesize{\textpm 0.0005} & 0.0677 & \hspace{-0.3cm}\footnotesize{\textpm 0.0039} & 1.00 & \hspace{-0.3cm}\footnotesize{\textpm 0.00} & 2500 & \multicolumn{3}{c}{$\{ 2500 \}$} \\ 
 && \phantom{0}5\,000 \footnotesize{(10\,000)} & 100 & 7.51 & 0.12 & 0.0580 & \hspace{-0.3cm}\footnotesize{\textpm 0.0003} & 0.0745 & \hspace{-0.3cm}\footnotesize{\textpm 0.0019} & 1.00 & \hspace{-0.3cm}\footnotesize{\textpm 0.00} & 5000 & \multicolumn{3}{c}{$\{ 5000 \}$} \\ 
 && 10\,000 \footnotesize{(10\,000)} & 100 & 29.75 & 0.25 & 0.0559 & \hspace{-0.3cm}\footnotesize{\textpm 0.0003} & 0.0563 & \hspace{-0.3cm}\footnotesize{\textpm 0.0028} & 1.00 & \hspace{-0.3cm}\footnotesize{\textpm 0.00} & 10000 & \multicolumn{3}{c}{$\{ 10000 \}$} \\ 

	\noalign{\bigskip}\noalign{\bigskip}
	&& data set sizes & runs & \multicolumn{1}{c}{train time} & \multicolumn{1}{c}{test time} & \multicolumn{2}{c}{test error} & \multicolumn{2}{c}{$L_2$-error} & \multicolumn{2}{c}{\# of ws} & \multicolumn{1}{c}{ws size: median} & \multicolumn{3}{c}{ws size: range}\\ 
	&&&&&&&&&&&&&&&\\[-0.35cm]\hline\hline&&&&&&&&&&&&&&&\\[-0.25cm] 
	\multirow{4}{*}{ \begin{turn}{90} RC-SVM  \end{turn}} &
	\multirow{4}{*}{ \hspace{-0.3cm} \begin{turn}{90} \# ws = 1  \end{turn}}
	  & \phantom{0}1\,000 \footnotesize{(10\,000)} & 100 & 0.35 & 0.07 & 0.0588 & \hspace{-0.3cm}\footnotesize{\textpm 0.0009} & 0.0801 & \hspace{-0.3cm}\footnotesize{\textpm 0.0058} & 1.00 & \hspace{-0.3cm}\footnotesize{\textpm 0.00} & 1000 & \multicolumn{3}{c}{$\{ 1000 \}$} \\ 
 && \phantom{0}2\,500 \footnotesize{(10\,000)} & 100 & 1.85 & 0.08 & 0.0572 & \hspace{-0.3cm}\footnotesize{\textpm 0.0005} & 0.0678 & \hspace{-0.3cm}\footnotesize{\textpm 0.0039} & 1.00 & \hspace{-0.3cm}\footnotesize{\textpm 0.00} & 2500 & \multicolumn{3}{c}{$\{ 2500 \}$} \\ 
 && \phantom{0}5\,000 \footnotesize{(10\,000)} & 100 & 7.73 & 0.12 & 0.0580 & \hspace{-0.3cm}\footnotesize{\textpm 0.0003} & 0.0745 & \hspace{-0.3cm}\footnotesize{\textpm 0.0019} & 1.00 & \hspace{-0.3cm}\footnotesize{\textpm 0.00} & 5000 & \multicolumn{3}{c}{$\{ 5000 \}$} \\ 
 && 10\,000 \footnotesize{(10\,000)} & 100 & 30.24 & 0.27 & 0.0559 & \hspace{-0.3cm}\footnotesize{\textpm 0.0003} & 0.0563 & \hspace{-0.3cm}\footnotesize{\textpm 0.0028} & 1.00 & \hspace{-0.3cm}\footnotesize{\textpm 0.00} & 10000 & \multicolumn{3}{c}{$\{ 10000 \}$} \\ 

	&&&&&&&&&&&&&&&\\[-0.35cm]\hline&&&&&&&&&&&&&&&\\[-0.25cm] 
	\multirow{4}{*}{ \begin{turn}{90} RC-SVM  \end{turn}} &
	\multirow{4}{*}{ \hspace{-0.3cm} \begin{turn}{90} \# ws = 2  \end{turn}}
	  & \phantom{0}1\,000 \footnotesize{(10\,000)} & 100 & 0.30 & 0.08 & 0.0596 & \hspace{-0.3cm}\footnotesize{\textpm 0.0009} & 0.0855 & \hspace{-0.3cm}\footnotesize{\textpm 0.0052} & 2.00 & \hspace{-0.3cm}\footnotesize{\textpm 0.00} & 500 & \multicolumn{3}{c}{$\{ 500 \}$} \\ 
 && \phantom{0}2\,500 \footnotesize{(10\,000)} & 100 & 1.01 & 0.11 & 0.0579 & \hspace{-0.3cm}\footnotesize{\textpm 0.0005} & 0.0736 & \hspace{-0.3cm}\footnotesize{\textpm 0.0033} & 2.00 & \hspace{-0.3cm}\footnotesize{\textpm 0.00} & 1250 & \multicolumn{3}{c}{$\{ 1250 \}$} \\ 
 && \phantom{0}5\,000 \footnotesize{(10\,000)} & 100 & 3.81 & 0.15 & 0.0573 & \hspace{-0.3cm}\footnotesize{\textpm 0.0005} & 0.0699 & \hspace{-0.3cm}\footnotesize{\textpm 0.0038} & 2.00 & \hspace{-0.3cm}\footnotesize{\textpm 0.00} & 2500 & \multicolumn{3}{c}{$\{ 2500 \}$} \\ 
 && 10\,000 \footnotesize{(10\,000)} & 100 & 15.22 & 0.24 & 0.0562 & \hspace{-0.3cm}\footnotesize{\textpm 0.0004} & 0.0598 & \hspace{-0.3cm}\footnotesize{\textpm 0.0035} & 2.00 & \hspace{-0.3cm}\footnotesize{\textpm 0.00} & 5000 & \multicolumn{3}{c}{$\{ 5000 \}$} \\ 

	&&&&&&&&&&&&&&&\\[-0.35cm]\hline&&&&&&&&&&&&&&&\\[-0.25cm] 
	\multirow{4}{*}{ \begin{turn}{90} RC-SVM  \end{turn}} &
	\multirow{4}{*}{ \hspace{-0.3cm} \begin{turn}{90} \# ws = 3  \end{turn}}
	  & \phantom{0}1\,000 \footnotesize{(10\,000)} & 100 & 0.32 & 0.12 & 0.0591 & \hspace{-0.3cm}\footnotesize{\textpm 0.0010} & 0.0825 & \hspace{-0.3cm}\footnotesize{\textpm 0.0067} & 3.00 & \hspace{-0.3cm}\footnotesize{\textpm 0.00} & 333 & $[ 333$ & \hspace{-0.5cm}, & \hspace{-0.5cm}$334 ]$ \\ 
 && \phantom{0}2\,500 \footnotesize{(10\,000)} & 100 & 0.82 & 0.14 & 0.0585 & \hspace{-0.3cm}\footnotesize{\textpm 0.0006} & 0.0776 & \hspace{-0.3cm}\footnotesize{\textpm 0.0038} & 3.00 & \hspace{-0.3cm}\footnotesize{\textpm 0.00} & 833 & $[ 833$ & \hspace{-0.5cm}, & \hspace{-0.5cm}$834 ]$ \\ 
 && \phantom{0}5\,000 \footnotesize{(10\,000)} & 100 & 2.52 & 0.18 & 0.0574 & \hspace{-0.3cm}\footnotesize{\textpm 0.0004} & 0.0710 & \hspace{-0.3cm}\footnotesize{\textpm 0.0031} & 3.00 & \hspace{-0.3cm}\footnotesize{\textpm 0.00} & 1667 & $[ 1666$ & \hspace{-0.5cm}, & \hspace{-0.5cm}$1667 ]$ \\ 
 && 10\,000 \footnotesize{(10\,000)} & 100 & 10.05 & 0.24 & 0.0564 & \hspace{-0.3cm}\footnotesize{\textpm 0.0004} & 0.0622 & \hspace{-0.3cm}\footnotesize{\textpm 0.0032} & 3.00 & \hspace{-0.3cm}\footnotesize{\textpm 0.00} & 3333 & $[ 3333$ & \hspace{-0.5cm}, & \hspace{-0.5cm}$3334 ]$ \\ 

	&&&&&&&&&&&&&&&\\[-0.35cm]\hline&&&&&&&&&&&&&&&\\[-0.25cm] 
	\multirow{4}{*}{ \begin{turn}{90} RC-SVM  \end{turn}} &
	\multirow{4}{*}{ \hspace{-0.3cm} \begin{turn}{90} \# ws = 4  \end{turn}}
	  & \phantom{0}1\,000 \footnotesize{(10\,000)} & 100 & 0.37 & 0.15 & 0.0589 & \hspace{-0.3cm}\footnotesize{\textpm 0.0011} & 0.0814 & \hspace{-0.3cm}\footnotesize{\textpm 0.0069} & 4.00 & \hspace{-0.3cm}\footnotesize{\textpm 0.00} & 250 & \multicolumn{3}{c}{$\{ 250 \}$} \\ 
 && \phantom{0}2\,500 \footnotesize{(10\,000)} & 100 & 0.77 & 0.16 & 0.0587 & \hspace{-0.3cm}\footnotesize{\textpm 0.0006} & 0.0793 & \hspace{-0.3cm}\footnotesize{\textpm 0.0038} & 4.00 & \hspace{-0.3cm}\footnotesize{\textpm 0.00} & 625 & \multicolumn{3}{c}{$\{ 625 \}$} \\ 
 && \phantom{0}5\,000 \footnotesize{(10\,000)} & 100 & 2.07 & 0.20 & 0.0579 & \hspace{-0.3cm}\footnotesize{\textpm 0.0004} & 0.0747 & \hspace{-0.3cm}\footnotesize{\textpm 0.0025} & 4.00 & \hspace{-0.3cm}\footnotesize{\textpm 0.00} & 1250 & \multicolumn{3}{c}{$\{ 1250 \}$} \\ 
 && 10\,000 \footnotesize{(10\,000)} & 100 & 7.41 & 0.27 & 0.0563 & \hspace{-0.3cm}\footnotesize{\textpm 0.0004} & 0.0616 & \hspace{-0.3cm}\footnotesize{\textpm 0.0036} & 4.00 & \hspace{-0.3cm}\footnotesize{\textpm 0.00} & 2500 & \multicolumn{3}{c}{$\{ 2500 \}$} \\ 

	&&&&&&&&&&&&&&&\\[-0.35cm]\hline&&&&&&&&&&&&&&&\\[-0.25cm] 
	\multirow{4}{*}{ \begin{turn}{90} RC-SVM  \end{turn}} &
	\multirow{4}{*}{ \hspace{-0.3cm} \begin{turn}{90} \# ws = 5  \end{turn}}
	  & \phantom{0}1\,000 \footnotesize{(10\,000)} & 100 & 0.46 & 0.17 & 0.0588 & \hspace{-0.3cm}\footnotesize{\textpm 0.0010} & 0.0811 & \hspace{-0.3cm}\footnotesize{\textpm 0.0064} & 5.00 & \hspace{-0.3cm}\footnotesize{\textpm 0.00} & 200 & \multicolumn{3}{c}{$\{ 200 \}$} \\ 
 && \phantom{0}2\,500 \footnotesize{(10\,000)} & 100 & 0.73 & 0.19 & 0.0586 & \hspace{-0.3cm}\footnotesize{\textpm 0.0007} & 0.0787 & \hspace{-0.3cm}\footnotesize{\textpm 0.0045} & 5.00 & \hspace{-0.3cm}\footnotesize{\textpm 0.00} & 500 & \multicolumn{3}{c}{$\{ 500 \}$} \\ 
 && \phantom{0}5\,000 \footnotesize{(10\,000)} & 100 & 1.87 & 0.24 & 0.0583 & \hspace{-0.3cm}\footnotesize{\textpm 0.0004} & 0.0775 & \hspace{-0.3cm}\footnotesize{\textpm 0.0027} & 5.00 & \hspace{-0.3cm}\footnotesize{\textpm 0.00} & 1000 & \multicolumn{3}{c}{$\{ 1000 \}$} \\ 
 && 10\,000 \footnotesize{(10\,000)} & 100 & 5.88 & 0.30 & 0.0563 & \hspace{-0.3cm}\footnotesize{\textpm 0.0003} & 0.0617 & \hspace{-0.3cm}\footnotesize{\textpm 0.0029} & 5.00 & \hspace{-0.3cm}\footnotesize{\textpm 0.00} & 2000 & \multicolumn{3}{c}{$\{ 2000 \}$} \\ 

	&&&&&&&&&&&&&&&\\[-0.35cm]\hline&&&&&&&&&&&&&&&\\[-0.25cm] 
	\multirow{4}{*}{ \begin{turn}{90} RC-SVM  \end{turn}} &
	\multirow{4}{*}{ \hspace{-0.3cm} \begin{turn}{90} \# ws = 6  \end{turn}}
	  & \phantom{0}1\,000 \footnotesize{(10\,000)} & 100 & 0.52 & 0.19 & 0.0586 & \hspace{-0.3cm}\footnotesize{\textpm 0.0010} & 0.0796 & \hspace{-0.3cm}\footnotesize{\textpm 0.0070} & 6.00 & \hspace{-0.3cm}\footnotesize{\textpm 0.00} & 167 & $[ 166$ & \hspace{-0.5cm}, & \hspace{-0.5cm}$167 ]$ \\ 
 && \phantom{0}2\,500 \footnotesize{(10\,000)} & 100 & 0.73 & 0.21 & 0.0586 & \hspace{-0.3cm}\footnotesize{\textpm 0.0006} & 0.0788 & \hspace{-0.3cm}\footnotesize{\textpm 0.0044} & 6.00 & \hspace{-0.3cm}\footnotesize{\textpm 0.00} & 417 & $[ 416$ & \hspace{-0.5cm}, & \hspace{-0.5cm}$417 ]$ \\ 
 && \phantom{0}5\,000 \footnotesize{(10\,000)} & 100 & 1.67 & 0.26 & 0.0586 & \hspace{-0.3cm}\footnotesize{\textpm 0.0005} & 0.0793 & \hspace{-0.3cm}\footnotesize{\textpm 0.0033} & 6.00 & \hspace{-0.3cm}\footnotesize{\textpm 0.00} & 833 & $[ 833$ & \hspace{-0.5cm}, & \hspace{-0.5cm}$834 ]$ \\ 
 && 10\,000 \footnotesize{(10\,000)} & 100 & 5.01 & 0.33 & 0.0565 & \hspace{-0.3cm}\footnotesize{\textpm 0.0003} & 0.0637 & \hspace{-0.3cm}\footnotesize{\textpm 0.0026} & 6.00 & \hspace{-0.3cm}\footnotesize{\textpm 0.00} & 1667 & $[ 1666$ & \hspace{-0.5cm}, & \hspace{-0.5cm}$1667 ]$ \\ 

	&&&&&&&&&&&&&&&\\[-0.35cm]\hline&&&&&&&&&&&&&&&\\[-0.25cm] 
	\multirow{4}{*}{ \begin{turn}{90} RC-SVM  \end{turn}} &
	\multirow{4}{*}{ \hspace{-0.3cm} \begin{turn}{90} \# ws = 10  \end{turn}}
	  & \phantom{0}1\,000 \footnotesize{(10\,000)} & 100 & 0.89 & 0.25 & 0.0593 & \hspace{-0.3cm}\footnotesize{\textpm 0.0011} & 0.0845 & \hspace{-0.3cm}\footnotesize{\textpm 0.0069} & 10.00 & \hspace{-0.3cm}\footnotesize{\textpm 0.00} & 100 & \multicolumn{3}{c}{$\{ 100 \}$} \\ 
 && \phantom{0}2\,500 \footnotesize{(10\,000)} & 100 & 0.99 & 0.33 & 0.0579 & \hspace{-0.3cm}\footnotesize{\textpm 0.0006} & 0.0744 & \hspace{-0.3cm}\footnotesize{\textpm 0.0043} & 10.00 & \hspace{-0.3cm}\footnotesize{\textpm 0.00} & 250 & \multicolumn{3}{c}{$\{ 250 \}$} \\ 
 && \phantom{0}5\,000 \footnotesize{(10\,000)} & 100 & 1.51 & 0.36 & 0.0590 & \hspace{-0.3cm}\footnotesize{\textpm 0.0005} & 0.0824 & \hspace{-0.3cm}\footnotesize{\textpm 0.0033} & 10.00 & \hspace{-0.3cm}\footnotesize{\textpm 0.00} & 500 & \multicolumn{3}{c}{$\{ 500 \}$} \\ 
 && 10\,000 \footnotesize{(10\,000)} & 100 & 3.59 & 0.44 & 0.0574 & \hspace{-0.3cm}\footnotesize{\textpm 0.0003} & 0.0705 & \hspace{-0.3cm}\footnotesize{\textpm 0.0024} & 10.00 & \hspace{-0.3cm}\footnotesize{\textpm 0.00} & 1000 & \multicolumn{3}{c}{$\{ 1000 \}$} \\ 

	&&&&&&&&&&&&&&&\\[-0.35cm]\hline&&&&&&&&&&&&&&&\\[-0.25cm] 
	\multirow{4}{*}{ \begin{turn}{90} RC-SVM  \end{turn}} &
	\multirow{4}{*}{ \hspace{-0.3cm} \begin{turn}{90} \# ws = 20  \end{turn}}
	  & \phantom{0}1\,000 \footnotesize{(10\,000)} & 100 & 1.90 & 0.38 & 0.0671 & \hspace{-0.3cm}\footnotesize{\textpm 0.0019} & 0.1238 & \hspace{-0.3cm}\footnotesize{\textpm 0.0082} & 20.00 & \hspace{-0.3cm}\footnotesize{\textpm 0.00} & 50 & \multicolumn{3}{c}{$\{ 50 \}$} \\ 
 && \phantom{0}2\,500 \footnotesize{(10\,000)} & 100 & 2.01 & 0.53 & 0.0578 & \hspace{-0.3cm}\footnotesize{\textpm 0.0007} & 0.0742 & \hspace{-0.3cm}\footnotesize{\textpm 0.0052} & 20.00 & \hspace{-0.3cm}\footnotesize{\textpm 0.00} & 125 & \multicolumn{3}{c}{$\{ 125 \}$} \\ 
 && \phantom{0}5\,000 \footnotesize{(10\,000)} & 100 & 2.19 & 0.64 & 0.0583 & \hspace{-0.3cm}\footnotesize{\textpm 0.0005} & 0.0779 & \hspace{-0.3cm}\footnotesize{\textpm 0.0031} & 20.00 & \hspace{-0.3cm}\footnotesize{\textpm 0.00} & 250 & \multicolumn{3}{c}{$\{ 250 \}$} \\ 
 && 10\,000 \footnotesize{(10\,000)} & 100 & 3.20 & 0.67 & 0.0582 & \hspace{-0.3cm}\footnotesize{\textpm 0.0003} & 0.0770 & \hspace{-0.3cm}\footnotesize{\textpm 0.0021} & 20.00 & \hspace{-0.3cm}\footnotesize{\textpm 0.00} & 500 & \multicolumn{3}{c}{$\{ 500 \}$} \\ 

      \end{tabular}
    } 
    \caption{Experimental results relating to the artificial data of Type \RM{3}}
    \label{tab::experiments::art_data_4_2}
  \end{minipage} 
  \hfill
  \begin{minipage}[t]{0.49\linewidth} \vspace{0pt}
    \resizebox{\textwidth}{!}{
        \begin{tabular}{cc||c|c||S[tabnumalign=centre]S[tabnumalign=centre]rlrlrlS[tabnumalign=centre]rcl}
	&& data set sizes & runs & \multicolumn{1}{c}{train time} & \multicolumn{1}{c}{test time} & \multicolumn{2}{c}{test error} & \multicolumn{2}{c}{$L_2$-error} & \multicolumn{2}{c}{\# of ws} & \multicolumn{1}{c}{ws size: median} & \multicolumn{3}{c}{ws size: range}\\ 
	&&&&&&&&&&&&&&&\\[-0.35cm]\hline\hline&&&&&&&&&&&&&&&\\[-0.25cm]  
	\multirow{4}{*}{ \begin{turn}{90} LS-SVM  \end{turn}}
	&  & \phantom{0}1\,000 \footnotesize{(10\,000)} & 100 & 0.61 & 0.07 & 0.0155 & \hspace{-0.3cm}\footnotesize{\textpm 0.0002} & 0.0848 & \hspace{-0.3cm}\footnotesize{\textpm 0.0012} & 1.00 &  & \\ 
 && \phantom{0}2\,500 \footnotesize{(10\,000)} & 100 & 3.28 & 0.16 & 0.0151 & \hspace{-0.3cm}\footnotesize{\textpm 0.0003} & 0.0821 & \hspace{-0.3cm}\footnotesize{\textpm 0.0016} & 1.00 &  & \\ 
 && \phantom{0}5\,000 \footnotesize{(10\,000)} & 100 & 11.89 & 0.63 & 0.0132 & \hspace{-0.3cm}\footnotesize{\textpm 0.0007} & 0.0694 & \hspace{-0.3cm}\footnotesize{\textpm 0.0049} & 1.00 &  & \\ 
 && 10\,000 \footnotesize{(10\,000)} & 100 & 45.94 & 0.58 & 0.0137 & \hspace{-0.3cm}\footnotesize{\textpm 0.0007} & 0.0730 & \hspace{-0.3cm}\footnotesize{\textpm 0.0046} & 1.00 &  & \\ 
 
	\noalign{\bigskip}\noalign{\bigskip}
	&& data set sizes & runs & \multicolumn{1}{c}{train time} & \multicolumn{1}{c}{test time} & \multicolumn{2}{c}{test error} & \multicolumn{2}{c}{$L_2$-error} & \multicolumn{2}{c}{\# of ws} & \multicolumn{1}{c}{ws size: median} & \multicolumn{3}{c}{ws size: range}\\ 
	&&&&&&&&&&&&&&&\\[-0.35cm]\hline\hline&&&&&&&&&&&&&&&\\[-0.25cm] 
	\multirow{4}{*}{ \begin{turn}{90} VP-SVM  \end{turn}} &
	\multirow{4}{*}{ \hspace{-0.3cm} \begin{turn}{90} $r$ =  0.25  \end{turn}}
	  & \phantom{0}1\,000 \footnotesize{(10\,000)} & 100 & 4.27 & 0.62 & 0.0172 & \hspace{-0.3cm}\footnotesize{\textpm 0.0005} & 0.0946 & \hspace{-0.3cm}\footnotesize{\textpm 0.0024} & 44.23 & \hspace{-0.3cm}\footnotesize{\textpm 2.19} & 23 & $[ 4$ & \hspace{-0.5cm}, & \hspace{-0.5cm}$54 ]$ \\ 
 && \phantom{0}2\,500 \footnotesize{(10\,000)} & 100 & 4.78 & 0.85 & 0.0142 & \hspace{-0.3cm}\footnotesize{\textpm 0.0002} & 0.0775 & \hspace{-0.3cm}\footnotesize{\textpm 0.0015} & 46.09 & \hspace{-0.3cm}\footnotesize{\textpm 2.58} & 54 & $[ 13$ & \hspace{-0.5cm}, & \hspace{-0.5cm}$108 ]$ \\ 
 && \phantom{0}5\,000 \footnotesize{(10\,000)} & 100 & 5.03 & 1.16 & 0.0129 & \hspace{-0.3cm}\footnotesize{\textpm 0.0002} & 0.0682 & \hspace{-0.3cm}\footnotesize{\textpm 0.0010} & 46.32 & \hspace{-0.3cm}\footnotesize{\textpm 2.61} & 108 & $[ 19$ & \hspace{-0.5cm}, & \hspace{-0.5cm}$244 ]$ \\ 
 && 10\,000 \footnotesize{(10\,000)} & 100 & 5.72 & 1.76 & 0.0119 & \hspace{-0.3cm}\footnotesize{\textpm 0.0001} & 0.0607 & \hspace{-0.3cm}\footnotesize{\textpm 0.0007} & 47.52 & \hspace{-0.3cm}\footnotesize{\textpm 2.72} & 210 & $[ 43$ & \hspace{-0.5cm}, & \hspace{-0.5cm}$416 ]$ \\ 

	&&&&&&&&&&&&&&&\\[-0.35cm]\hline&&&&&&&&&&&&&&&\\[-0.25cm] 
	\multirow{4}{*}{ \begin{turn}{90} VP-SVM  \end{turn}} &
	\multirow{4}{*}{ \hspace{-0.3cm} \begin{turn}{90} $r$ =  0.5  \end{turn}}
	  & \phantom{0}1\,000 \footnotesize{(10\,000)} & 100 & 1.45 & 0.30 & 0.0158 & \hspace{-0.3cm}\footnotesize{\textpm 0.0003} & 0.0867 & \hspace{-0.3cm}\footnotesize{\textpm 0.0019} & 14.50 & \hspace{-0.3cm}\footnotesize{\textpm 1.25} & 69 & $[ 18$ & \hspace{-0.5cm}, & \hspace{-0.5cm}$155 ]$ \\ 
 && \phantom{0}2\,500 \footnotesize{(10\,000)} & 100 & 1.62 & 0.49 & 0.0139 & \hspace{-0.3cm}\footnotesize{\textpm 0.0002} & 0.0752 & \hspace{-0.3cm}\footnotesize{\textpm 0.0013} & 15.22 & \hspace{-0.3cm}\footnotesize{\textpm 1.47} & 164 & $[ 40$ & \hspace{-0.5cm}, & \hspace{-0.5cm}$323 ]$ \\ 
 && \phantom{0}5\,000 \footnotesize{(10\,000)} & 100 & 2.17 & 0.78 & 0.0129 & \hspace{-0.3cm}\footnotesize{\textpm 0.0002} & 0.0683 & \hspace{-0.3cm}\footnotesize{\textpm 0.0011} & 15.00 & \hspace{-0.3cm}\footnotesize{\textpm 1.45} & 333 & $[ 78$ & \hspace{-0.5cm}, & \hspace{-0.5cm}$635 ]$ \\ 
 && 10\,000 \footnotesize{(10\,000)} & 100 & 4.52 & 1.24 & 0.0120 & \hspace{-0.3cm}\footnotesize{\textpm 0.0001} & 0.0611 & \hspace{-0.3cm}\footnotesize{\textpm 0.0009} & 15.72 & \hspace{-0.3cm}\footnotesize{\textpm 1.62} & 636 & $[ 144$ & \hspace{-0.5cm}, & \hspace{-0.5cm}$1261 ]$ \\ 

	&&&&&&&&&&&&&&&\\[-0.35cm]\hline&&&&&&&&&&&&&&&\\[-0.25cm] 
	\multirow{4}{*}{ \begin{turn}{90} VP-SVM  \end{turn}} &
	\multirow{4}{*}{ \hspace{-0.3cm} \begin{turn}{90} $r$ =  1  \end{turn}}
	  & \phantom{0}1\,000 \footnotesize{(10\,000)} & 100 & 0.49 & 0.18 & 0.0152 & \hspace{-0.3cm}\footnotesize{\textpm 0.0003} & 0.0833 & \hspace{-0.3cm}\footnotesize{\textpm 0.0017} & 5.16 & \hspace{-0.3cm}\footnotesize{\textpm 0.53} & 194 & $[ 69$ & \hspace{-0.5cm}, & \hspace{-0.5cm}$478 ]$ \\ 
 && \phantom{0}2\,500 \footnotesize{(10\,000)} & 100 & 0.99 & 0.32 & 0.0140 & \hspace{-0.3cm}\footnotesize{\textpm 0.0003} & 0.0753 & \hspace{-0.3cm}\footnotesize{\textpm 0.0019} & 5.40 & \hspace{-0.3cm}\footnotesize{\textpm 0.51} & 463 & $[ 156$ & \hspace{-0.5cm}, & \hspace{-0.5cm}$1260 ]$ \\ 
 && \phantom{0}5\,000 \footnotesize{(10\,000)} & 100 & 2.77 & 0.58 & 0.0132 & \hspace{-0.3cm}\footnotesize{\textpm 0.0003} & 0.0697 & \hspace{-0.3cm}\footnotesize{\textpm 0.0020} & 5.48 & \hspace{-0.3cm}\footnotesize{\textpm 0.54} & 912 & $[ 264$ & \hspace{-0.5cm}, & \hspace{-0.5cm}$2437 ]$ \\ 
 && 10\,000 \footnotesize{(10\,000)} & 100 & 10.03 & 0.80 & 0.0124 & \hspace{-0.3cm}\footnotesize{\textpm 0.0002} & 0.0638 & \hspace{-0.3cm}\footnotesize{\textpm 0.0013} & 5.56 & \hspace{-0.3cm}\footnotesize{\textpm 0.64} & 1799 & $[ 529$ & \hspace{-0.5cm}, & \hspace{-0.5cm}$4868 ]$ \\ 

	&&&&&&&&&&&&&&&\\[-0.35cm]\hline&&&&&&&&&&&&&&&\\[-0.25cm] 
	\multirow{4}{*}{ \begin{turn}{90} VP-SVM  \end{turn}} &
	\multirow{4}{*}{ \hspace{-0.3cm} \begin{turn}{90} $r$ =  2  \end{turn}}
	  & \phantom{0}1\,000 \footnotesize{(10\,000)} & 100 & 0.43 & 0.10 & 0.0155 & \hspace{-0.3cm}\footnotesize{\textpm 0.0002} & 0.0847 & \hspace{-0.3cm}\footnotesize{\textpm 0.0014} & 1.72 & \hspace{-0.3cm}\footnotesize{\textpm 0.45} & 581 & $[ 283$ & \hspace{-0.5cm}, & \hspace{-0.5cm}$1000 ]$ \\ 
 && \phantom{0}2\,500 \footnotesize{(10\,000)} & 100 & 2.29 & 0.15 & 0.0149 & \hspace{-0.3cm}\footnotesize{\textpm 0.0004} & 0.0808 & \hspace{-0.3cm}\footnotesize{\textpm 0.0024} & 1.55 & \hspace{-0.3cm}\footnotesize{\textpm 0.50} & 1613 & $[ 696$ & \hspace{-0.5cm}, & \hspace{-0.5cm}$2500 ]$ \\ 
 && \phantom{0}5\,000 \footnotesize{(10\,000)} & 100 & 7.96 & 0.42 & 0.0138 & \hspace{-0.3cm}\footnotesize{\textpm 0.0007} & 0.0739 & \hspace{-0.3cm}\footnotesize{\textpm 0.0049} & 1.67 & \hspace{-0.3cm}\footnotesize{\textpm 0.47} & 2994 & $[ 1345$ & \hspace{-0.5cm}, & \hspace{-0.5cm}$5000 ]$ \\ 
 && 10\,000 \footnotesize{(10\,000)} & 100 & 31.92 & 0.82 & 0.0126 & \hspace{-0.3cm}\footnotesize{\textpm 0.0009} & 0.0656 & \hspace{-0.3cm}\footnotesize{\textpm 0.0065} & 1.64 & \hspace{-0.3cm}\footnotesize{\textpm 0.48} & 6098 & $[ 2702$ & \hspace{-0.5cm}, & \hspace{-0.5cm}$10000 ]$ \\ 

	&&&&&&&&&&&&&&&\\[-0.35cm]\hline&&&&&&&&&&&&&&&\\[-0.25cm] 
	\multirow{4}{*}{ \begin{turn}{90} VP-SVM  \end{turn}} &
	\multirow{4}{*}{ \hspace{-0.3cm} \begin{turn}{90} $r$ =  3  \end{turn}}
	  & \phantom{0}1\,000 \footnotesize{(10\,000)} & 100 & 0.54 & 0.07 & 0.0155 & \hspace{-0.3cm}\footnotesize{\textpm 0.0002} & 0.0847 & \hspace{-0.3cm}\footnotesize{\textpm 0.0012} & 1.00 & \hspace{-0.3cm}\footnotesize{\textpm 0.00} & 1000 & \multicolumn{3}{c}{$\{ 1000 \}$} \\ 
 && \phantom{0}2\,500 \footnotesize{(10\,000)} & 100 & 3.09 & 0.12 & 0.0151 & \hspace{-0.3cm}\footnotesize{\textpm 0.0003} & 0.0822 & \hspace{-0.3cm}\footnotesize{\textpm 0.0016} & 1.00 & \hspace{-0.3cm}\footnotesize{\textpm 0.00} & 2500 & \multicolumn{3}{c}{$\{ 2500 \}$} \\ 
 && \phantom{0}5\,000 \footnotesize{(10\,000)} & 100 & 11.64 & 0.53 & 0.0132 & \hspace{-0.3cm}\footnotesize{\textpm 0.0007} & 0.0694 & \hspace{-0.3cm}\footnotesize{\textpm 0.0049} & 1.00 & \hspace{-0.3cm}\footnotesize{\textpm 0.00} & 5000 & \multicolumn{3}{c}{$\{ 5000 \}$} \\ 
 && 10\,000 \footnotesize{(10\,000)} & 100 & 45.40 & 0.54 & 0.0137 & \hspace{-0.3cm}\footnotesize{\textpm 0.0007} & 0.0730 & \hspace{-0.3cm}\footnotesize{\textpm 0.0046} & 1.00 & \hspace{-0.3cm}\footnotesize{\textpm 0.00} & 10000 & \multicolumn{3}{c}{$\{ 10000 \}$} \\ 

	\noalign{\bigskip}\noalign{\bigskip}
	&& data set sizes & runs & \multicolumn{1}{c}{train time} & \multicolumn{1}{c}{test time} & \multicolumn{2}{c}{test error} & \multicolumn{2}{c}{$L_2$-error} & \multicolumn{2}{c}{\# of ws} & \multicolumn{1}{c}{ws size: median} & \multicolumn{3}{c}{ws size: range}\\ 
	&&&&&&&&&&&&&&&\\[-0.35cm]\hline\hline&&&&&&&&&&&&&&&\\[-0.25cm] 
	\multirow{4}{*}{ \begin{turn}{90} RC-SVM  \end{turn}} &
	\multirow{4}{*}{ \hspace{-0.3cm} \begin{turn}{90} \# ws = 1  \end{turn}}
	  & \phantom{0}1\,000 \footnotesize{(10\,000)} & 100 & 0.53 & 0.06 & 0.0155 & \hspace{-0.3cm}\footnotesize{\textpm 0.0002} & 0.0847 & \hspace{-0.3cm}\footnotesize{\textpm 0.0012} & 1.00 & \hspace{-0.3cm}\footnotesize{\textpm 0.00} & 1000 & \multicolumn{3}{c}{$\{ 1000 \}$} \\ 
 && \phantom{0}2\,500 \footnotesize{(10\,000)} & 100 & 3.08 & 0.12 & 0.0151 & \hspace{-0.3cm}\footnotesize{\textpm 0.0003} & 0.0822 & \hspace{-0.3cm}\footnotesize{\textpm 0.0016} & 1.00 & \hspace{-0.3cm}\footnotesize{\textpm 0.00} & 2500 & \multicolumn{3}{c}{$\{ 2500 \}$} \\ 
 && \phantom{0}5\,000 \footnotesize{(10\,000)} & 100 & 11.67 & 0.52 & 0.0132 & \hspace{-0.3cm}\footnotesize{\textpm 0.0007} & 0.0694 & \hspace{-0.3cm}\footnotesize{\textpm 0.0049} & 1.00 & \hspace{-0.3cm}\footnotesize{\textpm 0.00} & 5000 & \multicolumn{3}{c}{$\{ 5000 \}$} \\ 
 && 10\,000 \footnotesize{(10\,000)} & 100 & 46.22 & 0.55 & 0.0137 & \hspace{-0.3cm}\footnotesize{\textpm 0.0007} & 0.0730 & \hspace{-0.3cm}\footnotesize{\textpm 0.0046} & 1.00 & \hspace{-0.3cm}\footnotesize{\textpm 0.00} & 10000 & \multicolumn{3}{c}{$\{ 10000 \}$} \\ 

	&&&&&&&&&&&&&&&\\[-0.35cm]\hline&&&&&&&&&&&&&&&\\[-0.25cm] 
	\multirow{4}{*}{ \begin{turn}{90} RC-SVM  \end{turn}} &
	\multirow{4}{*}{ \hspace{-0.3cm} \begin{turn}{90} \# ws = 2  \end{turn}}
	  & \phantom{0}1\,000 \footnotesize{(10\,000)} & 100 & 0.37 & 0.14 & 0.0153 & \hspace{-0.3cm}\footnotesize{\textpm 0.0002} & 0.0837 & \hspace{-0.3cm}\footnotesize{\textpm 0.0015} & 2.00 & \hspace{-0.3cm}\footnotesize{\textpm 0.00} & 500 & \multicolumn{3}{c}{$\{ 500 \}$} \\ 
 && \phantom{0}2\,500 \footnotesize{(10\,000)} & 100 & 1.61 & 0.13 & 0.0152 & \hspace{-0.3cm}\footnotesize{\textpm 0.0002} & 0.0830 & \hspace{-0.3cm}\footnotesize{\textpm 0.0012} & 2.00 & \hspace{-0.3cm}\footnotesize{\textpm 0.00} & 1250 & \multicolumn{3}{c}{$\{ 1250 \}$} \\ 
 && \phantom{0}5\,000 \footnotesize{(10\,000)} & 100 & 6.12 & 0.22 & 0.0151 & \hspace{-0.3cm}\footnotesize{\textpm 0.0002} & 0.0825 & \hspace{-0.3cm}\footnotesize{\textpm 0.0014} & 2.00 & \hspace{-0.3cm}\footnotesize{\textpm 0.00} & 2500 & \multicolumn{3}{c}{$\{ 2500 \}$} \\ 
 && 10\,000 \footnotesize{(10\,000)} & 100 & 23.51 & 0.97 & 0.0130 & \hspace{-0.3cm}\footnotesize{\textpm 0.0006} & 0.0680 & \hspace{-0.3cm}\footnotesize{\textpm 0.0046} & 2.00 & \hspace{-0.3cm}\footnotesize{\textpm 0.00} & 5000 & \multicolumn{3}{c}{$\{ 5000 \}$} \\ 

	&&&&&&&&&&&&&&&\\[-0.35cm]\hline&&&&&&&&&&&&&&&\\[-0.25cm] 
	\multirow{4}{*}{ \begin{turn}{90} RC-SVM  \end{turn}} &
	\multirow{4}{*}{ \hspace{-0.3cm} \begin{turn}{90} \# ws = 5 \end{turn}}
	  & \phantom{0}1\,000 \footnotesize{(10\,000)} & 100 & 0.46 & 0.22 & 0.0152 & \hspace{-0.3cm}\footnotesize{\textpm 0.0002} & 0.0830 & \hspace{-0.3cm}\footnotesize{\textpm 0.0014} & 5.00 & \hspace{-0.3cm}\footnotesize{\textpm 0.00} & 200 & \multicolumn{3}{c}{$\{ 200 \}$} \\ 
 && \phantom{0}2\,500 \footnotesize{(10\,000)} & 100 & 0.96 & 0.30 & 0.0150 & \hspace{-0.3cm}\footnotesize{\textpm 0.0002} & 0.0815 & \hspace{-0.3cm}\footnotesize{\textpm 0.0009} & 5.00 & \hspace{-0.3cm}\footnotesize{\textpm 0.00} & 500 & \multicolumn{3}{c}{$\{ 500 \}$} \\ 
 && \phantom{0}5\,000 \footnotesize{(10\,000)} & 100 & 2.70 & 0.28 & 0.0153 & \hspace{-0.3cm}\footnotesize{\textpm 0.0001} & 0.0833 & \hspace{-0.3cm}\footnotesize{\textpm 0.0008} & 5.00 & \hspace{-0.3cm}\footnotesize{\textpm 0.00} & 1000 & \multicolumn{3}{c}{$\{ 1000 \}$} \\ 
 && 10\,000 \footnotesize{(10\,000)} & 100 & 9.95 & 0.36 & 0.0152 & \hspace{-0.3cm}\footnotesize{\textpm 0.0001} & 0.0826 & \hspace{-0.3cm}\footnotesize{\textpm 0.0007} & 5.00 & \hspace{-0.3cm}\footnotesize{\textpm 0.00} & 2000 & \multicolumn{3}{c}{$\{ 2000 \}$} \\ 

	&&&&&&&&&&&&&&&\\[-0.35cm]\hline&&&&&&&&&&&&&&&\\[-0.25cm] 
	\multirow{4}{*}{ \begin{turn}{90} RC-SVM  \end{turn}} &
	\multirow{4}{*}{ \hspace{-0.3cm} \begin{turn}{90} \# ws = 10  \end{turn}}
	  & \phantom{0}1\,000 \footnotesize{(10\,000)} & 100 & 0.88 & 0.29 & 0.0154 & \hspace{-0.3cm}\footnotesize{\textpm 0.0002} & 0.0844 & \hspace{-0.3cm}\footnotesize{\textpm 0.0011} & 10.00 & \hspace{-0.3cm}\footnotesize{\textpm 0.00} & 100 & \multicolumn{3}{c}{$\{ 100 \}$} \\ 
 && \phantom{0}2\,500 \footnotesize{(10\,000)} & 100 & 1.02 & 0.46 & 0.0148 & \hspace{-0.3cm}\footnotesize{\textpm 0.0001} & 0.0808 & \hspace{-0.3cm}\footnotesize{\textpm 0.0007} & 10.00 & \hspace{-0.3cm}\footnotesize{\textpm 0.00} & 250 & \multicolumn{3}{c}{$\{ 250 \}$} \\ 
 && \phantom{0}5\,000 \footnotesize{(10\,000)} & 100 & 1.96 & 0.58 & 0.0150 & \hspace{-0.3cm}\footnotesize{\textpm 0.0001} & 0.0818 & \hspace{-0.3cm}\footnotesize{\textpm 0.0008} & 10.00 & \hspace{-0.3cm}\footnotesize{\textpm 0.00} & 500 & \multicolumn{3}{c}{$\{ 500 \}$} \\ 
 && 10\,000 \footnotesize{(10\,000)} & 100 & 5.67 & 0.53 & 0.0152 & \hspace{-0.3cm}\footnotesize{\textpm 0.0001} & 0.0827 & \hspace{-0.3cm}\footnotesize{\textpm 0.0005} & 10.00 & \hspace{-0.3cm}\footnotesize{\textpm 0.00} & 1000 & \multicolumn{3}{c}{$\{ 1000 \}$} \\ 

	&&&&&&&&&&&&&&&\\[-0.35cm]\hline&&&&&&&&&&&&&&&\\[-0.25cm] 
	\multirow{4}{*}{ \begin{turn}{90} RC-SVM  \end{turn}} &
	\multirow{4}{*}{ \hspace{-0.3cm} \begin{turn}{90} \# ws = 15  \end{turn}}
	  & \phantom{0}1\,000 \footnotesize{(10\,000)} & 100 & 1.50 & 0.34 & 0.0157 & \hspace{-0.3cm}\footnotesize{\textpm 0.0002} & 0.0864 & \hspace{-0.3cm}\footnotesize{\textpm 0.0014} & 15.00 & \hspace{-0.3cm}\footnotesize{\textpm 0.00} & 67 & $[ 66$ & \hspace{-0.5cm}, & \hspace{-0.5cm}$67 ]$ \\ 
 && \phantom{0}2\,500 \footnotesize{(10\,000)} & 100 & 1.62 & 0.53 & 0.0149 & \hspace{-0.3cm}\footnotesize{\textpm 0.0001} & 0.0814 & \hspace{-0.3cm}\footnotesize{\textpm 0.0007} & 15.00 & \hspace{-0.3cm}\footnotesize{\textpm 0.00} & 167 & $[ 166$ & \hspace{-0.5cm}, & \hspace{-0.5cm}$167 ]$ \\ 
 && \phantom{0}5\,000 \footnotesize{(10\,000)} & 100 & 2.10 & 0.84 & 0.0149 & \hspace{-0.3cm}\footnotesize{\textpm 0.0001} & 0.0812 & \hspace{-0.3cm}\footnotesize{\textpm 0.0006} & 15.00 & \hspace{-0.3cm}\footnotesize{\textpm 0.00} & 333 & $[ 333$ & \hspace{-0.5cm}, & \hspace{-0.5cm}$334 ]$ \\ 
 && 10\,000 \footnotesize{(10\,000)} & 100 & 4.80 & 0.87 & 0.0150 & \hspace{-0.3cm}\footnotesize{\textpm 0.0001} & 0.0819 & \hspace{-0.3cm}\footnotesize{\textpm 0.0005} & 15.00 & \hspace{-0.3cm}\footnotesize{\textpm 0.00} & 667 & $[ 666$ & \hspace{-0.5cm}, & \hspace{-0.5cm}$667 ]$ \\ 

	&&&&&&&&&&&&&&&\\[-0.35cm]\hline&&&&&&&&&&&&&&&\\[-0.25cm] 
	\multirow{4}{*}{ \begin{turn}{90} RC-SVM  \end{turn}} &
	\multirow{4}{*}{ \hspace{-0.3cm} \begin{turn}{90} \# ws = 20  \end{turn}}
	  & \phantom{0}1\,000 \footnotesize{(10\,000)} & 100 & 1.88 & 0.39 & 0.0162 & \hspace{-0.3cm}\footnotesize{\textpm 0.0003} & 0.0891 & \hspace{-0.3cm}\footnotesize{\textpm 0.0017} & 20.00 & \hspace{-0.3cm}\footnotesize{\textpm 0.00} & 50 & \multicolumn{3}{c}{$\{ 50 \}$} \\ 
 && \phantom{0}2\,500 \footnotesize{(10\,000)} & 100 & 1.99 & 0.59 & 0.0151 & \hspace{-0.3cm}\footnotesize{\textpm 0.0001} & 0.0822 & \hspace{-0.3cm}\footnotesize{\textpm 0.0006} & 20.00 & \hspace{-0.3cm}\footnotesize{\textpm 0.00} & 125 & \multicolumn{3}{c}{$\{ 125 \}$} \\ 
 && \phantom{0}5\,000 \footnotesize{(10\,000)} & 100 & 2.27 & 0.89 & 0.0149 & \hspace{-0.3cm}\footnotesize{\textpm 0.0001} & 0.0814 & \hspace{-0.3cm}\footnotesize{\textpm 0.0006} & 20.00 & \hspace{-0.3cm}\footnotesize{\textpm 0.00} & 250 & \multicolumn{3}{c}{$\{ 250 \}$} \\ 
 && 10\,000 \footnotesize{(10\,000)} & 100 & 4.24 & 1.14 & 0.0149 & \hspace{-0.3cm}\footnotesize{\textpm 0.0001} & 0.0812 & \hspace{-0.3cm}\footnotesize{\textpm 0.0005} & 20.00 & \hspace{-0.3cm}\footnotesize{\textpm 0.00} & 500 & \multicolumn{3}{c}{$\{ 500 \}$} \\ 

	&&&&&&&&&&&&&&&\\[-0.35cm]\hline&&&&&&&&&&&&&&&\\[-0.25cm] 
	\multirow{4}{*}{ \begin{turn}{90} RC-SVM  \end{turn}} &
	\multirow{4}{*}{ \hspace{-0.3cm} \begin{turn}{90} \# ws = 40  \end{turn}}
	  & \phantom{0}1\,000 \footnotesize{(10\,000)} & 100 & 3.96 & 0.59 & 0.0194 & \hspace{-0.3cm}\footnotesize{\textpm 0.0007} & 0.1053 & \hspace{-0.3cm}\footnotesize{\textpm 0.0033} & 40.00 & \hspace{-0.3cm}\footnotesize{\textpm 0.00} & 25 & \multicolumn{3}{c}{$\{ 25 \}$} \\ 
 && \phantom{0}2\,500 \footnotesize{(10\,000)} & 100 & 4.06 & 0.82 & 0.0157 & \hspace{-0.3cm}\footnotesize{\textpm 0.0002} & 0.0859 & \hspace{-0.3cm}\footnotesize{\textpm 0.0010} & 40.00 & \hspace{-0.3cm}\footnotesize{\textpm 0.00} & 62 & $[ 62$ & \hspace{-0.5cm}, & \hspace{-0.5cm}$63 ]$ \\ 
 && \phantom{0}5\,000 \footnotesize{(10\,000)} & 100 & 4.29 & 1.17 & 0.0152 & \hspace{-0.3cm}\footnotesize{\textpm 0.0001} & 0.0829 & \hspace{-0.3cm}\footnotesize{\textpm 0.0005} & 40.00 & \hspace{-0.3cm}\footnotesize{\textpm 0.00} & 125 & \multicolumn{3}{c}{$\{ 125 \}$} \\ 
 && 10\,000 \footnotesize{(10\,000)} & 100 & 4.93 & 1.75 & 0.0148 & \hspace{-0.3cm}\footnotesize{\textpm 0.0001} & 0.0808 & \hspace{-0.3cm}\footnotesize{\textpm 0.0004} & 40.00 & \hspace{-0.3cm}\footnotesize{\textpm 0.00} & 250 & \multicolumn{3}{c}{$\{ 250 \}$} \\ 

	&&&&&&&&&&&&&&&\\[-0.35cm]\hline&&&&&&&&&&&&&&&\\[-0.25cm] 
	\multirow{4}{*}{ \begin{turn}{90} RC-SVM  \end{turn}} &
	\multirow{4}{*}{ \hspace{-0.3cm} \begin{turn}{90} \# ws = 50  \end{turn}}
	  & \phantom{0}1\,000 \footnotesize{(10\,000)} & 100 & 4.96 & 0.69 & 0.0222 & \hspace{-0.3cm}\footnotesize{\textpm 0.0011} & 0.1181 & \hspace{-0.3cm}\footnotesize{\textpm 0.0045} & 50.00 & \hspace{-0.3cm}\footnotesize{\textpm 0.00} & 20 & \multicolumn{3}{c}{$\{ 20 \}$} \\ 
 && \phantom{0}2\,500 \footnotesize{(10\,000)} & 100 & 5.09 & 0.93 & 0.0161 & \hspace{-0.3cm}\footnotesize{\textpm 0.0002} & 0.0883 & \hspace{-0.3cm}\footnotesize{\textpm 0.0011} & 50.00 & \hspace{-0.3cm}\footnotesize{\textpm 0.00} & 50 & \multicolumn{3}{c}{$\{ 50 \}$} \\ 
 && \phantom{0}5\,000 \footnotesize{(10\,000)} & 100 & 5.30 & 1.29 & 0.0153 & \hspace{-0.3cm}\footnotesize{\textpm 0.0001} & 0.0839 & \hspace{-0.3cm}\footnotesize{\textpm 0.0005} & 50.00 & \hspace{-0.3cm}\footnotesize{\textpm 0.00} & 100 & \multicolumn{3}{c}{$\{ 100 \}$} \\ 
 && 10\,000 \footnotesize{(10\,000)} & 100 & 6.43 & 1.94 & 0.0149 & \hspace{-0.3cm}\footnotesize{\textpm 0.0001} & 0.0810 & \hspace{-0.3cm}\footnotesize{\textpm 0.0004} & 50.00 & \hspace{-0.3cm}\footnotesize{\textpm 0.00} & 200 & \multicolumn{3}{c}{$\{ 200 \}$} \\ 

      \end{tabular}
    } 
    \caption{Experimental results relating to the artificial data of Type \RM{4}}
    \label{tab::experiments::art_data_3_2}
  \end{minipage} 
\end{sidewaystable}

\begin{sidewaystable}[t!]
  \centering
  \begin{minipage}[t]{0.49\linewidth} \vspace{0pt}
    \resizebox{\textwidth}{!}{
        \begin{tabular}{cc||c|c||S[tabnumalign=centre]S[tabnumalign=centre]rlrlrlS[tabnumalign=centre]rcl}
	&& data set sizes & runs & \multicolumn{1}{c}{train time} & \multicolumn{1}{c}{test time} & \multicolumn{2}{c}{test error} & \multicolumn{2}{c}{$L_2$-error} & \multicolumn{2}{c}{\# of ws} & \multicolumn{1}{c}{ws size: median} & \multicolumn{3}{c}{ws size: range}\\ 
	&&&&&&&&&&&&&&&\\[-0.35cm]\hline\hline&&&&&&&&&&&&&&&\\[-0.25cm]  
	\multirow{4}{*}{ \begin{turn}{90} LS-SVM  \end{turn}}
	&  & \phantom{0}1\,000 \footnotesize{(10\,000)} & 100 & 0.53 & 0.05 & 0.0649 & \hspace{-0.3cm}\footnotesize{\textpm 0.0004} & 0.0370 & \hspace{-0.3cm}\footnotesize{\textpm 0.0055} & 1.00 &  & \\ 
 && \phantom{0}2\,500 \footnotesize{(10\,000)} & 100 & 3.26 & 0.09 & 0.0647 & \hspace{-0.3cm}\footnotesize{\textpm 0.0002} & 0.0330 & \hspace{-0.3cm}\footnotesize{\textpm 0.0032} & 1.00 &  & \\ 
 && \phantom{0}5\,000 \footnotesize{(10\,000)} & 100 & 12.51 & 0.15 & 0.0640 & \hspace{-0.3cm}\footnotesize{\textpm 0.0001} & 0.0240 & \hspace{-0.3cm}\footnotesize{\textpm 0.0015} & 1.00 &  & \\ 
 && 10\,000 \footnotesize{(10\,000)} & 100 & 49.29 & 0.28 & 0.0640 & \hspace{-0.3cm}\footnotesize{\textpm 0.0001} & 0.0223 & \hspace{-0.3cm}\footnotesize{\textpm 0.0015} & 1.00 &  & \\ 
 
	\noalign{\bigskip}\noalign{\bigskip}
	&& data set sizes & runs & \multicolumn{1}{c}{train time} & \multicolumn{1}{c}{test time} & \multicolumn{2}{c}{test error} & \multicolumn{2}{c}{$L_2$-error} & \multicolumn{2}{c}{\# of ws} & \multicolumn{1}{c}{ws size: median} & \multicolumn{3}{c}{ws size: range}\\ 
	&&&&&&&&&&&&&&&\\[-0.35cm]\hline\hline&&&&&&&&&&&&&&&\\[-0.25cm] 
	\multirow{4}{*}{ \begin{turn}{90} VP-SVM  \end{turn}} &
	\multirow{4}{*}{ \hspace{-0.3cm} \begin{turn}{90} $r$ =  0.25  \end{turn}}
	  & \phantom{0}1\,000 \footnotesize{(10\,000)} & 100 & 4.85 & 0.66 & 0.0797 & \hspace{-0.3cm}\footnotesize{\textpm 0.0021} & 0.1258 & \hspace{-0.3cm}\footnotesize{\textpm 0.0076} & 44.56 & \hspace{-0.3cm}\footnotesize{\textpm 2.07} & 22 & $[ 6$ & \hspace{-0.5cm}, & \hspace{-0.5cm}$54 ]$ \\ 
 && \phantom{0}2\,500 \footnotesize{(10\,000)} & 100 & 4.76 & 0.77 & 0.0719 & \hspace{-0.3cm}\footnotesize{\textpm 0.0009} & 0.0882 & \hspace{-0.3cm}\footnotesize{\textpm 0.0043} & 46.18 & \hspace{-0.3cm}\footnotesize{\textpm 2.55} & 54 & $[ 14$ & \hspace{-0.5cm}, & \hspace{-0.5cm}$116 ]$ \\ 
 && \phantom{0}5\,000 \footnotesize{(10\,000)} & 100 & 4.88 & 0.95 & 0.0682 & \hspace{-0.3cm}\footnotesize{\textpm 0.0005} & 0.0663 & \hspace{-0.3cm}\footnotesize{\textpm 0.0033} & 47.45 & \hspace{-0.3cm}\footnotesize{\textpm 2.61} & 105 & $[ 20$ & \hspace{-0.5cm}, & \hspace{-0.5cm}$214 ]$ \\ 
 && 10\,000 \footnotesize{(10\,000)} & 100 & 5.77 & 1.14 & 0.0657 & \hspace{-0.3cm}\footnotesize{\textpm 0.0003} & 0.0460 & \hspace{-0.3cm}\footnotesize{\textpm 0.0027} & 48.05 & \hspace{-0.3cm}\footnotesize{\textpm 2.78} & 208 & $[ 38$ & \hspace{-0.5cm}, & \hspace{-0.5cm}$407 ]$ \\ 

	&&&&&&&&&&&&&&&\\[-0.35cm]\hline&&&&&&&&&&&&&&&\\[-0.25cm] 
	\multirow{4}{*}{ \begin{turn}{90} VP-SVM  \end{turn}} &
	\multirow{4}{*}{ \hspace{-0.3cm} \begin{turn}{90} $r$ =  0.5  \end{turn}}
	  & \phantom{0}1\,000 \footnotesize{(10\,000)} & 100 & 1.65 & 0.30 & 0.0703 & \hspace{-0.3cm}\footnotesize{\textpm 0.0011} & 0.0820 & \hspace{-0.3cm}\footnotesize{\textpm 0.0068} & 14.98 & \hspace{-0.3cm}\footnotesize{\textpm 1.44} & 67 & $[ 16$ & \hspace{-0.5cm}, & \hspace{-0.5cm}$129 ]$ \\ 
 && \phantom{0}2\,500 \footnotesize{(10\,000)} & 100 & 1.64 & 0.37 & 0.0670 & \hspace{-0.3cm}\footnotesize{\textpm 0.0006} & 0.0565 & \hspace{-0.3cm}\footnotesize{\textpm 0.0045} & 15.17 & \hspace{-0.3cm}\footnotesize{\textpm 1.52} & 165 & $[ 40$ & \hspace{-0.5cm}, & \hspace{-0.5cm}$334 ]$ \\ 
 && \phantom{0}5\,000 \footnotesize{(10\,000)} & 100 & 2.26 & 0.45 & 0.0652 & \hspace{-0.3cm}\footnotesize{\textpm 0.0003} & 0.0404 & \hspace{-0.3cm}\footnotesize{\textpm 0.0035} & 15.69 & \hspace{-0.3cm}\footnotesize{\textpm 1.47} & 319 & $[ 74$ & \hspace{-0.5cm}, & \hspace{-0.5cm}$644 ]$ \\ 
 && 10\,000 \footnotesize{(10\,000)} & 100 & 4.81 & 0.62 & 0.0645 & \hspace{-0.3cm}\footnotesize{\textpm 0.0002} & 0.0307 & \hspace{-0.3cm}\footnotesize{\textpm 0.0023} & 15.83 & \hspace{-0.3cm}\footnotesize{\textpm 1.49} & 632 & $[ 167$ & \hspace{-0.5cm}, & \hspace{-0.5cm}$1343 ]$ \\ 

	&&&&&&&&&&&&&&&\\[-0.35cm]\hline&&&&&&&&&&&&&&&\\[-0.25cm] 
	\multirow{4}{*}{ \begin{turn}{90} VP-SVM  \end{turn}} &
	\multirow{4}{*}{ \hspace{-0.3cm} \begin{turn}{90} $r$ =  1  \end{turn}}
	  & \phantom{0}1\,000 \footnotesize{(10\,000)} & 100 & 0.56 & 0.16 & 0.0670 & \hspace{-0.3cm}\footnotesize{\textpm 0.0008} & 0.0578 & \hspace{-0.3cm}\footnotesize{\textpm 0.0064} & 5.33 & \hspace{-0.3cm}\footnotesize{\textpm 0.47} & 188 & $[ 68$ & \hspace{-0.5cm}, & \hspace{-0.5cm}$464 ]$ \\ 
 && \phantom{0}2\,500 \footnotesize{(10\,000)} & 100 & 1.05 & 0.19 & 0.0656 & \hspace{-0.3cm}\footnotesize{\textpm 0.0004} & 0.0431 & \hspace{-0.3cm}\footnotesize{\textpm 0.0042} & 5.33 & \hspace{-0.3cm}\footnotesize{\textpm 0.47} & 469 & $[ 171$ & \hspace{-0.5cm}, & \hspace{-0.5cm}$1162 ]$ \\ 
 && \phantom{0}5\,000 \footnotesize{(10\,000)} & 100 & 3.01 & 0.26 & 0.0643 & \hspace{-0.3cm}\footnotesize{\textpm 0.0002} & 0.0290 & \hspace{-0.3cm}\footnotesize{\textpm 0.0025} & 5.42 & \hspace{-0.3cm}\footnotesize{\textpm 0.52} & 923 & $[ 297$ & \hspace{-0.5cm}, & \hspace{-0.5cm}$2445 ]$ \\ 
 && 10\,000 \footnotesize{(10\,000)} & 100 & 10.81 & 0.41 & 0.0641 & \hspace{-0.3cm}\footnotesize{\textpm 0.0001} & 0.0243 & \hspace{-0.3cm}\footnotesize{\textpm 0.0022} & 5.54 & \hspace{-0.3cm}\footnotesize{\textpm 0.61} & 1805 & $[ 555$ & \hspace{-0.5cm}, & \hspace{-0.5cm}$4840 ]$ \\ 

	&&&&&&&&&&&&&&&\\[-0.35cm]\hline&&&&&&&&&&&&&&&\\[-0.25cm] 
	\multirow{4}{*}{ \begin{turn}{90} VP-SVM  \end{turn}} &
	\multirow{4}{*}{ \hspace{-0.3cm} \begin{turn}{90} $r$ =  2  \end{turn}}
	  & \phantom{0}1\,000 \footnotesize{(10\,000)} & 100 & 0.49 & 0.13 & 0.0653 & \hspace{-0.3cm}\footnotesize{\textpm 0.0006} & 0.0427 & \hspace{-0.3cm}\footnotesize{\textpm 0.0061} & 1.65 & \hspace{-0.3cm}\footnotesize{\textpm 0.48} & 606 & $[ 266$ & \hspace{-0.5cm}, & \hspace{-0.5cm}$1000 ]$ \\ 
 && \phantom{0}2\,500 \footnotesize{(10\,000)} & 100 & 2.21 & 0.11 & 0.0649 & \hspace{-0.3cm}\footnotesize{\textpm 0.0003} & 0.0357 & \hspace{-0.3cm}\footnotesize{\textpm 0.0042} & 1.70 & \hspace{-0.3cm}\footnotesize{\textpm 0.48} & 1471 & $[ 603$ & \hspace{-0.5cm}, & \hspace{-0.5cm}$2500 ]$ \\ 
 && \phantom{0}5\,000 \footnotesize{(10\,000)} & 100 & 9.32 & 0.18 & 0.0641 & \hspace{-0.3cm}\footnotesize{\textpm 0.0001} & 0.0248 & \hspace{-0.3cm}\footnotesize{\textpm 0.0018} & 1.54 & \hspace{-0.3cm}\footnotesize{\textpm 0.50} & 3247 & $[ 1348$ & \hspace{-0.5cm}, & \hspace{-0.5cm}$5000 ]$ \\ 
 && 10\,000 \footnotesize{(10\,000)} & 100 & 35.16 & 0.31 & 0.0640 & \hspace{-0.3cm}\footnotesize{\textpm 0.0001} & 0.0227 & \hspace{-0.3cm}\footnotesize{\textpm 0.0018} & 1.62 & \hspace{-0.3cm}\footnotesize{\textpm 0.49} & 6173 & $[ 2608$ & \hspace{-0.5cm}, & \hspace{-0.5cm}$10000 ]$ \\ 

	&&&&&&&&&&&&&&&\\[-0.35cm]\hline&&&&&&&&&&&&&&&\\[-0.25cm] 
	\multirow{4}{*}{ \begin{turn}{90} VP-SVM  \end{turn}} &
	\multirow{4}{*}{ \hspace{-0.3cm} \begin{turn}{90} $r$ =  3  \end{turn}}
	  & \phantom{0}1\,000 \footnotesize{(10\,000)} & 100 & 0.54 & 0.07 & 0.0649 & \hspace{-0.3cm}\footnotesize{\textpm 0.0004} & 0.0371 & \hspace{-0.3cm}\footnotesize{\textpm 0.0055} & 1.00 & \hspace{-0.3cm}\footnotesize{\textpm 0.00} & 1000 & \multicolumn{3}{c}{$\{ 1000 \}$} \\ 
 && \phantom{0}2\,500 \footnotesize{(10\,000)} & 100 & 3.25 & 0.09 & 0.0647 & \hspace{-0.3cm}\footnotesize{\textpm 0.0002} & 0.0331 & \hspace{-0.3cm}\footnotesize{\textpm 0.0032} & 1.00 & \hspace{-0.3cm}\footnotesize{\textpm 0.00} & 2500 & \multicolumn{3}{c}{$\{ 2500 \}$} \\ 
 && \phantom{0}5\,000 \footnotesize{(10\,000)} & 100 & 12.54 & 0.16 & 0.0640 & \hspace{-0.3cm}\footnotesize{\textpm 0.0001} & 0.0240 & \hspace{-0.3cm}\footnotesize{\textpm 0.0015} & 1.00 & \hspace{-0.3cm}\footnotesize{\textpm 0.00} & 5000 & \multicolumn{3}{c}{$\{ 5000 \}$} \\ 
 && 10\,000 \footnotesize{(10\,000)} & 100 & 49.35 & 0.29 & 0.0640 & \hspace{-0.3cm}\footnotesize{\textpm 0.0001} & 0.0223 & \hspace{-0.3cm}\footnotesize{\textpm 0.0015} & 1.00 & \hspace{-0.3cm}\footnotesize{\textpm 0.00} & 10000 & \multicolumn{3}{c}{$\{ 10000 \}$} \\ 

	\noalign{\bigskip}\noalign{\bigskip}
	&& data set sizes & runs & \multicolumn{1}{c}{train time} & \multicolumn{1}{c}{test time} & \multicolumn{2}{c}{test error} & \multicolumn{2}{c}{$L_2$-error} & \multicolumn{2}{c}{\# of ws} & \multicolumn{1}{c}{ws size: median} & \multicolumn{3}{c}{ws size: range}\\ 
	&&&&&&&&&&&&&&&\\[-0.35cm]\hline\hline&&&&&&&&&&&&&&&\\[-0.25cm] 
	\multirow{4}{*}{ \begin{turn}{90} RC-SVM  \end{turn}} &
	\multirow{4}{*}{ \hspace{-0.3cm} \begin{turn}{90} \# ws = 1  \end{turn}}
	  & \phantom{0}1\,000 \footnotesize{(10\,000)} & 100 & 0.53 & 0.07 & 0.0649 & \hspace{-0.3cm}\footnotesize{\textpm 0.0004} & 0.0373 & \hspace{-0.3cm}\footnotesize{\textpm 0.0056} & 1.00 & \hspace{-0.3cm}\footnotesize{\textpm 0.00} & 1000 & \multicolumn{3}{c}{$\{ 1000 \}$} \\ 
 && \phantom{0}2\,500 \footnotesize{(10\,000)} & 100 & 3.27 & 0.09 & 0.0647 & \hspace{-0.3cm}\footnotesize{\textpm 0.0002} & 0.0331 & \hspace{-0.3cm}\footnotesize{\textpm 0.0033} & 1.00 & \hspace{-0.3cm}\footnotesize{\textpm 0.00} & 2500 & \multicolumn{3}{c}{$\{ 2500 \}$} \\ 
 && \phantom{0}5\,000 \footnotesize{(10\,000)} & 100 & 12.55 & 0.16 & 0.0640 & \hspace{-0.3cm}\footnotesize{\textpm 0.0001} & 0.0240 & \hspace{-0.3cm}\footnotesize{\textpm 0.0015} & 1.00 & \hspace{-0.3cm}\footnotesize{\textpm 0.00} & 5000 & \multicolumn{3}{c}{$\{ 5000 \}$} \\ 
 && 10\,000 \footnotesize{(10\,000)} & 100 & 50.04 & 0.29 & 0.0640 & \hspace{-0.3cm}\footnotesize{\textpm 0.0001} & 0.0223 & \hspace{-0.3cm}\footnotesize{\textpm 0.0015} & 1.00 & \hspace{-0.3cm}\footnotesize{\textpm 0.00} & 10000 & \multicolumn{3}{c}{$\{ 10000 \}$} \\ 

	&&&&&&&&&&&&&&&\\[-0.35cm]\hline&&&&&&&&&&&&&&&\\[-0.25cm] 
	\multirow{4}{*}{ \begin{turn}{90} RC-SVM  \end{turn}} &
	\multirow{4}{*}{ \hspace{-0.3cm} \begin{turn}{90} \# ws = 2  \end{turn}}
	  & \phantom{0}1\,000 \footnotesize{(10\,000)} & 100 & 0.38 & 0.09 & 0.0652 & \hspace{-0.3cm}\footnotesize{\textpm 0.0005} & 0.0409 & \hspace{-0.3cm}\footnotesize{\textpm 0.0056} & 2.00 & \hspace{-0.3cm}\footnotesize{\textpm 0.00} & 500 & \multicolumn{3}{c}{$\{ 500 \}$} \\ 
 && \phantom{0}2\,500 \footnotesize{(10\,000)} & 100 & 1.71 & 0.12 & 0.0649 & \hspace{-0.3cm}\footnotesize{\textpm 0.0003} & 0.0352 & \hspace{-0.3cm}\footnotesize{\textpm 0.0037} & 2.00 & \hspace{-0.3cm}\footnotesize{\textpm 0.00} & 1250 & \multicolumn{3}{c}{$\{ 1250 \}$} \\ 
 && \phantom{0}5\,000 \footnotesize{(10\,000)} & 100 & 6.39 & 0.18 & 0.0640 & \hspace{-0.3cm}\footnotesize{\textpm 0.0001} & 0.0242 & \hspace{-0.3cm}\footnotesize{\textpm 0.0019} & 2.00 & \hspace{-0.3cm}\footnotesize{\textpm 0.00} & 2500 & \multicolumn{3}{c}{$\{ 2500 \}$} \\ 
 && 10\,000 \footnotesize{(10\,000)} & 100 & 25.47 & 0.30 & 0.0640 & \hspace{-0.3cm}\footnotesize{\textpm 0.0001} & 0.0221 & \hspace{-0.3cm}\footnotesize{\textpm 0.0014} & 2.00 & \hspace{-0.3cm}\footnotesize{\textpm 0.00} & 5000 & \multicolumn{3}{c}{$\{ 5000 \}$} \\ 

	&&&&&&&&&&&&&&&\\[-0.35cm]\hline&&&&&&&&&&&&&&&\\[-0.25cm] 
	\multirow{4}{*}{ \begin{turn}{90} RC-SVM  \end{turn}} &
	\multirow{4}{*}{ \hspace{-0.3cm} \begin{turn}{90} \# ws = 5 \end{turn}}
	  & \phantom{0}1\,000 \footnotesize{(10\,000)} & 100 & 0.48 & 0.17 & 0.0655 & \hspace{-0.3cm}\footnotesize{\textpm 0.0005} & 0.0440 & \hspace{-0.3cm}\footnotesize{\textpm 0.0056} & 5.00 & \hspace{-0.3cm}\footnotesize{\textpm 0.00} & 200 & \multicolumn{3}{c}{$\{ 200 \}$} \\ 
 && \phantom{0}2\,500 \footnotesize{(10\,000)} & 100 & 1.02 & 0.19 & 0.0654 & \hspace{-0.3cm}\footnotesize{\textpm 0.0004} & 0.0426 & \hspace{-0.3cm}\footnotesize{\textpm 0.0044} & 5.00 & \hspace{-0.3cm}\footnotesize{\textpm 0.00} & 500 & \multicolumn{3}{c}{$\{ 500 \}$} \\ 
 && \phantom{0}5\,000 \footnotesize{(10\,000)} & 100 & 2.82 & 0.24 & 0.0643 & \hspace{-0.3cm}\footnotesize{\textpm 0.0002} & 0.0295 & \hspace{-0.3cm}\footnotesize{\textpm 0.0026} & 5.00 & \hspace{-0.3cm}\footnotesize{\textpm 0.00} & 1000 & \multicolumn{3}{c}{$\{ 1000 \}$} \\ 
 && 10\,000 \footnotesize{(10\,000)} & 100 & 10.63 & 0.37 & 0.0641 & \hspace{-0.3cm}\footnotesize{\textpm 0.0001} & 0.0248 & \hspace{-0.3cm}\footnotesize{\textpm 0.0017} & 5.00 & \hspace{-0.3cm}\footnotesize{\textpm 0.00} & 2000 & \multicolumn{3}{c}{$\{ 2000 \}$} \\ 

	&&&&&&&&&&&&&&&\\[-0.35cm]\hline&&&&&&&&&&&&&&&\\[-0.25cm] 
	\multirow{4}{*}{ \begin{turn}{90} RC-SVM  \end{turn}} &
	\multirow{4}{*}{ \hspace{-0.3cm} \begin{turn}{90} \# ws = 10  \end{turn}}
	  & \phantom{0}1\,000 \footnotesize{(10\,000)} & 100 & 0.85 & 0.25 & 0.0657 & \hspace{-0.3cm}\footnotesize{\textpm 0.0005} & 0.0471 & \hspace{-0.3cm}\footnotesize{\textpm 0.0054} & 10.00 & \hspace{-0.3cm}\footnotesize{\textpm 0.00} & 100 & \multicolumn{3}{c}{$\{ 100 \}$} \\ 
 && \phantom{0}2\,500 \footnotesize{(10\,000)} & 100 & 1.05 & 0.33 & 0.0656 & \hspace{-0.3cm}\footnotesize{\textpm 0.0004} & 0.0448 & \hspace{-0.3cm}\footnotesize{\textpm 0.0041} & 10.00 & \hspace{-0.3cm}\footnotesize{\textpm 0.00} & 250 & \multicolumn{3}{c}{$\{ 250 \}$} \\ 
 && \phantom{0}5\,000 \footnotesize{(10\,000)} & 100 & 2.04 & 0.35 & 0.0646 & \hspace{-0.3cm}\footnotesize{\textpm 0.0002} & 0.0335 & \hspace{-0.3cm}\footnotesize{\textpm 0.0032} & 10.00 & \hspace{-0.3cm}\footnotesize{\textpm 0.00} & 500 & \multicolumn{3}{c}{$\{ 500 \}$} \\ 
 && 10\,000 \footnotesize{(10\,000)} & 100 & 6.34 & 0.51 & 0.0644 & \hspace{-0.3cm}\footnotesize{\textpm 0.0001} & 0.0299 & \hspace{-0.3cm}\footnotesize{\textpm 0.0019} & 10.00 & \hspace{-0.3cm}\footnotesize{\textpm 0.00} & 1000 & \multicolumn{3}{c}{$\{ 1000 \}$} \\ 

	&&&&&&&&&&&&&&&\\[-0.35cm]\hline&&&&&&&&&&&&&&&\\[-0.25cm] 
	\multirow{4}{*}{ \begin{turn}{90} RC-SVM  \end{turn}} &
	\multirow{4}{*}{ \hspace{-0.3cm} \begin{turn}{90} \# ws = 15  \end{turn}}
	  & \phantom{0}1\,000 \footnotesize{(10\,000)} & 100 & 1.47 & 0.32 & 0.0664 & \hspace{-0.3cm}\footnotesize{\textpm 0.0006} & 0.0537 & \hspace{-0.3cm}\footnotesize{\textpm 0.0059} & 15.00 & \hspace{-0.3cm}\footnotesize{\textpm 0.00} & 67 & $[ 66$ & \hspace{-0.5cm}, & \hspace{-0.5cm}$67 ]$ \\ 
 && \phantom{0}2\,500 \footnotesize{(10\,000)} & 100 & 1.63 & 0.43 & 0.0656 & \hspace{-0.3cm}\footnotesize{\textpm 0.0004} & 0.0449 & \hspace{-0.3cm}\footnotesize{\textpm 0.0040} & 15.00 & \hspace{-0.3cm}\footnotesize{\textpm 0.00} & 167 & $[ 166$ & \hspace{-0.5cm}, & \hspace{-0.5cm}$167 ]$ \\ 
 && \phantom{0}5\,000 \footnotesize{(10\,000)} & 100 & 2.18 & 0.49 & 0.0647 & \hspace{-0.3cm}\footnotesize{\textpm 0.0002} & 0.0350 & \hspace{-0.3cm}\footnotesize{\textpm 0.0030} & 15.00 & \hspace{-0.3cm}\footnotesize{\textpm 0.00} & 333 & $[ 333$ & \hspace{-0.5cm}, & \hspace{-0.5cm}$334 ]$ \\ 
 && 10\,000 \footnotesize{(10\,000)} & 100 & 5.23 & 0.58 & 0.0646 & \hspace{-0.3cm}\footnotesize{\textpm 0.0001} & 0.0335 & \hspace{-0.3cm}\footnotesize{\textpm 0.0022} & 15.00 & \hspace{-0.3cm}\footnotesize{\textpm 0.00} & 667 & $[ 666$ & \hspace{-0.5cm}, & \hspace{-0.5cm}$667 ]$ \\ 

	&&&&&&&&&&&&&&&\\[-0.35cm]\hline&&&&&&&&&&&&&&&\\[-0.25cm] 
	\multirow{4}{*}{ \begin{turn}{90} RC-SVM  \end{turn}} &
	\multirow{4}{*}{ \hspace{-0.3cm} \begin{turn}{90} \# ws = 20  \end{turn}}
	  & \phantom{0}1\,000 \footnotesize{(10\,000)} & 100 & 1.92 & 0.38 & 0.0671 & \hspace{-0.3cm}\footnotesize{\textpm 0.0007} & 0.0601 & \hspace{-0.3cm}\footnotesize{\textpm 0.0054} & 20.00 & \hspace{-0.3cm}\footnotesize{\textpm 0.00} & 50 & \multicolumn{3}{c}{$\{ 50 \}$} \\ 
 && \phantom{0}2\,500 \footnotesize{(10\,000)} & 100 & 2.00 & 0.52 & 0.0659 & \hspace{-0.3cm}\footnotesize{\textpm 0.0004} & 0.0473 & \hspace{-0.3cm}\footnotesize{\textpm 0.0037} & 20.00 & \hspace{-0.3cm}\footnotesize{\textpm 0.00} & 125 & \multicolumn{3}{c}{$\{ 125 \}$} \\ 
 && \phantom{0}5\,000 \footnotesize{(10\,000)} & 100 & 2.32 & 0.61 & 0.0647 & \hspace{-0.3cm}\footnotesize{\textpm 0.0002} & 0.0356 & \hspace{-0.3cm}\footnotesize{\textpm 0.0029} & 20.00 & \hspace{-0.3cm}\footnotesize{\textpm 0.00} & 250 & \multicolumn{3}{c}{$\{ 250 \}$} \\ 
 && 10\,000 \footnotesize{(10\,000)} & 100 & 4.75 & 0.69 & 0.0647 & \hspace{-0.3cm}\footnotesize{\textpm 0.0002} & 0.0353 & \hspace{-0.3cm}\footnotesize{\textpm 0.0021} & 20.00 & \hspace{-0.3cm}\footnotesize{\textpm 0.00} & 500 & \multicolumn{3}{c}{$\{ 500 \}$} \\ 

	&&&&&&&&&&&&&&&\\[-0.35cm]\hline&&&&&&&&&&&&&&&\\[-0.25cm] 
	\multirow{4}{*}{ \begin{turn}{90} RC-SVM  \end{turn}} &
	\multirow{4}{*}{ \hspace{-0.3cm} \begin{turn}{90} \# ws = 40  \end{turn}}
	  & \phantom{0}1\,000 \footnotesize{(10\,000)} & 100 & 3.89 & 0.60 & 0.0707 & \hspace{-0.3cm}\footnotesize{\textpm 0.0011} & 0.0850 & \hspace{-0.3cm}\footnotesize{\textpm 0.0065} & 40.00 & \hspace{-0.3cm}\footnotesize{\textpm 0.00} & 25 & \multicolumn{3}{c}{$\{ 25 \}$} \\ 
 && \phantom{0}2\,500 \footnotesize{(10\,000)} & 100 & 4.05 & 0.81 & 0.0669 & \hspace{-0.3cm}\footnotesize{\textpm 0.0005} & 0.0577 & \hspace{-0.3cm}\footnotesize{\textpm 0.0039} & 40.00 & \hspace{-0.3cm}\footnotesize{\textpm 0.00} & 62 & $[ 62$ & \hspace{-0.5cm}, & \hspace{-0.5cm}$63 ]$ \\ 
 && \phantom{0}5\,000 \footnotesize{(10\,000)} & 100 & 4.47 & 1.00 & 0.0650 & \hspace{-0.3cm}\footnotesize{\textpm 0.0002} & 0.0388 & \hspace{-0.3cm}\footnotesize{\textpm 0.0030} & 40.00 & \hspace{-0.3cm}\footnotesize{\textpm 0.00} & 125 & \multicolumn{3}{c}{$\{ 125 \}$} \\ 
 && 10\,000 \footnotesize{(10\,000)} & 100 & 5.66 & 1.24 & 0.0649 & \hspace{-0.3cm}\footnotesize{\textpm 0.0002} & 0.0379 & \hspace{-0.3cm}\footnotesize{\textpm 0.0022} & 40.00 & \hspace{-0.3cm}\footnotesize{\textpm 0.00} & 250 & \multicolumn{3}{c}{$\{ 250 \}$} \\ 

	&&&&&&&&&&&&&&&\\[-0.35cm]\hline&&&&&&&&&&&&&&&\\[-0.25cm] 
	\multirow{4}{*}{ \begin{turn}{90} RC-SVM  \end{turn}} &
	\multirow{4}{*}{ \hspace{-0.3cm} \begin{turn}{90} \# ws = 50  \end{turn}}
	  & \phantom{0}1\,000 \footnotesize{(10\,000)} & 100 & 4.70 & 0.69 & 0.0726 & \hspace{-0.3cm}\footnotesize{\textpm 0.0010} & 0.0960 & \hspace{-0.3cm}\footnotesize{\textpm 0.0051} & 50.00 & \hspace{-0.3cm}\footnotesize{\textpm 0.00} & 20 & \multicolumn{3}{c}{$\{ 20 \}$} \\ 
 && \phantom{0}2\,500 \footnotesize{(10\,000)} & 100 & 5.06 & 0.90 & 0.0675 & \hspace{-0.3cm}\footnotesize{\textpm 0.0005} & 0.0627 & \hspace{-0.3cm}\footnotesize{\textpm 0.0038} & 50.00 & \hspace{-0.3cm}\footnotesize{\textpm 0.00} & 50 & \multicolumn{3}{c}{$\{ 50 \}$} \\ 
 && \phantom{0}5\,000 \footnotesize{(10\,000)} & 100 & 5.33 & 1.15 & 0.0651 & \hspace{-0.3cm}\footnotesize{\textpm 0.0002} & 0.0410 & \hspace{-0.3cm}\footnotesize{\textpm 0.0027} & 50.00 & \hspace{-0.3cm}\footnotesize{\textpm 0.00} & 100 & \multicolumn{3}{c}{$\{ 100 \}$} \\ 
 && 10\,000 \footnotesize{(10\,000)} & 100 & 6.57 & 1.49 & 0.0649 & \hspace{-0.3cm}\footnotesize{\textpm 0.0002} & 0.0383 & \hspace{-0.3cm}\footnotesize{\textpm 0.0020} & 50.00 & \hspace{-0.3cm}\footnotesize{\textpm 0.00} & 200 & \multicolumn{3}{c}{$\{ 200 \}$} \\ 

      \end{tabular}
    } 
    \caption{Experimental results relating to the artificial data of Type \RM{5}}
    \label{tab::experiments::art_data_5_2}
  \end{minipage} 
\end{sidewaystable}

\end{document}